\documentclass[runningheads]{llncs}
\usepackage[T1]{fontenc}
\usepackage{graphicx}
\usepackage[hyphens]{url}  
\usepackage{graphicx,xcolor} 
\usepackage{algorithm}

\usepackage{caption} 
\usepackage{dsfont}
\usepackage{amsfonts}
\usepackage{amsmath}
\usepackage{booktabs}
\usepackage{algpseudocode}
\usepackage{makecell}
\usepackage{arydshln}

\newcommand{\RNum}[1]{\expandafter{\romannumeral #1\relax}}
\newcommand{\etal}{\textit{et al.}}

\begin{document}
\title{Gradient Adjusting Networks for\\ Domain Inversion}
%
%

\author{Erez Sheffi \and Michael Rotman \and Lior Wolf}
\authorrunning{E. Sheffi et al.}
\institute{Blavatnik School of Computer Science, Tel Aviv University}
%
\maketitle              

\begin{abstract}
StyleGAN2 was demonstrated to be a powerful image generation engine that supports semantic editing. However, in order to manipulate a real-world image, one first needs to be able to retrieve its corresponding latent representation in StyleGAN's latent space that is decoded to an image as close as possible to the desired image. For many real-world images, a latent representation does not exist, which necessitates the tuning of the generator network. We present a per-image optimization method that tunes a StyleGAN2 generator such that it achieves a local edit to the generator's weights, resulting in almost perfect inversion, while still allowing image editing, by keeping the rest of the mapping between an input latent representation tensor and an output image relatively intact. The method is based on a one-shot training of a set of shallow update networks (aka. Gradient Modification Modules) that modify the layers of the generator. After training the Gradient Modification Modules, a modified generator is obtained by a single application of these networks to the original parameters, and the previous editing capabilities of the generator are maintained.  Our experiments show a sizable gap in performance over the current state of the art in this very active domain. Our code is available at \url{https://github.com/sheffier/gani}.

\end{abstract}

\section{Introduction}
\label{sec:intro}
The ability to distinguish between synthetic images and real ones has become increasingly challenging since the introduction of Generative Adversarial Networks (GAN)~\cite{goodfellow2014generative}.
Although the images produced by this framework are often indistinguishable from real ones, one lacks the ability to control the specific outcome. Most relevant to our work is the StyleGAN~\cite{karras2019style} family of generators, which can produce, for example, realistic faces based on random input vectors. However, for most real-world face images, one cannot find an input vector that would result in exactly the same image.

This is a key limitation of StyleGANs (and other GANs), with far-reaching implications at the application level. It has been repeatedly shown that the StyleGAN latent space displays semantic properties that make it especially suited for image editing applications, see~\cite{bermano2022state} for a survey. However, without the ability to embed real-world images within this space, these capabilities are limited mostly to synthetic images.

A set of techniques were, therefore, developed in order to tune the StyleGAN generator $G$ such that it would produce the desired image~\cite{karras2019style,roich2021pivotal,alaluf2021hyperstyle}. This has to be done carefully, since performing this tuning too aggressively would lead to the loss of the semantic properties. While being a very active research domain, fuelled by many concrete applications, results are still lacking and the generated copy is still distinctively different from the desired image.

Existing methods either (\RNum{1}) optimize one image at a time, or (\RNum{2}) employ pretrained feedforward networks to produce a modification of $G$, given an input image $I$. The first category is believed to be more accurate, but slower at inference time than the second.

Our work combines elements from both approaches -- we propose an optimization procedure for a single image, and our procedure employs trainable networks. The role of these networks is to estimate the change one wishes to apply to the parameters of $G$. By optimizing the networks to control this change rather than applying it directly, we regularize the change being applied. The per-layer networks we train adapt the parameters of $G$ based on previous parameter variations. Thus, these changes are applied in a very local manner, separately to each layer. Through weight sharing, the capacity of these networks is limited and ensures that this mapping is relatively simple.

In addition to a novel architecture, we also modify the loss term that is used and show that a face-parsing network provides a strong signal, side by side with a face-identification network that is widely used in the relevant literature.

Our results demonstrate: (\RNum{1}) a far more faithful inversion in comparison to the state-of-the-art methods, (\RNum{2}) a more limited effect on the result of applying $G$ to other vectors in its input space, (\RNum{3}) a superb ability to support downstream editing applications.

\section{Related Work}
\label{sec:related_work}
We describe adversarial image generation methods and ways to control the generated image to fit an input image.

{\bf Generative Adversarial Networks \quad}
Generative Adversarial Networks (GAN)~\cite{goodfellow2014generative} are a family of generative models composed of two neural networks: a generator and a discriminator. The generator is tasked with learning a mapping from a latent space to a given data distribution, whereas the discriminator aims to distinguish between generated samples and real ones. GANs have been widely applied in many computer vision tasks, such as generating super-resolution images~\cite{ledig2017photo,wang2018esrgan}, image-to-image translation~\cite{zhu2017unpaired}, and face generation~\cite{karras2017progressive,karras2019style,karras2020analyzing}.

Once trained, given an input vector, the generator produces a realistic image. Since the mapping between the input vector and the image space is not trivial, it is hard to predict the generated image from the input vector.
One way of controlling the synthesis is by feeding additional information to the GAN during the training phase, for instance, by adding additional discrete~\cite{mirza2014conditional,odena2017conditional}, or continuous~\cite{ding2021ccgan} labels as inputs.
A major caveat in this approach is that it requires additional supervision. To bypass this limitation, other approaches constrain the input vector space directly, either by applying tools from information theory~\cite{chen2016infogan} or by limiting this space to a non-trivial topology~\cite{rotman2022unsupervised}. As a result, the input vector space is disentangled: different entries of the input vector control a different aspect or "generative factor" of the generated image.
It has been observed~\cite{radford2015unsupervised,karras2019style} that a continuous translation between two vectors in the GAN latent space leads to a continuous change in the generated image. Shen \etal~\cite{shen2020interpreting} has further expanded this observation to face generation, where it was shown that facial attributes lie in different hyperplanes, and are therefore controllable using vector arithmetic in the latent space. Further investigation into the latent space structure has been performed by applying unsupervised methods, such as Principal Component Analysis (PCA)~\cite{harkonen2020ganspace} or eigenvalue decomposition~\cite{shen2021closed}, or by using semantic labels~\cite{shen2020interpreting,abdal2021styleflow}.
The existence of an underlying structure, and the presence of a semantic algebra, make it possible to edit the generated images.

\noindent
{\bf StyleGAN \quad}
StyleGAN~\cite{karras2019style} is a style-based generator architecture.
Unlike the traditional GAN, which maps a noisy signal directly to images, it splits the noisy signal into two:
(\RNum{1}) a style generating signal $z\in \mathbb{R}^{512}$ to a style latent space, $w \in W = \mathbb{R}^{512}$,
which is used globally as a set of layer-wise parameters for the GAN
(\RNum{2}) additional noise, which is added to the feature maps after each convolutional layer.
This design benefits from both stochastic variation and scale-specific feature synthesis.
StyleGAN2~\cite{karras2020analyzing} introduced a path-length regularization term that enables smoother style mapping between $Z$ and $W$,
together with a better normalization scheme for the generator.

\noindent
{\bf Image Editing \quad}
By understanding the influence of the different entries in the latent space, image synthesis can be controlled~\cite{chen2016infogan}.
Specifically for face generation, this mapping has been investigated in two directions, unsupervised and supervised.
The unsupervised path aims to unveil the domain's structure by applying PCA~\cite{harkonen2020ganspace} or eigenvalue decomposition~\cite{shen2021closed}.
Other works directly influence this mapping by conditioning it on supervised labels~\cite{shen2020interpreting,abdal2021styleflow}.
The existence of such an underlying structure, and the understanding of its algebra, enables the semantic modification of generated images,
for instance, for facial editing~\cite{tewari2020stylerig}.
The ability to control the generated samples is a key aspect of making GANs useful for real-life applications.

\noindent
{\bf Image Inversion \quad}
In order to edit a real image using a latent space modification, the originating point in the latent space has to be identified.
The solutions to the inverse problem can be divided into three families: (\RNum{1}) optimization-based (\RNum{2}) encoder-based (\RNum{3}) generator-modifying.
Unlike the last approach, the first two families do not alter any of the generator parameters.
In the optimization-based approach, a latent code $w^*$ is evaluated given an input image, $I$, in an iterative manner~\cite{creswell2018inverting,ling2021editgan},
such that, $I \approx G(w^*, \theta) $, where $\theta$ are the generator's parameters.

For the task of face generation, Karras \etal~\cite{karras2020analyzing} proposed an optimization-based inversion scheme for StyleGAN2,
where both the latent code $w$ and the injected noise are optimized together,
combined with a regularization term that minimizes the auto-correlation of the injected noise at different scales.
Abdal~\etal~\cite{Abdal2019Image2StyleGANHT} extended this direction, by expanding latent space $W$ to $W^+$.
To accomplish this, instead of constraining the latent code, $w\in W$, to be identical for all convolutional layers,
each layer is now fed with a different set of parameters.
This modification extended the dimensionality of the latent code from $\mathbb{R}^{512}$ to $\mathbb{R}^{18\times 512}$.
To preserve spatial details during inversion, Zhang\etal~\cite{zhang2021sals} considered a spatially-structured latent space,
replacing the original one-dimensional representation, $W$.

The second family of solutions is encoder-based. These methods utilize an encoder, $E$, that maps between the image space and the latent space,
$w^* = E\left( I\right)$. Unlike the iterative family, these encoders are trained on a set of samples~\cite{guan2020collaborative,luo2017learning}.
These image encoders have also been applied for face generation, where they are employed during training,
by combining the auto-encoder framework with a GAN~\cite{pidhorskyi2020adversarial}, or more commonly, on pre-trained generative models,
which require far less training data. These approaches focus on mapping an image to an initial latent code, $w\in W$,
which may later be fine-tuned using an optimization method~\cite{zhu2020domain}.
pSp~\cite{richardson2020encoding} employs a different scheme, in which an additional fine-tuning of $w$ is not performed;
instead, a pyramid-based encoder is designed for each style vector of the StyleGAN2 framework.
e4e~\cite{tov2021designing} further expands on this idea by limiting the hierarchical structure of the $w$ codes to be of a residual type,
so that each style vector is the sum of a basis style vector and a residual part.
Further improvement was achieved for the inversion problem by iteratively encoding the image onto the latent space and feeding the generated image back to the encoder as an input~\cite{alaluf2021restyle}.

In the last approach, an initial $w$ latent code is evaluated using an encoder. As a second step, the generator is tuned to produce the required image from the $w$~\cite{roich2021pivotal}.
This tuning procedure has been further improved by employing hypernetworks~\cite{alaluf2021hyperstyle}.
In this scenario, instead of directly modifying the generator weights, a set of residual weights is evaluated using an additional neural network.
The input to this network is the target image. Therefore, given a new image, a new set of weights is computed.

A faithful edit is also required to preserve the original identity. Liang \etal~\cite{liang2021ssflow} applied Neural Spline Networks to find faithful editing directions.
Editing local aspects was accomplished by manipulating specific parts of the feature maps throughout the generation process~\cite{wang2021attribute}.

\section{Method}
\label{sec:method}
Our approach adapts a StyleGAN generator for one image at a time by adding a small correction to the generator’s parameters.
This correction is computed using the novel gradient modification modules, a set of small neural networks that map between the gradients of the loss criteria with respect to the generator's parameters to the parameters' corrections.
During training, only the parameters of the gradient modification modules are optimized (the generator parameters are kept frozen) with respect to the loss criteria.
The output of these modules is used to update the parameters of StyleGAN's generator, resulting in a new generator that is capable of faithfully generating the input image.
Note that the gradient modification modules are not added to StyleGAN and that the structure of StyleGAN is not modified at all.

Given a candidate target image $I$ to be edited, a corresponding latent code, $w \in W^+ = \mathbb{R}^{18 \times 512}$, is estimated using the off-the-shelf encoder e4e~\cite{tov2021designing}.
The latent code, $w$, is fed into the pre-trained generator to produce a reconstructed image, $G(w) = G(w,\theta)$,
where the right-hand side explicitly states the parameters $\theta$ of network $G$. Since $w$ is not an exact solution to the inverse problem,
the reconstructed image is usually of poor quality. In our method, $w$ does not change.
Instead, we tune the generator parameters $\theta$ to improve the generated image, obtaining $G(w,\theta')\sim I$, where $\theta'$ are the tuned parameters.

Consider the following image similarity loss function,
\begin{equation}
    \textstyle
    \mathcal{L}\left(I_1, I_2\right) = \lambda_1\mathcal{L}_{\text{rec}} + \lambda_2 \mathcal{L}_{lpips} + \lambda_3 \mathcal{L}_{sim} + \lambda_4 \mathcal{L}_{FP} \,,
    \label{eq:specloss}
\end{equation}
which is the sum of four terms. The first term is a pixel-wise reconstruction loss, the $L_2$ or the $L_{1}^{\text{smooth}}$~\cite{girshick2015fast} distance between the images $I_1$ and $I_2$.
The second term, $\mathcal{L}_{lpips}$, is a perceptual similarity loss~\cite{zhang2018unreasonable}, that relies on feature maps from a pre-trained AlexNet~\cite{simonyan2014very} on the ImageNet dataset.
$\mathcal{L}_{sim}$ is an identity-preserving similarity loss that is applied on a pair of real and reconstructed images. This loss term accounts for the cosine distance of feature vectors extracted from a pre-trained ArcFace~\cite{deng2019arcface} facial recognition network for the facial domain, and a pre-trained MoCo~\cite{Chen2020ImprovedBW} for the non-facial domains, following ~\cite{tov2021designing,alaluf2021hyperstyle}.
The last term is a multi-layer face-parsing loss, $\mathcal{L}_{FP}$. Similarly to $\mathcal{L}_{sim}$, it measures the layer-wise aggregated cosine distance of all the feature vectors from the contracting path of the pre-trained facial parsing network P~\cite{CelebAMask-HQ}, with a U-Net~\cite{ronneberger2015u} backbone. In total, 5 feature vectors are used for the cosine distance evaluation.

Tuning the $G$'s parameters, $\theta$, involves the estimation of the modified parameters, $\theta'$, using a set of feed-forward networks.  Let $l_i$ be the $i$th layer of $G$, and let
$\theta_i$ and $\frac{\partial \mathcal{L}}{\partial\theta_i}$ be its learnable parameters and the gradients of the objective function w.r.t to these parameters, respectively. Unlike PTI~\cite{roich2021pivotal}, which applies a regular gradient step to update $\theta_i$, the gradient updates are based on the mapping,
\begin{equation}
    \textstyle
    \theta_i' = \theta_i \odot \left(1 + \Delta\theta_i \right)=  \theta_i \odot \left(1 + M_i \left( \frac{\partial \mathcal{L}}{\partial\theta_i}\right)\right) \,,
\end{equation}
where $\mathcal{M}=\left\{M_i \right\}$ is a set of gradient modification modules, each mapping between the original gradients, $\frac{\partial\mathcal{L}}{\partial \theta_i}$, and the parameter correction,  $\Delta\theta_i$.

Each module $M_i$ contains a sequential set of $l=1\dots L$ residual blocks~\cite{he2016identity}. Let $y_l$ be the input to the $l$th residual block. The output of block $l$, $y_{l+1}$ is then:
\begin{align}
    \textstyle
    r       & =  W_2^l \sigma\left(\text{SN}^l_2 \left( W_1^l \sigma\left(\text{SN}^l_1\left(y_l\right)\right)\right)\right) + b^l \\
    y_{l+1} & =  y_l + r \label{eq:residual} \,,
\end{align}
where $\sigma$ is the LeakyReLU~\cite{maas2013rectifier} activation function, with a slope of $0.01$ and $\text{SN}$ is the Scale Normalization (SN)~\cite{nguyen2019transformers}.
The learnable parameters of each block reside in two linear operators, $W_1^l$ and $W_2^l$, in the bias term, $b^l$, and in the scale coefficient of the SN layer.
The parameters of $W^l_1$ and $W^l_2$ are initialized according to~\cite{he2015delving} from the normal distribution, with an initial standard deviation factored by $0.1$, whereas the bias, $b^l$ is sampled uniformly.
\begin{figure}[t]
    \centering
    \includegraphics[height=0.20\paperheight]{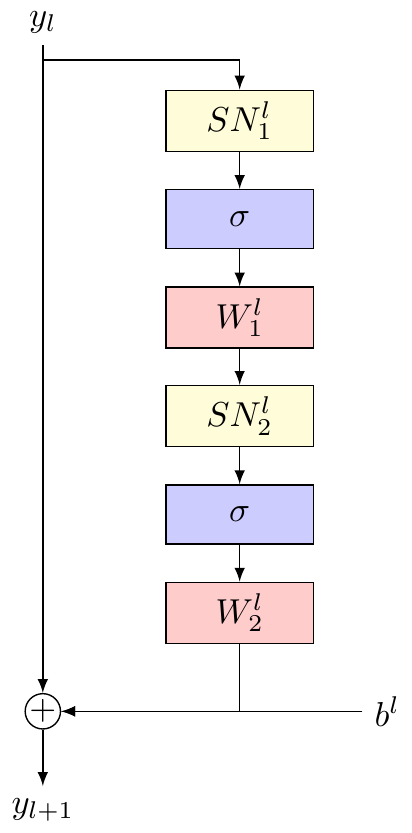}
    \caption{%
        The structure of one residual block of a gradient modification module, $\mathcal{M}_i$. The input to the first block, $y_0$ is, $\frac{\partial \mathcal{L}}{ \partial \theta_i}$.}
    \label{fig:Mblock}
\end{figure}
All the $M_i$ networks assigned to convolutional layers with the number of channels, $c_{in} = c_{out} =512$, share the same parameters, $\psi_i = \left\{ W^l_1, W^l_2, b^l, \text{SN}^l_1, \text{SN}^l_2\right\}_{l=1}^{L}$.
During optimization, (only) the parameters $\left\{\psi_i\right\}$ of the set $\mathcal{M}$ are optimized to minimize the loss function,
\begin{equation}
    \textstyle
    \mathcal{L}_{total} = \lambda_e \mathcal{L}\left( I,G(w,\theta') \right) + \lambda_{l}\mathcal{L}\left( G(f(z),\theta),G(f(z),\theta')\right) \,
    \label{eq:loss}
\end{equation}
where $z \sim \mathcal{N} \left( 0, \mathds{1} \right) \in \mathbb{R}^{512}$, and $f$ is the style mapping of StyleGAN~\cite{karras2019style} so style mapping, $f(z)$ lies in $W$ (and not in $W^+$). The first term in Eq.~\eqref{eq:loss} is responsible for generating a high-quality image.
It is crucial that the optimization procedure should not dramatically alter the mapping $G: w \rightarrow I$ as a large variation in the mapping would require the re-identification of the editing directions. The second term in Eq.~\eqref{eq:loss}, a localization regularizer~\cite{roich2021pivotal,mitchell2021fast}, prevents the generator from drifting by forcing the generator to produce identical images for randomly sampled latent codes.
An outline of our method appears in Algorithm~\ref{alg:cap}.

\begin{algorithm}[t]
    \caption{Method Outline}\label{alg:cap}
    \begin{algorithmic}
        \Require Image $I$, generator $G_\theta$, encoder $E$, $\mathcal{M}$ \vspace{2pt}
        \Ensure $G(w,\theta')$
        \State $w \leftarrow E(I)$ \Comment{Obtain $w$ from a pre-trained encoder}
        \For{$i \in  1\dots n_{\text{iterations}}$}\vspace{5pt}
        \State Compute $\frac{\partial \mathcal{L}(I,G(w,\theta))}{\partial \theta} $
        \For{$M_i \in \mathcal{M}$}
        \State $\Delta \theta_i = M_i \left( \frac{\partial \mathcal{L}}{\partial\theta_i}\right)$ \Comment{$\mathcal{L}$ in Eq.~\eqref{eq:specloss}}
        \EndFor
        \State Sample $z \sim \mathcal{N} \left( 0, \mathds{1} \right) \in \mathbb{R}^{512}$ \Comment{Localization}
        \State Compute $\frac{\partial \mathcal{L}_{total}}{\partial \psi}$ \Comment{$\mathcal{L}_{total}$ in Eq.~\eqref{eq:loss}}
        \State Update the parameters $\psi$ of $\mathcal{M}$
        \EndFor
    \end{algorithmic}
\end{algorithm}

\section{Experiments}
\label{sec:experiments}

We conduct an extensive set of experiments on various image synthesis datasets. For all image domains, our method utilizes a pre-trained StyleGAN2. For the facial domain, this pre-trained network was optimized for generating facial images distributed according to the FFHQ~\cite{karras2019style} dataset, whereas the inversion and editing capabilities of our approach are evaluated over images from the test set of the Celeb-HQ~\cite{karras2017progressive} dataset.
Our method is also evaluated on Church and Horses from the LSUN~\cite{yu2015lsun} dataset, on automobiles from Stanford Cars~\cite{KrauseStarkDengFei-Fei_3DRR2013}, and on wildlife images from AFHQ-WILD~\cite{choi2020starganv2}. The pre-trained models for these domains were acquired from the e4e~\cite{tov2021designing} and HyperStyle~\cite{alaluf2021hyperstyle} official GitHub repositories. The initial inverted vectors, $w\in W^+$, were also obtained using the e4e encoders in these repositories. Since there is no pre-trained encoder to $W^+$ space for the AFHQ-WILD dataset, our method is evaluated on this using the $W$ latent space.

For all experiments, the Ranger~\cite{Ranger} optimizer was used, with a learning rate of $0.001$. The number of iterations and the loss coefficients used for our method appears in Table~\ref{table:expsetup}. For the CelebA-HQ and AFHQ-Wild datasets, $\mathcal{L}_{\text{pixel-wise}} = L_{2}$ and for Stanford Cars, LSUN Church, and LSUN Horses,   $\mathcal{L}_{\text{pixel-wise}} = L_{1}^{\text{smooth}}$ with $\beta=0.1$. The difference in the number of iterations depends on the quality of the reconstruction of the original generator $G$ from $w$. Both LSUN Church and LSUN Horses resulted in a poor initial reconstruction and required twice as many iterations.

\begin{table}[t]
    \centering
    \caption{The loss coefficients and number of running iterations that were used by our method for each of the different datasets.}
    \label{table:expsetup}
    \begin{tabular}{@{}lccccccc@{}}
        \toprule
                      & Iterations & $\lambda_{e}$ & $\lambda_{l}$ & $\lambda_1 $ & $\lambda_2 $ & $\lambda_3 $ & $\lambda_4 $ \\
        \midrule
        CelebA-HQ     & 300        & 1.0           & 0.2           & 1.0          & 0.8          & 0.1          & 1.0          \\
        AFHQ-Wild     & 300        & 1.0           & 0.2           & 1.0          & 0.8          & 0.1          & 1.0          \\
        Stanford Cars & 400        & 1.0           & 0.2           & 1.0          & 0.8          & 0.1          & -            \\
        LSUN Horses   & 800        & 1.0           & 0.2           & 1.0          & 0.8          & 0.1          & -            \\
        LSUN Church   & 800        & 1.0           & 0.2           & 1.0          & 0.8          & 0.1          & -            \\
        \bottomrule
    \end{tabular}
\end{table}

Table~\ref{table:results_celeba_hq} presents four evaluation criteria for reconstruction quality over the CelebA-HQ dataset. The four metrics compare the input image $I$ with the one generated by the inversion method, $G(w,\theta')$. These metrics are the pixel-wise similarity, which is the Euclidean norm $L_2$, the perceptual similarity, LPIPS~\cite{zhang2018unreasonable}, the structural similarity score, MS-SSIM~\cite{wang2003multiscale}, and identity similarity, ID~\cite{huang2020curricularface}, between the input image, $I$, and its reconstruction, $G(w,\theta')$. Our method outperforms all other methods in all metrics except LPIPS, where the original StyleGAN2 inversion approach matches ours.

\begin{table}[t]
\centering
        \caption{Reconstruction metrics on the CelebA-HQ test set.}
        \label{table:results_celeba_hq}
        \begin{tabular}{@{}l@{\hspace{2pt}}c@{\hspace{2pt}}c@{\hspace{2pt}}c@{\hspace{2pt}}c@{}}
            \toprule
            Method                                                 & $\uparrow$ID    & $\uparrow$MS-SSIM & $\downarrow$LPIPS & $\downarrow$ $L_2$ \\ \midrule
            StyleGAN2~\cite{karras2020analyzing}                   & $0.78$          & $0.90$            & $0.020$           & $0.090$            \\
            PTI~\cite{roich2021pivotal}                            & $0.85$          & $0.92$            & $0.090$           & $0.015$            \\
            IDInvert~\cite{zhu2020domain}                          & $0.18$          & $0.68$            & $0.220$           & $0.061$            \\
            pSp~\cite{richardson2020encoding}                      & $0.56$          & $0.76$            & $0.170$           & $0.034$            \\
            e4e~\cite{tov2021designing}                            & $0.50$          & $0.72$            & $0.200$           & $0.052$            \\
            $\text{ReStyle}_{\text{pSp}}$~\cite{alaluf2021restyle} & $0.66$          & $0.79$            & $0.130$           & $0.030$            \\
            $\text{ReStyle}_{\text{e4e}}$~\cite{alaluf2021restyle} & $0.52$          & $0.74$            & $0.190$           & $0.041$            \\
            NP-GAN-I~\cite{Feng2022NearPG}                         & $-$             & $-$               & $0.283$           & $0.004$            \\
            HyperStyle~\cite{alaluf2021hyperstyle}                 & $0.76$          & $0.84$            & $0.090$           & $0.019$            \\
            HFGI~\cite{Bai2022HighfidelityGI}                      & $-$             & $-$               & $0.100$           & $0.021$            \\
            HyperInverter~\cite{Dinh2021HyperInverterIS}           & $0.60$          & $0.67$            & $0.105$           & $0.024$            \\
            Feature-Style Encoder~\cite{Yao2022FeatureStyleEF}     & $0.87$          & $-$               & $0.066$           & $0.019$            \\
            Ours                                  & $\mathbf{0.99}$ & $\mathbf{0.97}$   & $\mathbf{0.020}$  & $\mathbf{0.003}$   \\
            \bottomrule
        \end{tabular}
\end{table}

Table~\ref{table:results_stanford_cars} evaluates the reconstruction quality for the Stanford Cars dataset, Table~\ref{table:results_afhq_wild} for the AFHQ-Wild dataset, Table~\ref{table:results_lsun_church} for the LSUN church dataset and Table~\ref{table:results_lsun_horses} for the LSUN horses dataset, using three evaluation criteria: the pixel-wise Euclidean norm, $L_2$, the perceptual similarity, LPIPS, and the MS-SSIM~\cite{wang2003multiscale} score.
Our method outperforms all other methods on the AFHQ-Wild and LSUN church datasets. In the Stanford Cars and the LSUN horses datasets, our approach surpasses other approaches in all evaluation metrics, except for the $L_2$ metric, where it is second to Near Perfect GAN Inversion (NP-GAN-I)~\cite{Feng2022NearPG}.

\begin{table}[t]
\centering
        \caption{Reconstruction metrics on the Stanford Cars dataset~\cite{KrauseStarkDengFei-Fei_3DRR2013} test set.}
        \label{table:results_stanford_cars}
        \begin{tabular}{@{}lccc@{}}
            \toprule
            Method                                                 & $\uparrow$MS-SSIM & $\downarrow$LPIPS & $\downarrow$ $L_2$ \\
            \midrule
            StyleGAN2~\cite{karras2020analyzing}                   & $0.79$            & $0.16$            & $0.060$            \\
            PTI~\cite{roich2021pivotal}                            & $0.93$            & $0.11$            & $0.010$            \\
            pSp~\cite{richardson2020encoding}                      & $0.58$            & $0.29$            & $0.100$            \\
            e4e~\cite{tov2021designing}                            & $0.53$            & $0.32$            & $0.120$            \\
            $\text{ReStyle}_{\text{pSp}}$~\cite{alaluf2021restyle} & $0.66$            & $0.25$            & $0.070$            \\
            $\text{ReStyle}_{\text{e4e}}$~\cite{alaluf2021restyle} & $0.60$            & $0.29$            & $0.090$            \\
            HyperStyle~\cite{alaluf2021hyperstyle}                 & $0.67$            & $0.27$            & $0.070$            \\
            NP-GAN-I~\cite{Feng2022NearPG}                         & $-$               & $0.15$            & $\mathbf{0.006}$   \\
            Ours                                  & $\mathbf{0.94}$   & $\mathbf{0.03}$   & $0.010$            \\
            \bottomrule
        \end{tabular}
\end{table}

\begin{table}[t]
\centering
        \caption{Reconstruction metrics for AFHQ-Wild test set.}
        \label{table:results_afhq_wild}
        \begin{tabular}{@{}lccc@{}}
            \toprule
            Method                                                 & $\uparrow$MS-SSIM & $\downarrow$LPIPS & $\downarrow$ $L_2$ \\
            \midrule
            StyleGAN2~\cite{karras2020analyzing}                   & $0.82$            & $0.13$            & $0.030$            \\
            PTI~\cite{roich2021pivotal}                            & $0.93$            & $0.08$            & $0.010$            \\
            pSp~\cite{richardson2020encoding}                      & $0.51$            & $0.35$            & $0.130$            \\
            e4e~\cite{tov2021designing}                            & $0.47$            & $0.36$            & $0.140$            \\
            $\text{ReStyle}_{\text{pSp}}$~\cite{alaluf2021restyle} & $0.57$            & $0.21$            & $0.050$            \\
            $\text{ReStyle}_{\text{e4e}}$~\cite{alaluf2021restyle} & $0.52$            & $0.25$            & $0.070$            \\
            HyperStyle~\cite{alaluf2021hyperstyle}                 & $0.56$            & $0.24$            & $0.060$            \\
            NP-GAN-I~\cite{Feng2022NearPG}                         & $-$               & $0.38$            & $0.014$            \\
            Ours                                  & $\mathbf{0.96}$   & $\mathbf{0.03}$   & $\mathbf{0.006}$   \\
            \bottomrule
        \end{tabular}
\end{table}

\begin{table}[t]
\centering
        \caption{Reconstruction metrics on the LSUN churches~\cite{yu2015lsun} (outdoor)  test set.}
        \label{table:results_lsun_church}
        \begin{tabular}{@{}lccc@{}}
            \toprule
            Method                                     & $\uparrow$MS-SSIM & $\downarrow$LPIPS & $\downarrow$ $L_2$ \\
            \midrule
            StyleGAN2~\cite{karras2020analyzing}       & $0.4797$          & $0.325$           & $0.167$            \\
            PTI~\cite{roich2021pivotal}                & $0.6968$          & $0.097$           & $0.051$            \\
            pSp~\cite{richardson2020encoding}          & $0.4070$          & $0.310$           & $0.130$            \\
            e4e~\cite{karras2020analyzing}             & $0.3481$          & $0.420$           & $0.140$            \\
            $\text{ReStyle}$~\cite{alaluf2021restyle}  & $0.3878$          & $0.377$           & $0.127$            \\
            HFGI~\cite{Bai2022HighfidelityGI}          & $-$               & $0.220$           & $0.090$            \\
            HyperInverter~\cite{Yao2022FeatureStyleEF} & $0.5762$          & $0.223$           & $0.091$            \\
            Ours                      & $\mathbf{0.9479}$ & $\mathbf{0.015}$  & $\mathbf{0.014}$   \\
            \bottomrule
        \end{tabular}

\end{table}

\begin{table}[t]
    \centering
    \caption{Reconstruction metrics for LSUN horses.}
    \label{table:results_lsun_horses}
    \begin{tabular}{@{}lccc@{}}
        \toprule
        Method                                    & $\uparrow$MS-SSIM & $\downarrow$LPIPS & $\downarrow$ $L_2$ \\
        \midrule
        e4e~\cite{karras2020analyzing}            & $0.25$            & $0.431$           & $0.1740$           \\
        $\text{ReStyle}$~\cite{alaluf2021restyle} & $-$               & $0.525$           & $0.159$            \\
        NP-GAN-I~\cite{Feng2022NearPG}            & $-$               & $0.141$           & $\mathbf{0.005}$   \\
        Ours                     & $\mathbf{0.94}$   & $\mathbf{0.016}$  & $0.010$            \\
        \bottomrule
    \end{tabular}
\end{table}

\begin{figure}[t]
    \centering
    \footnotesize
    \renewcommand{\arraystretch}{0.0}
    \begin{tabular}{@{}c@{}c@{}c@{}c@{}c@{}c@{}c@{}}
        \includegraphics[width=0.135\linewidth]{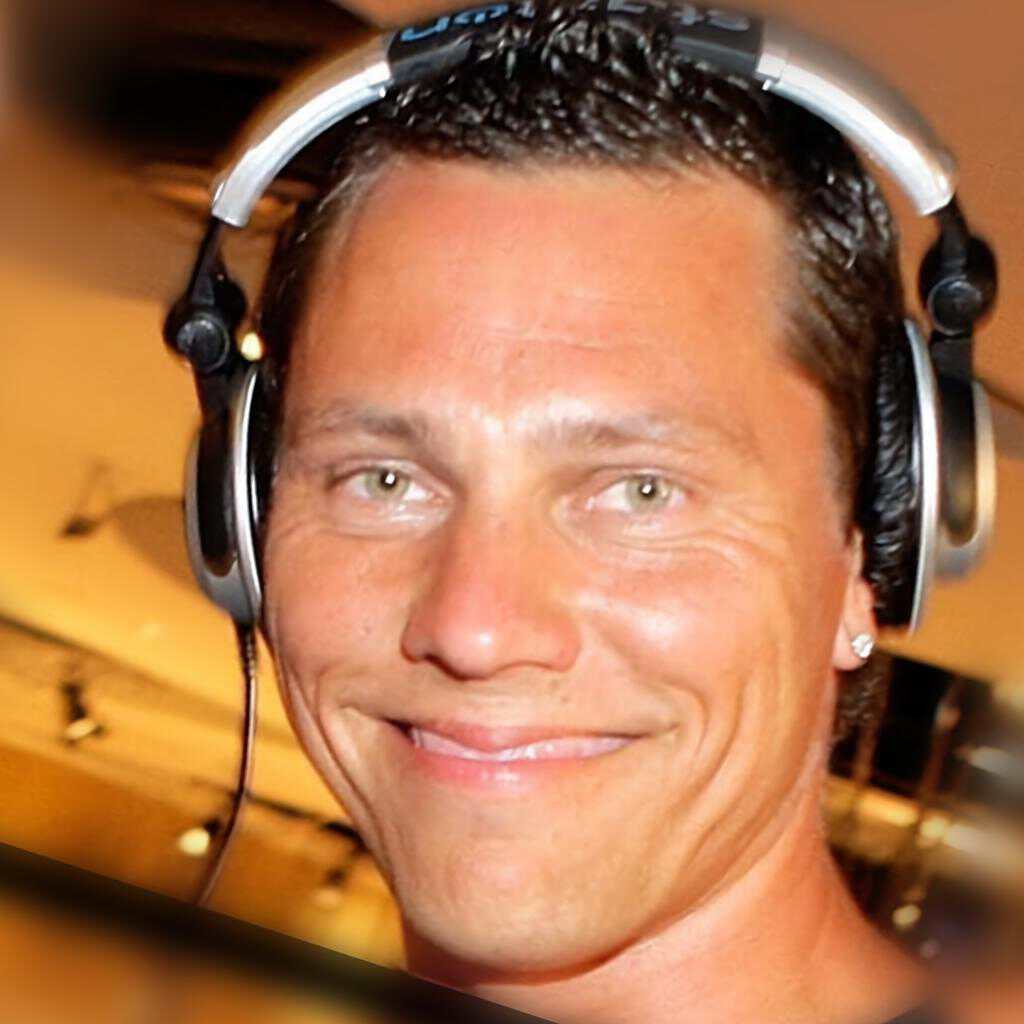}                    &
        \includegraphics[width=0.135\linewidth]{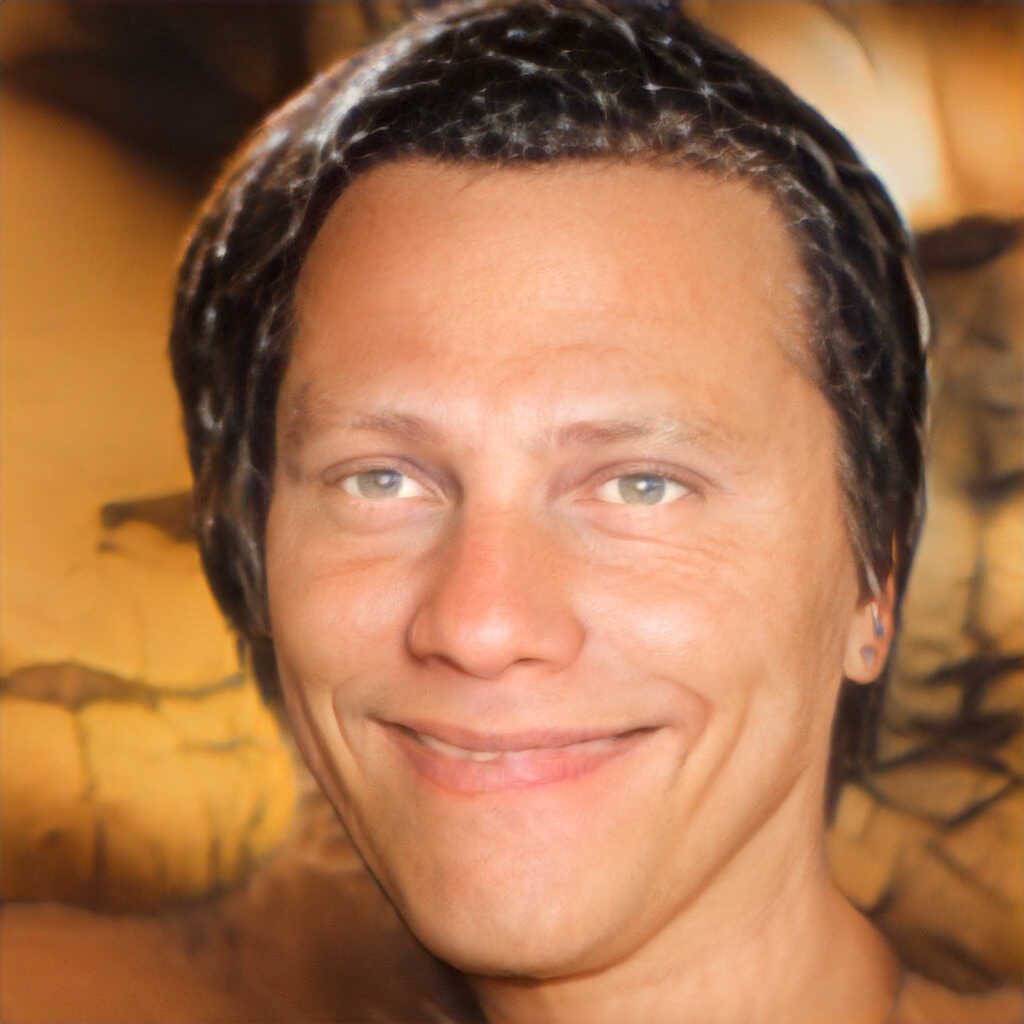}         &
        \includegraphics[width=0.135\linewidth]{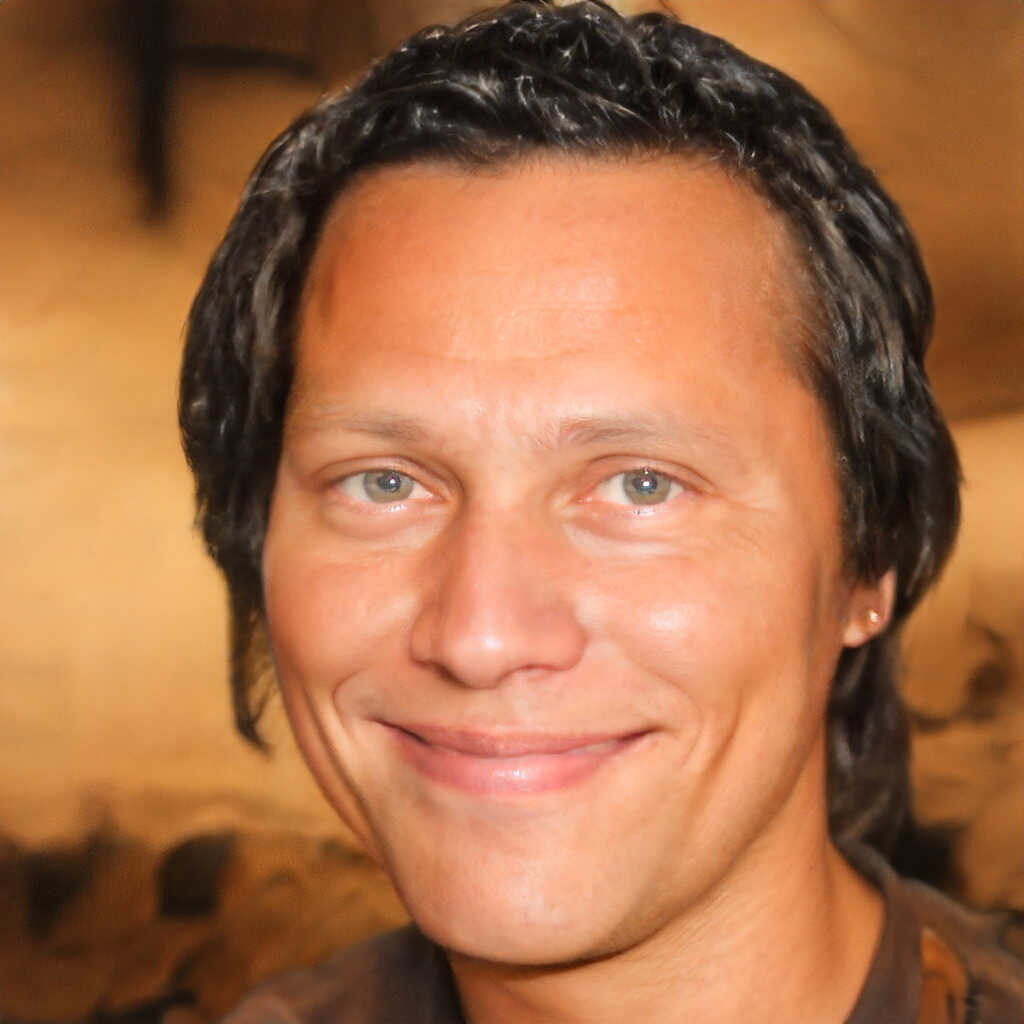}         &
        \includegraphics[width=0.135\linewidth]{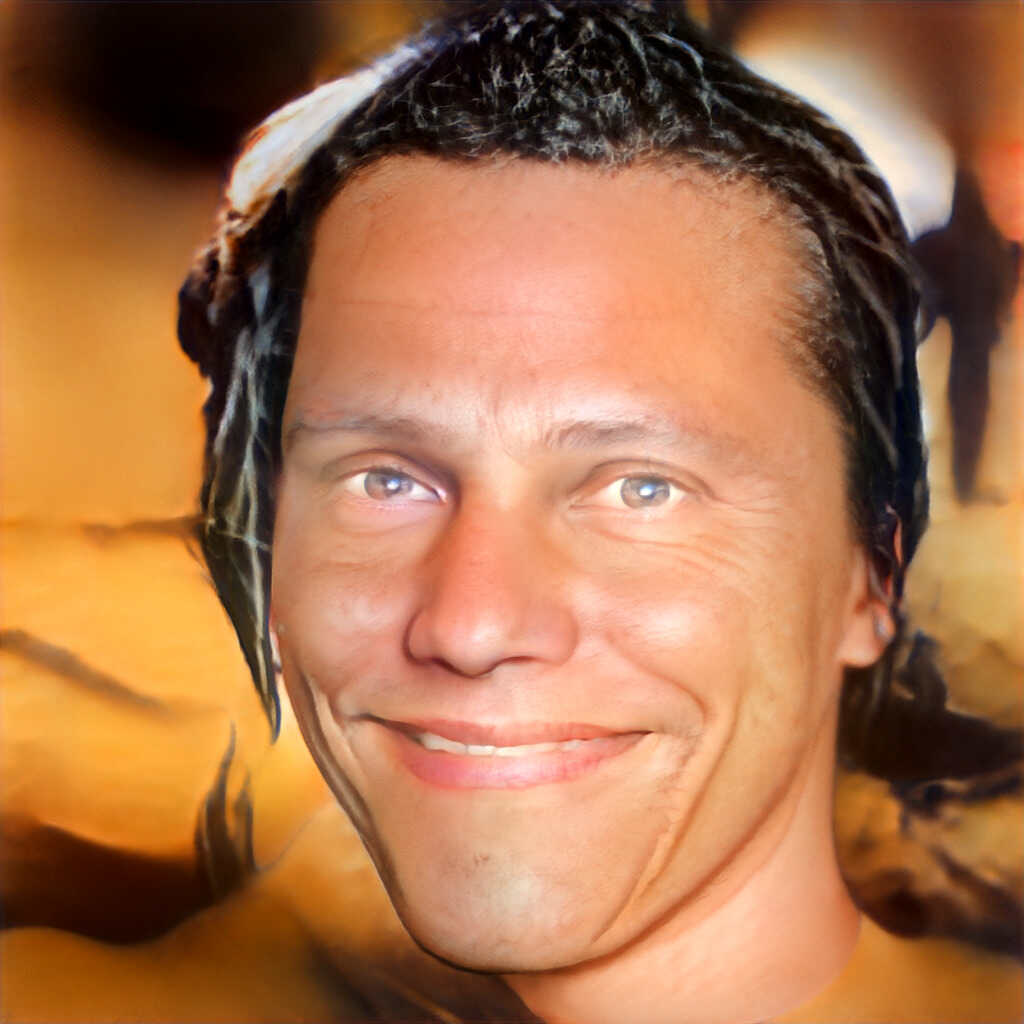} &
        \includegraphics[width=0.135\linewidth]{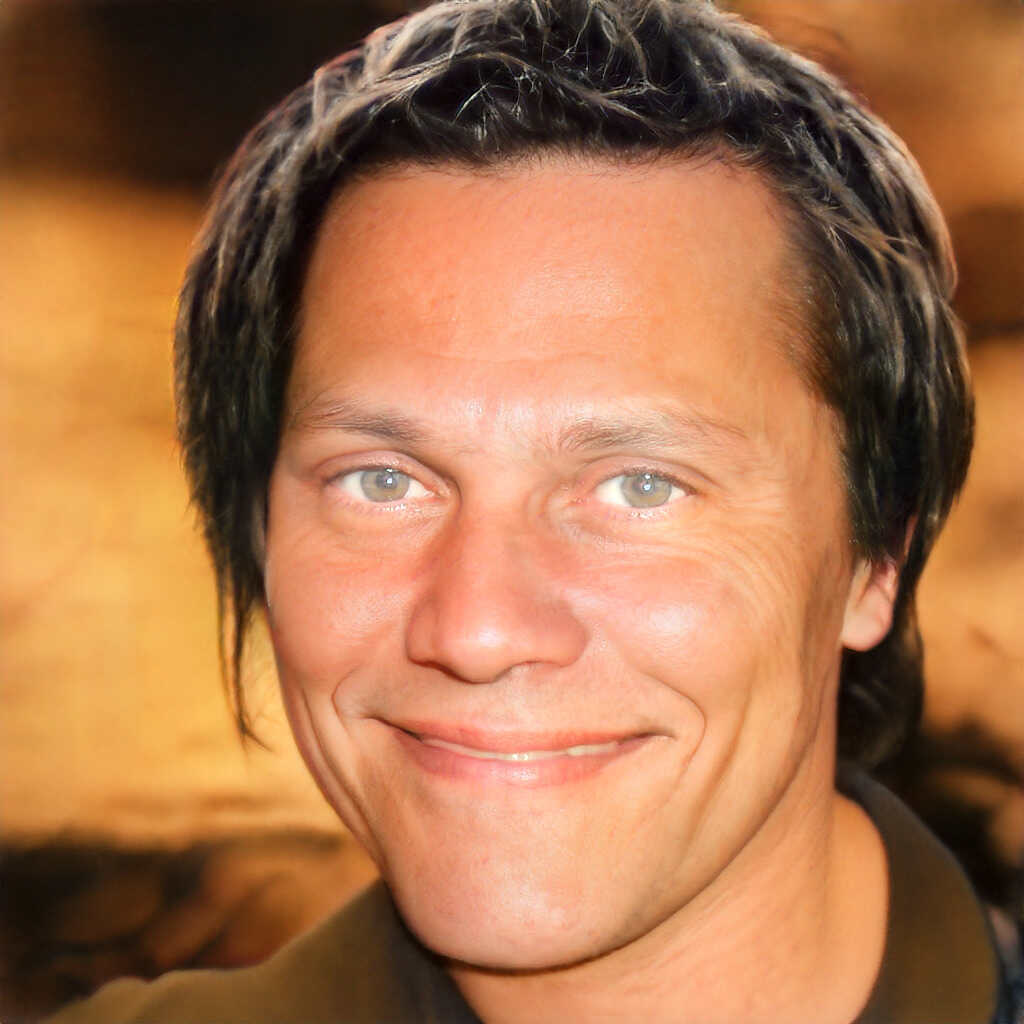} &
        \includegraphics[width=0.135\linewidth]{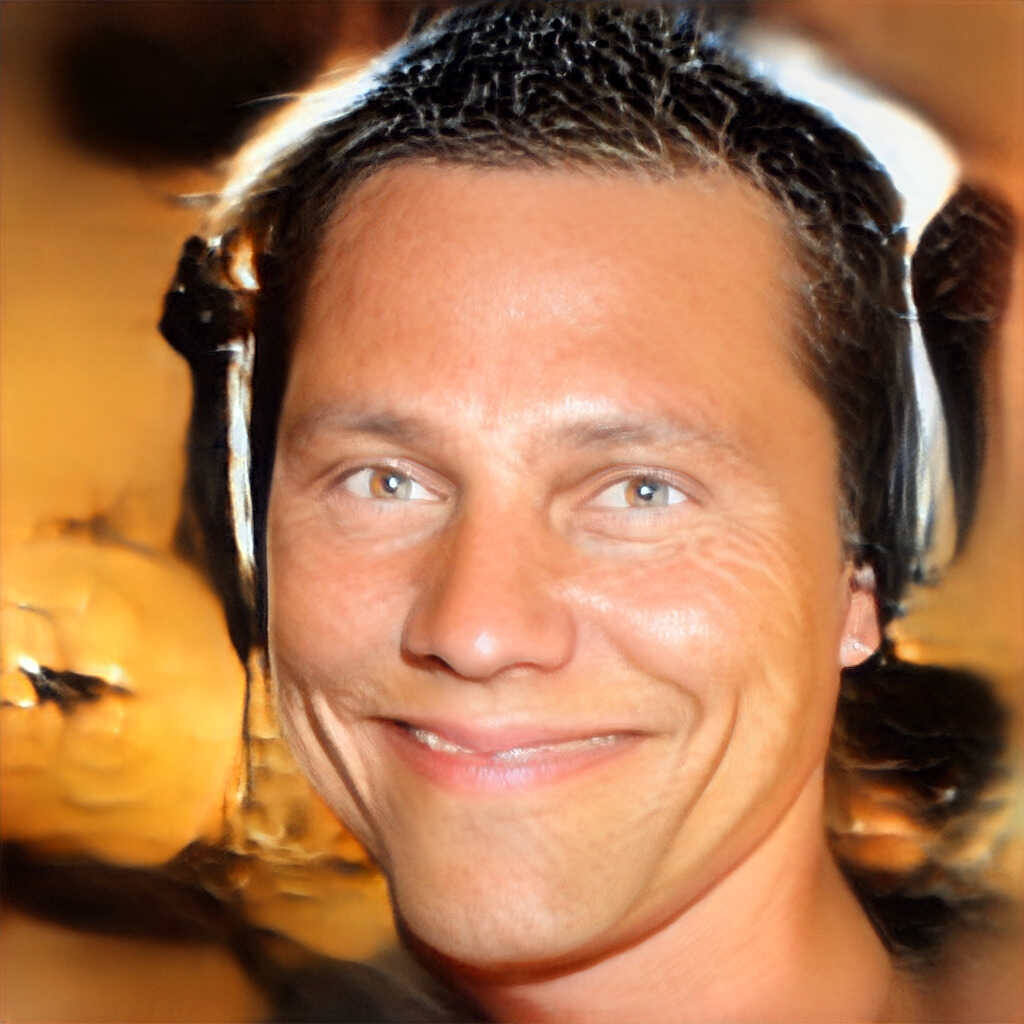}  &
        \includegraphics[width=0.135\linewidth]{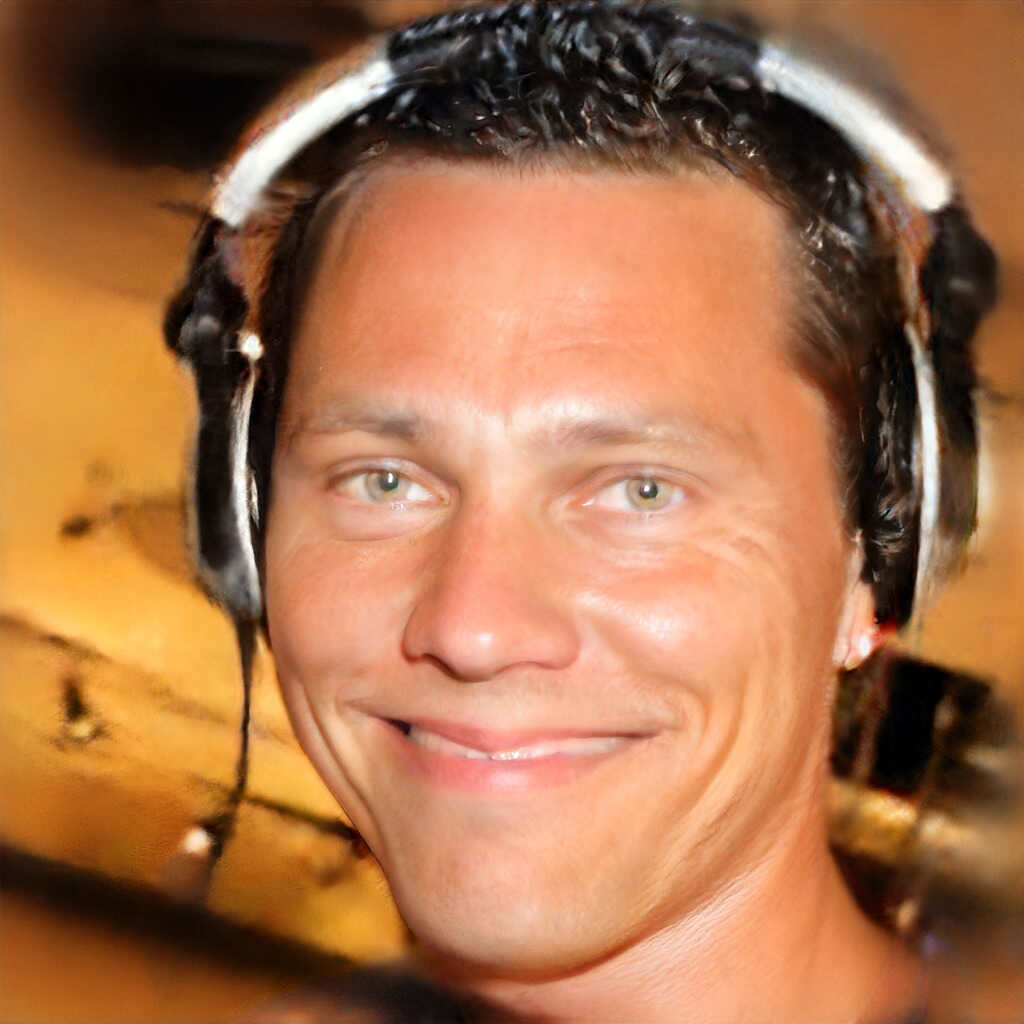}                                                                                                                                                   \\
        \includegraphics[width=0.135\linewidth]{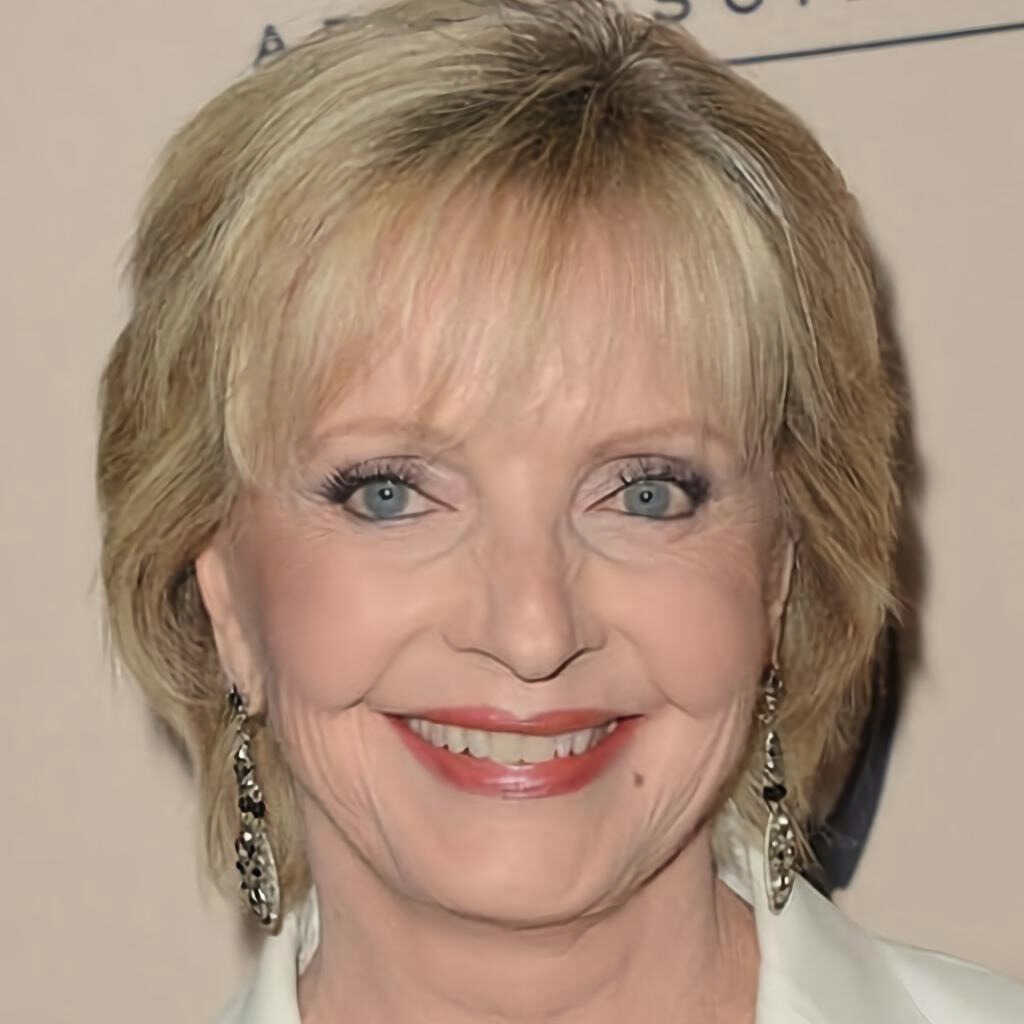}                    &
        \includegraphics[width=0.135\linewidth]{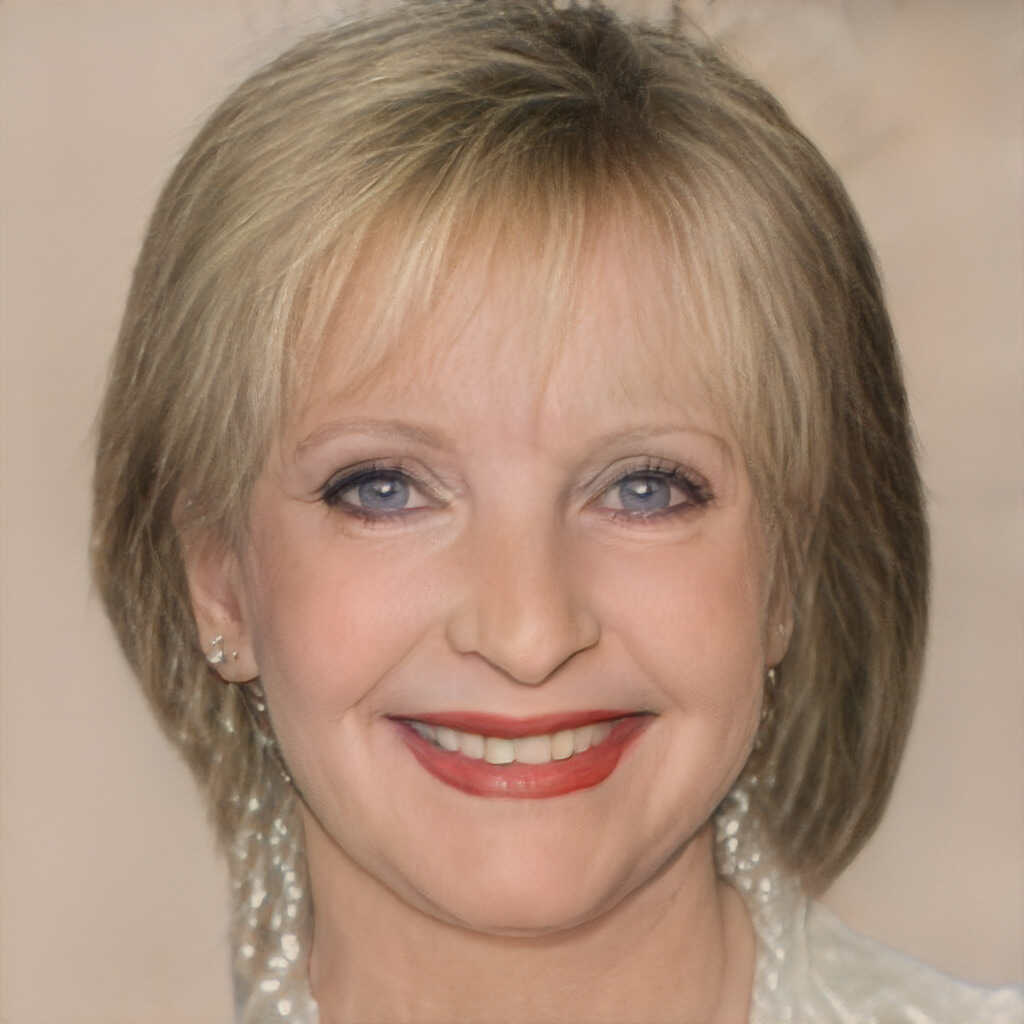}         &
        \includegraphics[width=0.135\linewidth]{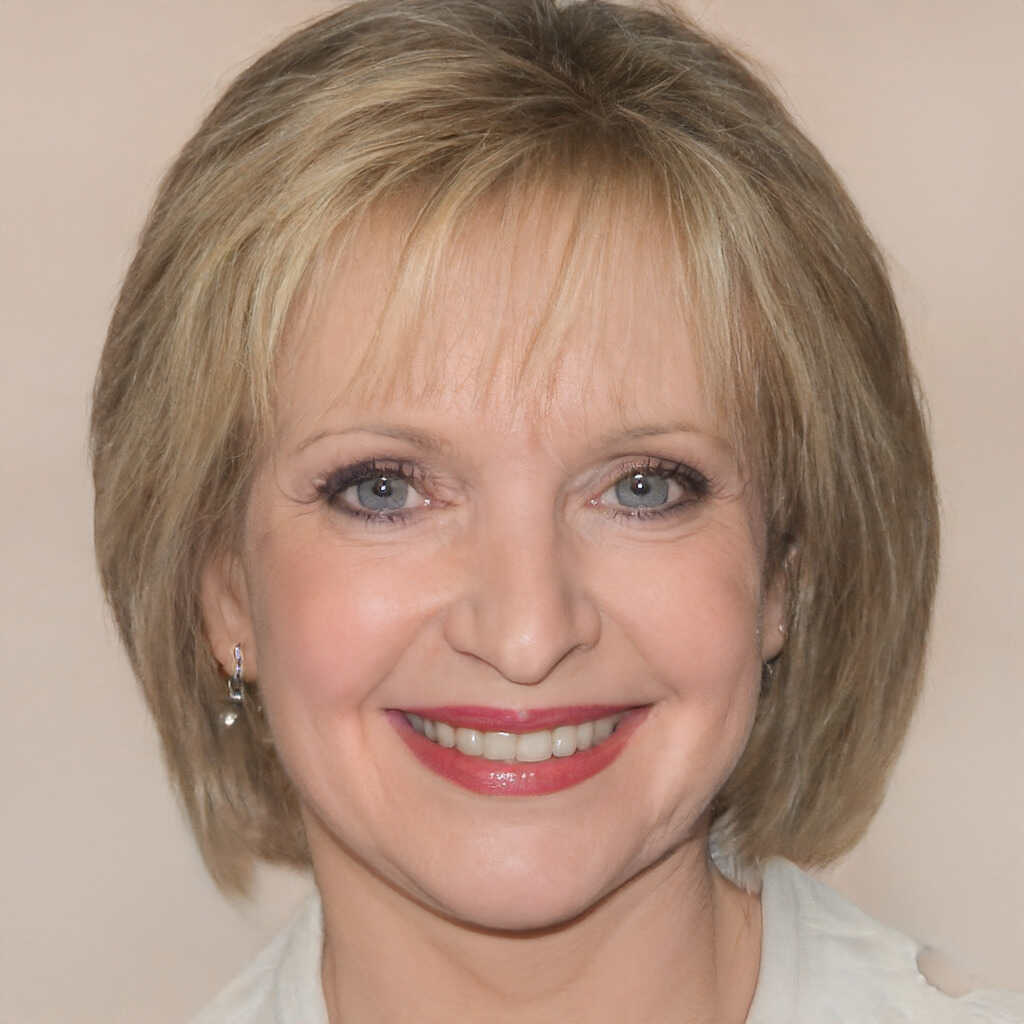}         &
        \includegraphics[width=0.135\linewidth]{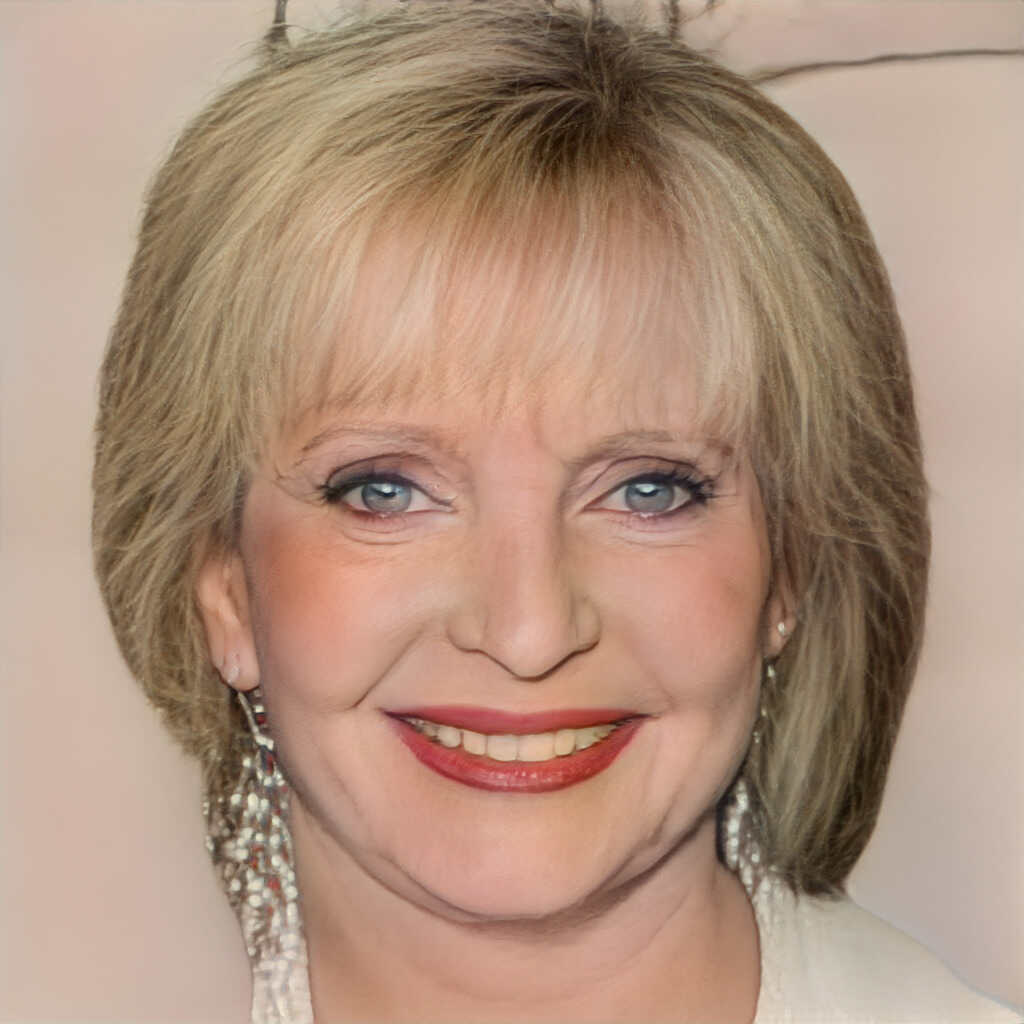} &
        \includegraphics[width=0.135\linewidth]{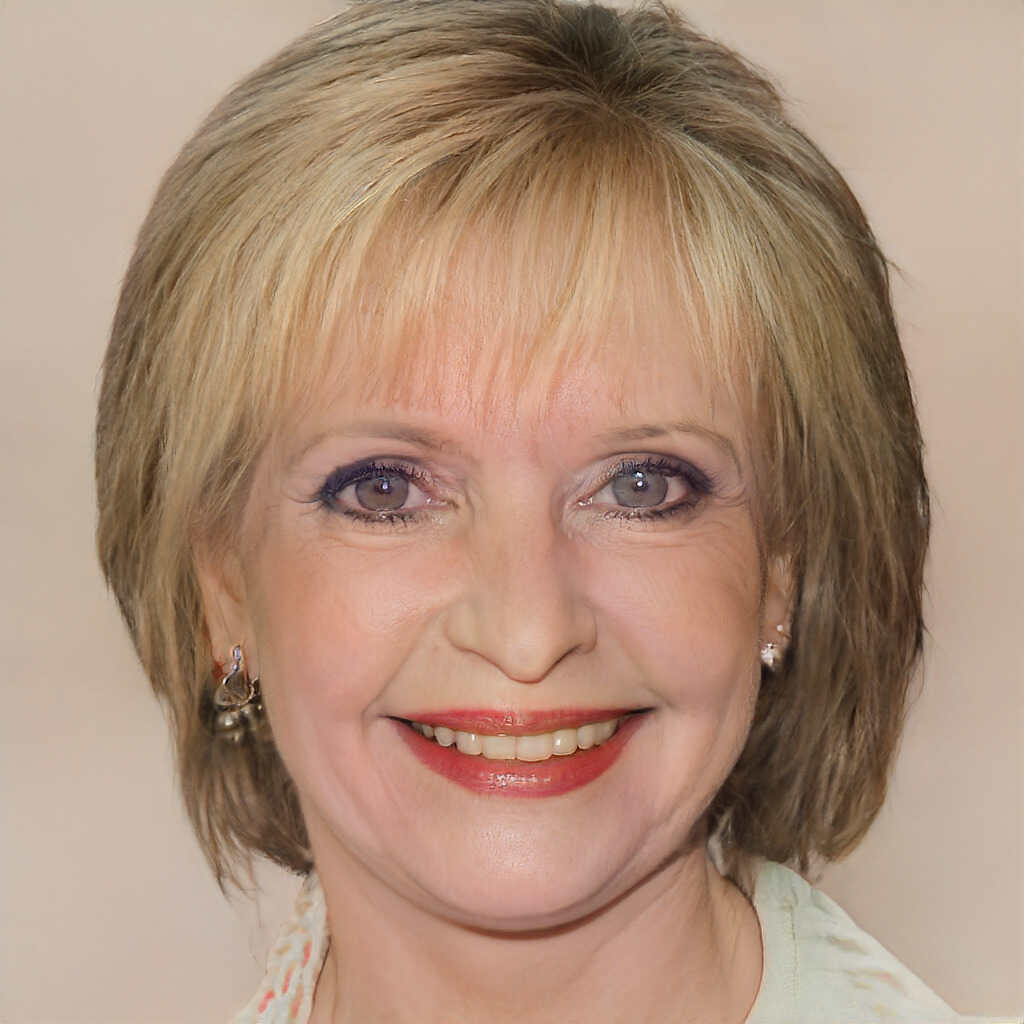} &
        \includegraphics[width=0.135\linewidth]{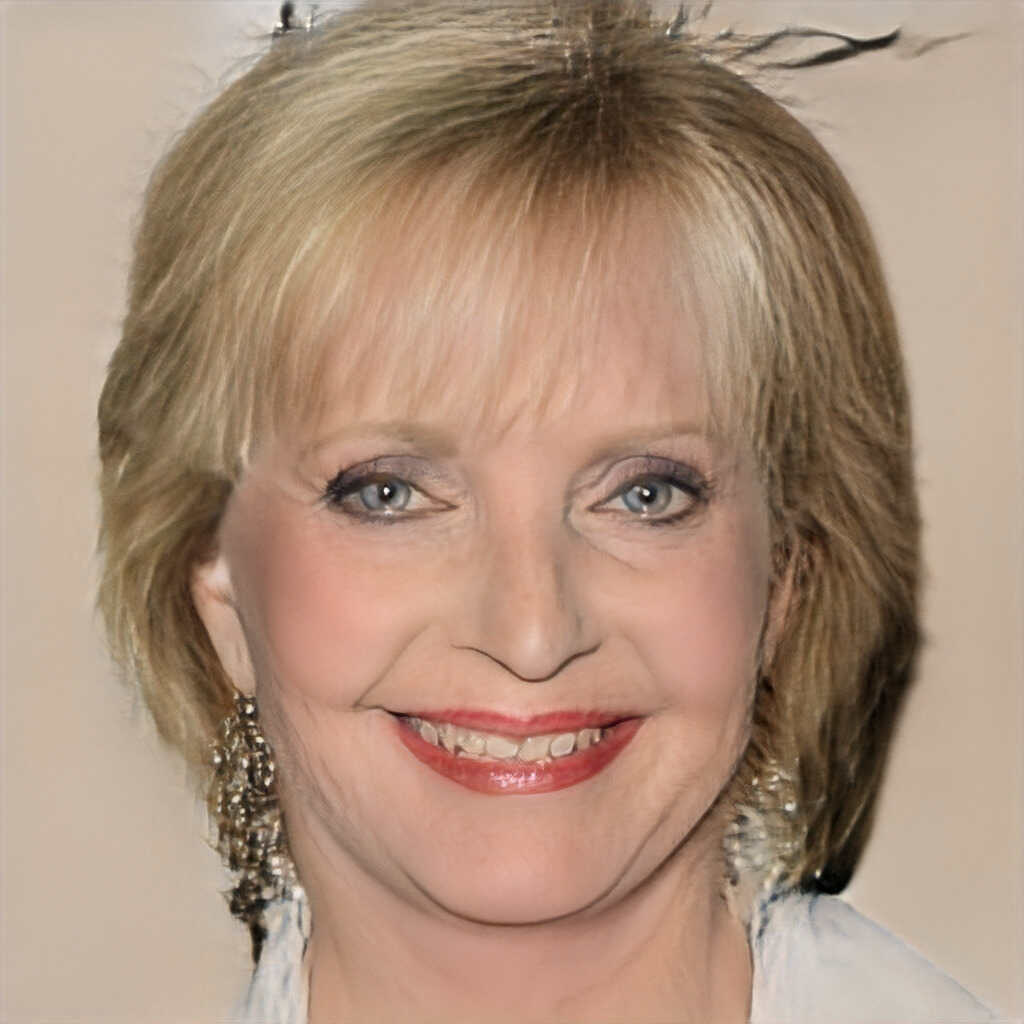}  &
        \includegraphics[width=0.135\linewidth]{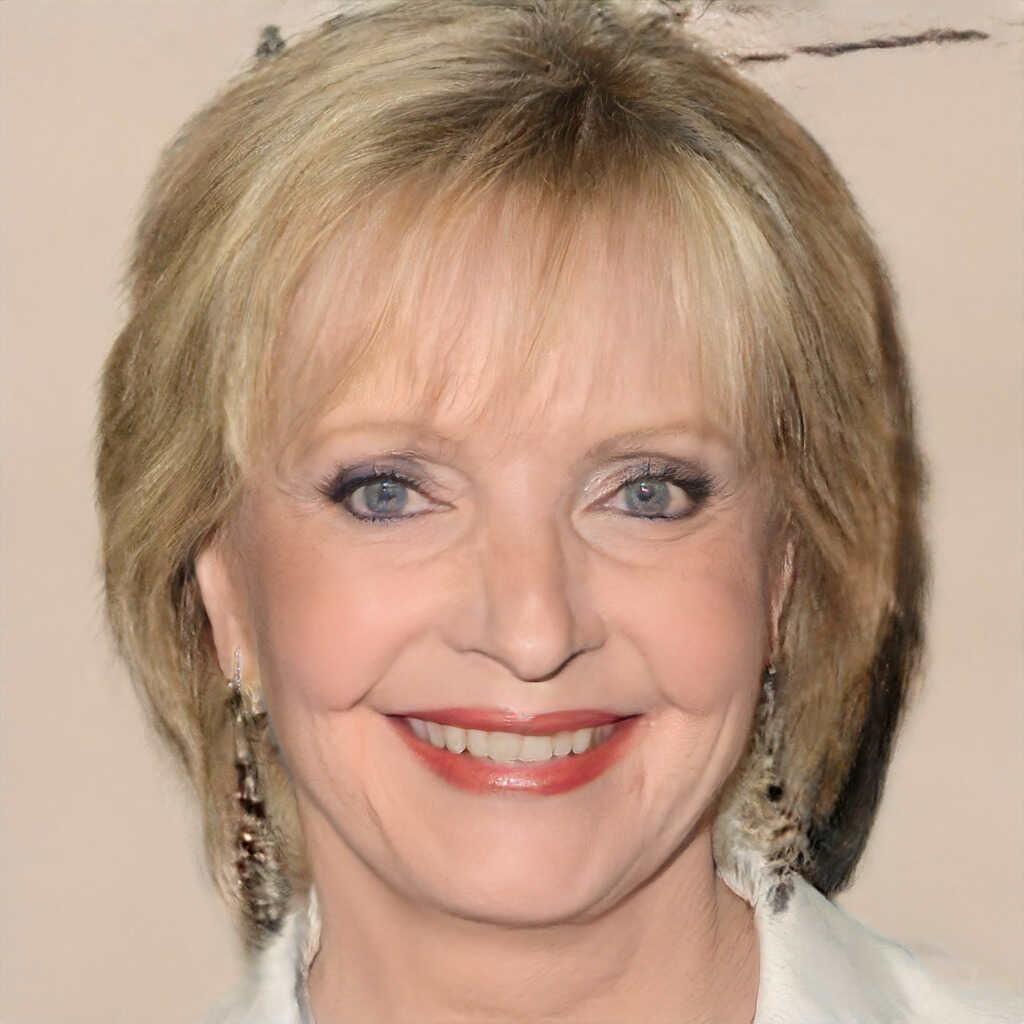}                                                                                                                                                   \\
        \includegraphics[width=0.135\linewidth]{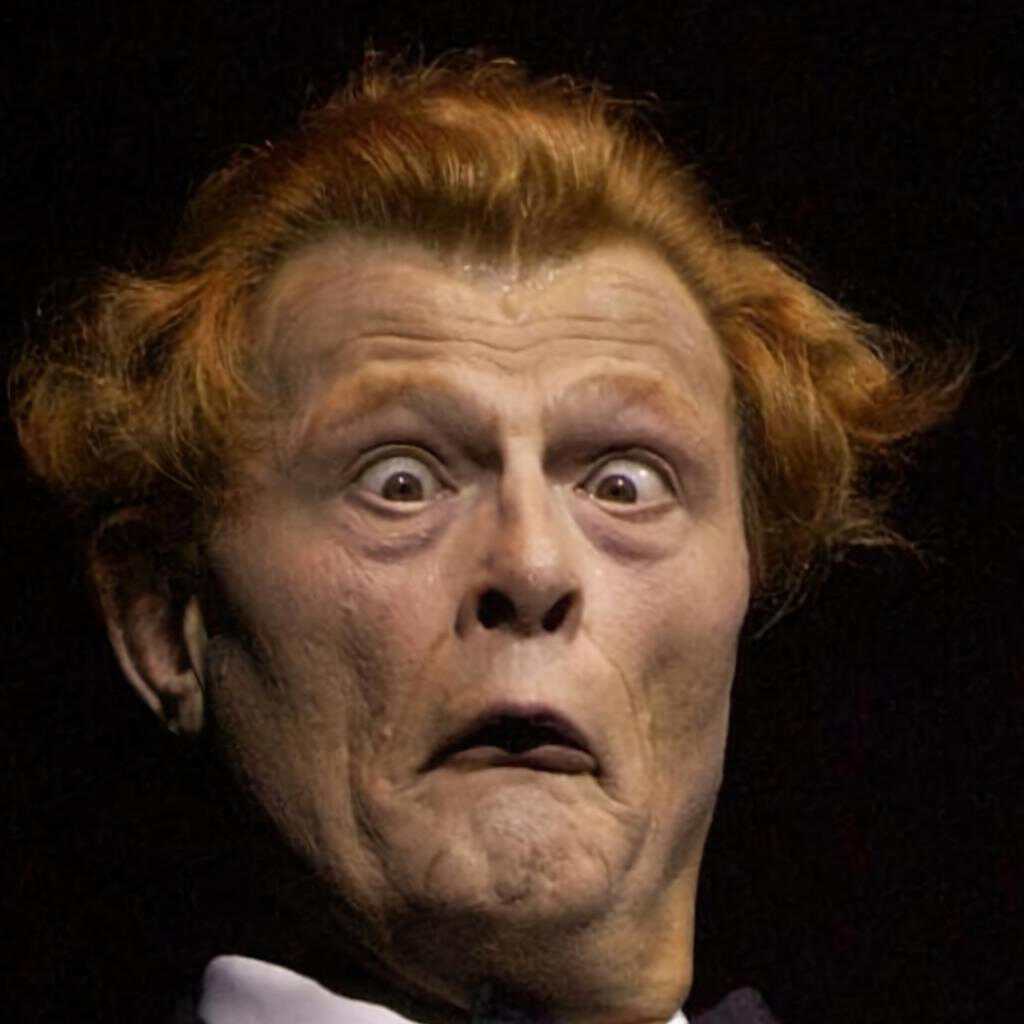}                    &
        \includegraphics[width=0.135\linewidth]{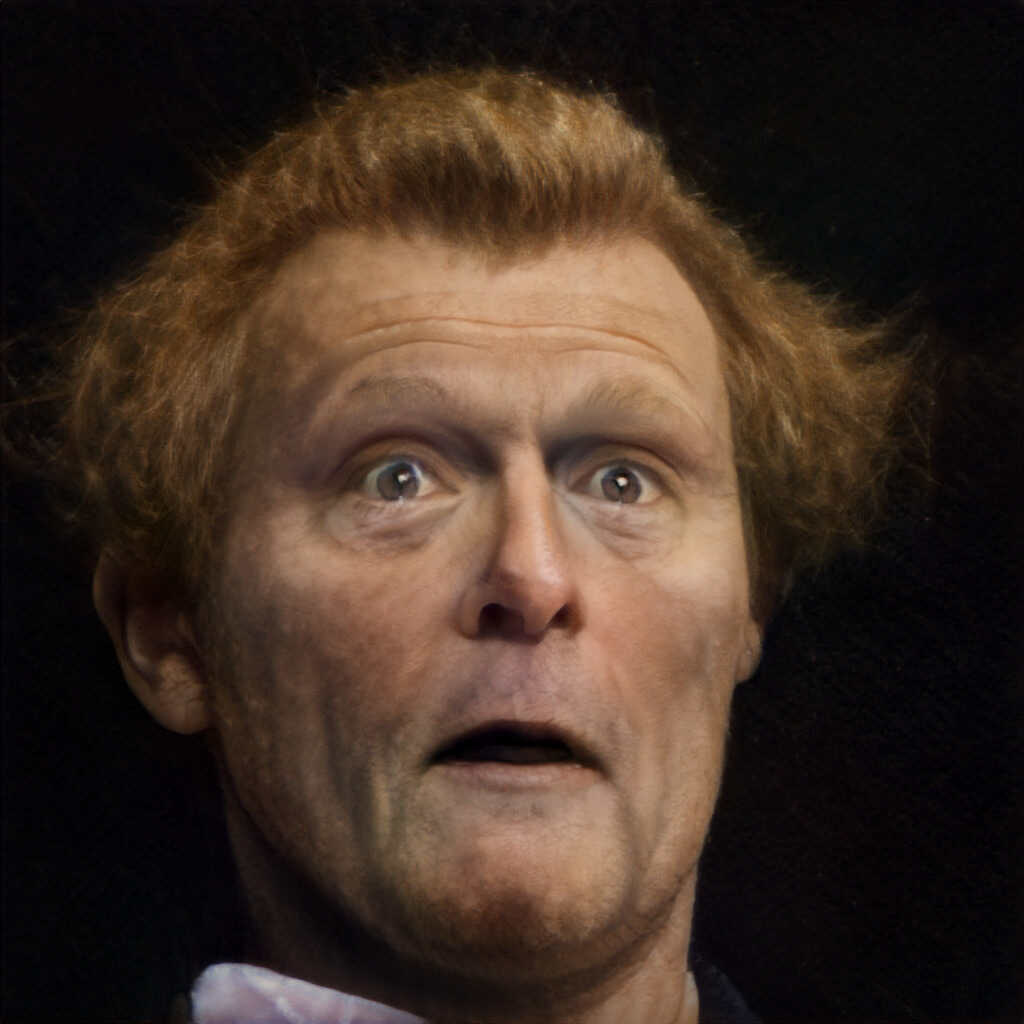}         &
        \includegraphics[width=0.135\linewidth]{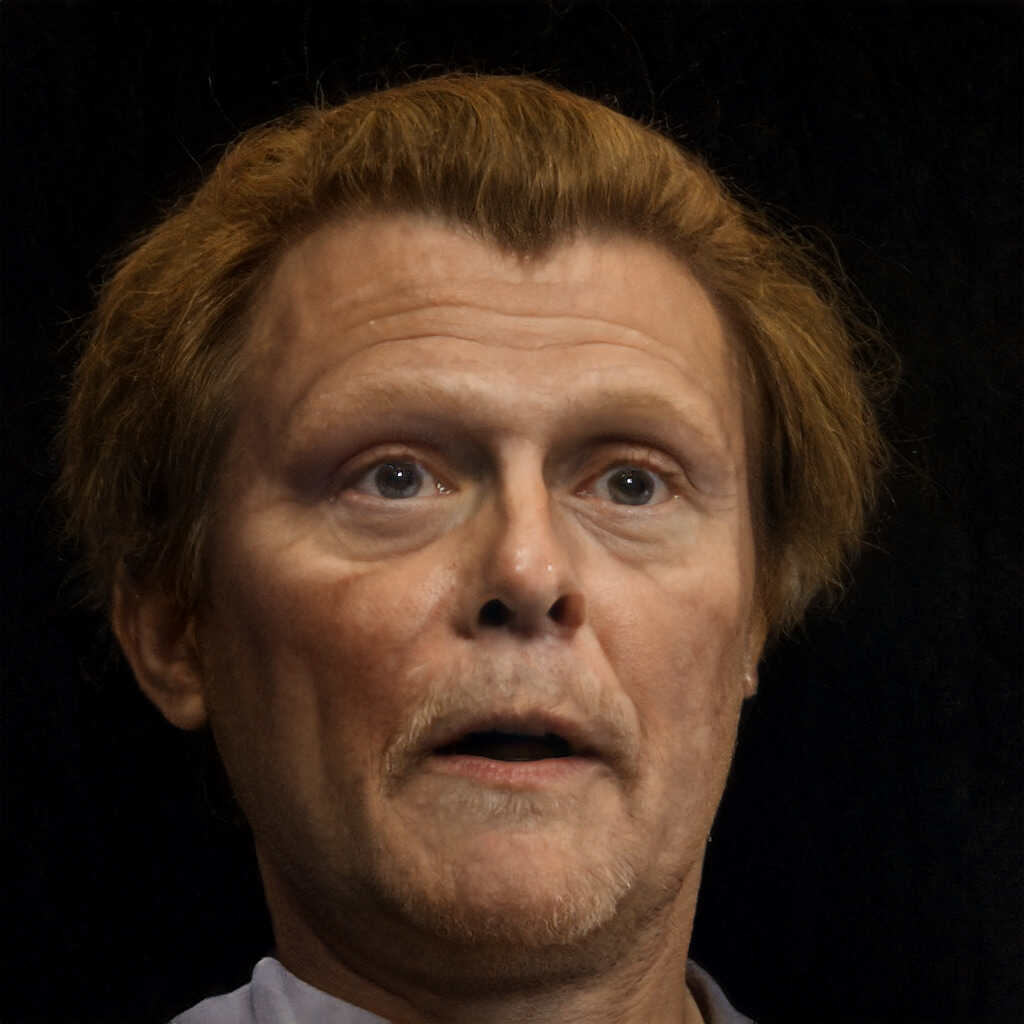}         &
        \includegraphics[width=0.135\linewidth]{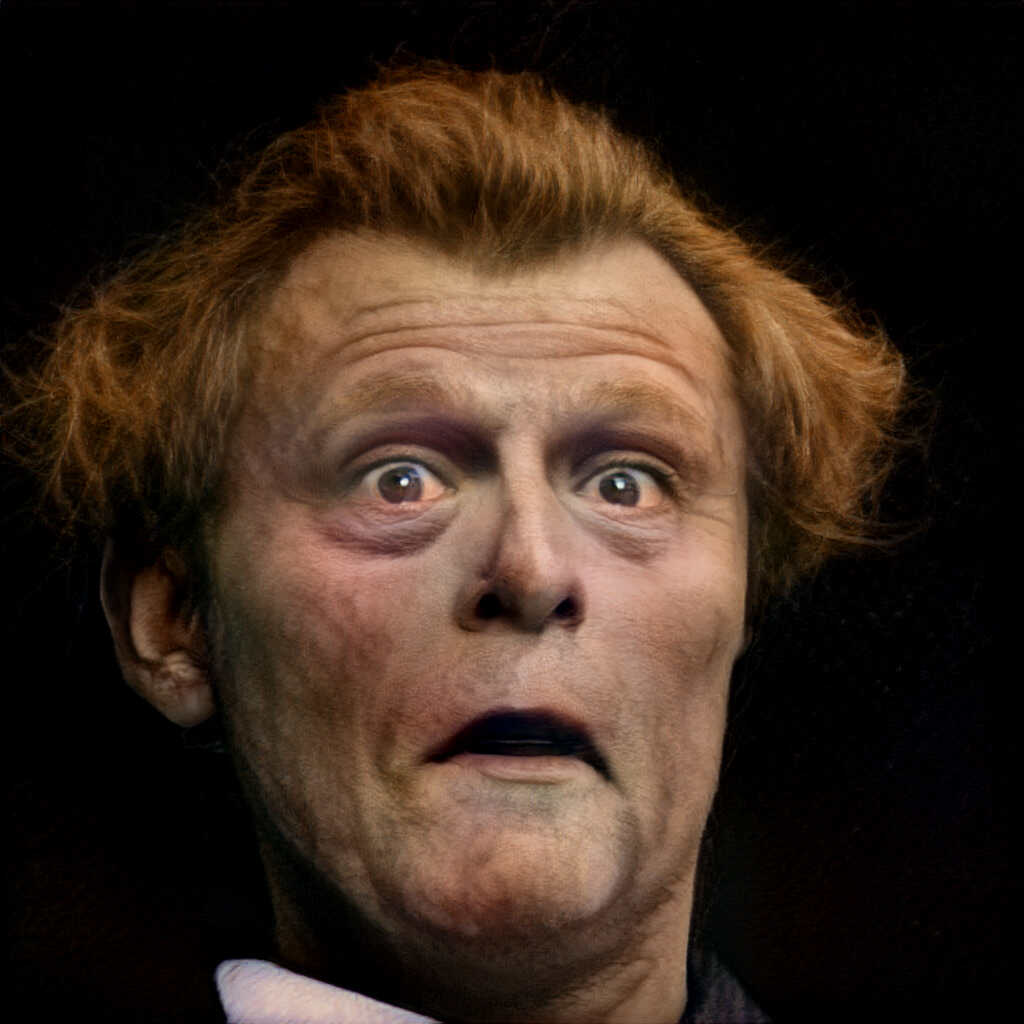} &
        \includegraphics[width=0.135\linewidth]{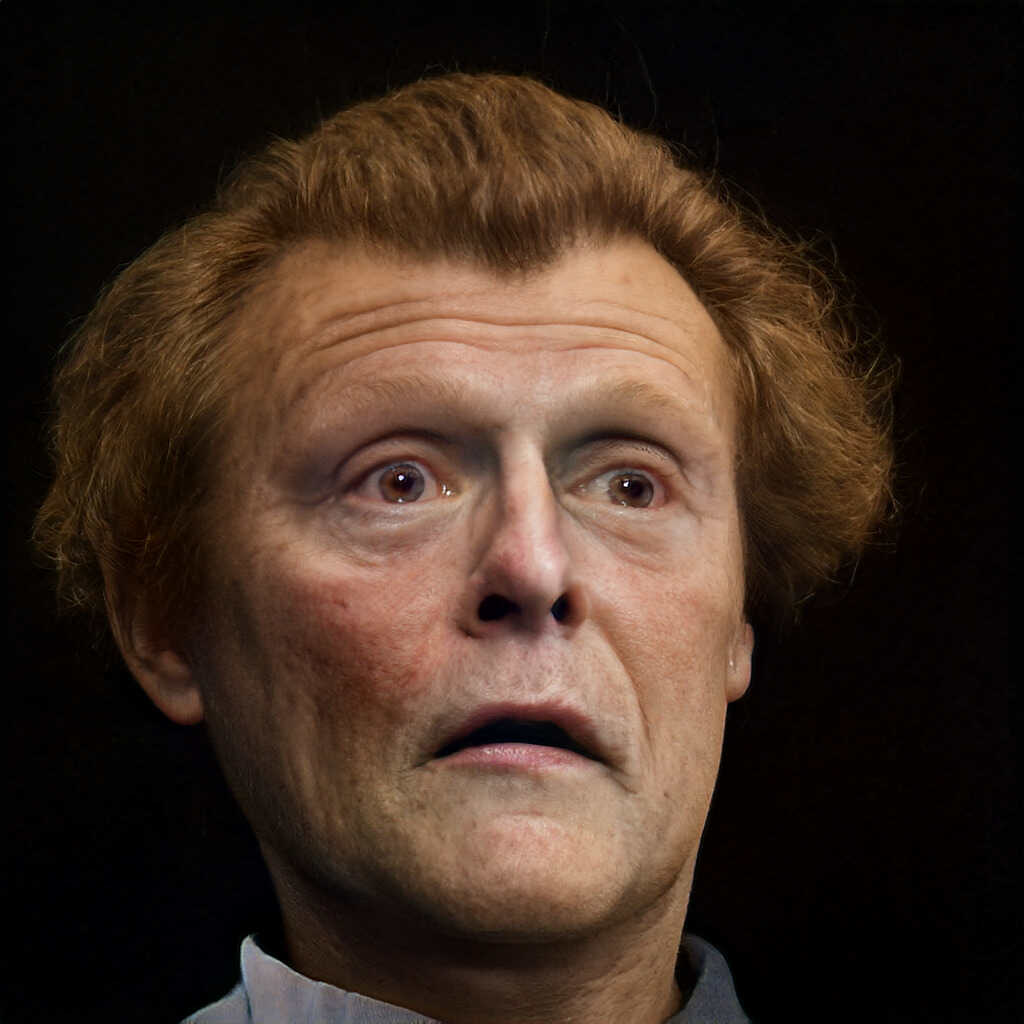} &
        \includegraphics[width=0.135\linewidth]{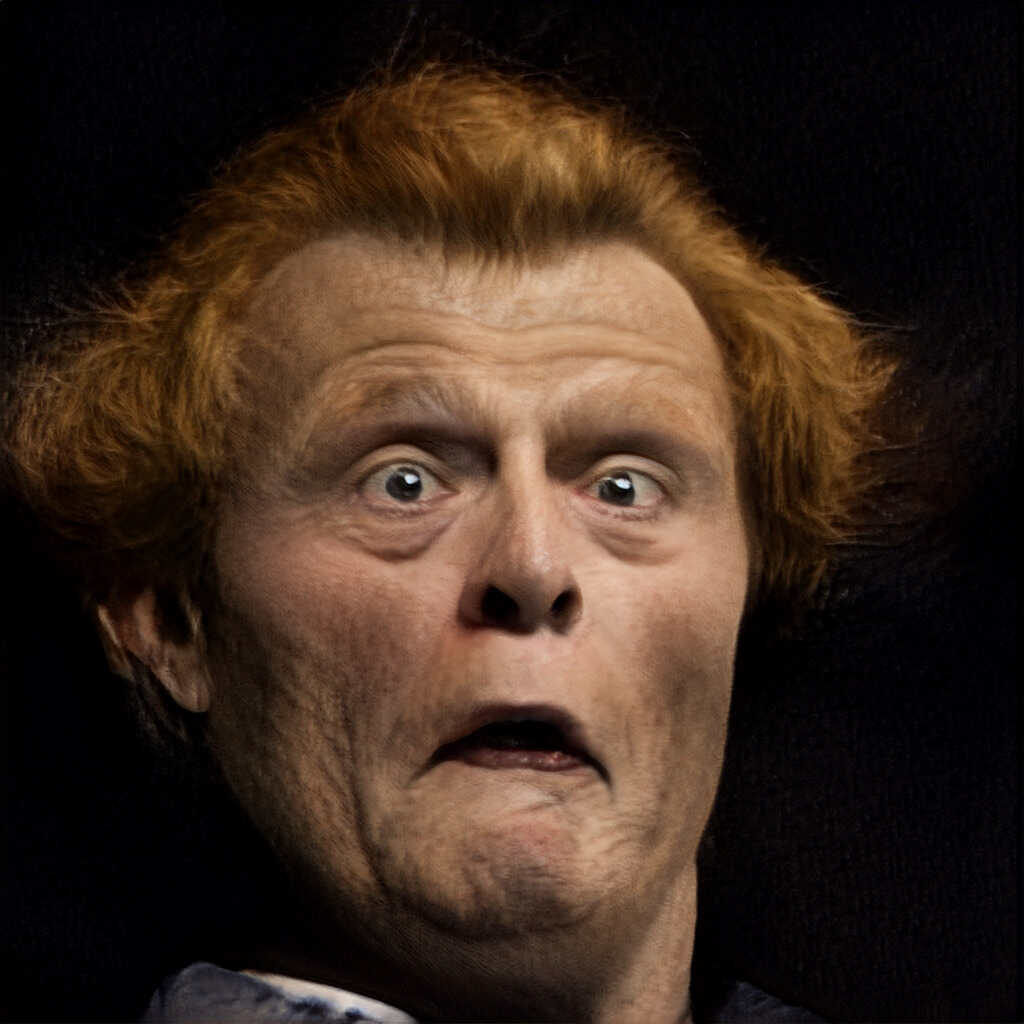}  &
        \includegraphics[width=0.135\linewidth]{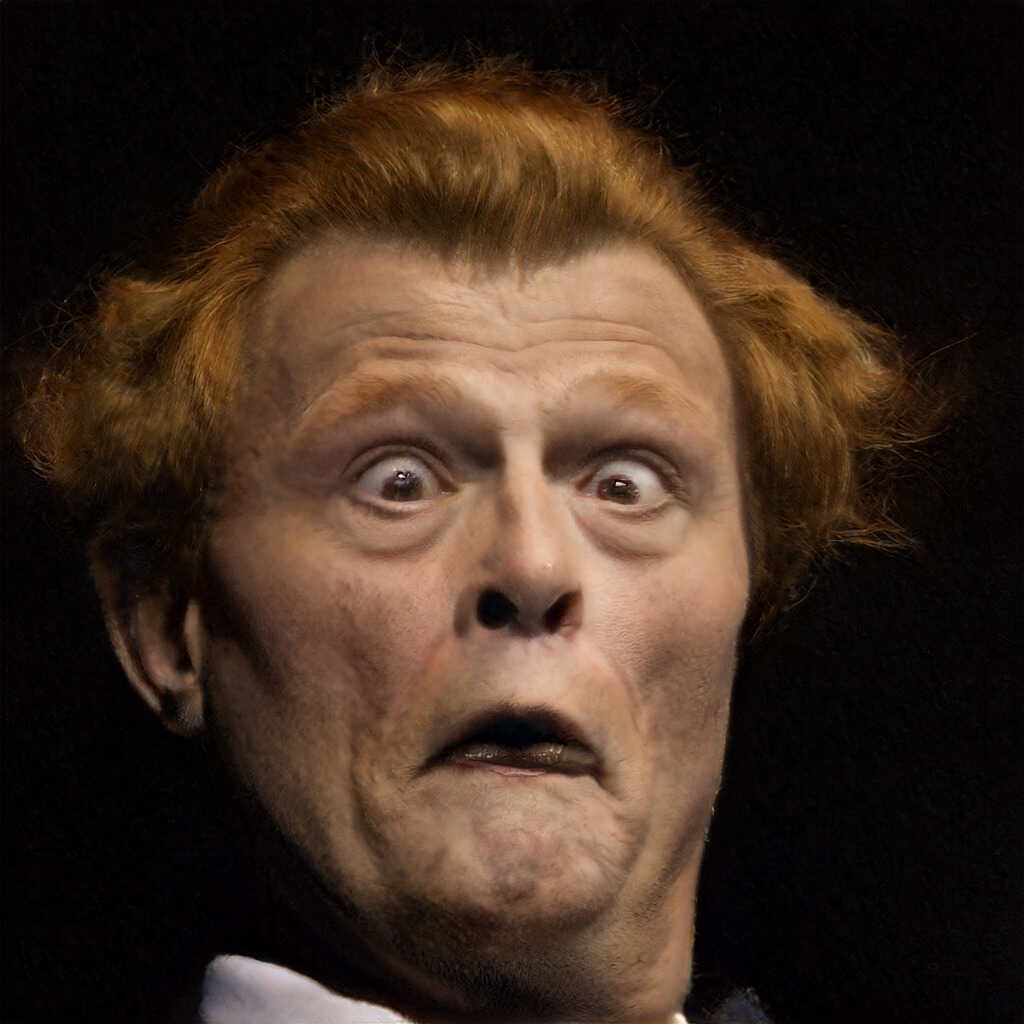}                                                                                                                                                   \\[5pt]
        GT & pSp & e4e & $\text{ReStyle}_{\text{pSp}}$ & $\text{ReStyle}_{\text{e4e}}$ & HyperStyle & Ours \\
    \end{tabular}
    \caption{%
        Visual comparison of image reconstruction quality on CelebA-HQ.}
    \label{fig:celeba_hq_inversion_baseline}
\end{figure}

\begin{figure}[t]
    \centering
    \footnotesize
    \renewcommand{\arraystretch}{0.0}
    \begin{tabular}{@{}c@{}c@{}c@{}c@{}c@{}c@{}c@{}}
        \includegraphics[width=0.142\linewidth]{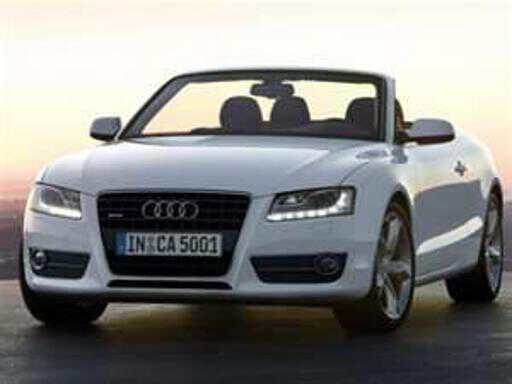}           &
        \includegraphics[width=0.142\linewidth]{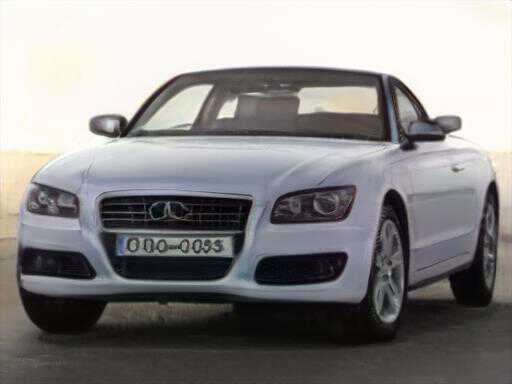}         &
        \includegraphics[width=0.142\linewidth]{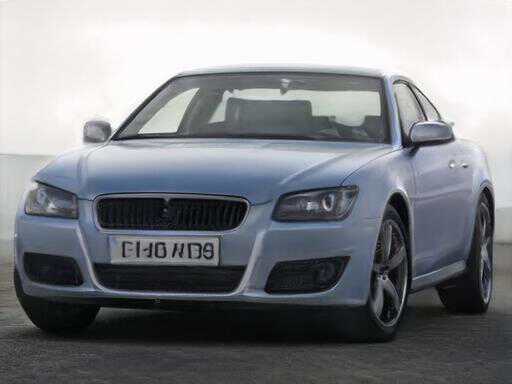}         &
        \includegraphics[width=0.142\linewidth]{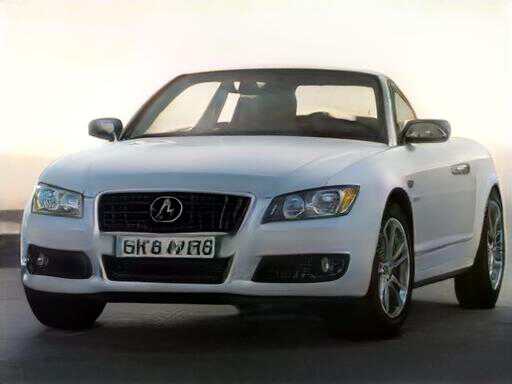} &
        \includegraphics[width=0.142\linewidth]{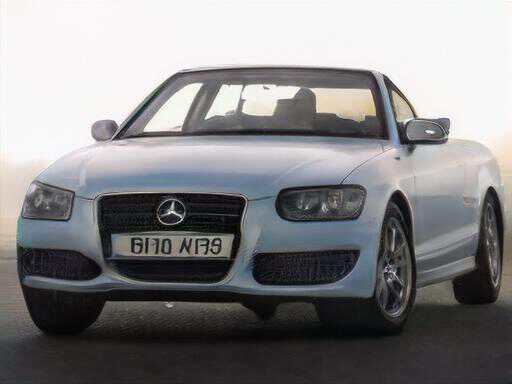} &
        \includegraphics[width=0.142\linewidth]{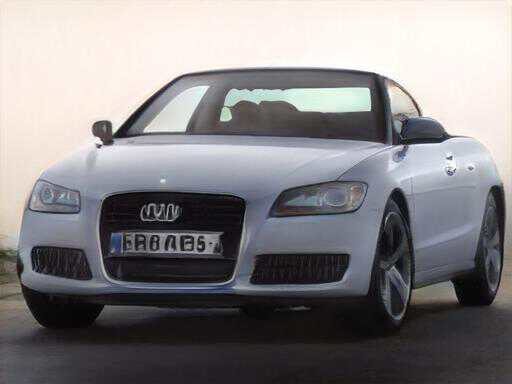}  &
        \includegraphics[width=0.142\linewidth]{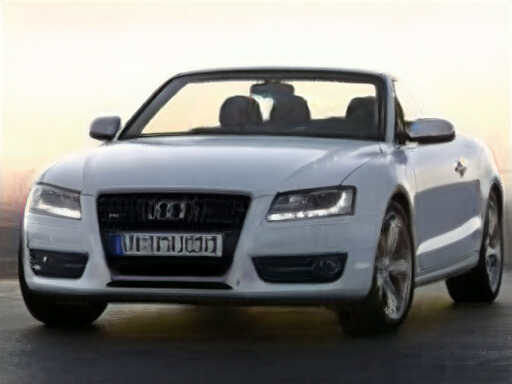}                                                                                                                                        \\
        \includegraphics[width=0.142\linewidth]{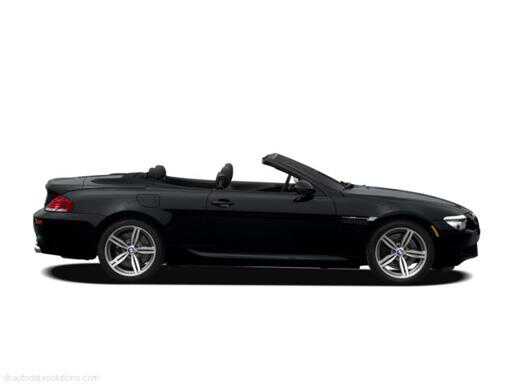}          &
        \includegraphics[width=0.142\linewidth]{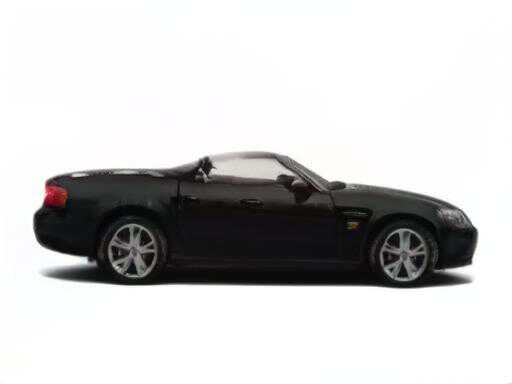}         &
        \includegraphics[width=0.142\linewidth]{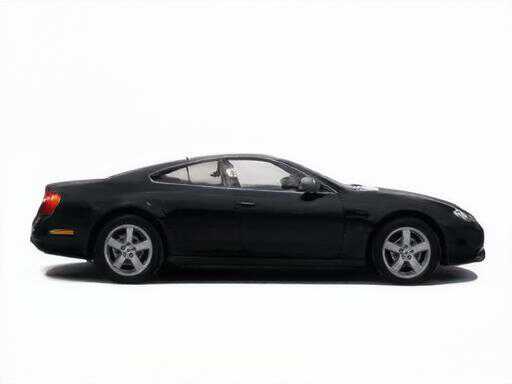}         &
        \includegraphics[width=0.142\linewidth]{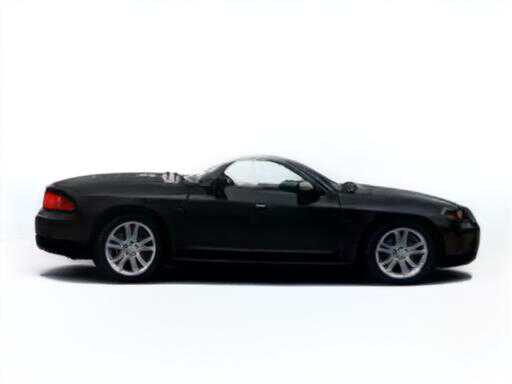} &
        \includegraphics[width=0.142\linewidth]{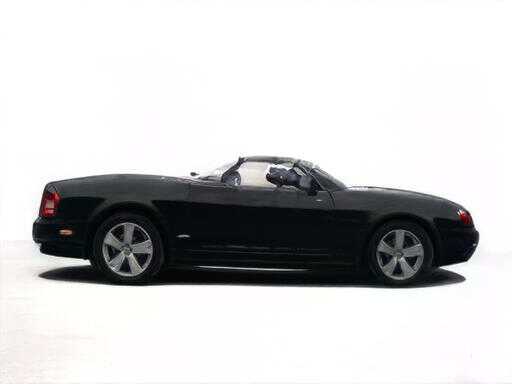} &
        \includegraphics[width=0.142\linewidth]{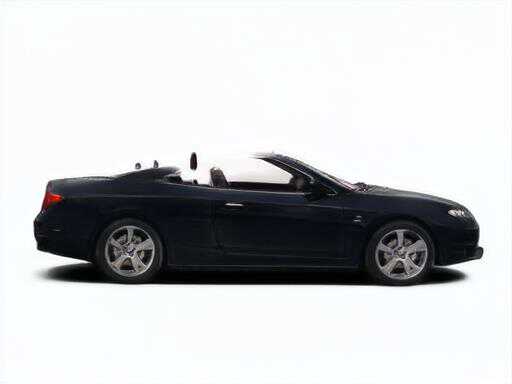}  &
        \includegraphics[width=0.142\linewidth]{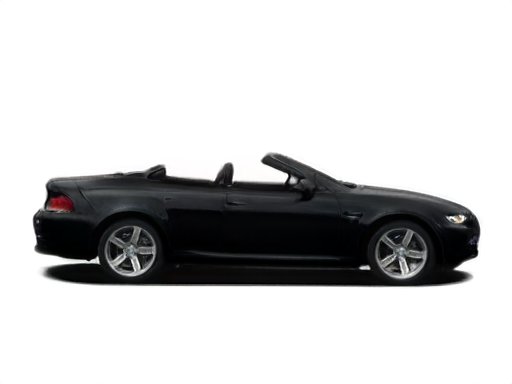}                                                                                                                                       \\
        \includegraphics[width=0.142\linewidth]{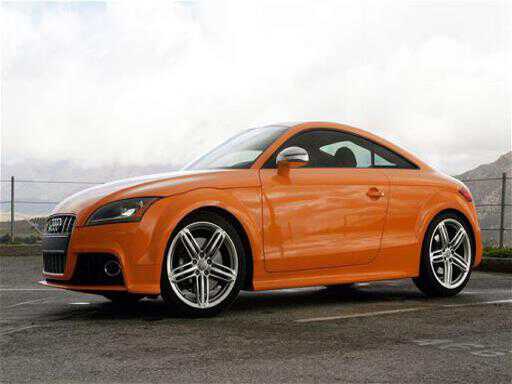}          &
        \includegraphics[width=0.142\linewidth]{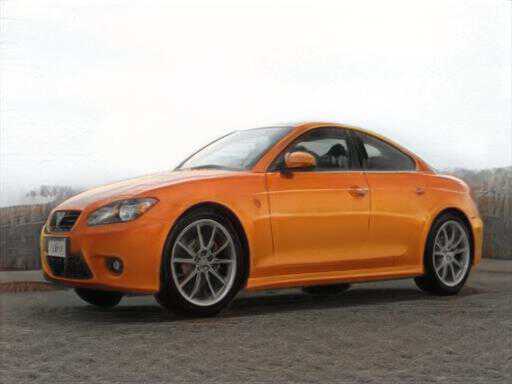}         &
        \includegraphics[width=0.142\linewidth]{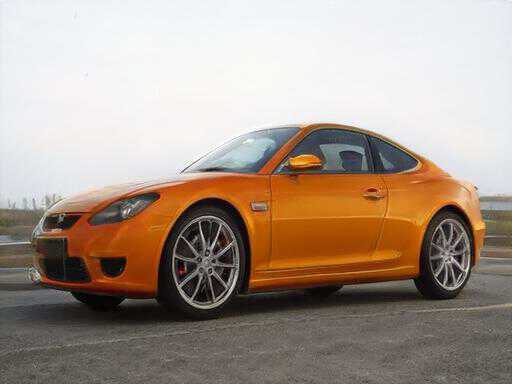}         &
        \includegraphics[width=0.142\linewidth]{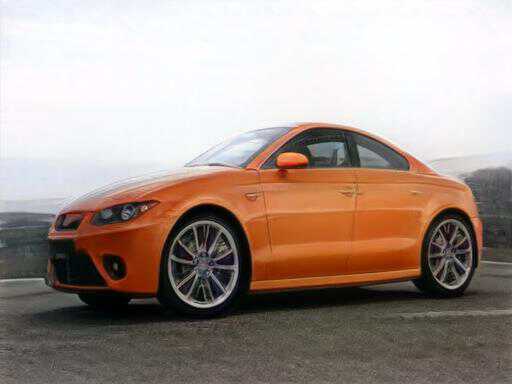} &
        \includegraphics[width=0.142\linewidth]{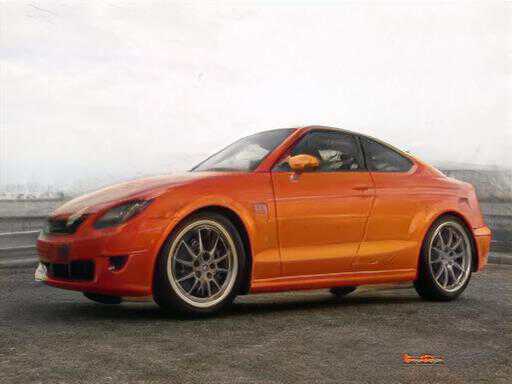} &
        \includegraphics[width=0.142\linewidth]{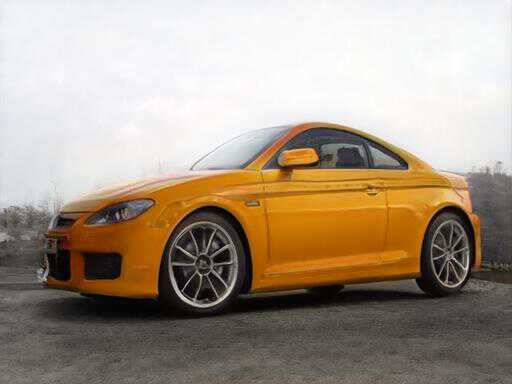}  &
        \includegraphics[width=0.142\linewidth]{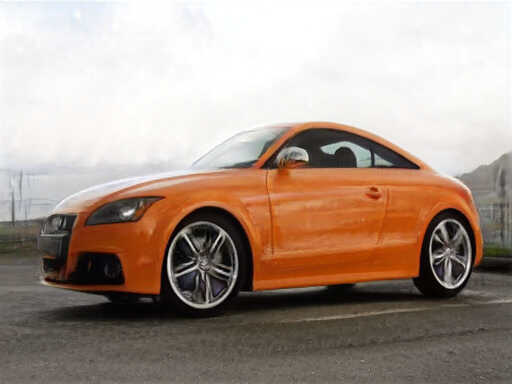}                                                                                                                                        \\[2pt]
        GT & pSp & e4e & $\text{ReStyle}_{\text{pSp}}$ & $\text{ReStyle}_{\text{e4e}}$ & HyperStyle & ours \\
    \end{tabular}
    \caption{%
        Visual comparison of image reconstruction quality on the Stanford Cars.}
    \label{fig:standford_cars_inversion_baseline}
\end{figure}

\noindent
{\bf Inversion Quality \quad}
We begin with a qualitative evaluation of reconstructed images. Fig.~\ref{fig:celeba_hq_inversion_baseline} demonstrates the reconstruction of facial images taken from the CelebA-HQ~\cite{karras2017progressive} dataset, and Fig.~\ref{fig:standford_cars_inversion_baseline} demonstrates the reconstruction of car images taken from the Stanford Cars~\cite{KrauseStarkDengFei-Fei_3DRR2013} dataset. We compare the reconstructed images produced by our method with the following previous approaches: pSp~\cite{richardson2020encoding}, e4e~\cite{tov2021designing}, $\text{ReStyle}_{\text{pSp}}$~\cite{alaluf2021restyle}, $\text{ReStyle}_{\text{e4e}}$~\cite{alaluf2021restyle}, HyperStyle~\cite{alaluf2021hyperstyle}.

As can be seen in Fig.~\ref{fig:celeba_hq_inversion_baseline}, our method is able to generate almost identical reconstructions. The first and third rows demonstrate the reconstruction of difficult examples. Our method is able to produce near-identical reconstruction, whereas all other methods struggle to achieve inversion of good quality. The second row demonstrates the reconstruction of a relatively easy example.
Although all methods are able to produce meaningful reconstruction, only our method is truly able to preserve identity and properly reconstruct fine details (such as gaze, eye color, dimples, etc.).

As can be seen in Fig.~\ref{fig:standford_cars_inversion_baseline}, there is a large gap between our method and the other ones. Specifically, all baseline methods struggle to reconstruct various elements (fine or coarse). In all three rows, only our method is able to reconstruct the coarse shape of the car properly. In the first row, we can see that only our method is able to reconstruct the following elements properly: (\RNum{1}) the coarse shape of the car, (\RNum{2}) the shape of the headlight, (\RNum{3}) the fact that the lights are on, (\RNum{4}) the absence of a roof, (\RNum{5}) the car's logo. In the second row, only our method is able to reconstruct the following elements properly: (\RNum{1}) shape of the headlights, (\RNum{2}) side-mirrors, (\RNum{3}) color tone, (\RNum{4}) reflection. In the third row, only our method is able to reconstruct the following elements properly: (\RNum{1}) wheels, (\RNum{2}) placement and size of the passenger window, (\RNum{3}) side-mirror shape and color.

\begin{figure}[t]
    \centering
    \footnotesize
    \renewcommand{\arraystretch}{0.0}
    \begin{tabular}{@{}c@{}c@{}c@{}c@{}c@{}c@{}c@{}}
        \includegraphics[width=0.142\linewidth]{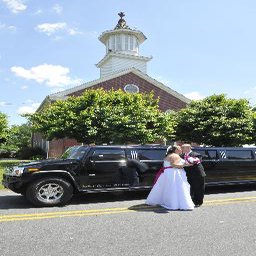}               &
        \includegraphics[width=0.142\linewidth]{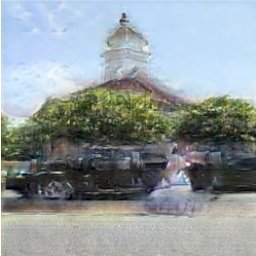}           &
        \includegraphics[width=0.142\linewidth]{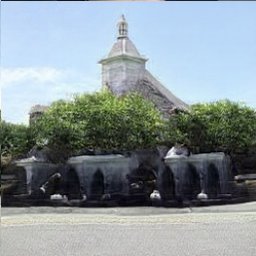}           &
        \includegraphics[width=0.142\linewidth]{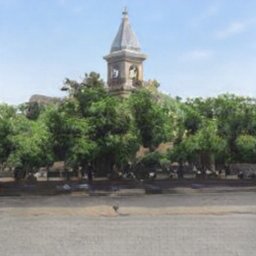}           &
        \includegraphics[width=0.142\linewidth]{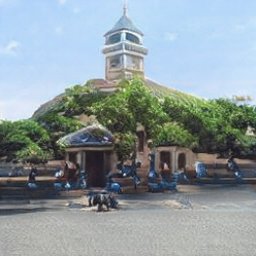}       &
        \includegraphics[width=0.142\linewidth]{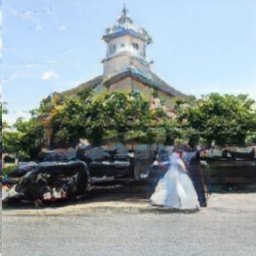} &
        \includegraphics[width=0.142\linewidth]{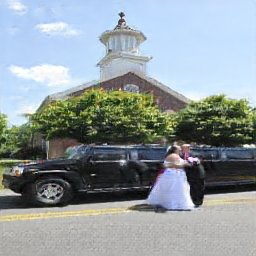}                                                                                                   \\ [5pt]
        GT & PTI & StyleGAN2 & e4e & ReStyle & HI & Ours \\
    \end{tabular}
    \caption{%
        Visual comparison of image reconstruction quality on the LSUN Church. HI stands for HyperInverter~\cite{Yao2022FeatureStyleEF}}
    \label{fig:lsun_church_inversion_baseline}
\end{figure}

To test the effect of the number of residual blocks, $L$, of the gradient modification modules, 
the image reconstruction performance of our approach was evaluated for 200 randomly sampled images from the CelebA-HQ test set. As Table~\ref{table:ablation_residual_blocks} shows, adding more blocks slightly improves the image quality scores of the MS-SSIM, LPIPS, and $L_2$ metrics, whereas it hardly degrades the identity preservation score.

\begin{table}[t]
    \centering
    \caption{The effect of the number of residual blocks, $L$, on image reconstruction. These results are obtained on a fixed evaluation set consisting of 200 randomly sampled images from the CelebA-HQ test set.}
    \label{table:ablation_residual_blocks}
    \begin{tabular}{cccccc}
        \toprule
        $L$ & \# Parameters & $\uparrow$ID       & $\uparrow$MS-SSIM & $\downarrow$LPIPS & $\downarrow$ $L_2$ \\
        \midrule
        $1$ & $1,396,658$   & $\mathbf{0.99773}$ & $0.9739$          & $0.0201$          & $0.0033$           \\
        $2$ & $2,793,316$   & $0.99769$          & $0.9778$          & $0.0177$          & $0.0028$           \\
        $3$ & $4,189,974$   & $0.99751$          & $0.9785$          & $0.0170$          & $0.0027$           \\
        $4$ & $5,586,632$   & $0.99729$          & $\mathbf{0.9790}$ & $\mathbf{0.0167}$ & $\mathbf{0.0026}$  \\
        \bottomrule
    \end{tabular}
\end{table}

In order to assess the contribution of the normalization scheme utilized by our approach our model was evaluated on the first 200 samples from the test set of CelebA-HQ.
First, we test whether the inputs to the module $M_i$ have to be normalized using SN,
Instance Normalization (IN)~\cite{ulyanov2016instance}, or kept untouched.
Additionally, we test which normalization scheme should be applied inside each residual block.
Table~\ref{table:grads_as_inputablation} summarizes the different normalization choices.
As can be seen, not applying normalization in the residual block harms all evaluation metrics,
whereas normalizing the inputs prior to feeding them into the gradient modification modules, $M_i$, is not necessary.

\begin{table}[t]
    \centering
    \caption{Different normalization schemes applied to 200 samples from the test set of the CelebA-HQ dataset. PreNorm stands for normalizing the inputs prior to the application of $M_i$, and PostNorm stands for the normalization scheme applied in each residual block. SN is the Scale Normalization~\cite{nguyen2019transformers} and IN is the one-dimensional Instance Normalization~\cite{ulyanov2016instance}.}
    \label{table:grads_as_inputablation}
    \begin{tabular}{lcccccc}
        \toprule
                     & PreNorm  & PostNorm  & $\uparrow$ID     & $\uparrow$MS-SSIM & $\downarrow$LPIPS & $\downarrow$ $L_2$ \\
        \midrule
        (a)          & $-$ & $-$ & $0.834$          & $0.765$           & $0.172$           & $0.039$            \\
        (b)          & $-$ & IN  & $0.996$          & $0.996$           & $0.030$           & $0.005$            \\
        \textbf{(c)} & $-$ & SN  & $\mathbf{0.998}$ & $\mathbf{0.998}$  & $\mathbf{0.020}$  & $\mathbf{0.003}$   \\
        (d)          & IN  & $-$ & $0.895$          & $0.802$           & $0.151$           & $0.032$            \\
        (e)          & IN  & IN  & $0.997$          & $0.974$           & $0.021$           & $0.003$            \\
        (f)          & IN  & SN  & $0.998$          & $0.973$           & $0.021$           & $0.003$            \\
        (g)          & SN  & $-$ & $0.997$          & $0.778$           & $0.168$           & $0.037$            \\
        (h)          & SN  & IN  & $0.857$          & $0.972$           & $0.023$           & $0.004$            \\
        (i)          & SN  & SN  & $0.997$          & $0.974$           & $0.021$           & $0.003$            \\
        \bottomrule
    \end{tabular}
\end{table}

\begin{figure}[ht!]
    \centering
    \begin{tabular}{@{}c:c@{}c@{}c@{}c:c@{}c@{}c@{}c:c@{}c@{}c@{}c@{}}
        {Source}                                                                                & \multicolumn{4}{c}{Pose} & \multicolumn{4}{c}{Smile} & \multicolumn{4}{c}{Age} \\[2pt]
        \includegraphics[width=0.07\linewidth]{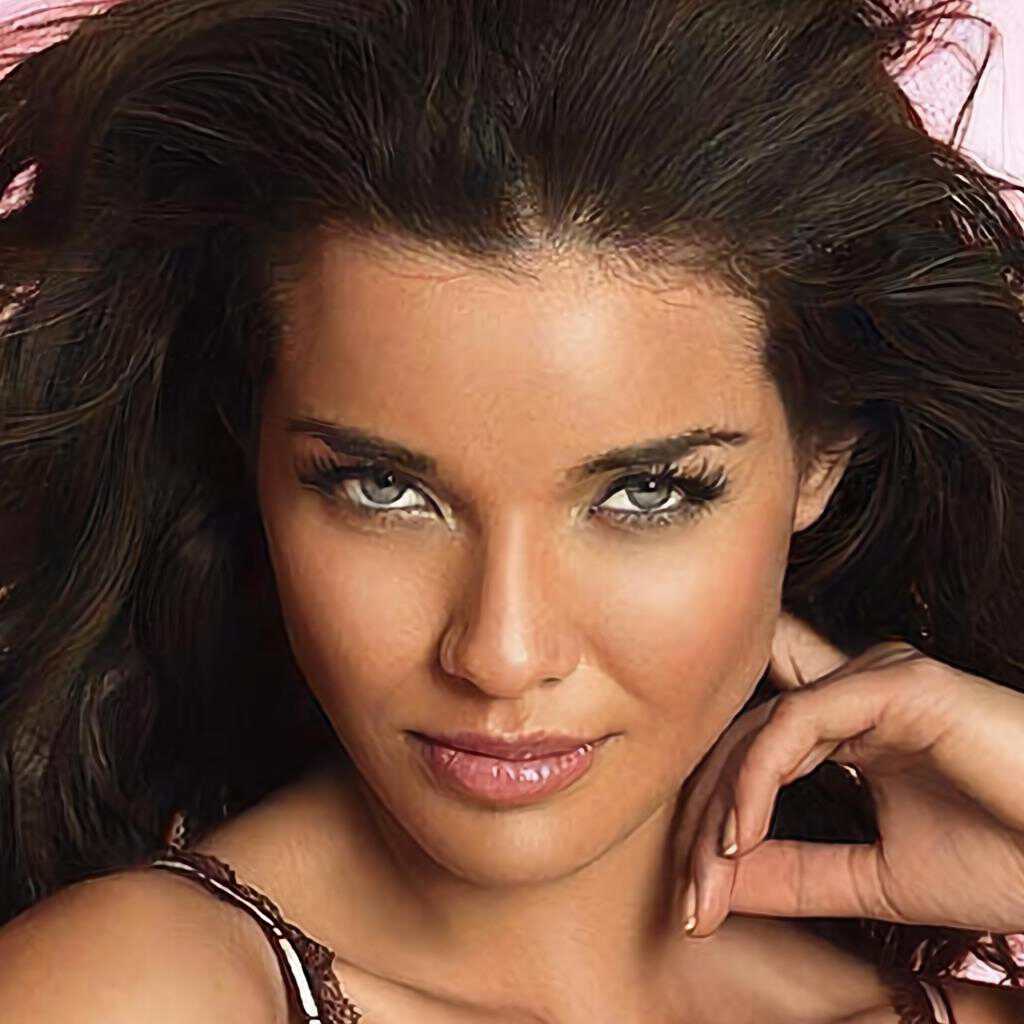}                  &
        \includegraphics[width=0.07\linewidth]{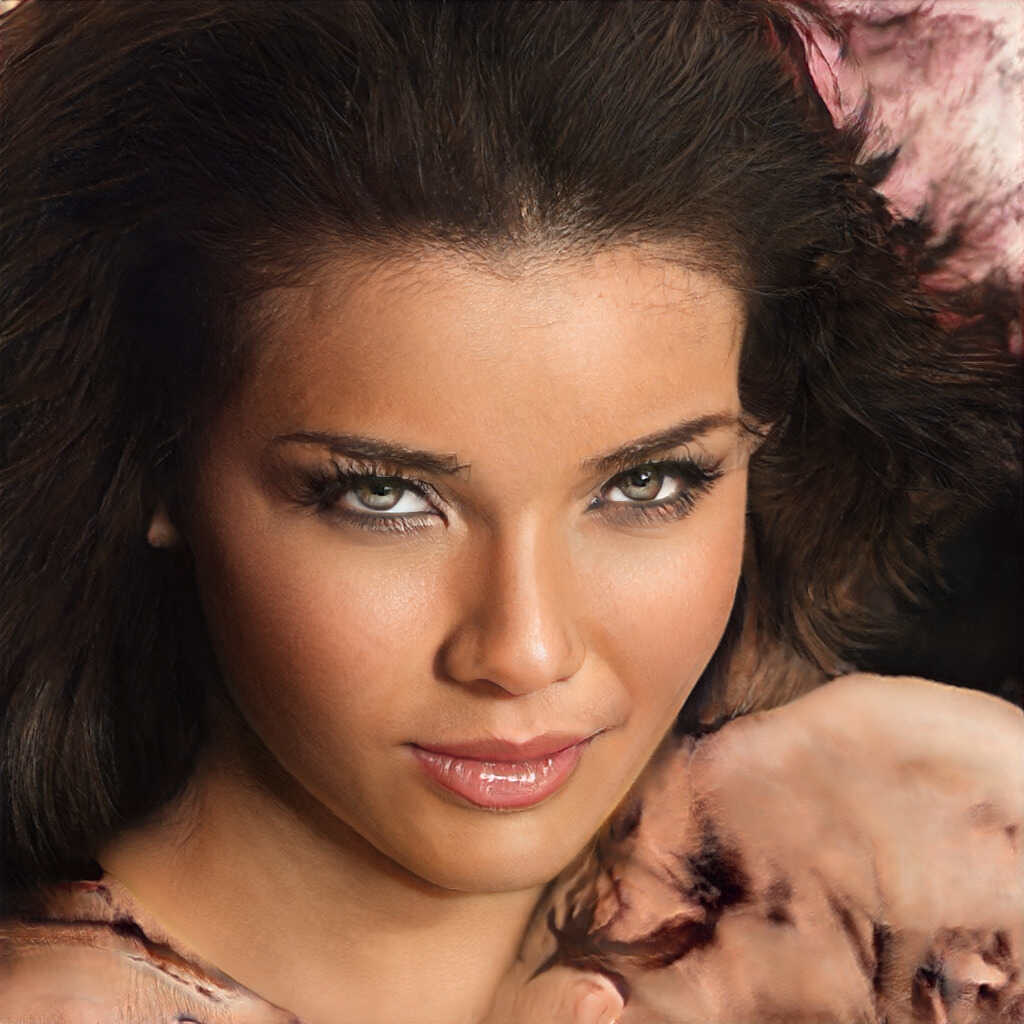}    &
        \includegraphics[width=0.07\linewidth]{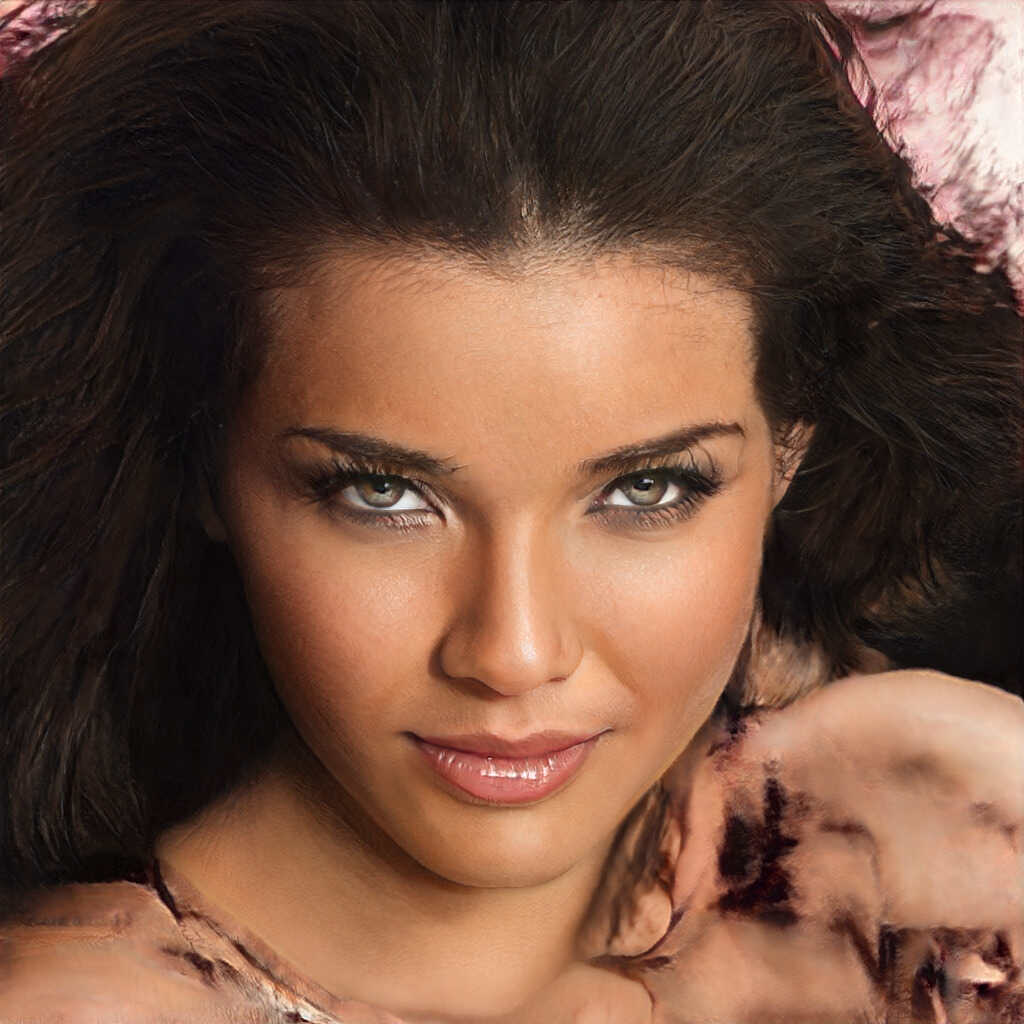}    &
        \includegraphics[width=0.07\linewidth]{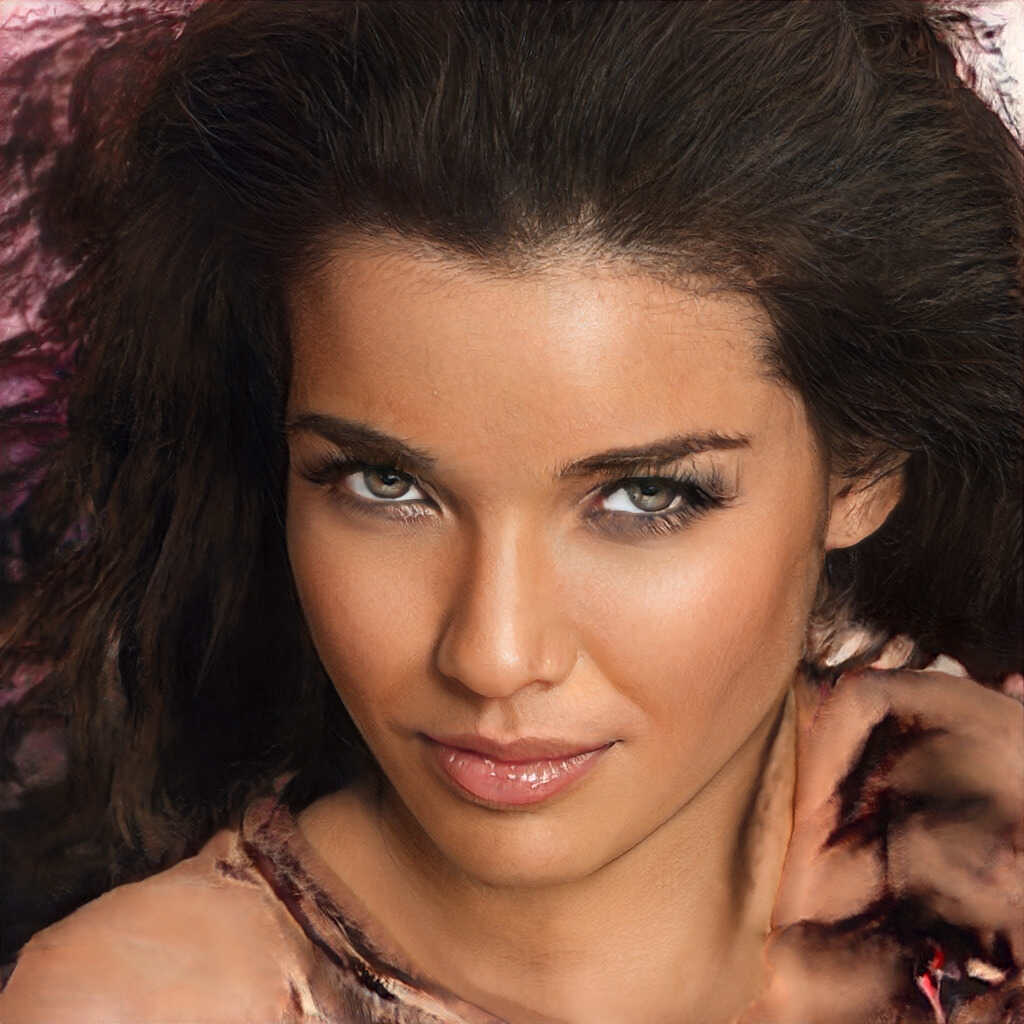}     &
        \includegraphics[width=0.07\linewidth]{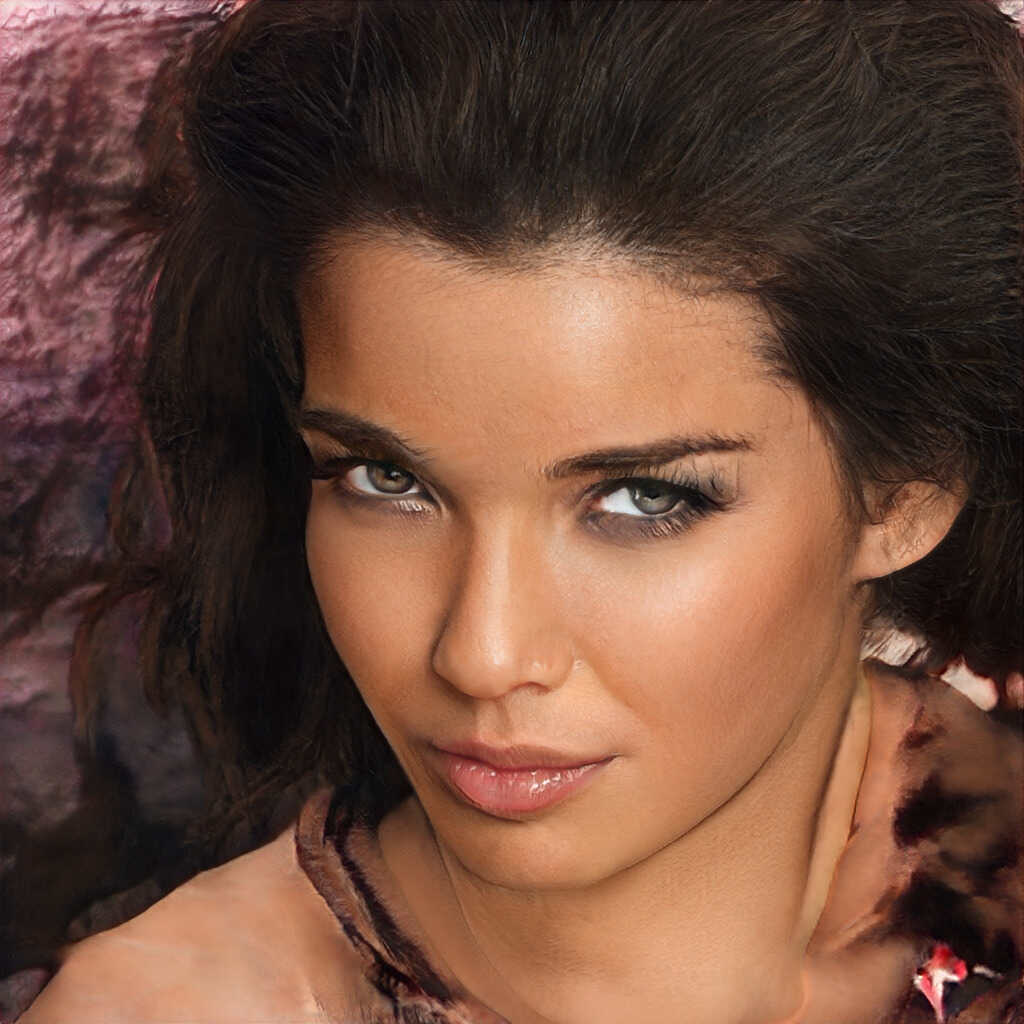}     &
        \includegraphics[width=0.07\linewidth]{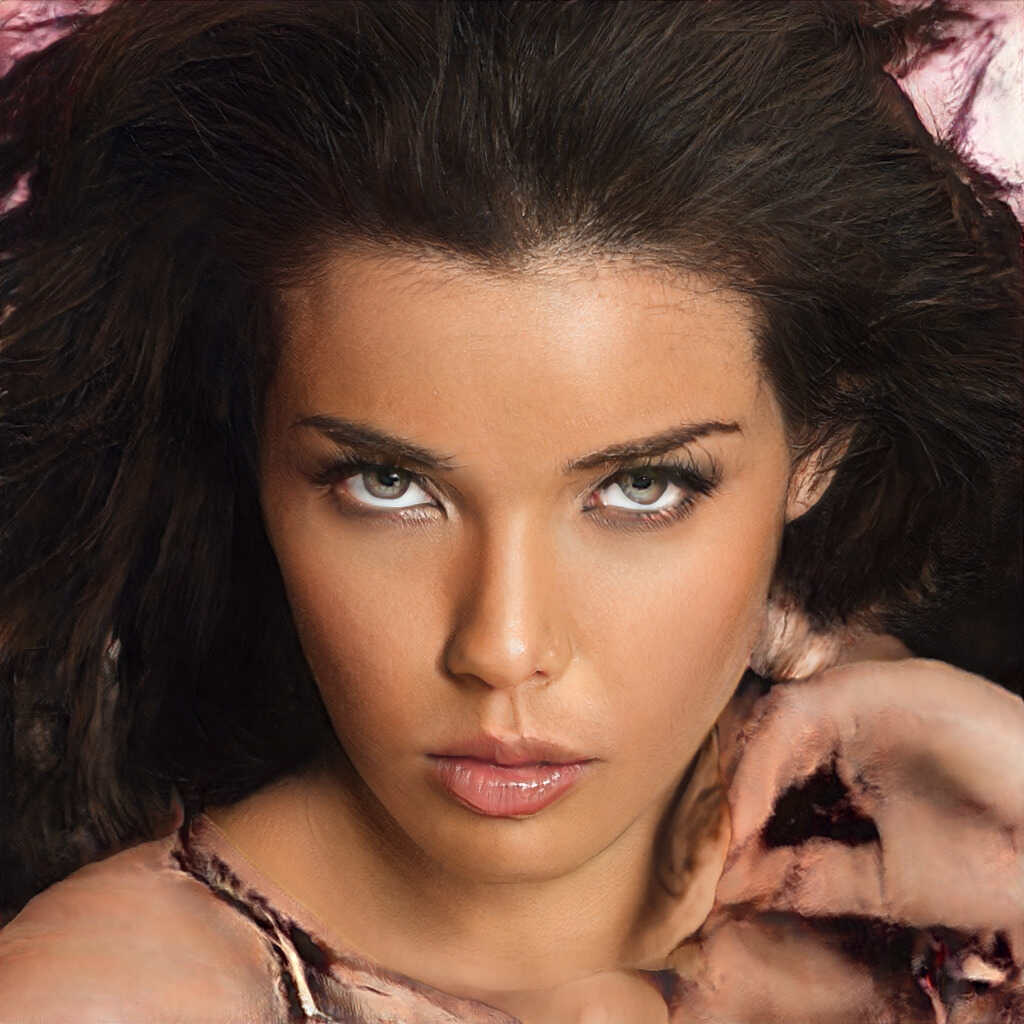} &
        \includegraphics[width=0.07\linewidth]{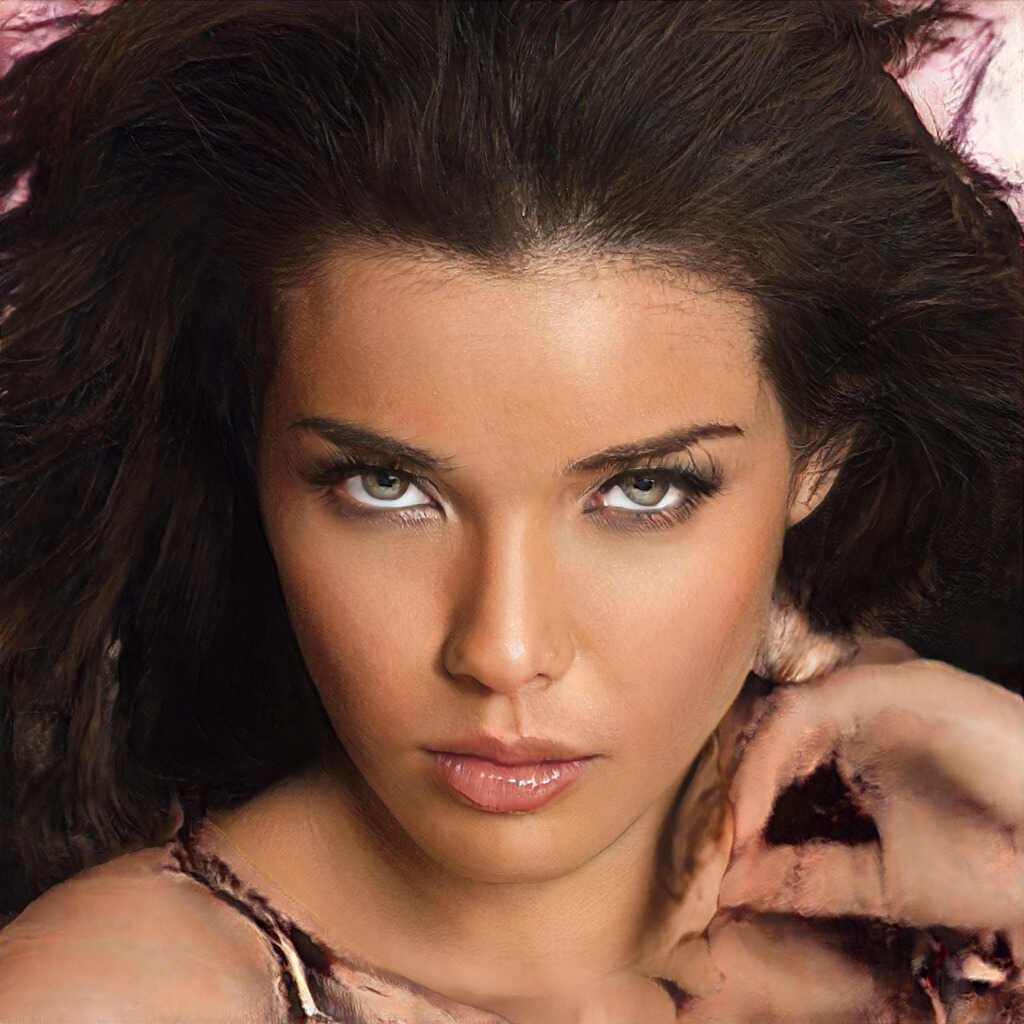}   &
        \includegraphics[width=0.07\linewidth]{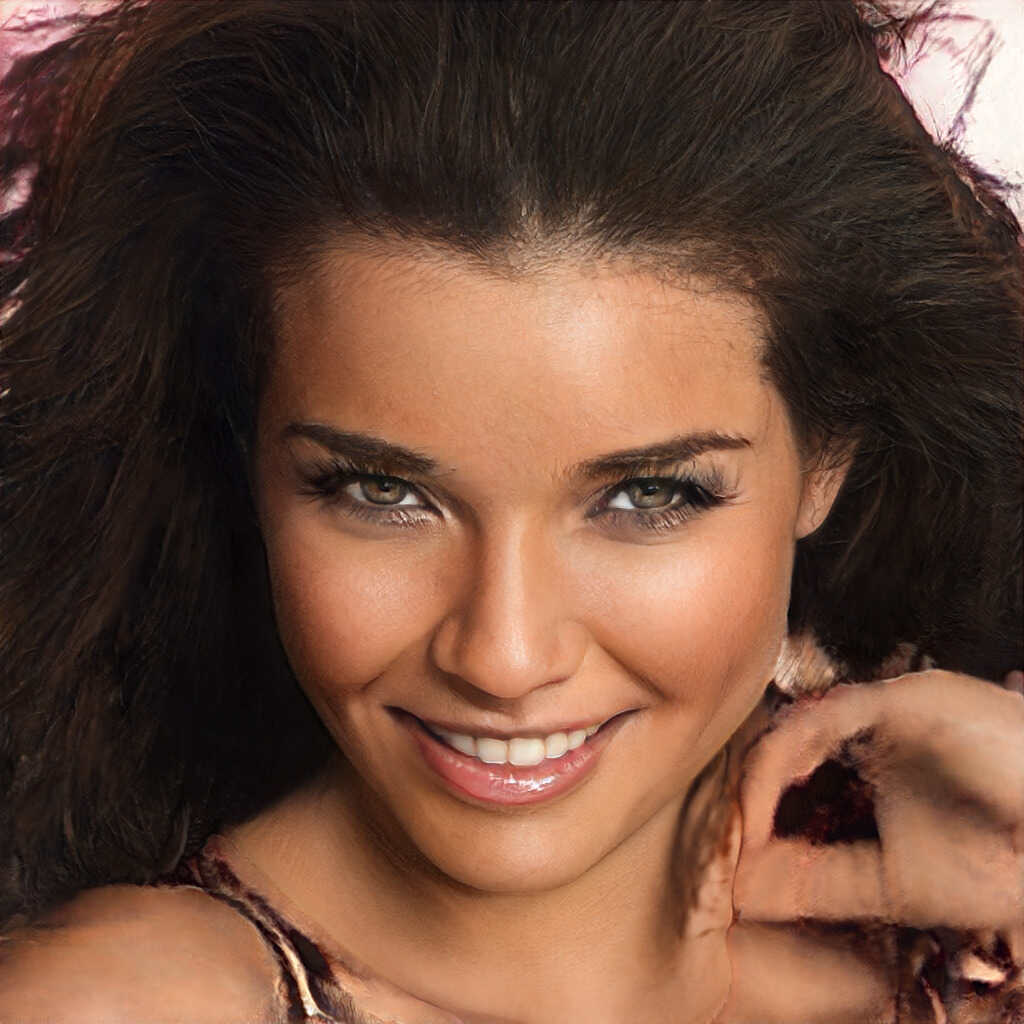}    &
        \includegraphics[width=0.07\linewidth]{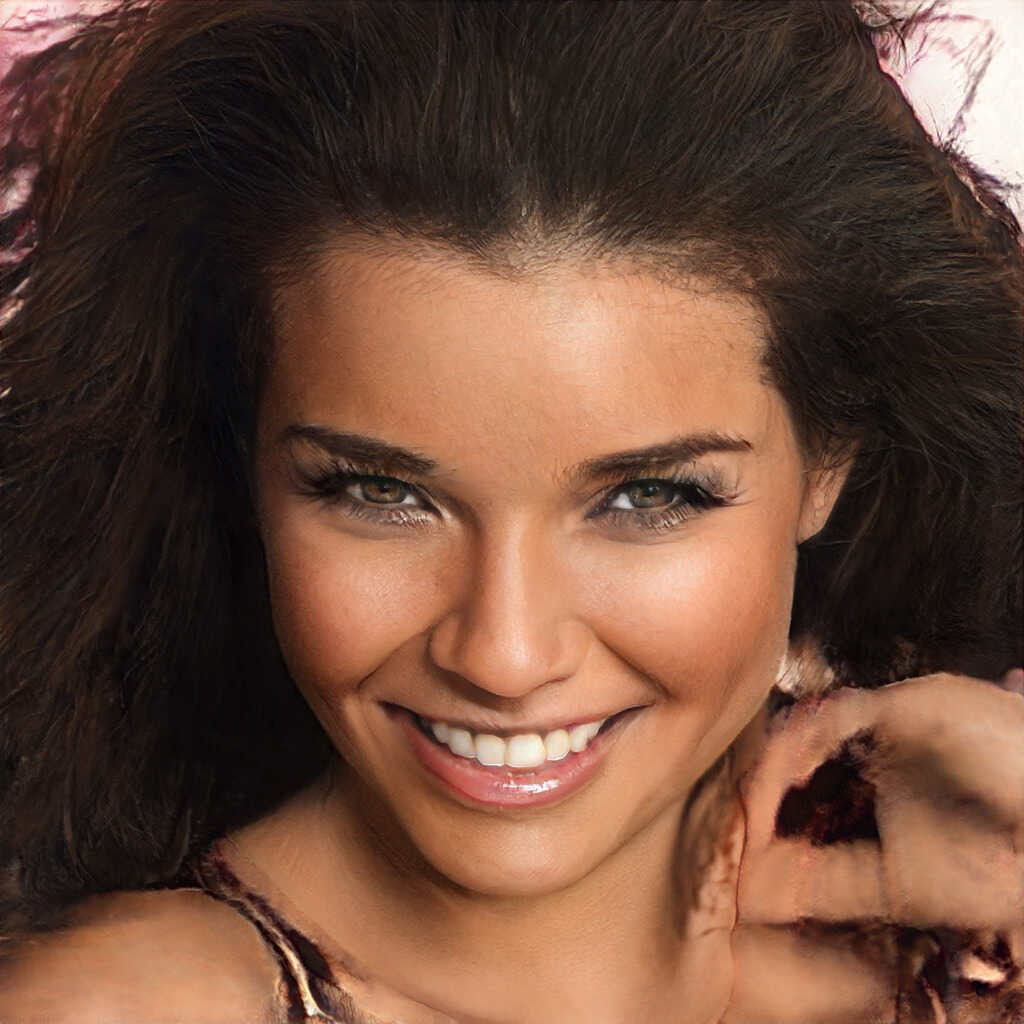}  &
        \includegraphics[width=0.07\linewidth]{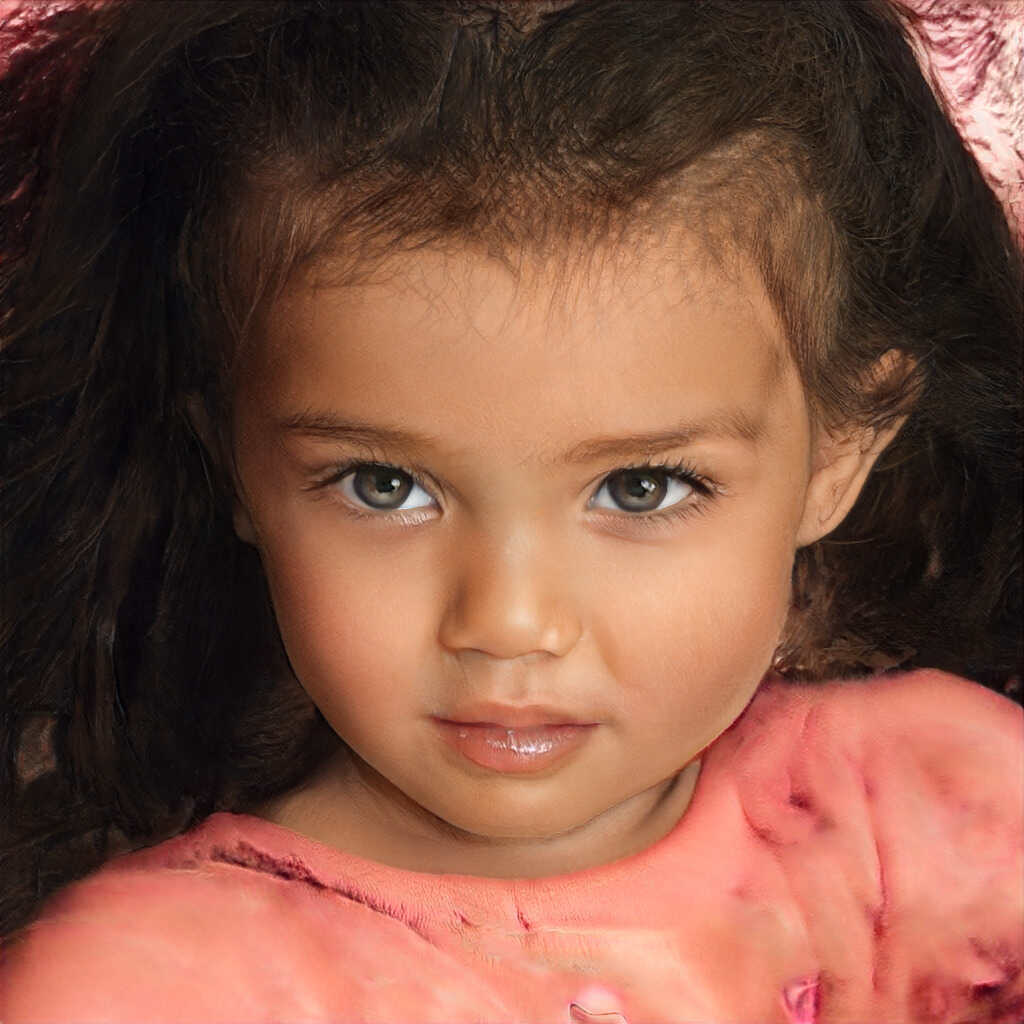}     &
        \includegraphics[width=0.07\linewidth]{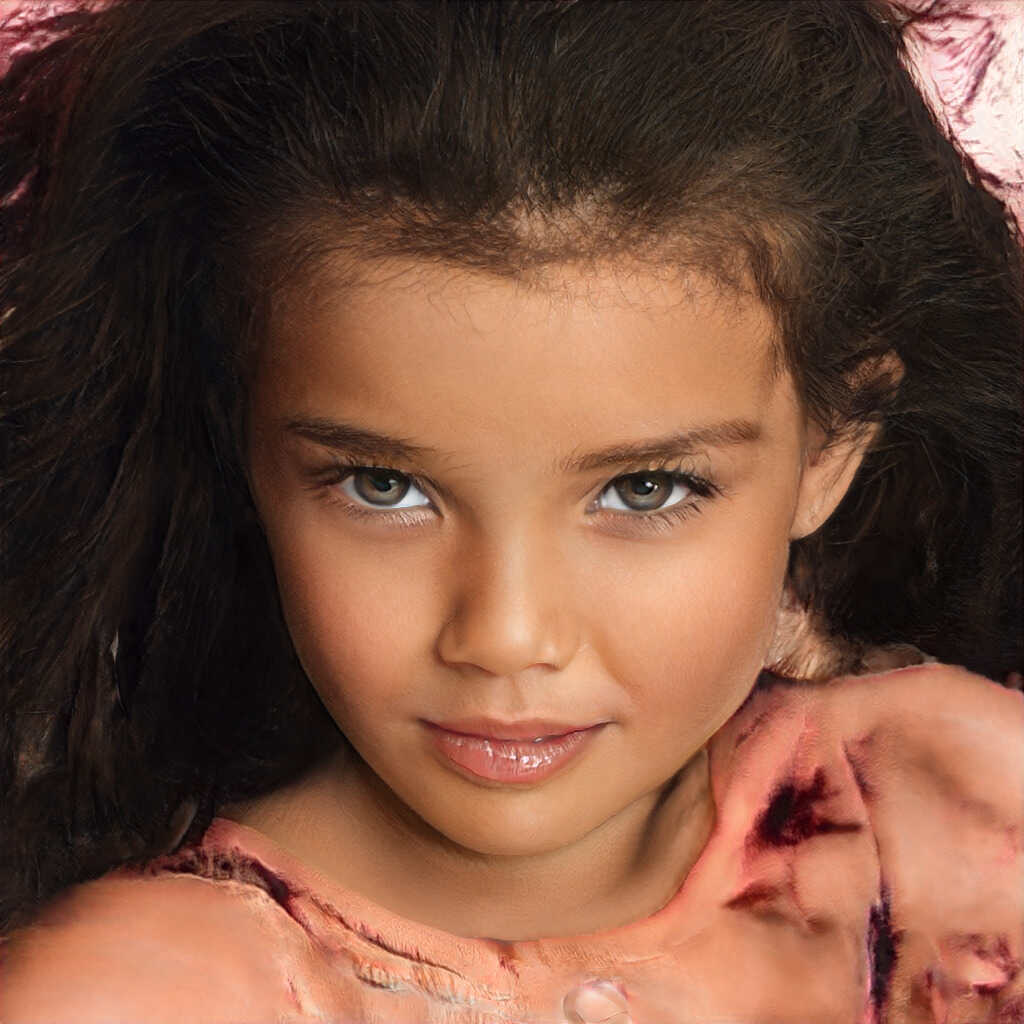}     &
        \includegraphics[width=0.07\linewidth]{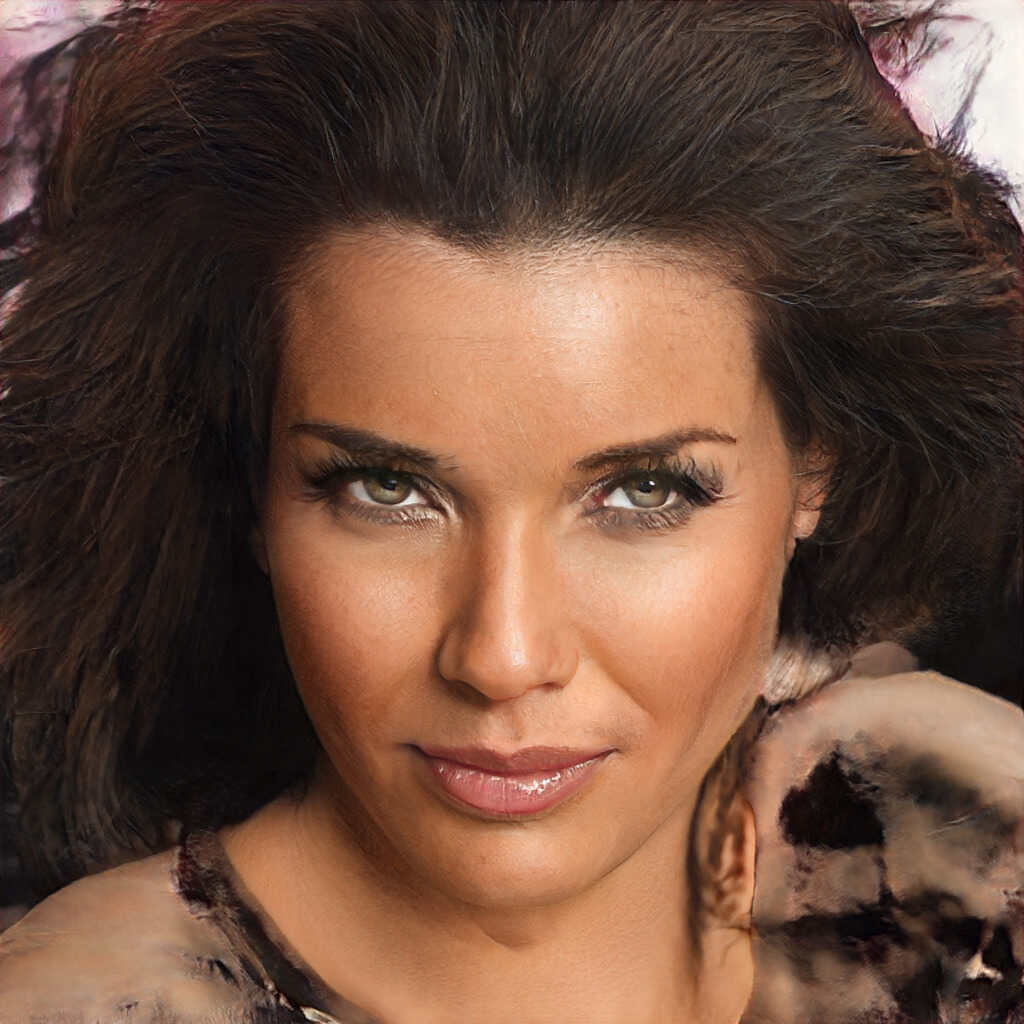}      &
        \includegraphics[width=0.07\linewidth]{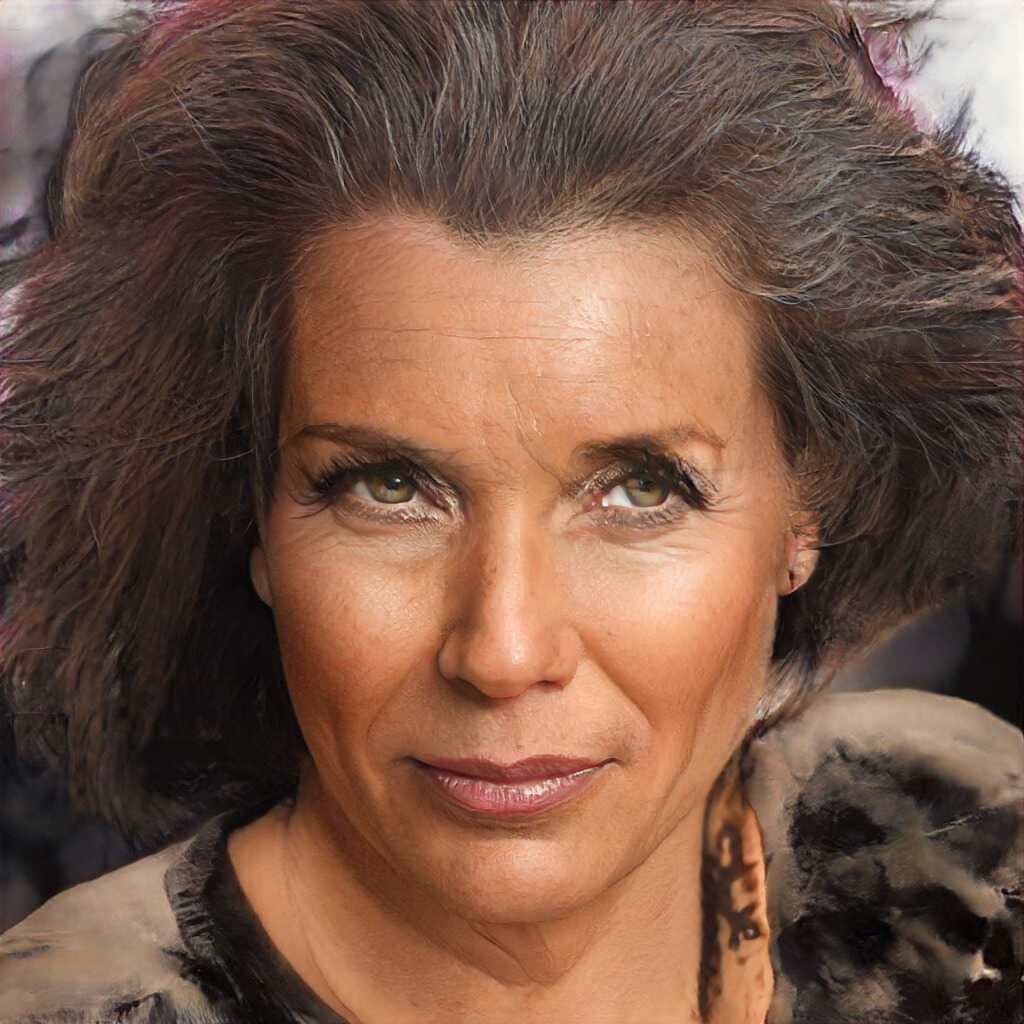}                                                                                                               \\
        \includegraphics[width=0.07\linewidth]{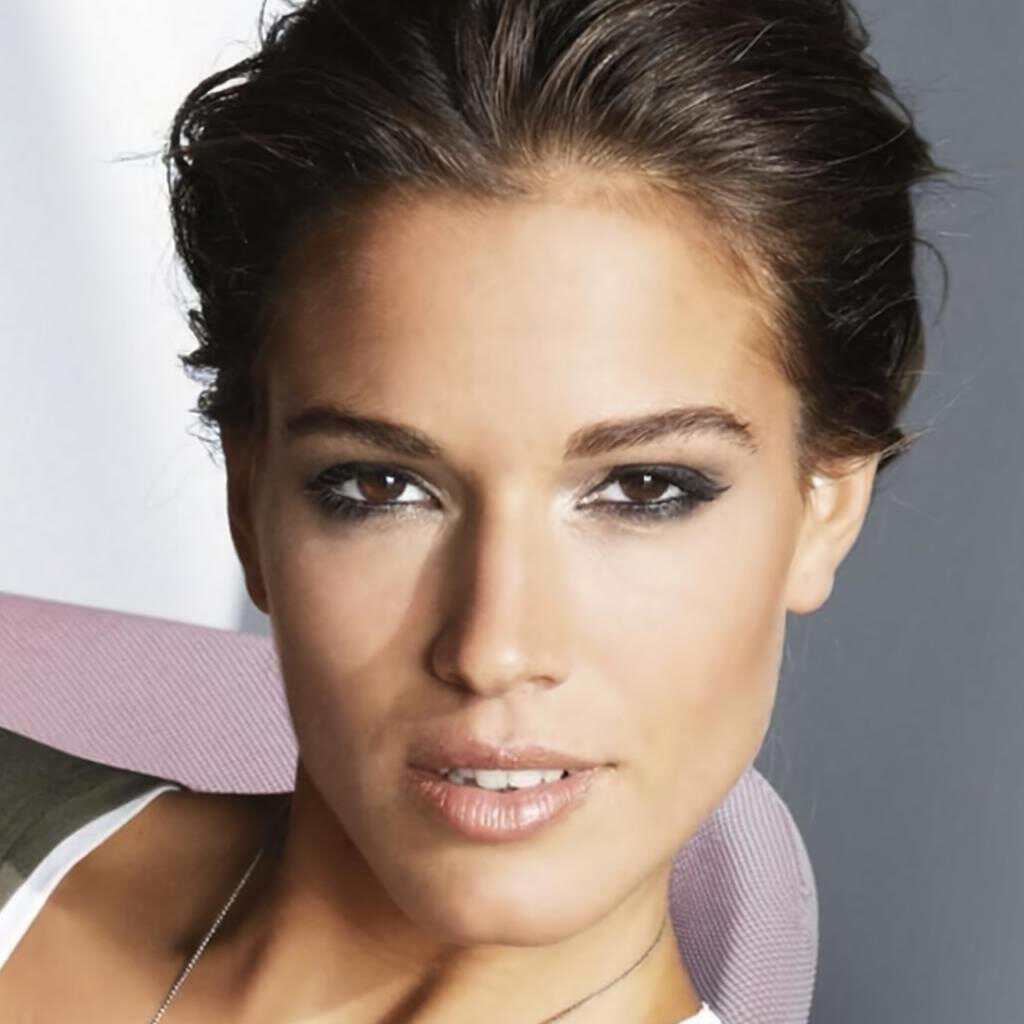}                  &
        \includegraphics[width=0.07\linewidth]{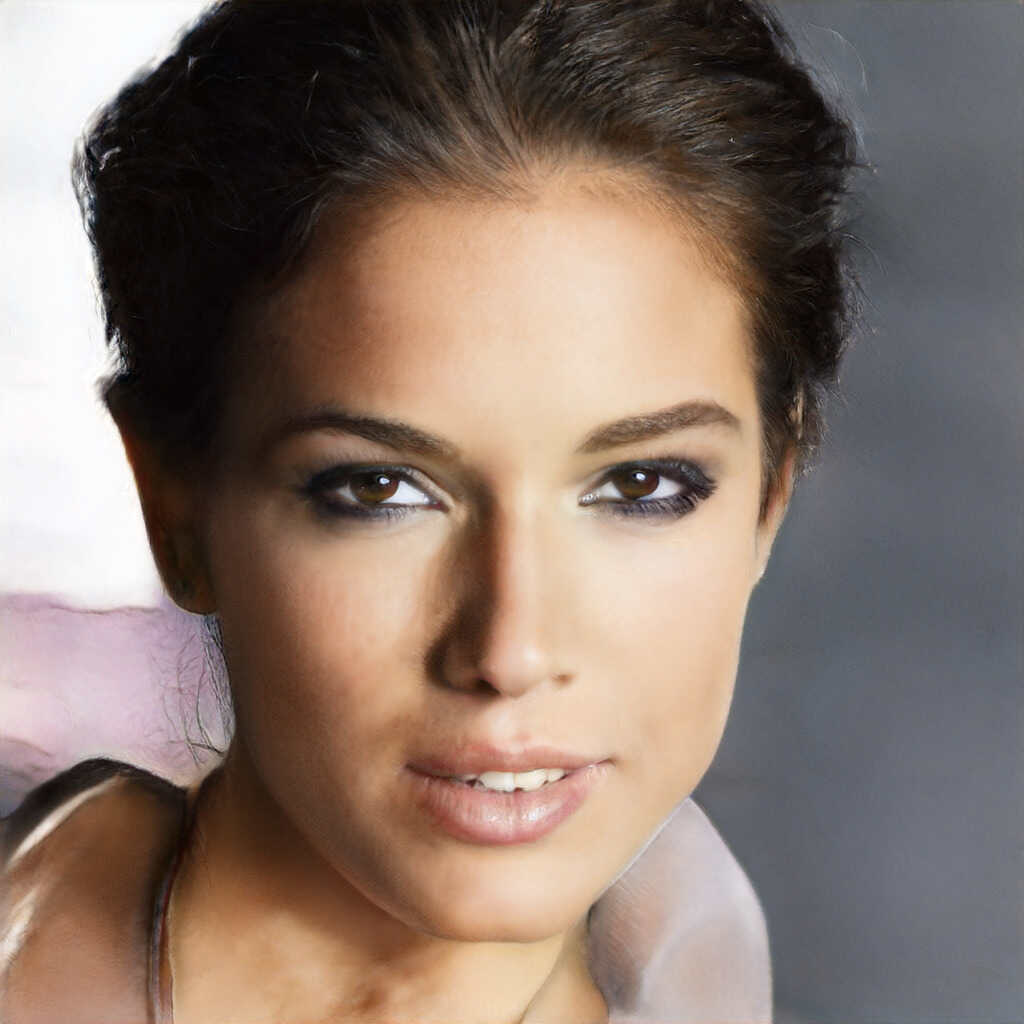}    &
        \includegraphics[width=0.07\linewidth]{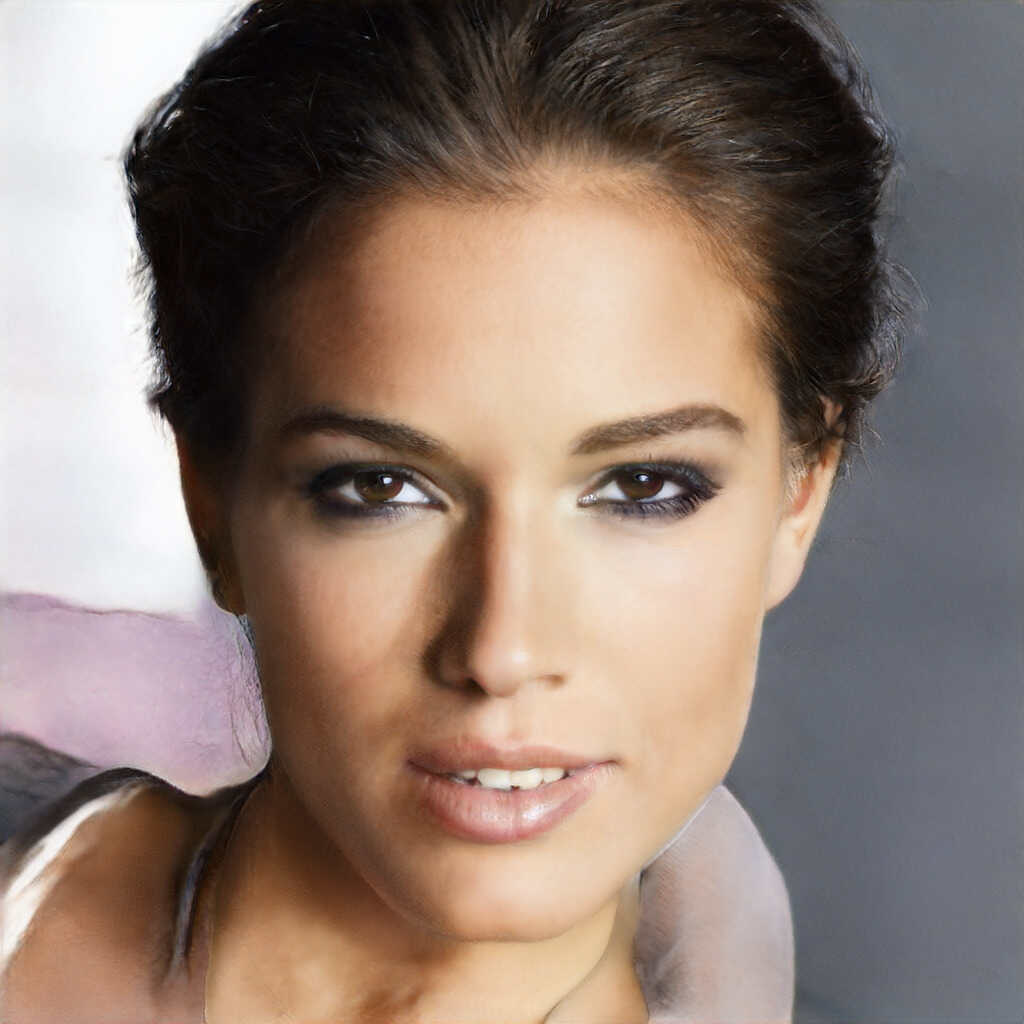}    &
        \includegraphics[width=0.07\linewidth]{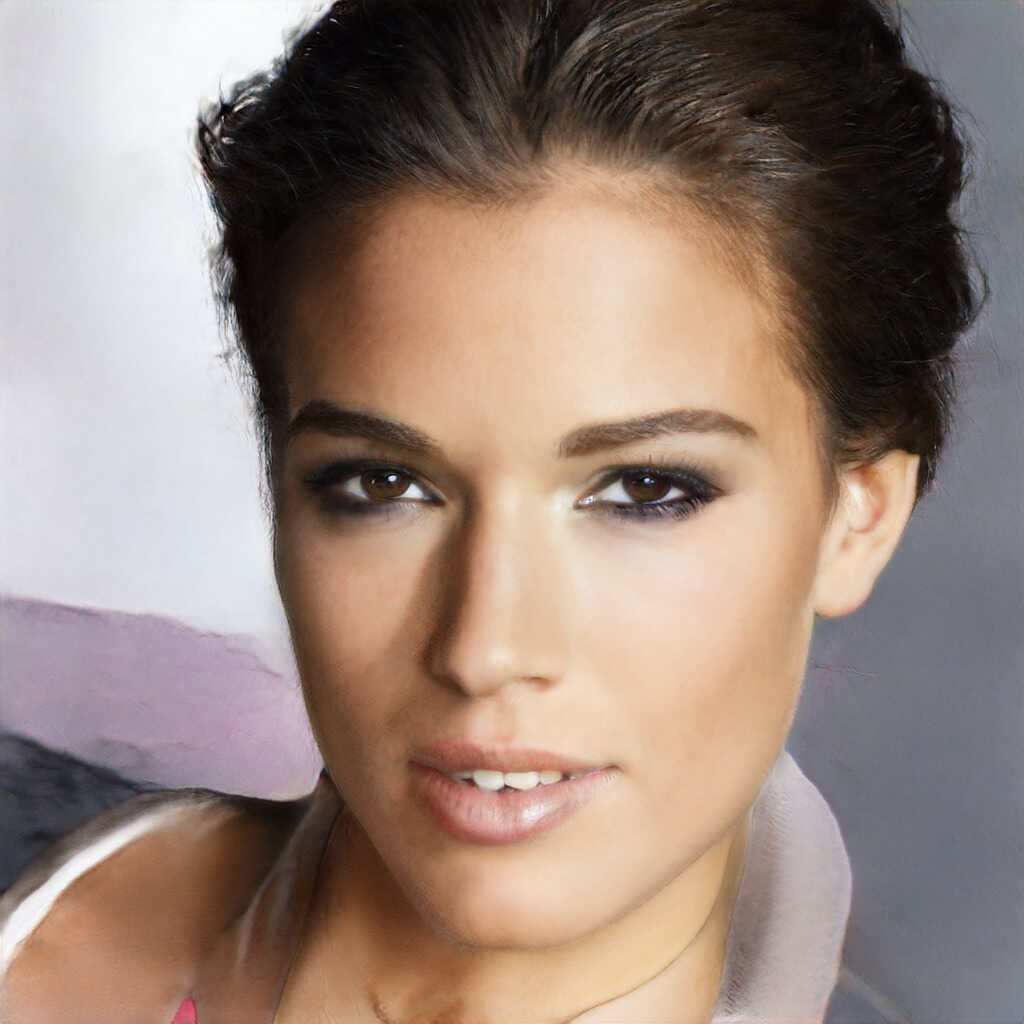}     &
        \includegraphics[width=0.07\linewidth]{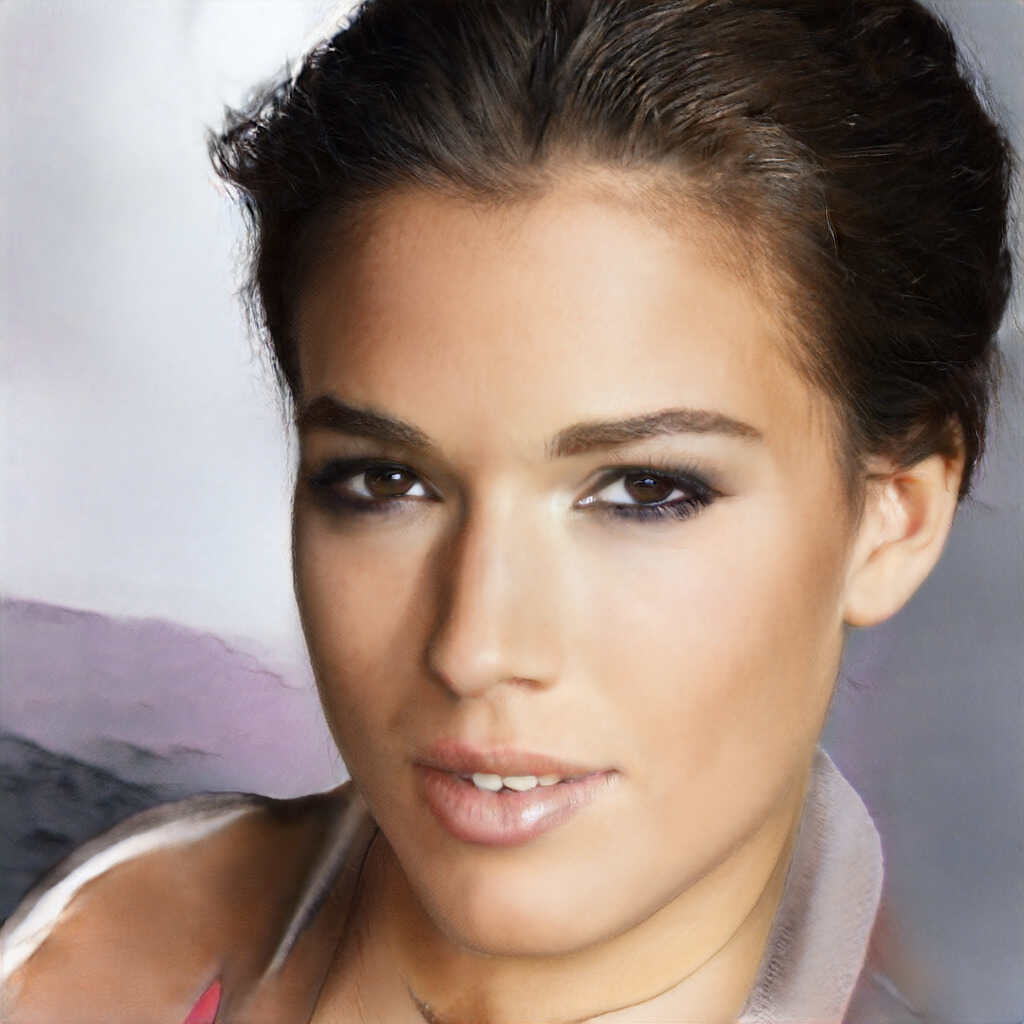}     &
        \includegraphics[width=0.07\linewidth]{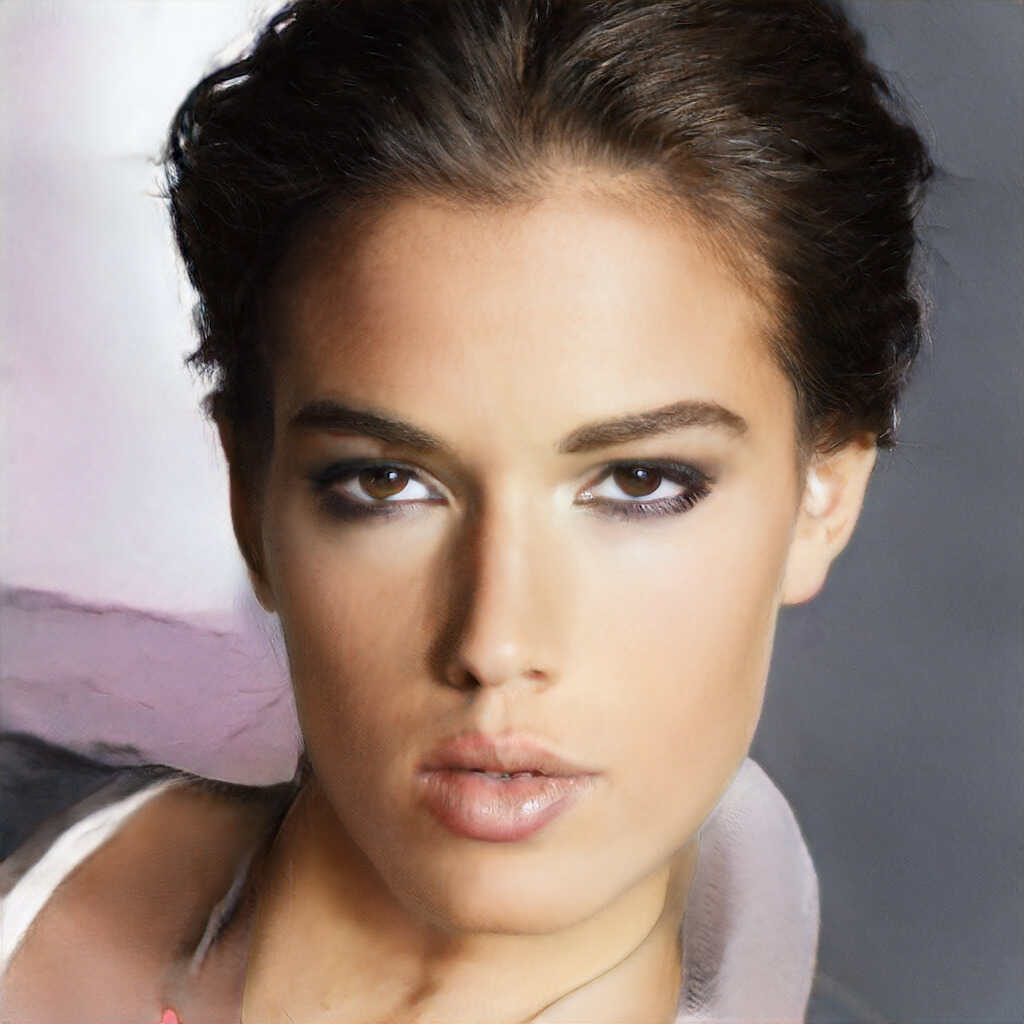} &
        \includegraphics[width=0.07\linewidth]{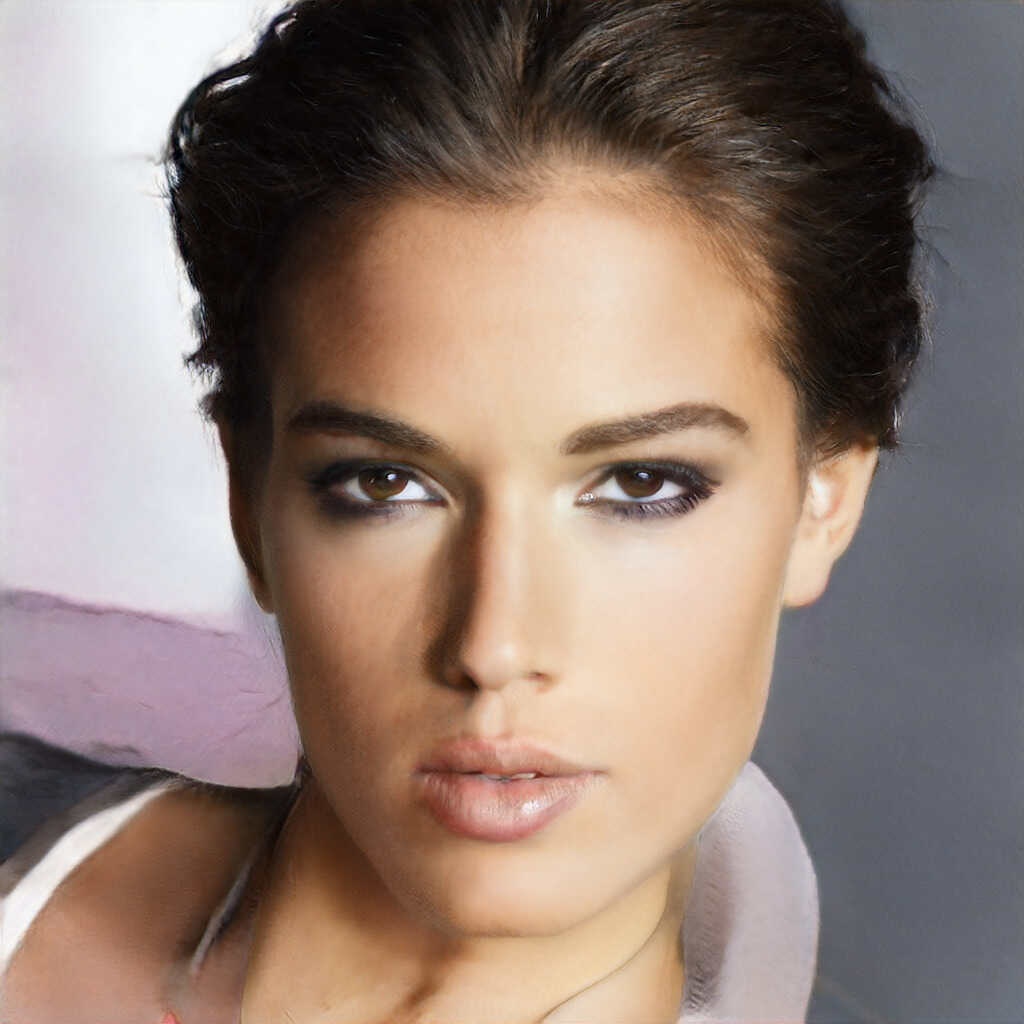}   &
        \includegraphics[width=0.07\linewidth]{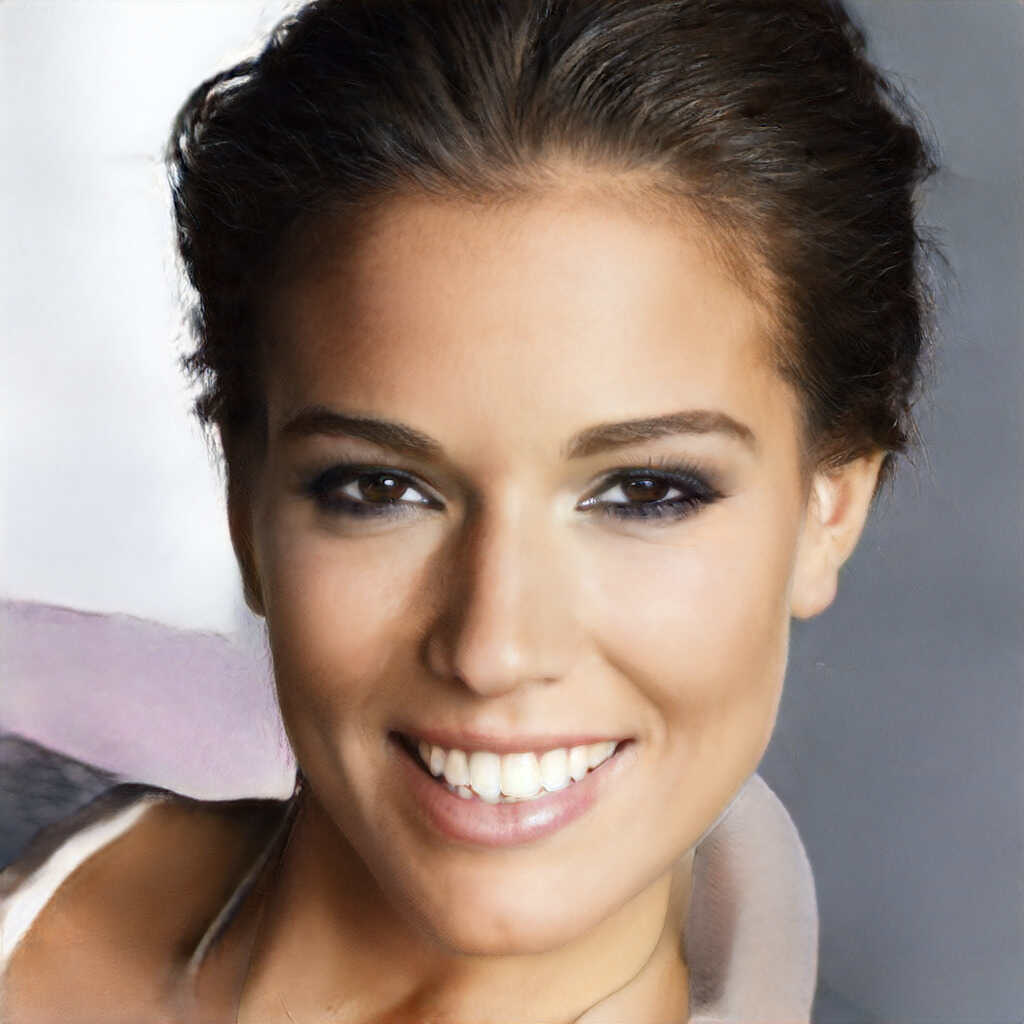}    &
        \includegraphics[width=0.07\linewidth]{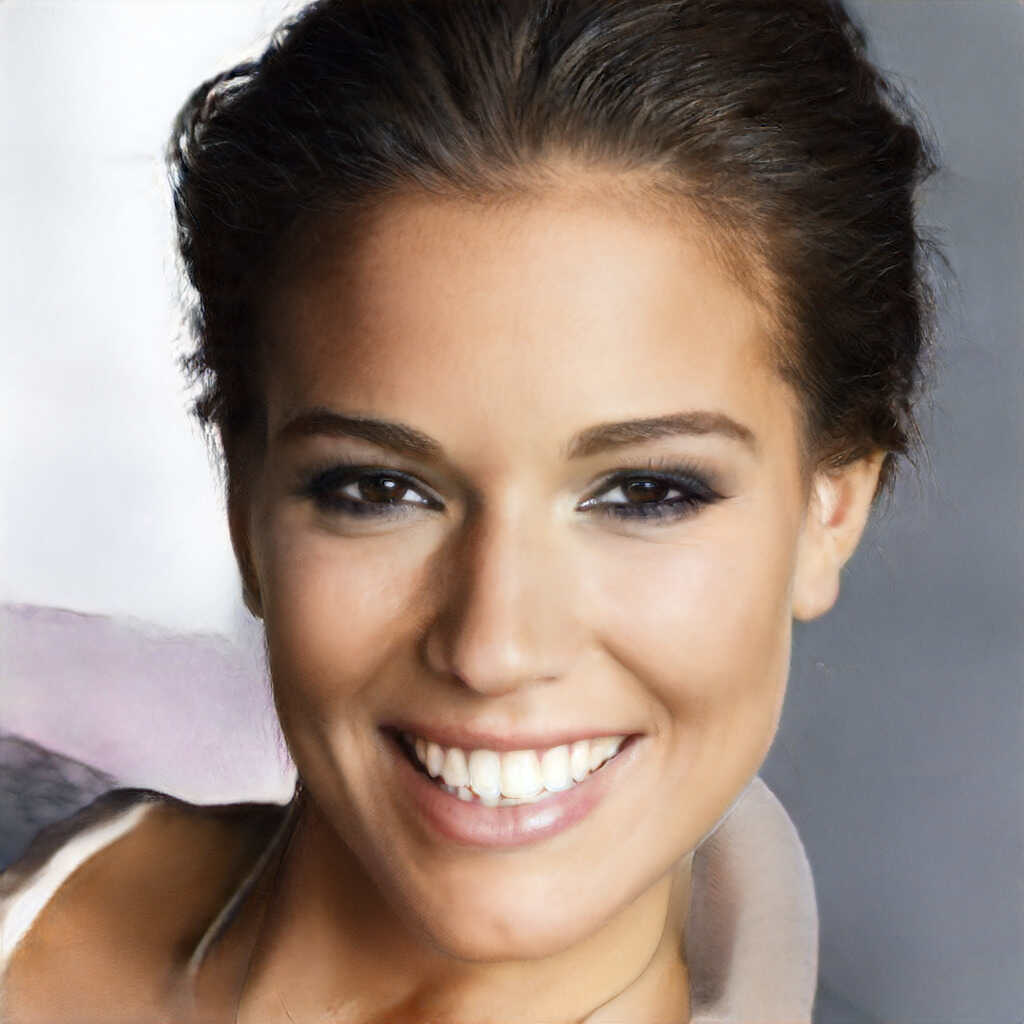}  &
        \includegraphics[width=0.07\linewidth]{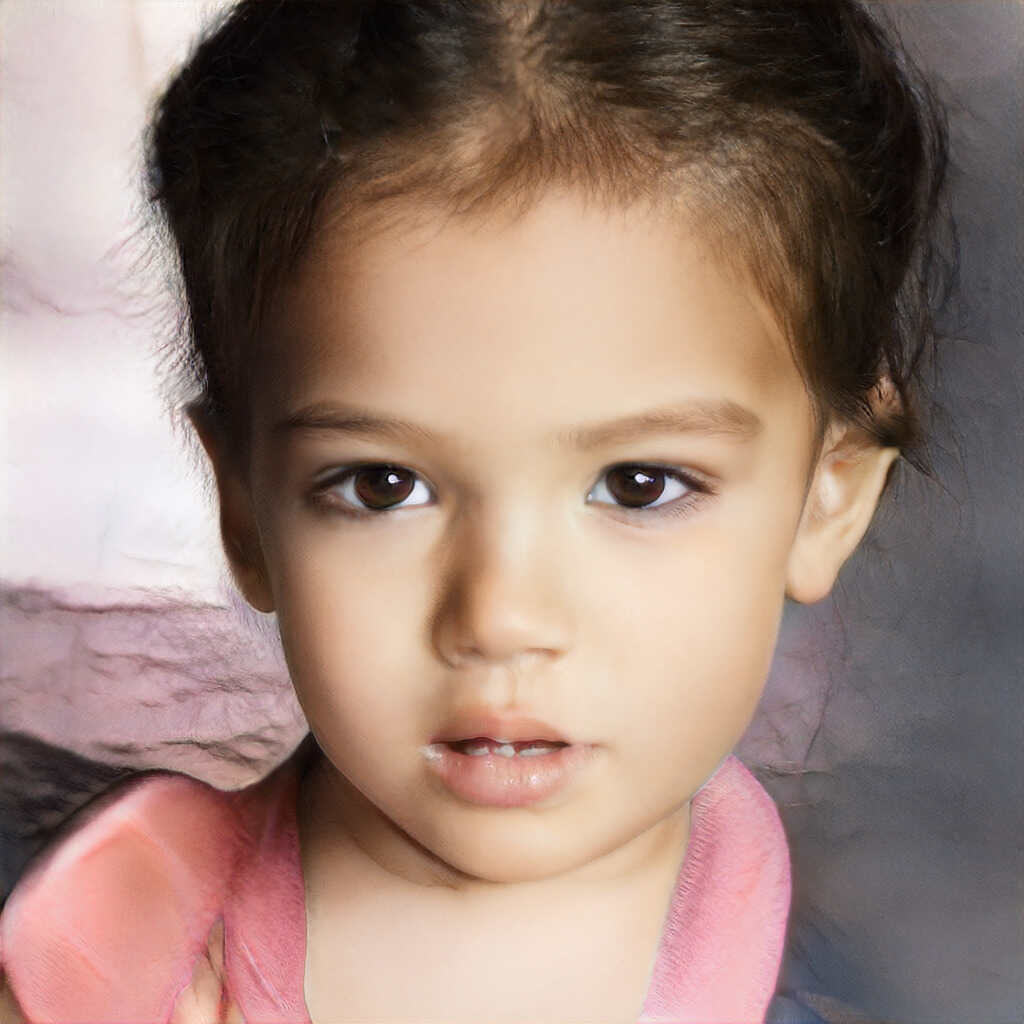}     &
        \includegraphics[width=0.07\linewidth]{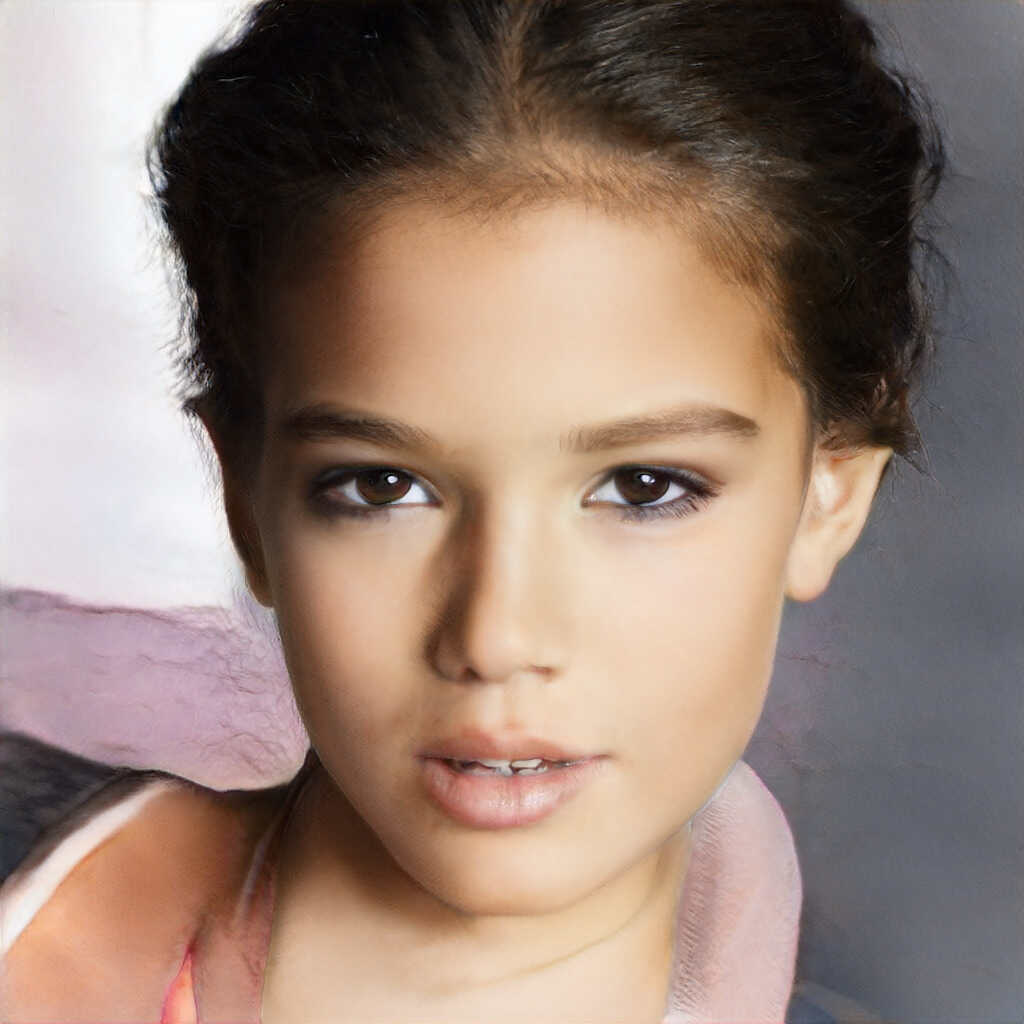}     &
        \includegraphics[width=0.07\linewidth]{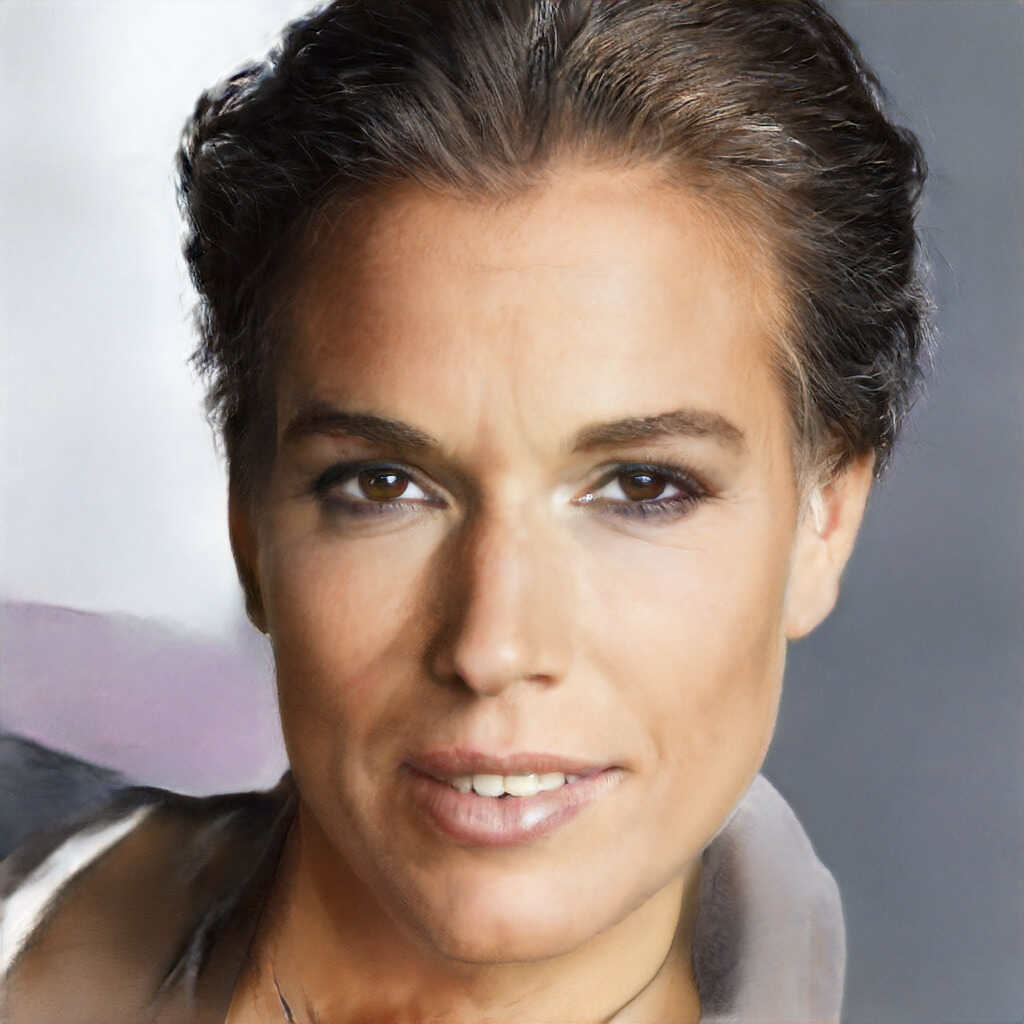}      &
        \includegraphics[width=0.07\linewidth]{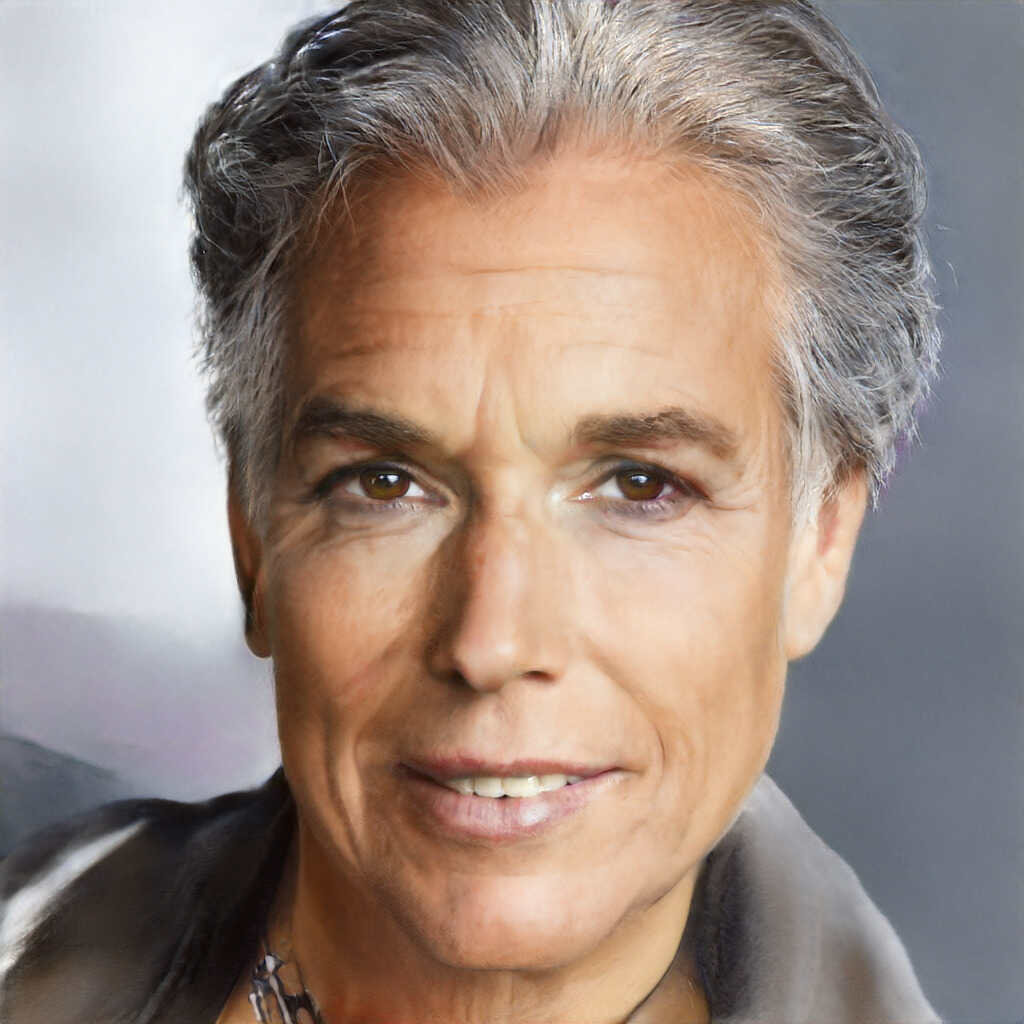}                                                                                                               \\
        \includegraphics[width=0.07\linewidth]{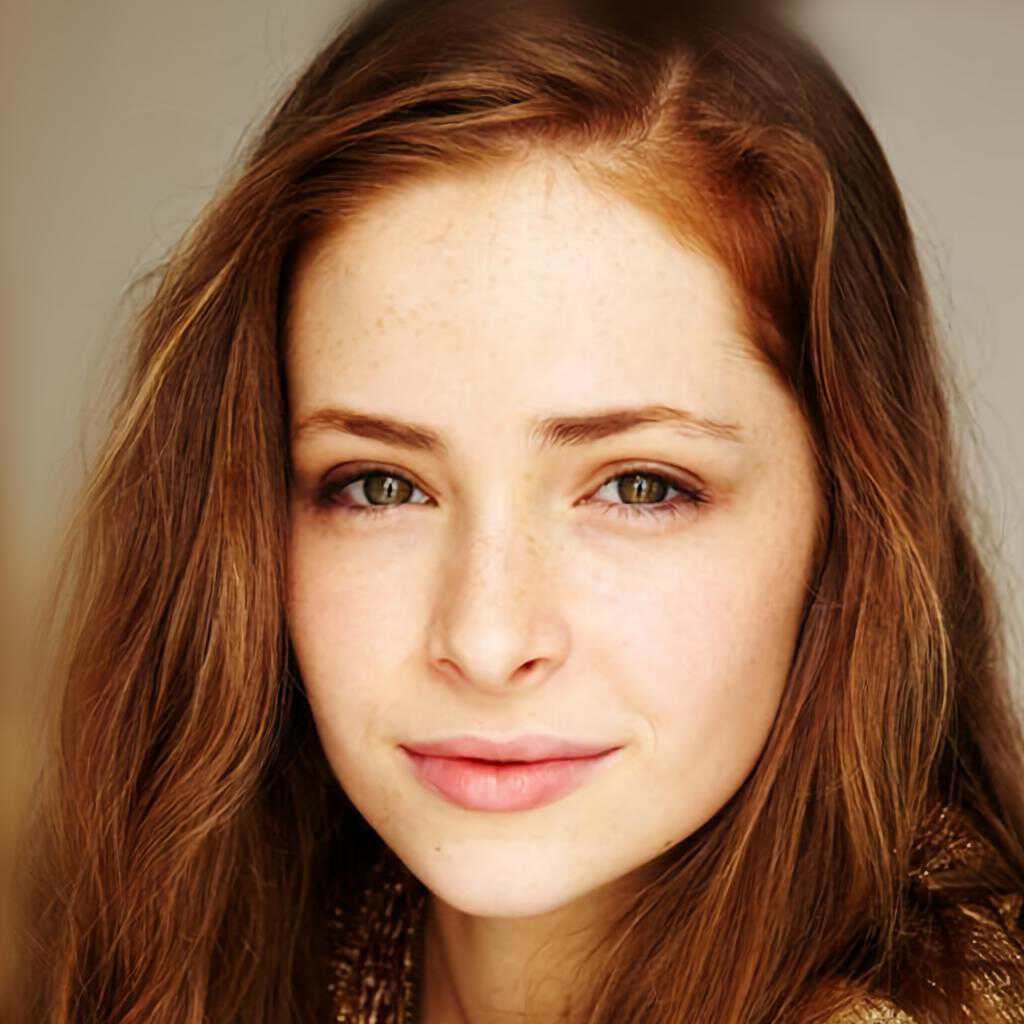}                  &
        \includegraphics[width=0.07\linewidth]{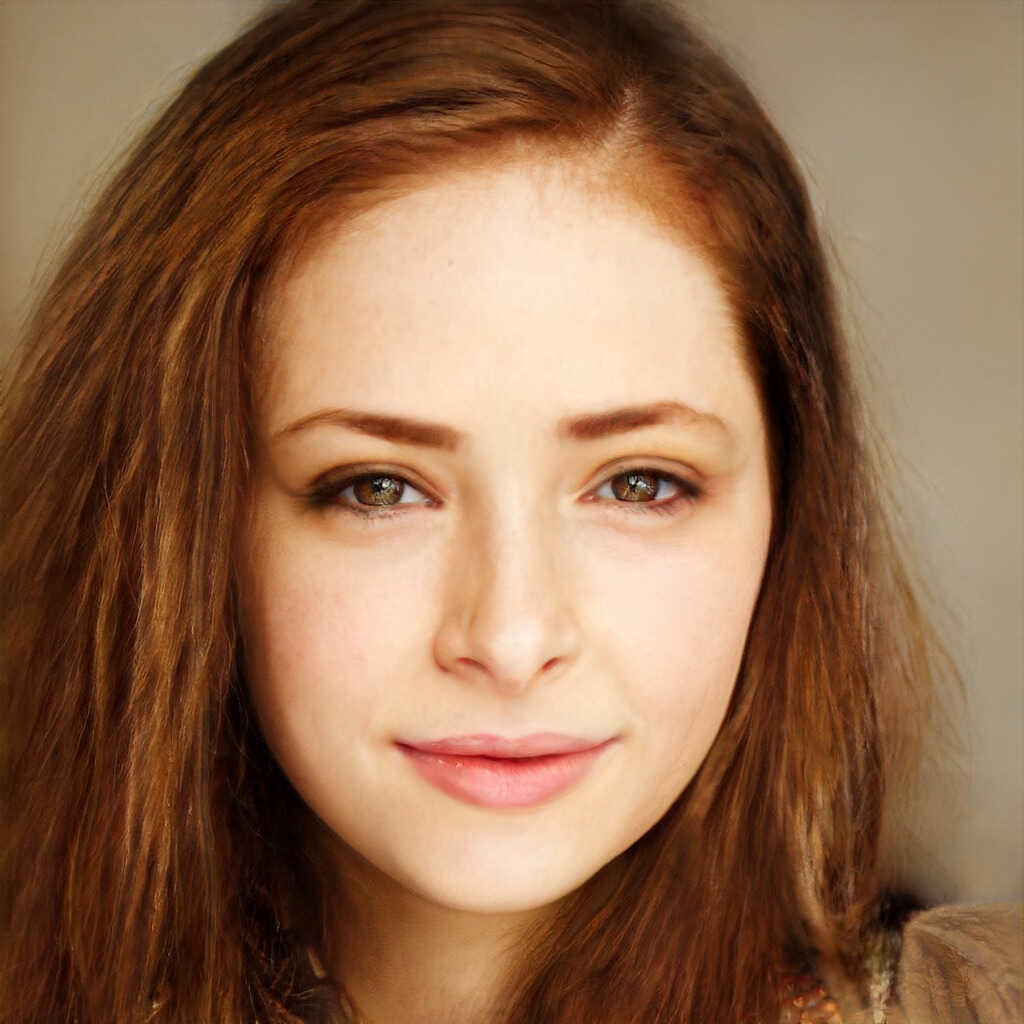}    &
        \includegraphics[width=0.07\linewidth]{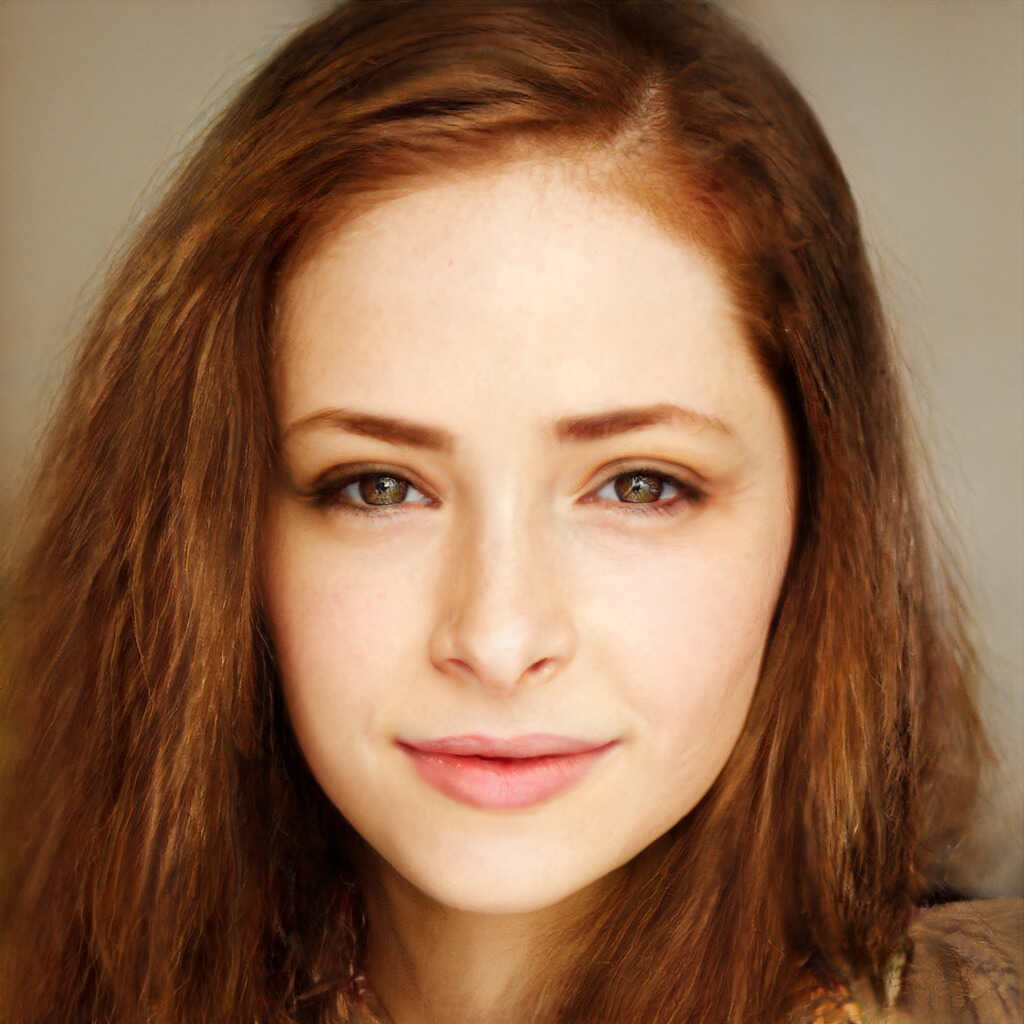}    &
        \includegraphics[width=0.07\linewidth]{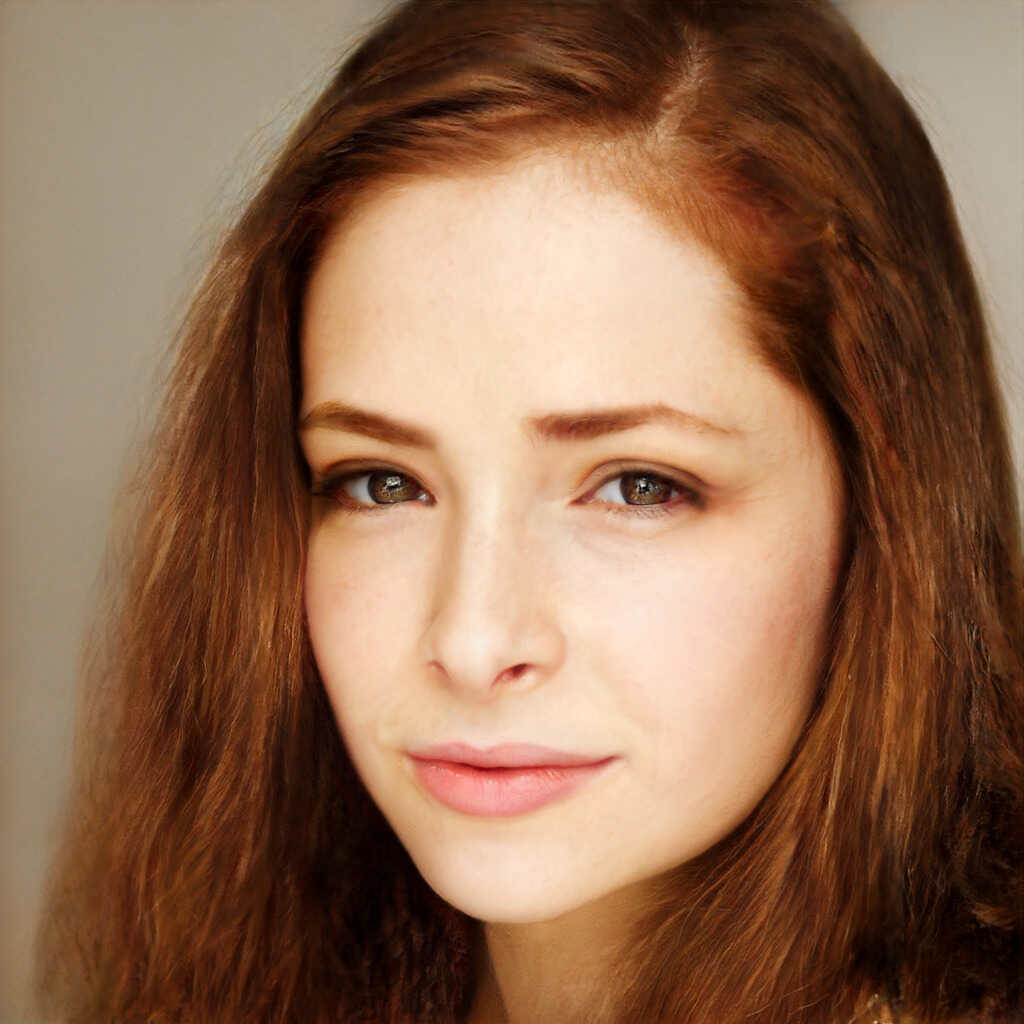}     &
        \includegraphics[width=0.07\linewidth]{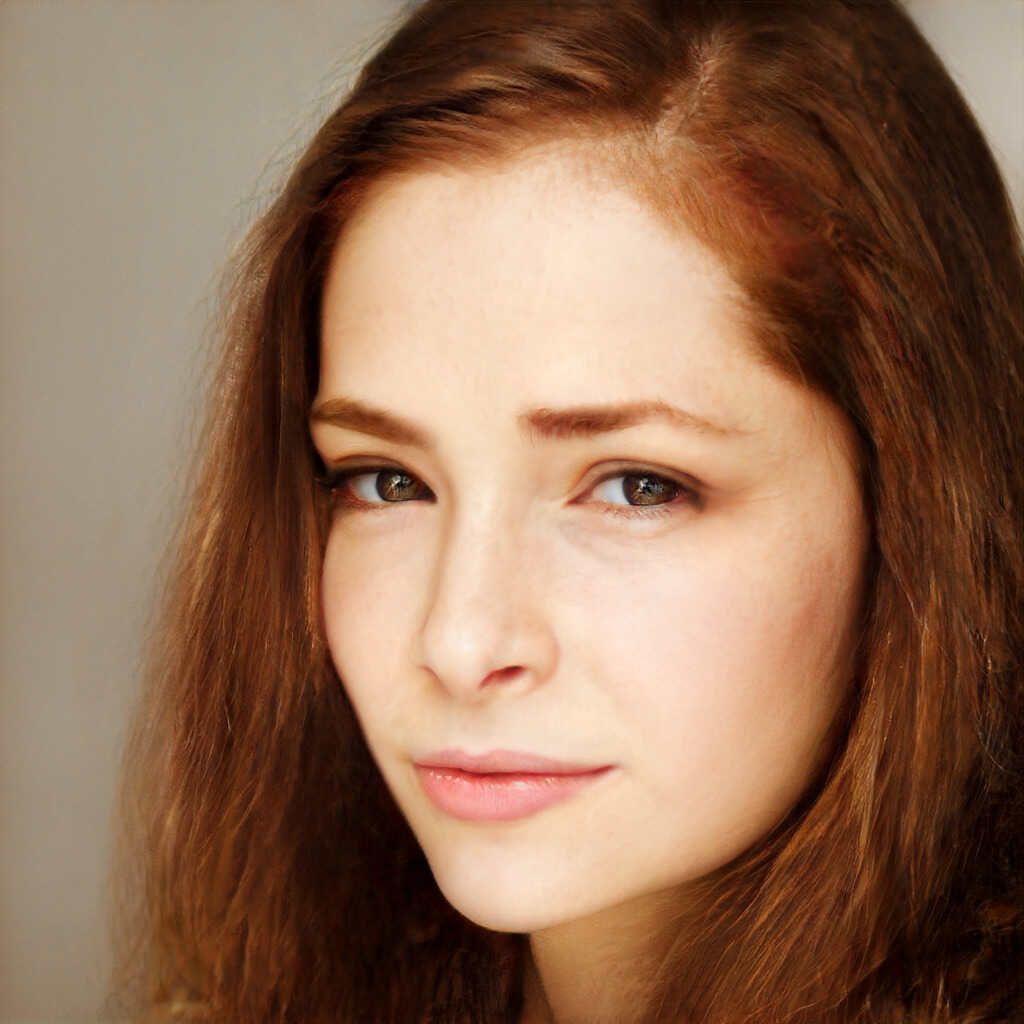}     &
        \includegraphics[width=0.07\linewidth]{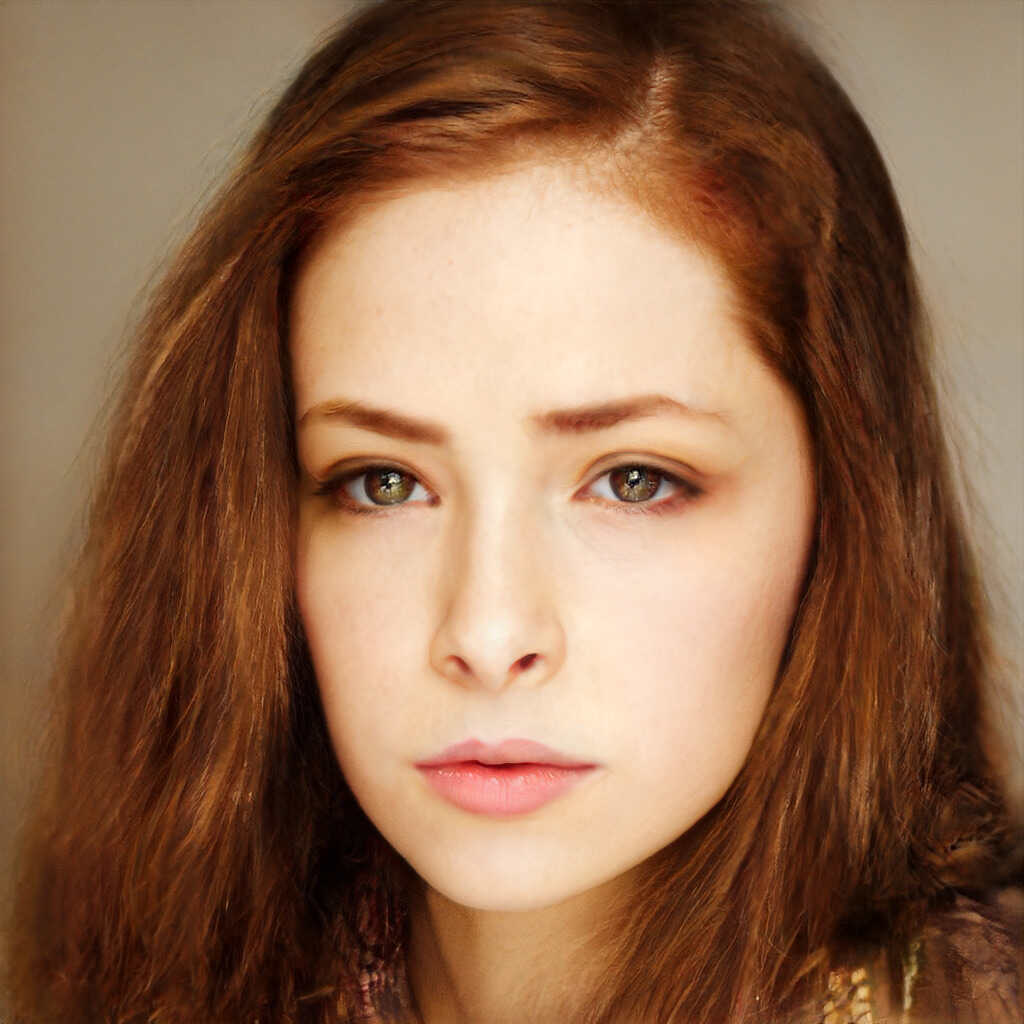} &
        \includegraphics[width=0.07\linewidth]{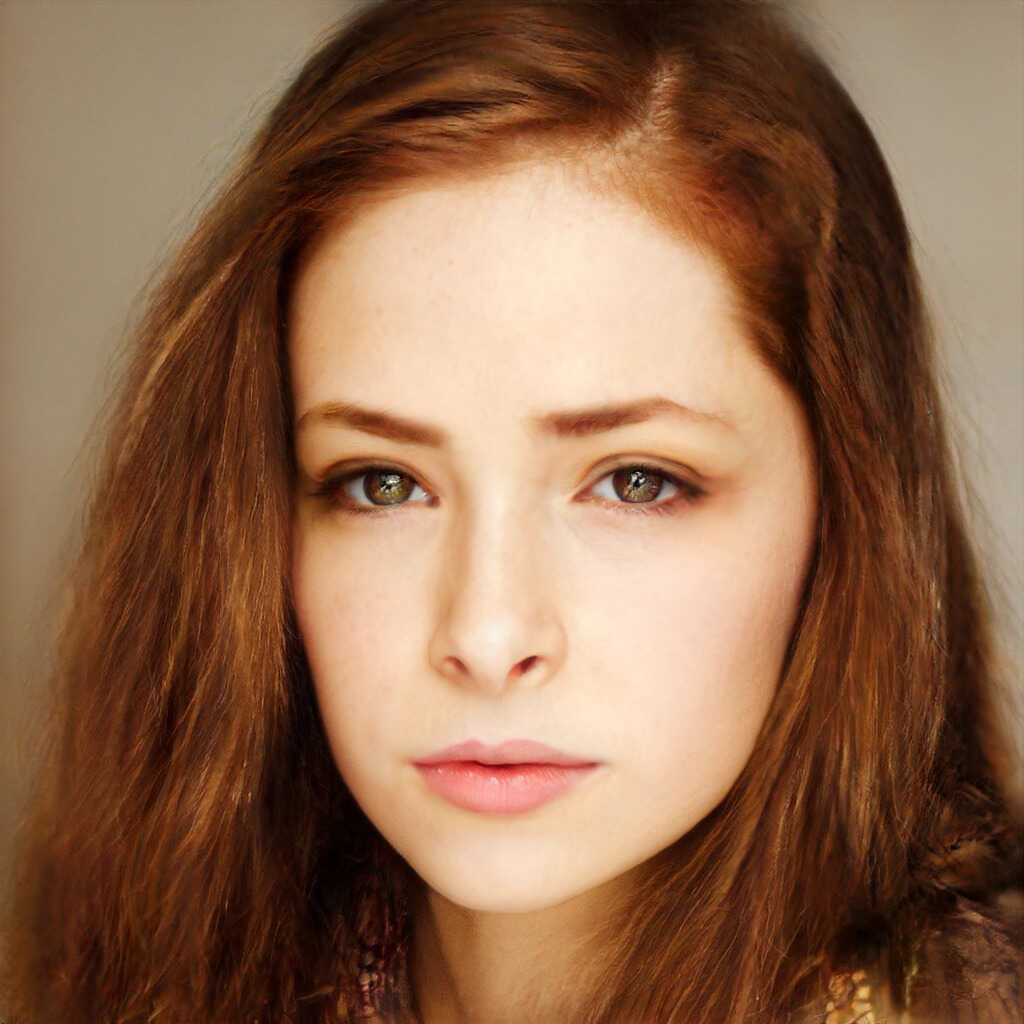}   &
        \includegraphics[width=0.07\linewidth]{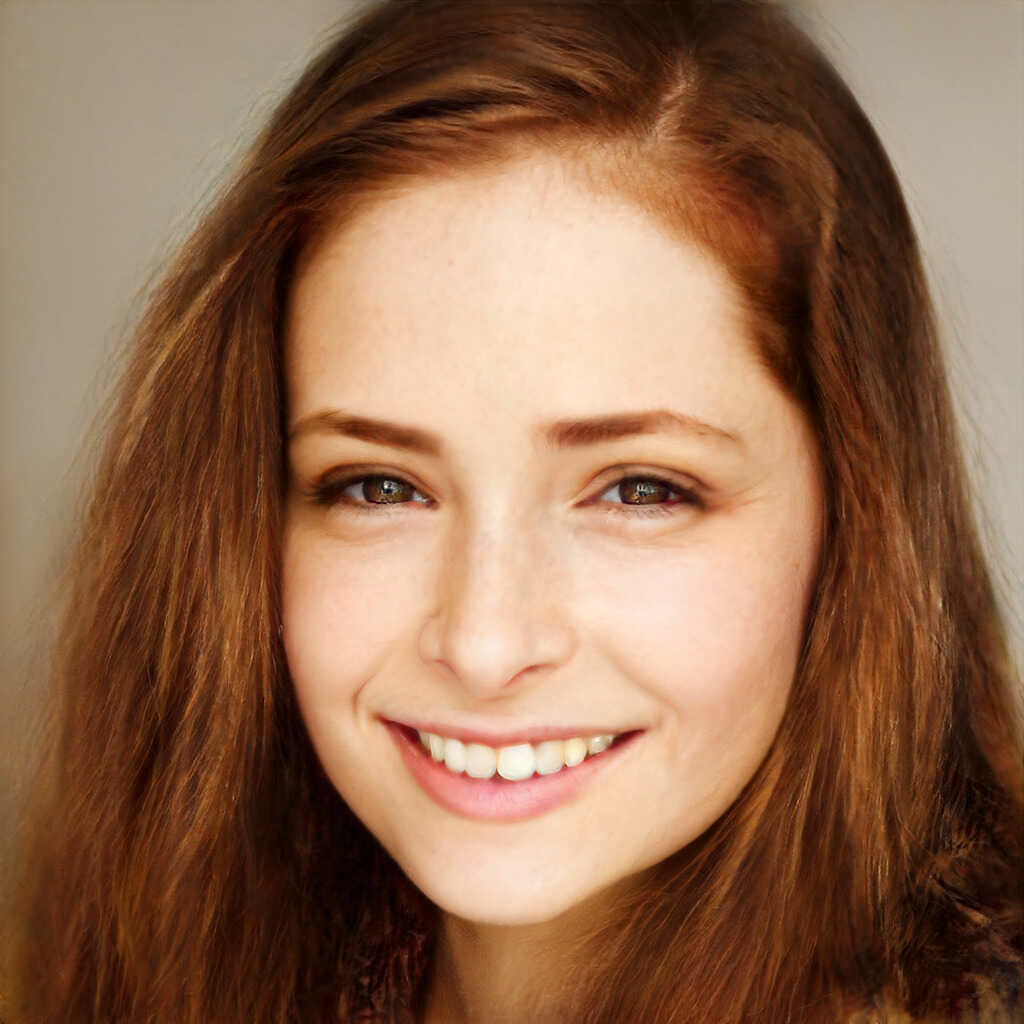}    &
        \includegraphics[width=0.07\linewidth]{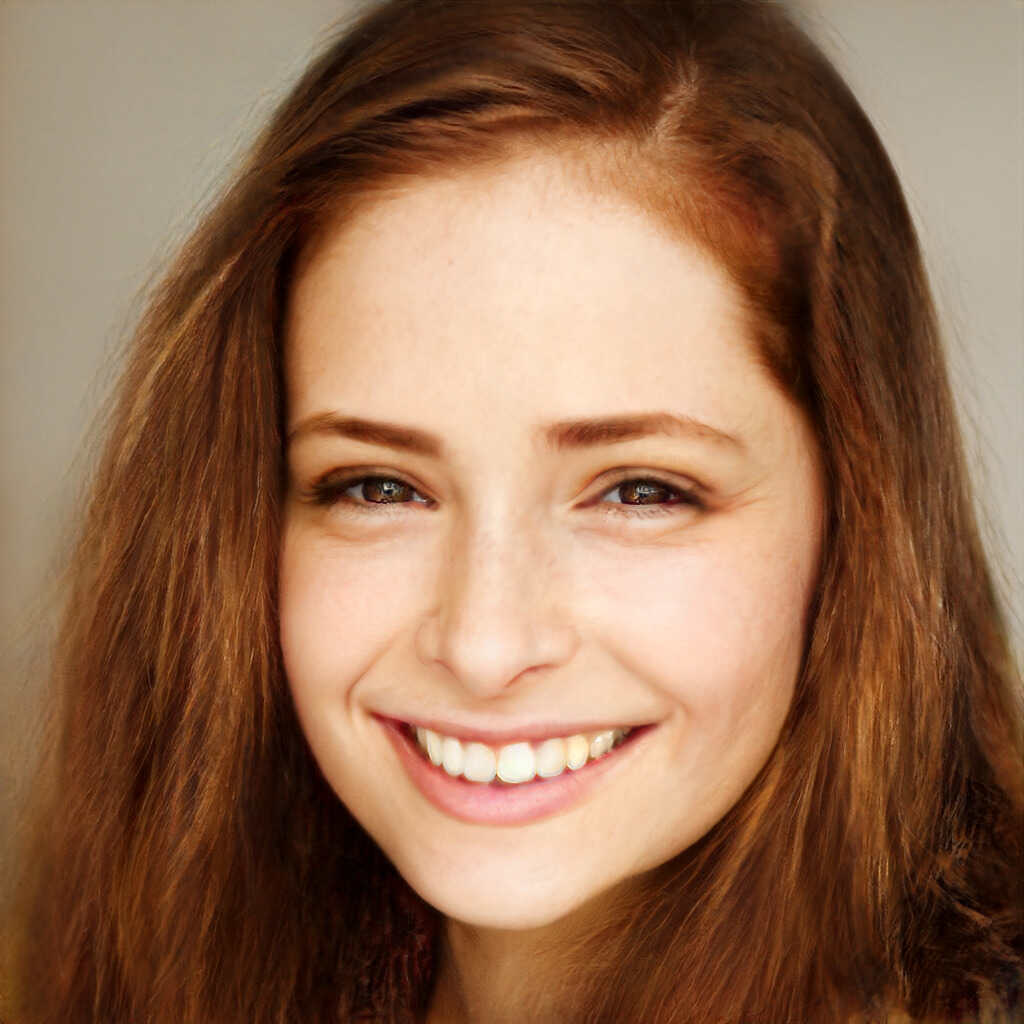}  &
        \includegraphics[width=0.07\linewidth]{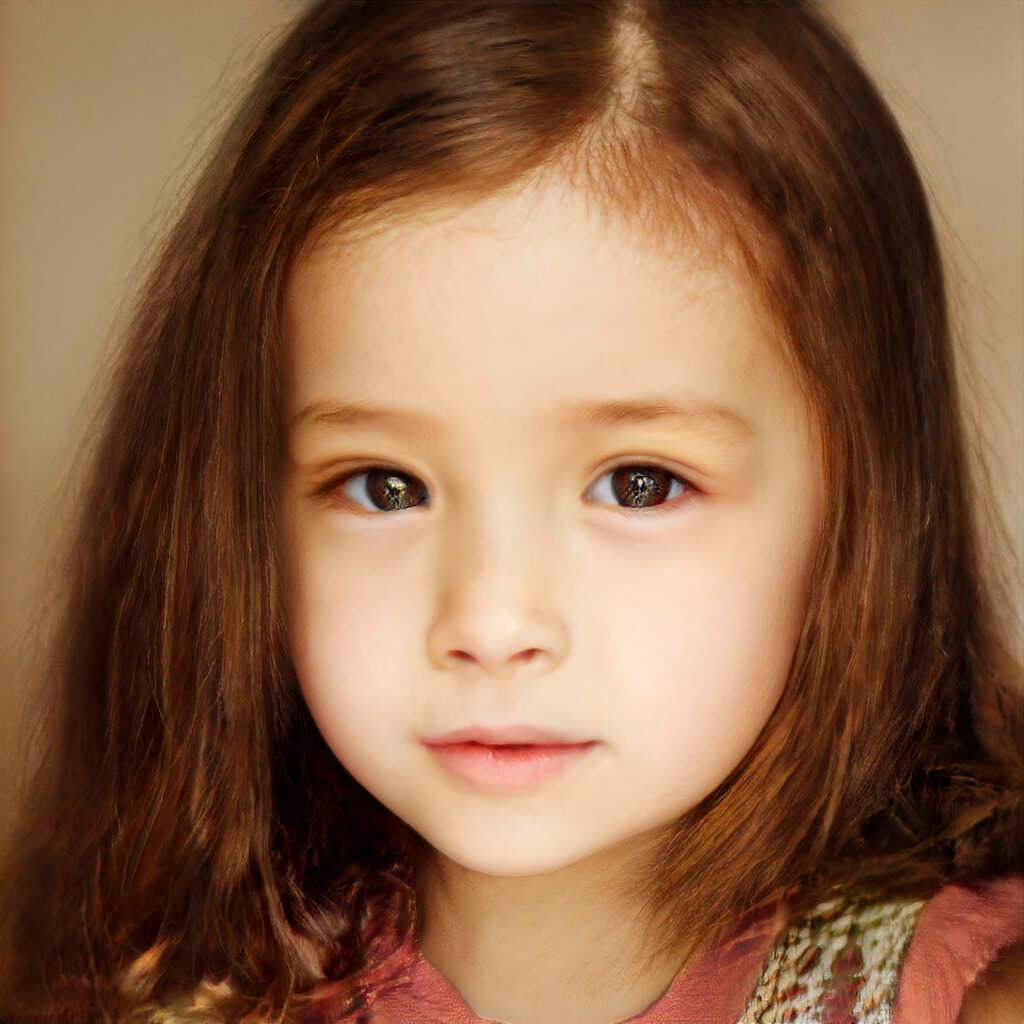}     &
        \includegraphics[width=0.07\linewidth]{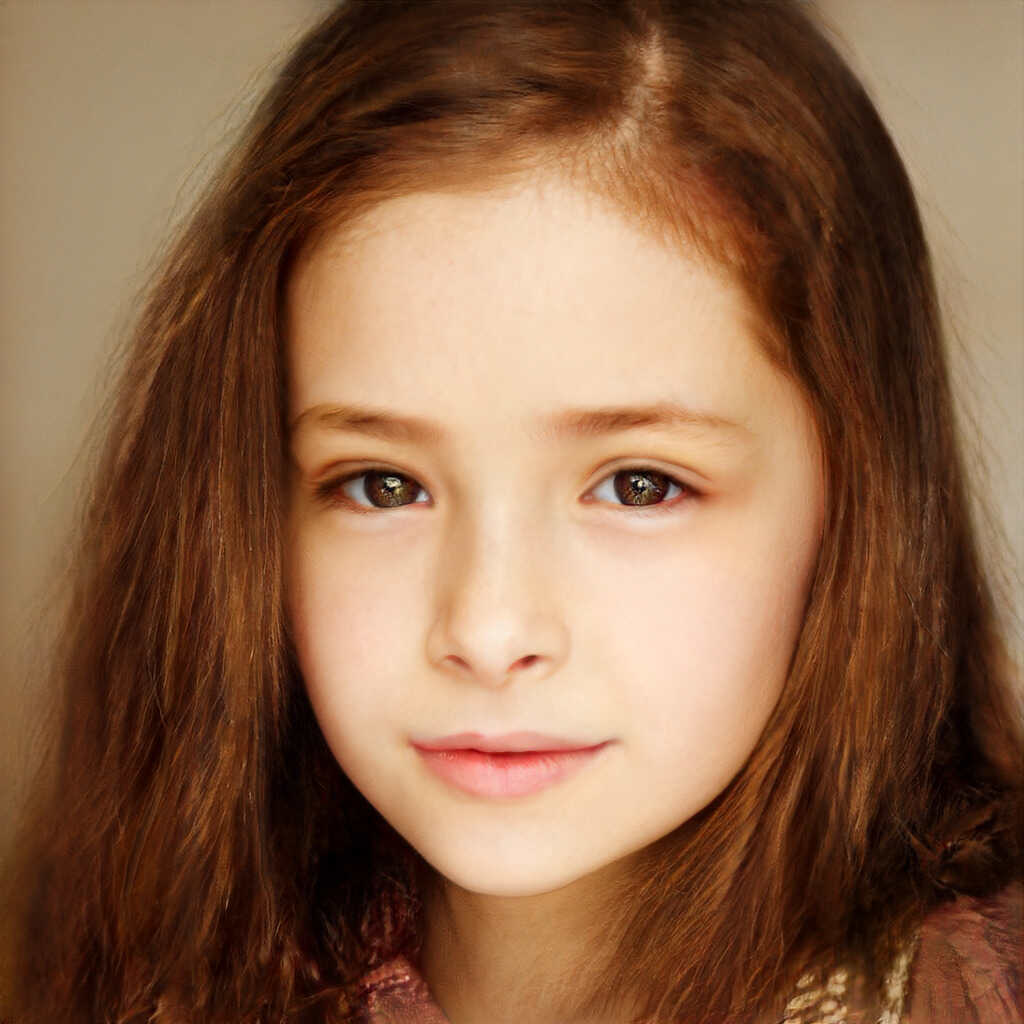}     &
        \includegraphics[width=0.07\linewidth]{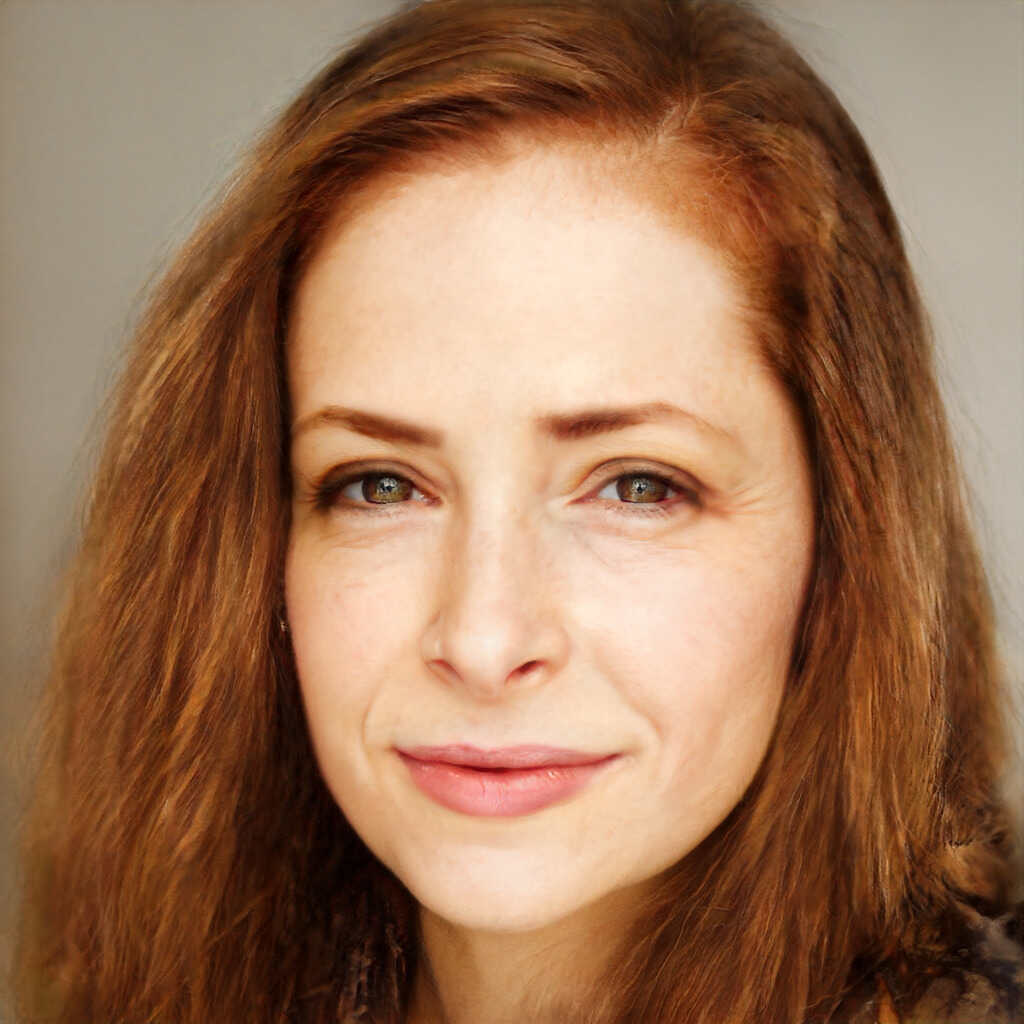}      &
        \includegraphics[width=0.07\linewidth]{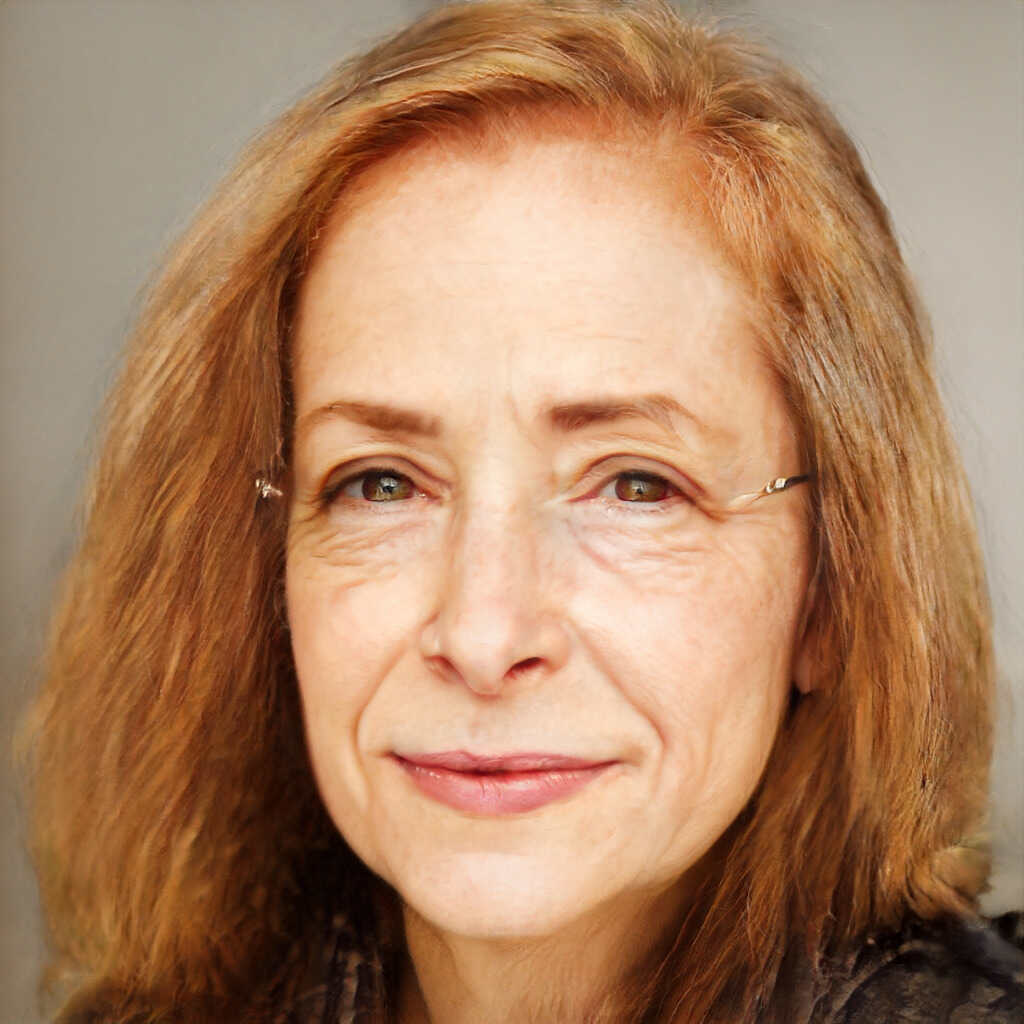}                                                                                                               \\
        \includegraphics[width=0.07\linewidth]{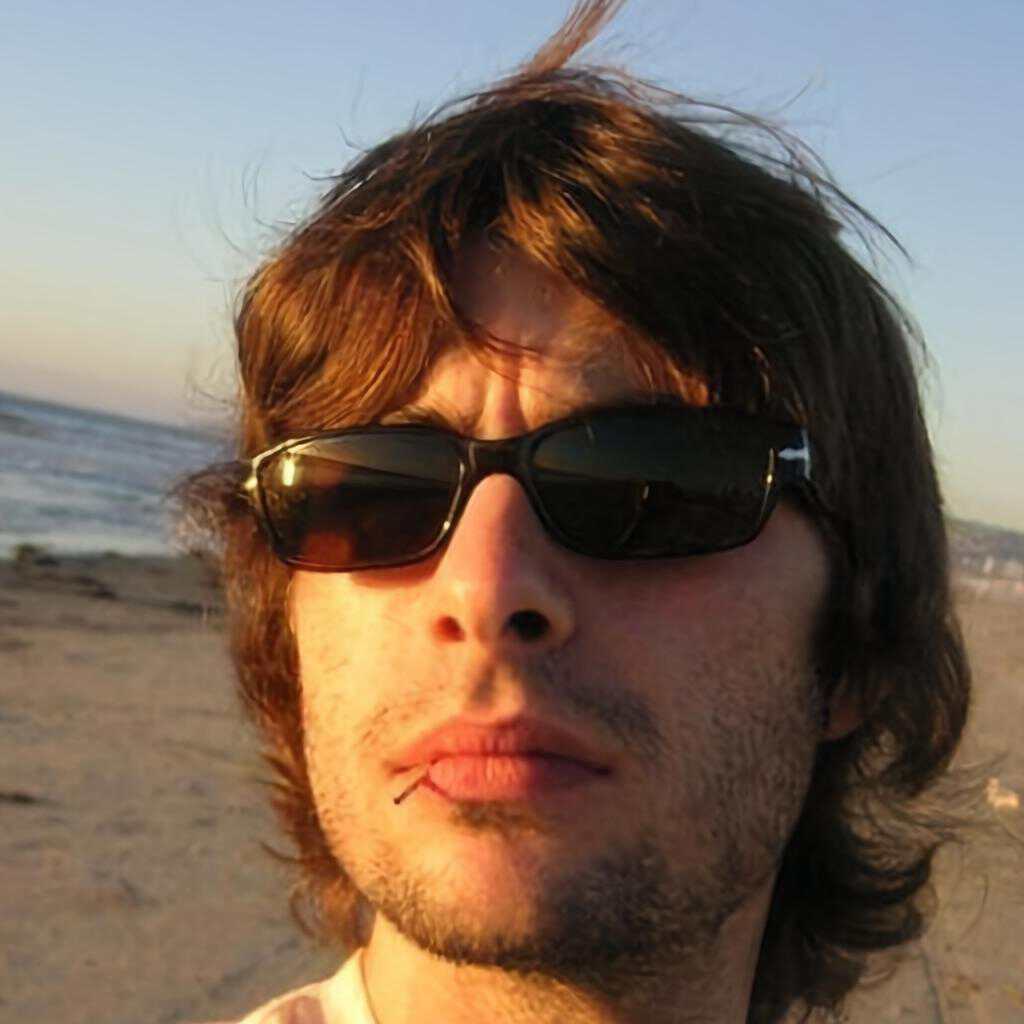}                  &
        \includegraphics[width=0.07\linewidth]{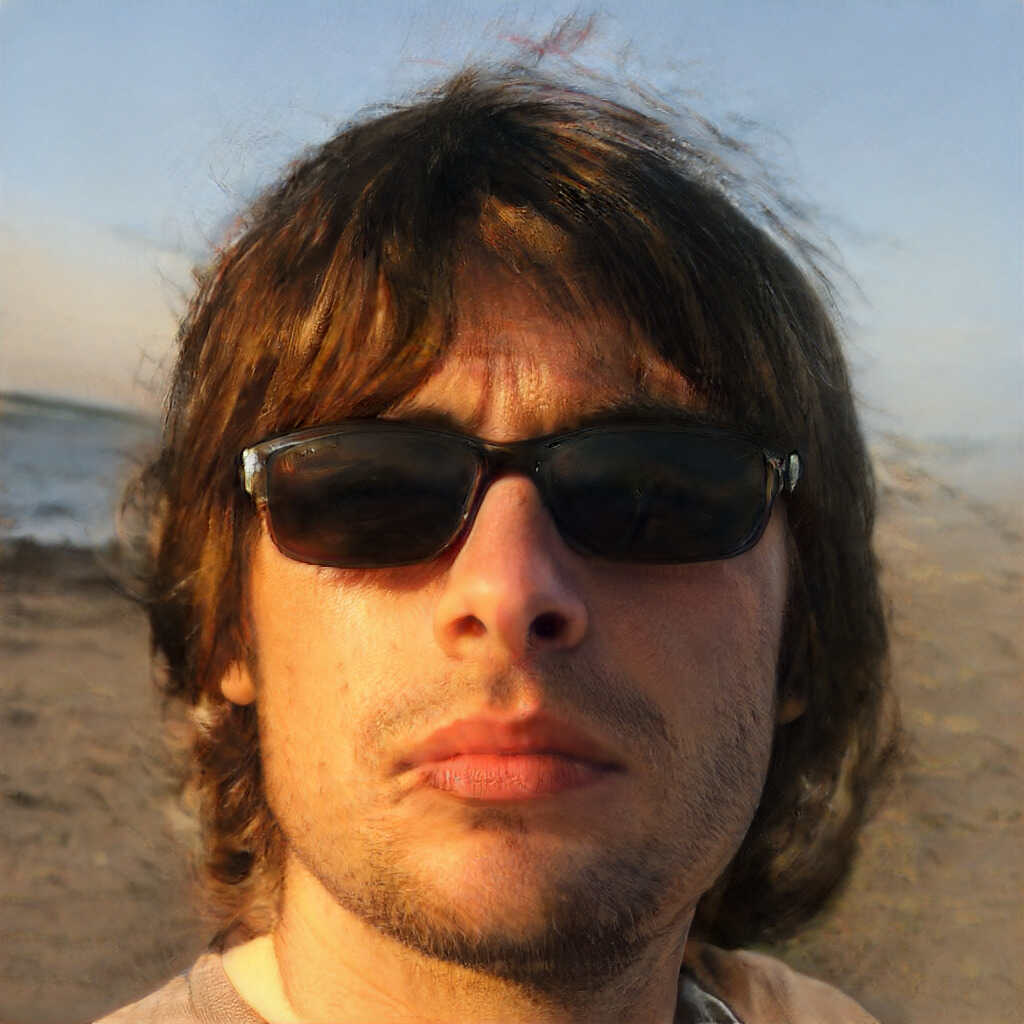}    &
        \includegraphics[width=0.07\linewidth]{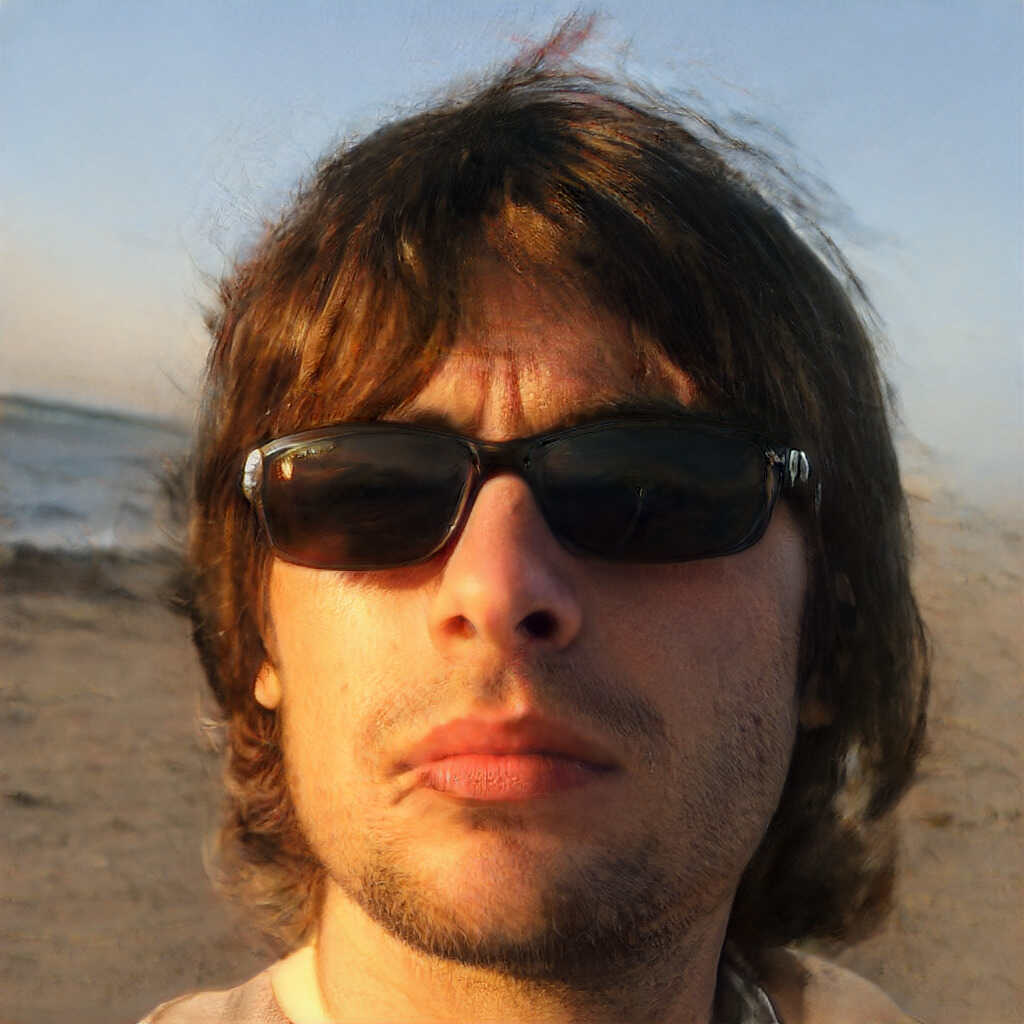}    &
        \includegraphics[width=0.07\linewidth]{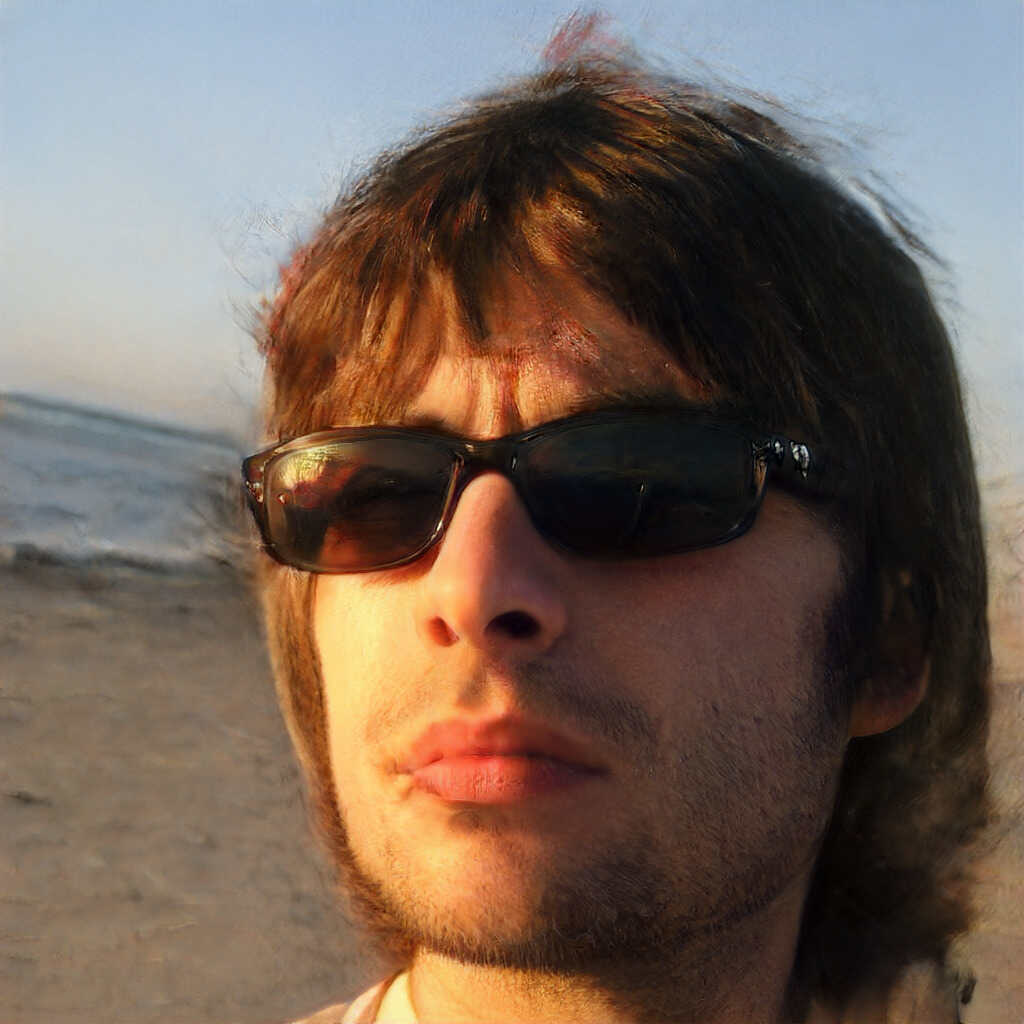}     &
        \includegraphics[width=0.07\linewidth]{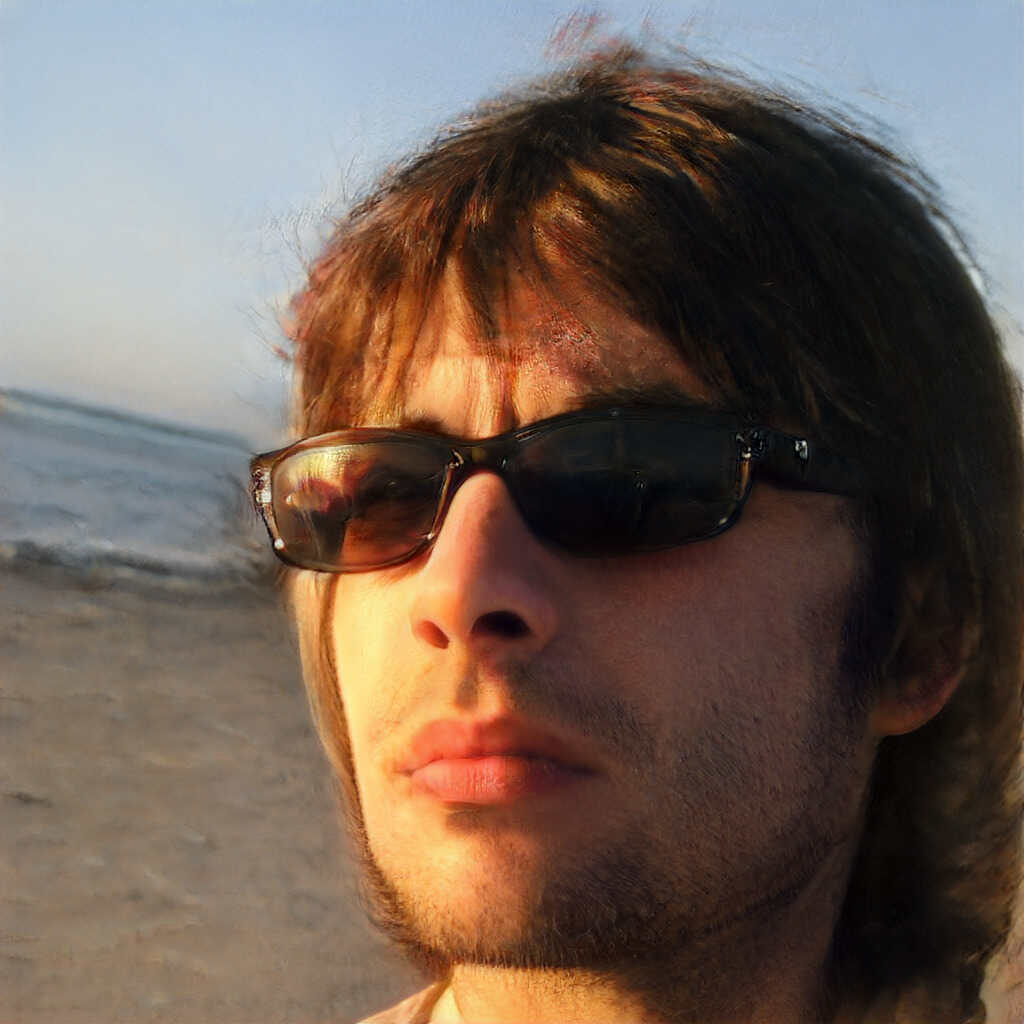}     &
        \includegraphics[width=0.07\linewidth]{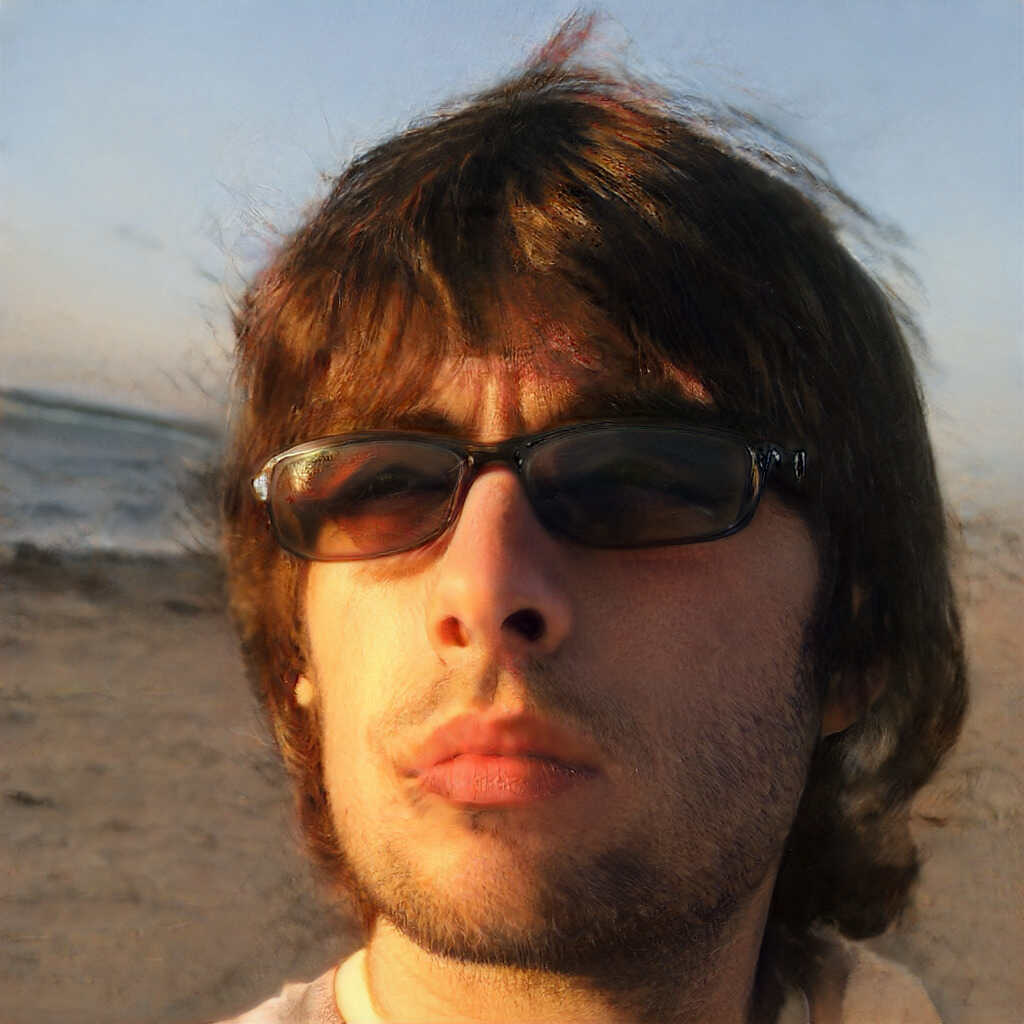} &
        \includegraphics[width=0.07\linewidth]{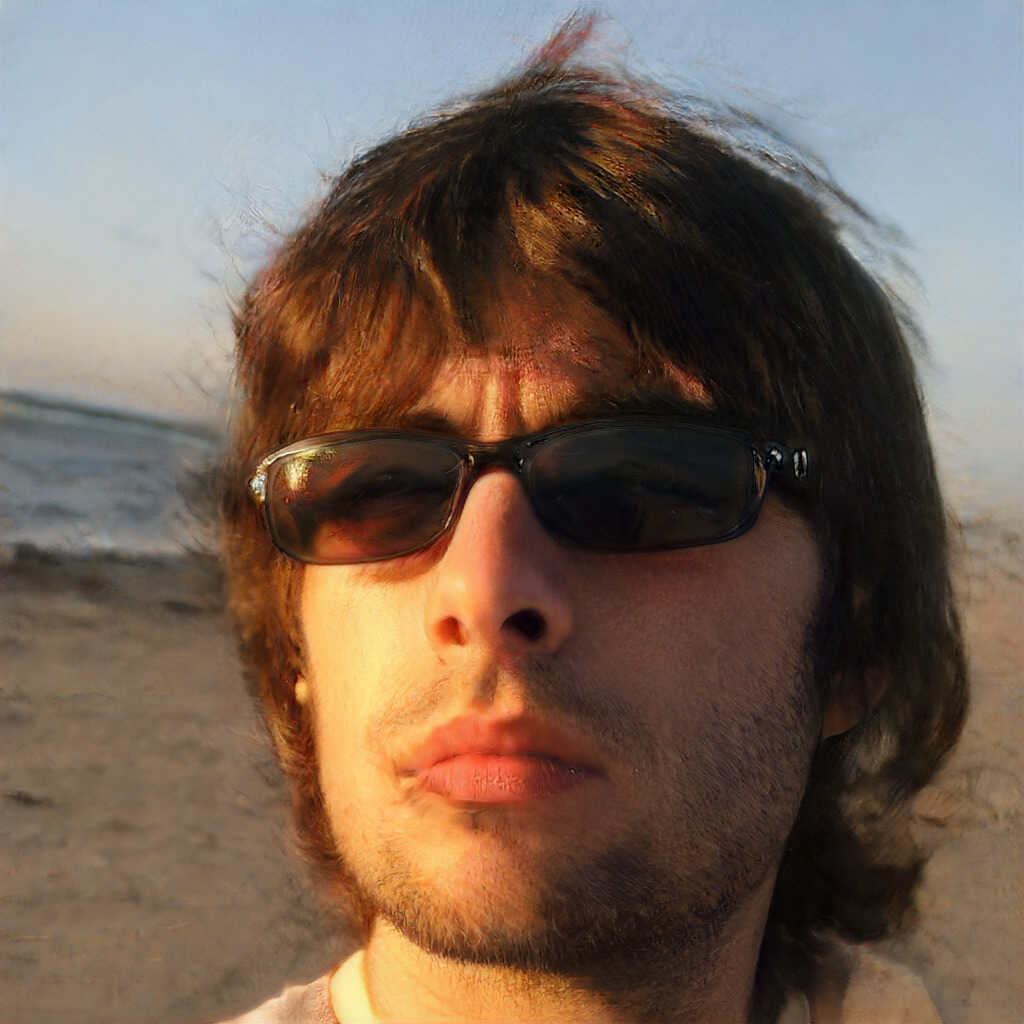}   &
        \includegraphics[width=0.07\linewidth]{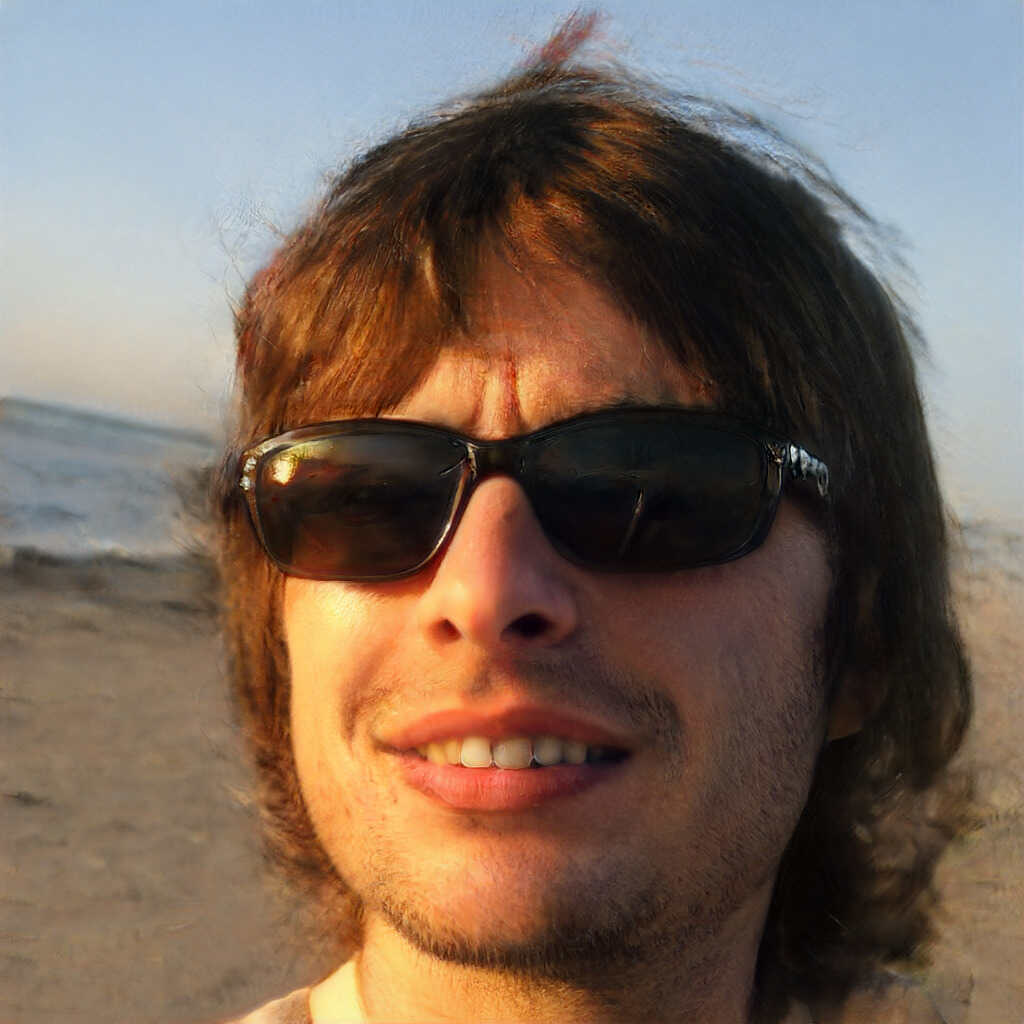}    &
        \includegraphics[width=0.07\linewidth]{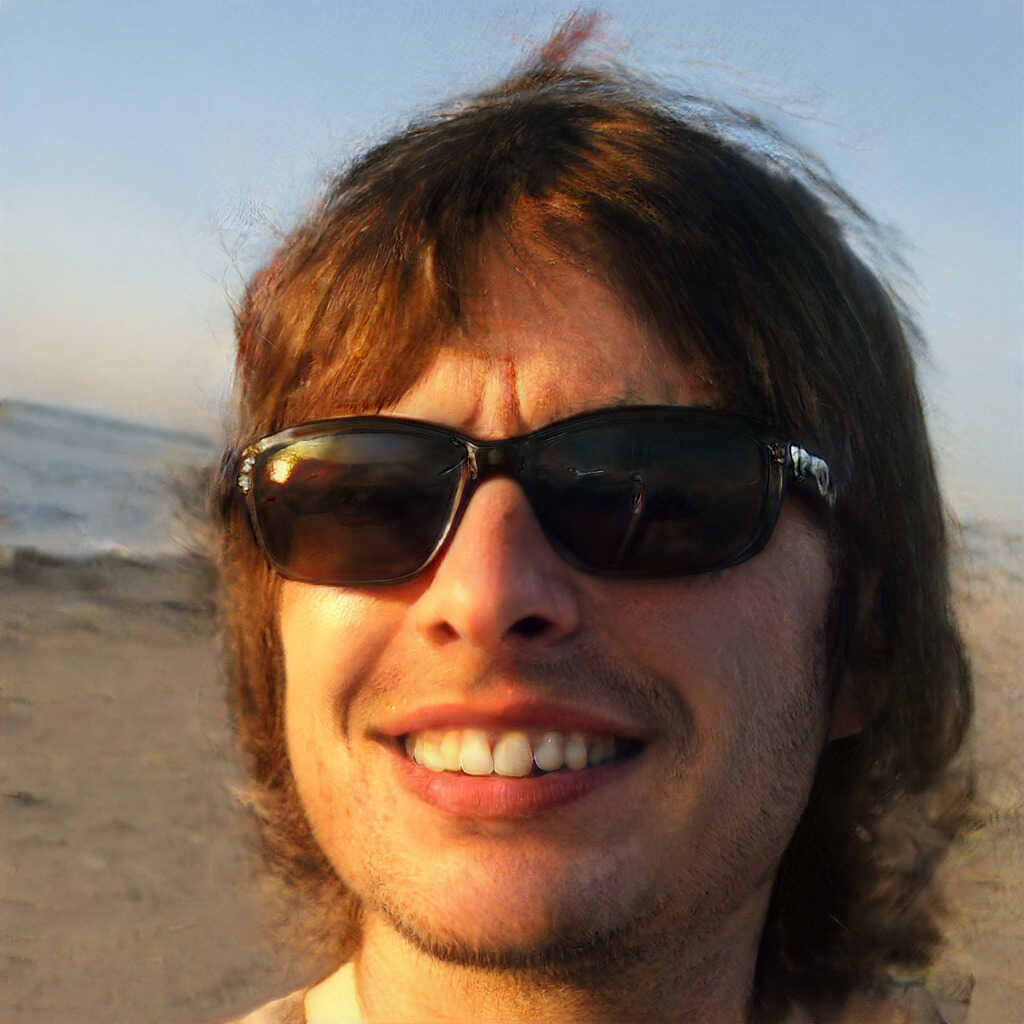}  &
        \includegraphics[width=0.07\linewidth]{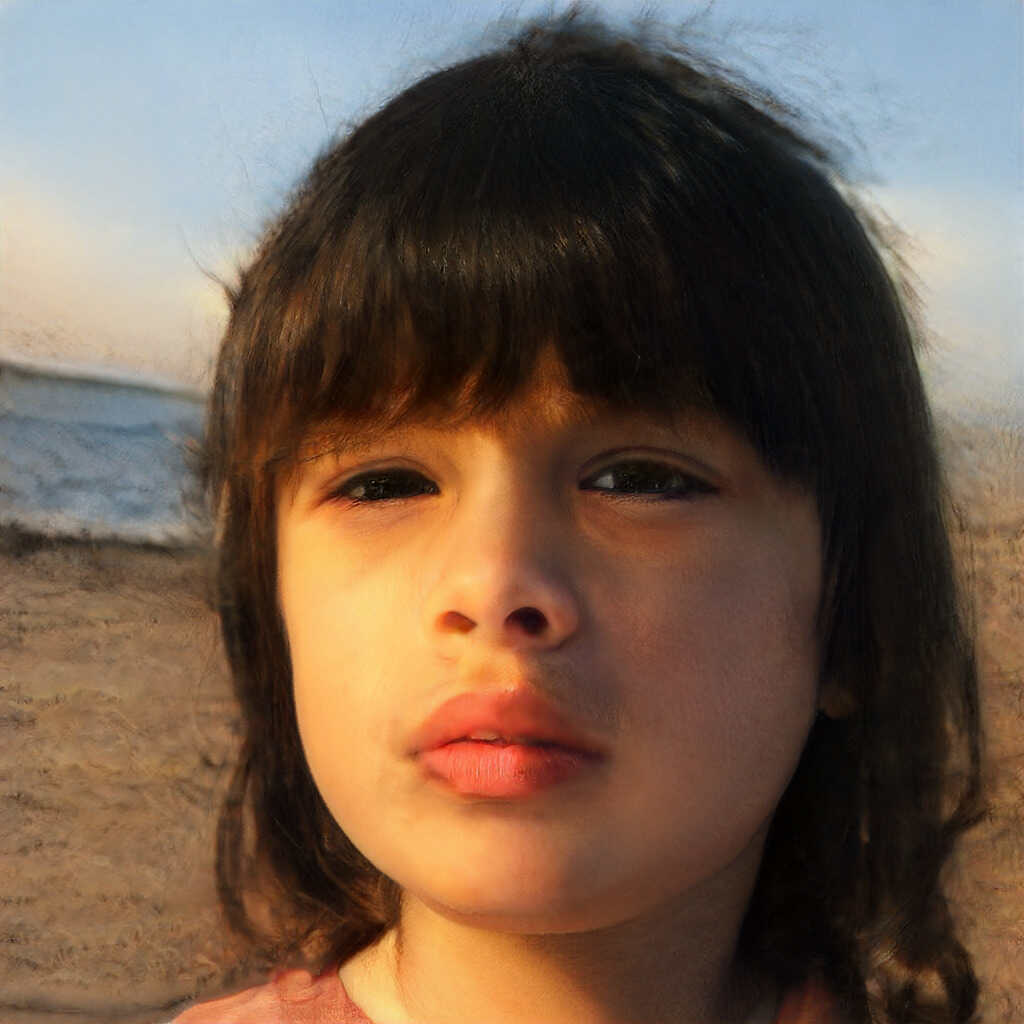}     &
        \includegraphics[width=0.07\linewidth]{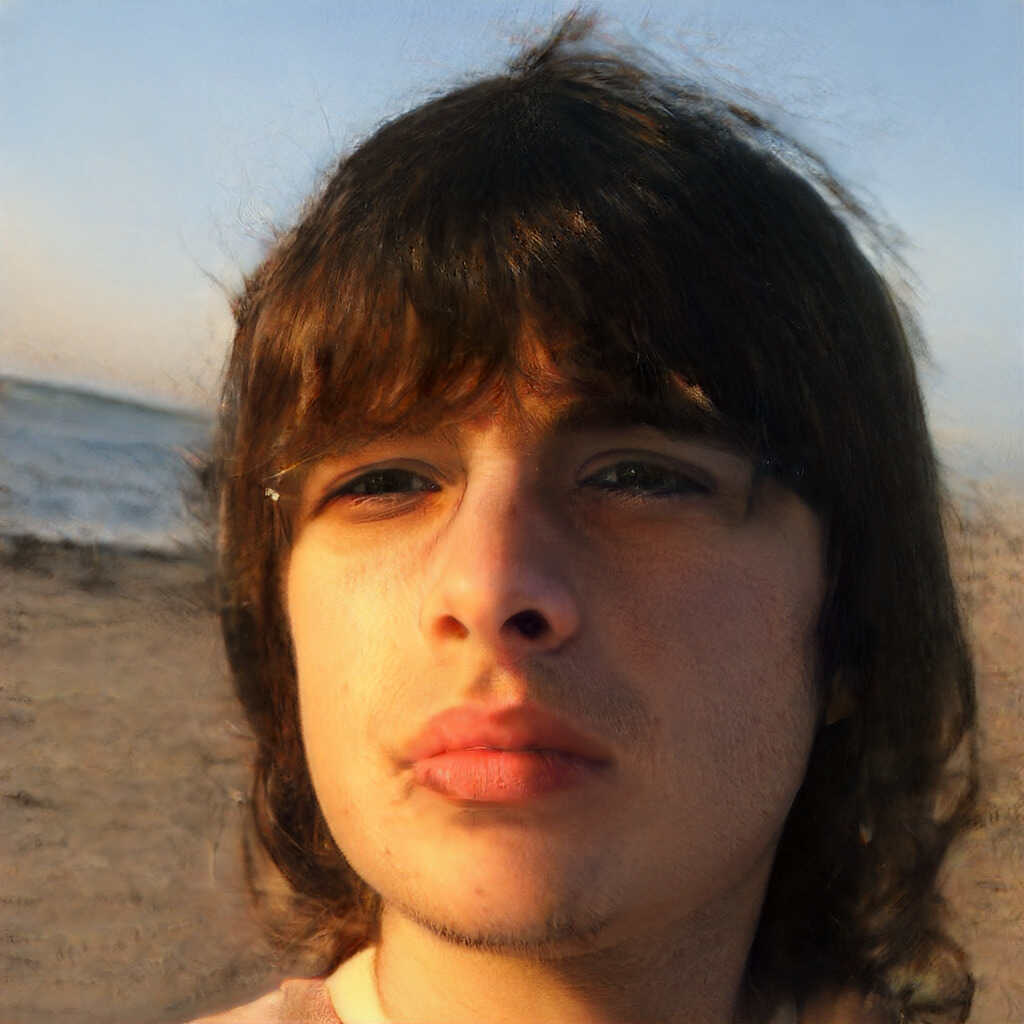}     &
        \includegraphics[width=0.07\linewidth]{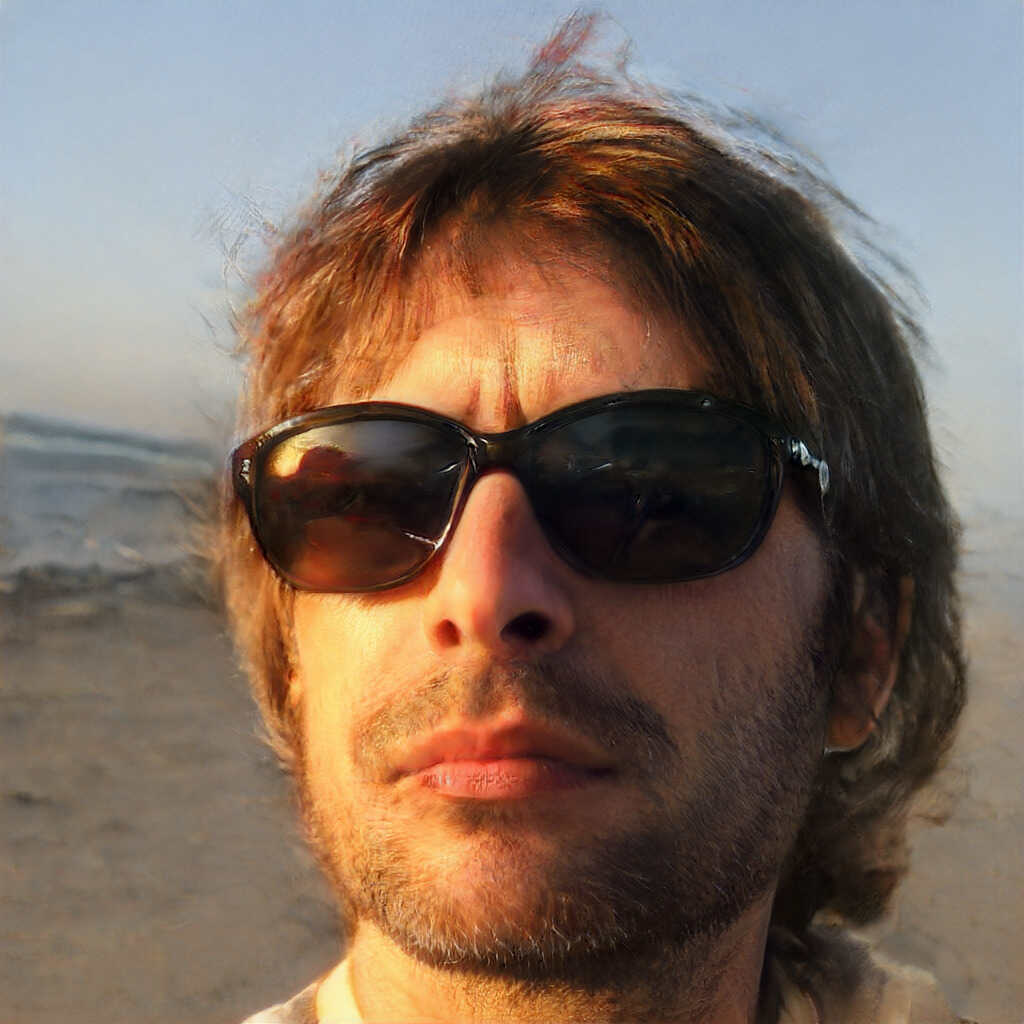}      &
        \includegraphics[width=0.07\linewidth]{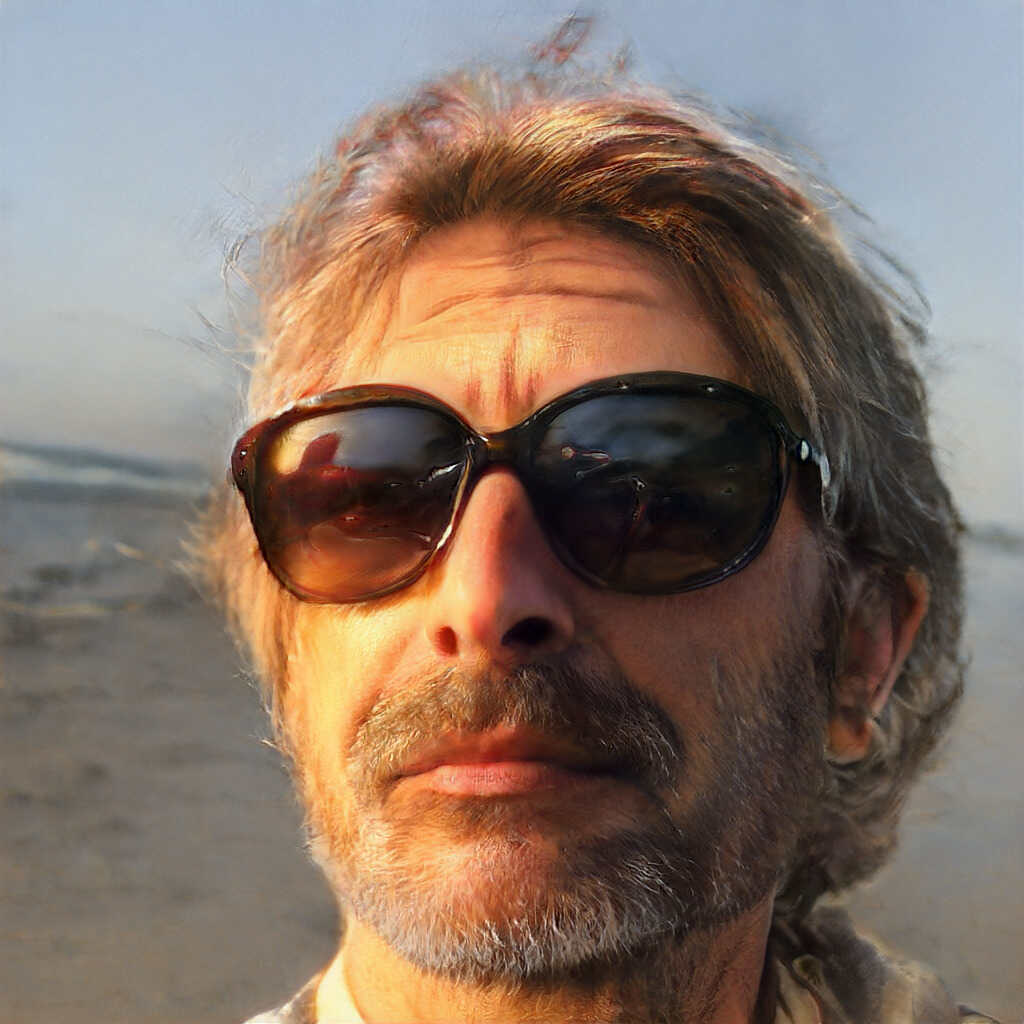}                                                                                                               
    \end{tabular}
    \caption{Pose, smile and age editing using InterFaceGAN~\cite{shen2020interpreting} on CelebA-HQ.}
    \label{fig:edit_celeba_hq_pose_smile_edit}
\end{figure}

\begin{figure}[ht!]
\centering
\footnotesize
\renewcommand{\arraystretch}{0.0}
\begin{tabular}{@{}c@{\hspace{2pt}}c@{\hspace{2pt}}:c@{}c@{}c@{}c@{}}
  \rotatebox{90}{\hspace{6pt} -Blue Skies}                                                                              &
  \includegraphics[width=0.18\linewidth]{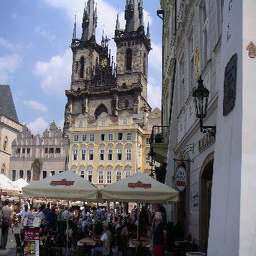}                                           &
  \includegraphics[width=0.18\linewidth]{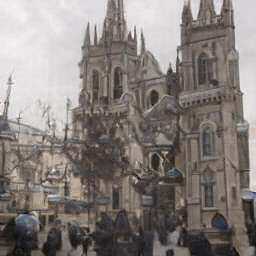}         &
  \includegraphics[width=0.18\linewidth]{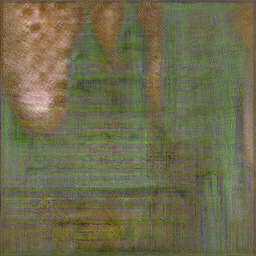} &
  \includegraphics[width=0.18\linewidth]{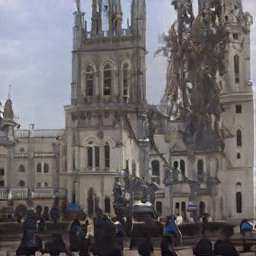} &
  \includegraphics[width=0.18\linewidth]{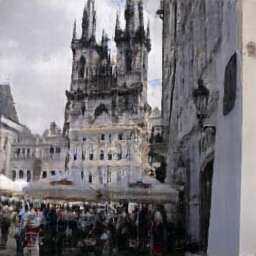}                                                                                                                               \\
  \rotatebox{90}{\hspace{8pt} -Clouds}                                                                                 &
  \includegraphics[width=0.18\linewidth]{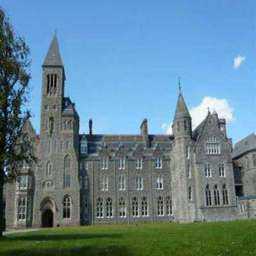}                                           &
  \includegraphics[width=0.18\linewidth]{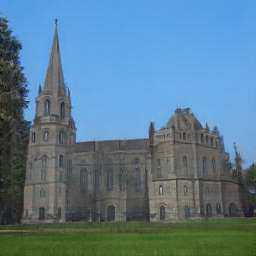}              &
  \includegraphics[width=0.18\linewidth]{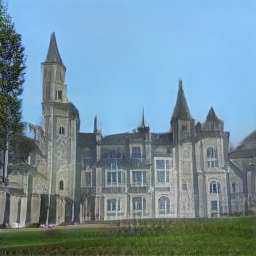}      &
  \includegraphics[width=0.18\linewidth]{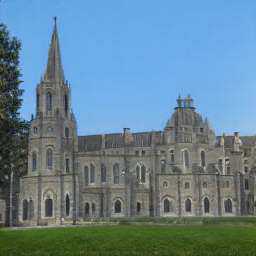}      &
  \includegraphics[width=0.18\linewidth]{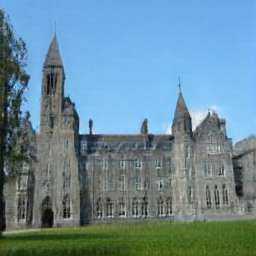}                                                                                                                                    \\
  \rotatebox{90}{\hspace{8pt} +Clouds}                                                                                 &
  \includegraphics[width=0.18\linewidth]{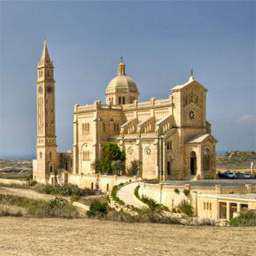}                                           &
  \includegraphics[width=0.18\linewidth]{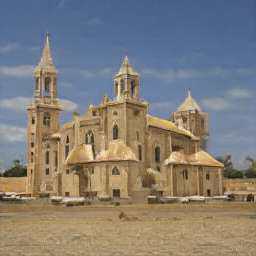}            &
  \includegraphics[width=0.18\linewidth]{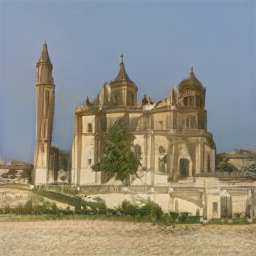}    &
  \includegraphics[width=0.18\linewidth]{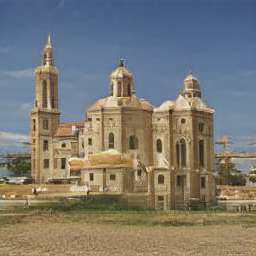}    &
  \includegraphics[width=0.18\linewidth]{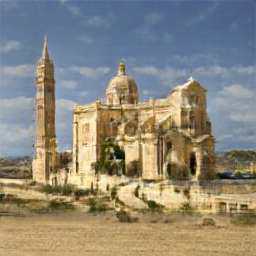}                                                                                                                                  \\
  \rotatebox{90}{\hspace{8pt} -Vibrant}                                                                                &
  \includegraphics[width=0.18\linewidth]{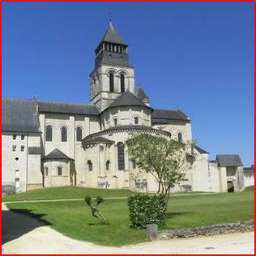}                                           &
  \includegraphics[width=0.18\linewidth]{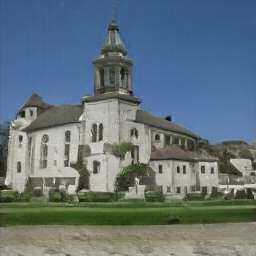}            &
  \includegraphics[width=0.18\linewidth]{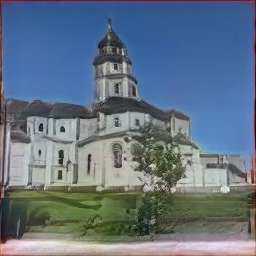}    &
  \includegraphics[width=0.18\linewidth]{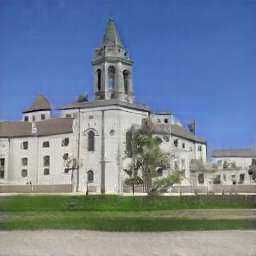}    &
  \includegraphics[width=0.18\linewidth]{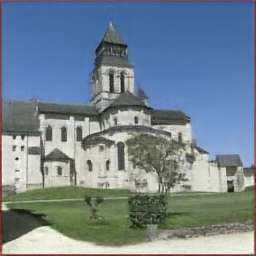}                                                                                                                                  \\
  \rotatebox{90}{\hspace{6pt} +Vibrant}                                                                                &
  \includegraphics[width=0.18\linewidth]{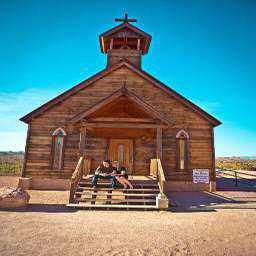}                                           &
  \includegraphics[width=0.18\linewidth]{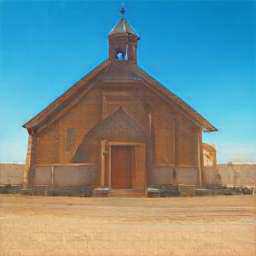}            &
  \includegraphics[width=0.18\linewidth]{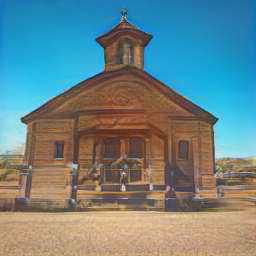}    &
  \includegraphics[width=0.18\linewidth]{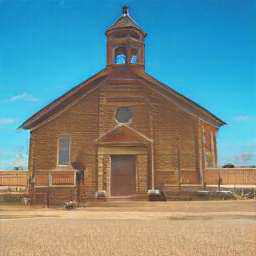}    &
  \includegraphics[width=0.18\linewidth]{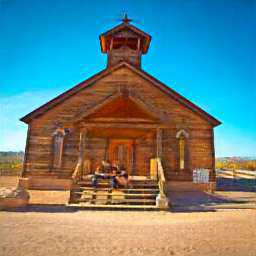}                                                                                                                                  \\ [5pt]
                                                                                                                        & GT & e4e & $\text{ReStyle}_{\text{pSp}}$ & $\text{ReStyle}_{\text{e4e}}$ & {Ours} \\
\end{tabular}
\caption{%
  Editing quality on LSUN Church using GANSpace~\cite{harkonen2020ganspace}}
\label{fig:additional_lsun_church_editing_baseline}
\end{figure}

\begin{figure}[ht!]
    \centering
    \footnotesize
    \begin{tabular}{@{}c@{\hspace{5pt}}c:c@{}c@{}c@{}c@{}c@{}c@{}}
        \rotatebox{90}{+Color}                                                               &
        \includegraphics[width=0.142\linewidth]{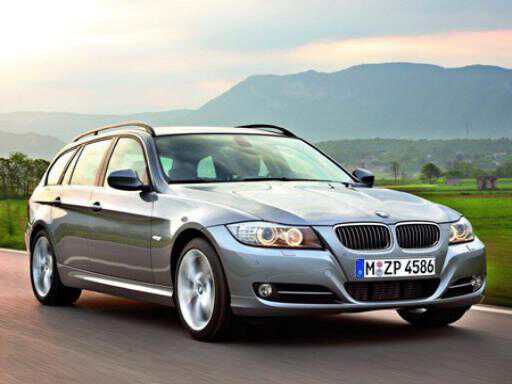}             &
        \includegraphics[width=0.142\linewidth]{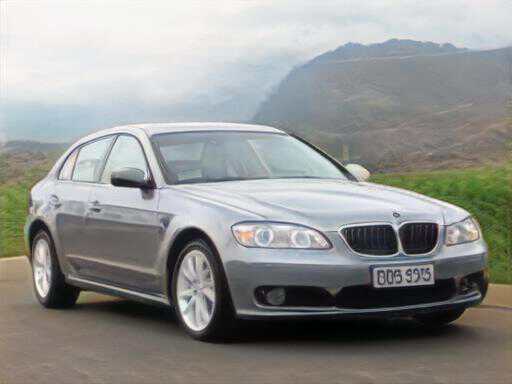}         &
        \includegraphics[width=0.142\linewidth]{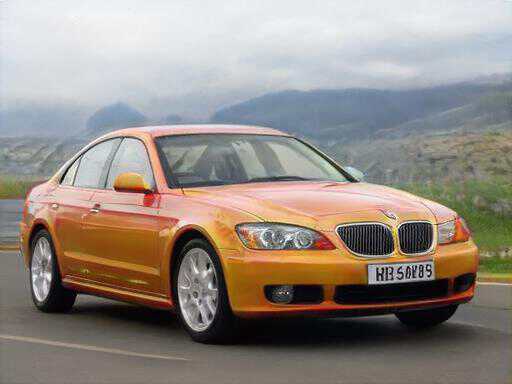}         &
        \includegraphics[width=0.142\linewidth]{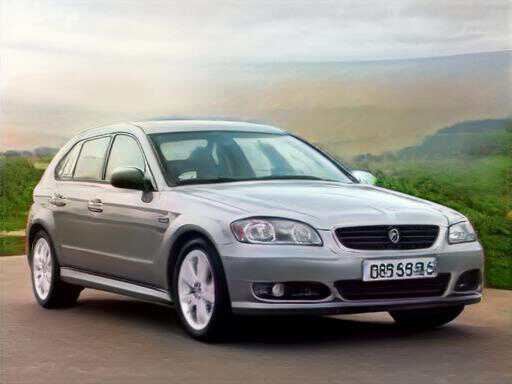} &
        \includegraphics[width=0.142\linewidth]{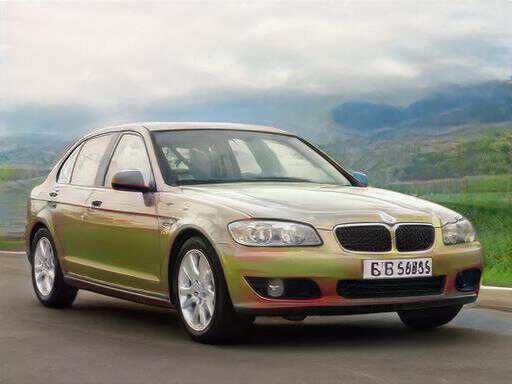} &
        \includegraphics[width=0.142\linewidth]{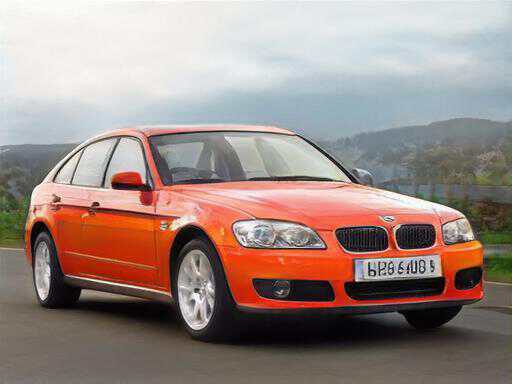}  &
        \includegraphics[width=0.142\linewidth]{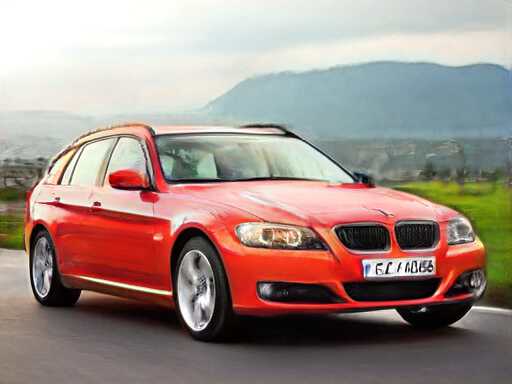}                                                                                                                                                           \\
        \rotatebox{90}{+Cube}                                                                &
        \includegraphics[width=0.142\linewidth]{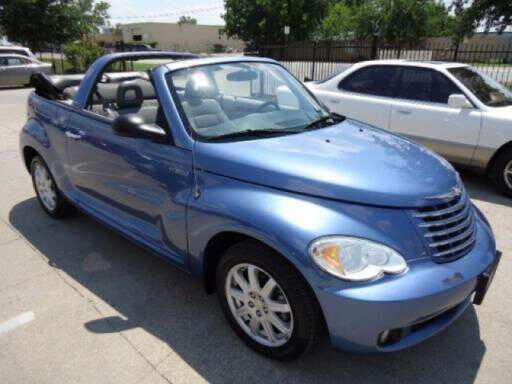}           &
        \includegraphics[width=0.142\linewidth]{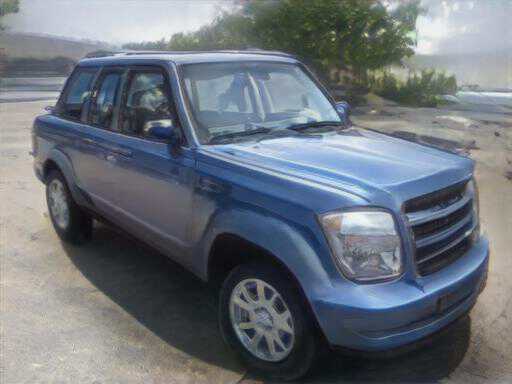}          &
        \includegraphics[width=0.142\linewidth]{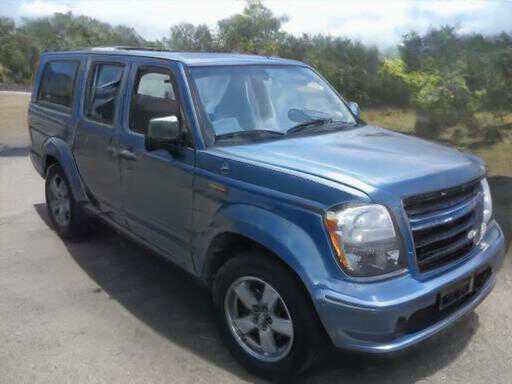}          &
        \includegraphics[width=0.142\linewidth]{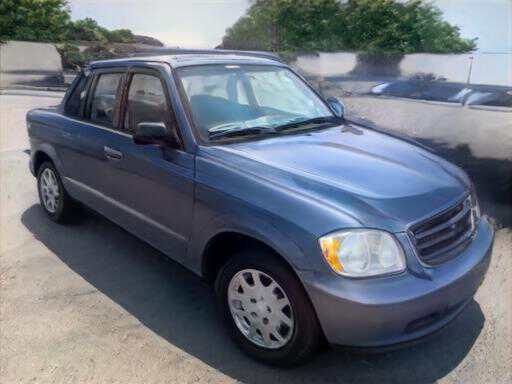}  &
        \includegraphics[width=0.142\linewidth]{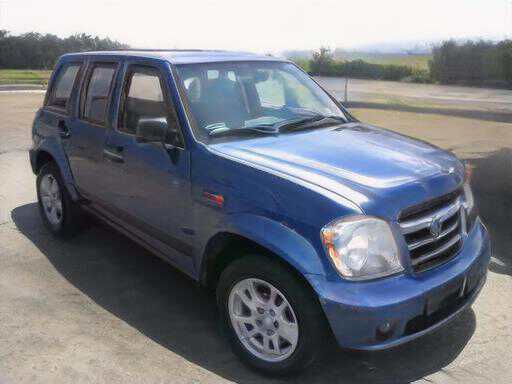}  &
        \includegraphics[width=0.142\linewidth]{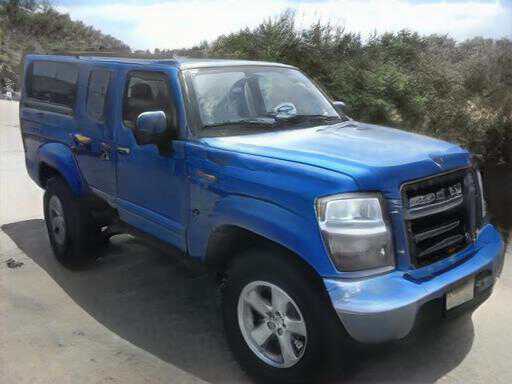}   &
        \includegraphics[width=0.142\linewidth]{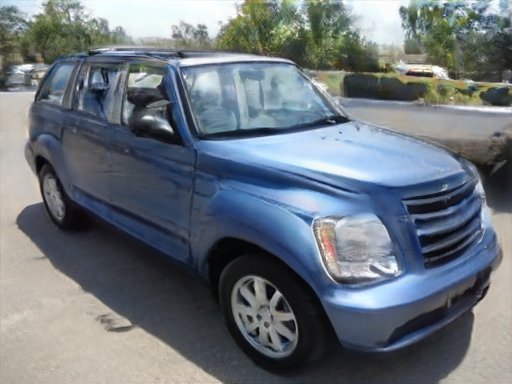}                                                                                                                                                           \\
                                                                                                           & GT & pSp & e4e &  $\text{ReStyle}_{\text{pSp}}$ & $\text{ReStyle}_{\text{e4e}}$ & HyperStyle & Ours \\
    \end{tabular}
    \caption{%
        Editing quality on the Stanford Cars using GANSpace~\cite{harkonen2020ganspace}}
    \label{fig:standford_cars_editing_baseline}
\end{figure}

\noindent
{\bf Editing Quality \quad} The ability of our method to retain the original editing directions of StyleGAN is seen in Fig.~\ref{fig:edit_celeba_hq_pose_smile_edit} for the CelebA-HQ dataset,
editing directions and tools were taken from InterFaceGAN~\cite{shen2020interpreting}.
A visual comparison to other approaches for Stanford Cars and LSUN Church datasets appears in Fig.~\ref{fig:standford_cars_editing_baseline} and Fig.~\ref{fig:additional_lsun_church_editing_baseline}, respectively. For these datasets, the editing directions were taken from GANSpace~\cite{harkonen2020ganspace}. Evidently, unlike other approaches, our method is able to retain the editing direction without harming image quality, for instance, on the automotive images, it is the only approach that modifies the color uniformly, whereas in the outdoor image in Fig.~\ref{fig:additional_lsun_church_editing_baseline}, it does not introduce artifacts or change the image's content.

\begin{figure}[t]
    \centering
    \normalsize
    \renewcommand{\arraystretch}{0.0}
    \begin{tabular}{@{}c@{}c@{}c@{}c@{}}
        \rotatebox{90}{\hspace{30pt} CelebA-HQ}                                                                     &
        \includegraphics[width=0.20\linewidth]{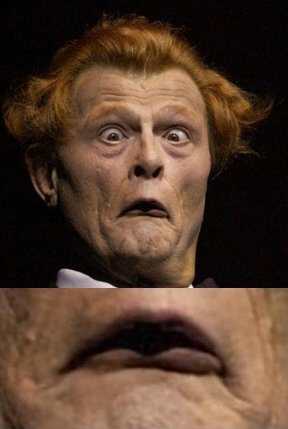}                &
        \includegraphics[width=0.20\linewidth]{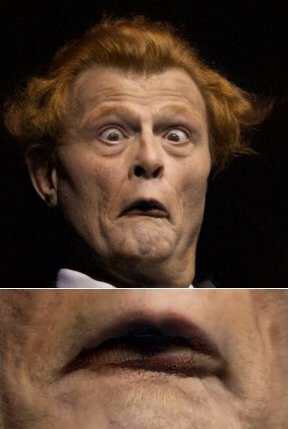} &
        \includegraphics[width=0.20\linewidth]{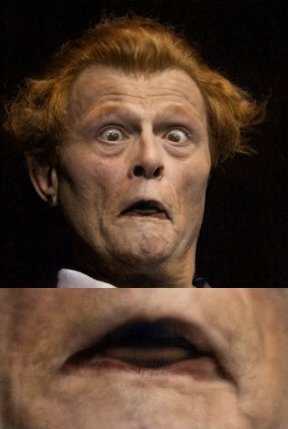}                \\
        \rotatebox{90}{\hspace{30pt} AFHQ-Wild}                                                                     &
        \includegraphics[width=0.20\linewidth]{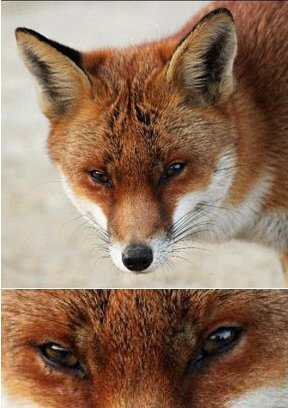}                  &
        \includegraphics[width=0.20\linewidth]{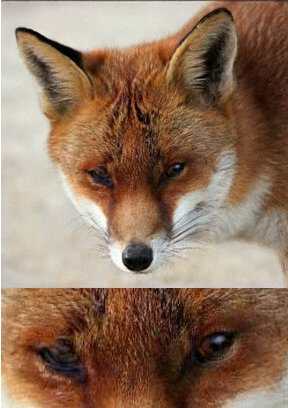}   &
        \includegraphics[width=0.20\linewidth]{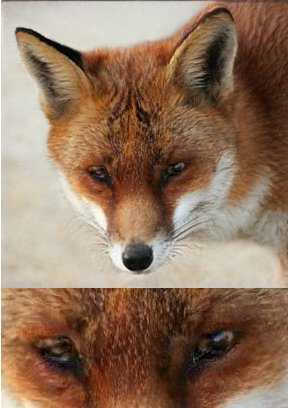}                  \\ [2pt]
                                                                                                                    & (a) & (b) & (c) \\
    \end{tabular}
    \caption{%
        Visual comparison of the effect of training with $\mathcal{L}_{FP}$. (a) Ground truth (b) with $\mathcal{L}_{FP}$ (c) without $\mathcal{L}_{FP}$.}
    \label{fig:parsing_ablation}
\end{figure}

\noindent
{\bf Face-Parsing Loss\quad} Fig.~\ref{fig:parsing_ablation} shows the benefits of applying the face-parsing loss to images from the CelebA-HQ and AFHQ-Wild datasets. As can be seen, for the facial image, the reconstructed image retains the fine details in the lips when the parsing loss is applied.
Furthermore, even though the parsing loss utilizes a pre-trained network, trained on the task of human facial semantic segmentation, it also improves the reconstruction quality for other non-human facial domains.

\begin{figure}[ht!]
    \centering
    \normalsize
    \renewcommand{\arraystretch}{0.0}
    \begin{tabular}{@{}c@{}c@{}}
        \includegraphics[width=0.3\linewidth]{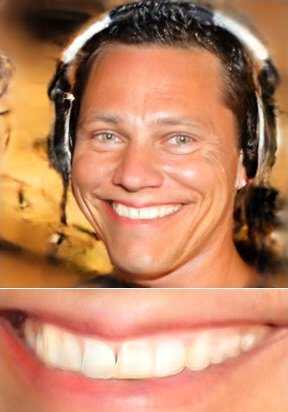} &
        \includegraphics[width=0.3\linewidth]{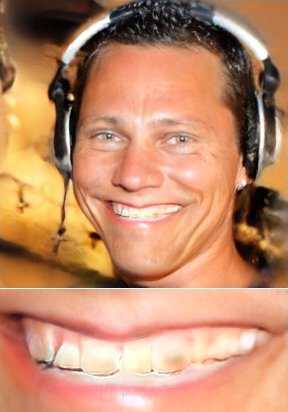}    \\
        (a)                                                                                      & (b) \\
    \end{tabular}
    \caption{%
        Image reconstruction. (a) with localization. (b) without localization
    }
    \label{fig:localization_ablation}
\end{figure}

\noindent
{\bf Localization Regularization\quad}
The localization term prevents model drifting, which is often manifested as additional artifacts. As seen in Fig.~\ref{fig:localization_ablation}, it is helpful in preventing  the diffusion of the skin color to the teeth.

\section{Conclusions}

The proliferation of work devoted to effectively inverting StyleGANs indicates an acute need for such technologies, in order to perform image editing and semantic image manipulations. In this work, we present a novel method that employs learned mappings between the loss gradient of a layer and a suggested shift to the layer's parameters. 
These shifts are used within an iterative optimization process, in order to fine-tune the StyleGAN generator. The learning of the mapping networks and the optimization of the generator occur concurrently and on a single sample.

We conduct a set of experiments that is considerably more extensive than what has been presented in recent relevant contributions, showing that the modified generator produces the target image more accurately than all other methods in this very active field. Moreover, it supports downstream editing tasks more convincingly than the alternatives.

\subsubsection{Acknowledgements} 
This project has received funding from the European Research Council (ERC) under the European Unions Horizon 2020 research, innovation program (grant ERC CoG 725974). 

\clearpage
\bibliographystyle{splncs04}
\bibliography{stylegan}

\appendix
\section{Additional qualitative results}
\label{sec:additional_qual_results}

\begin{figure}[h]
\centering
\footnotesize
\renewcommand{\arraystretch}{0.0}
\begin{tabular}{@{}c@{}c@{}c@{}c@{}c@{}}
  \includegraphics[width=0.145\linewidth]{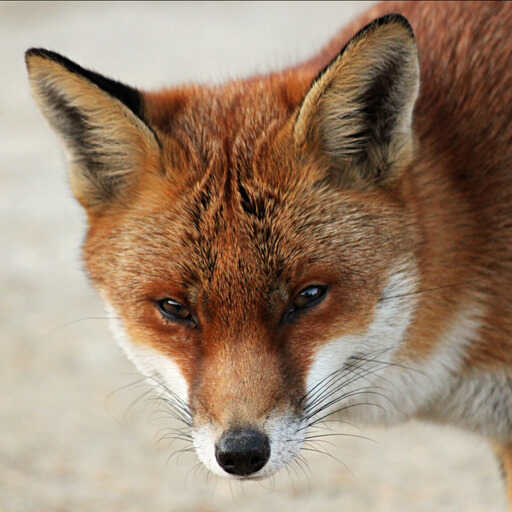} &
  \includegraphics[width=0.145\linewidth]{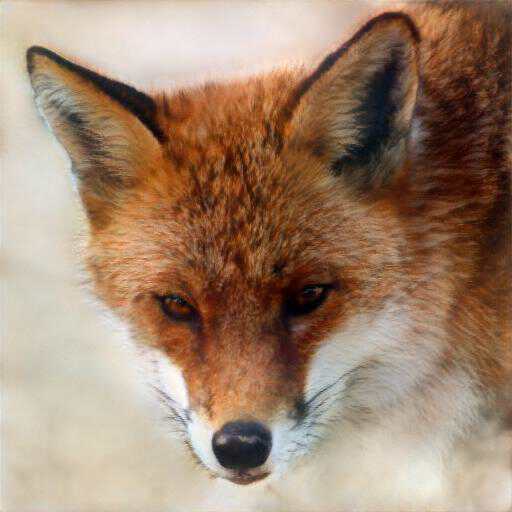}              &
  \includegraphics[width=0.145\linewidth]{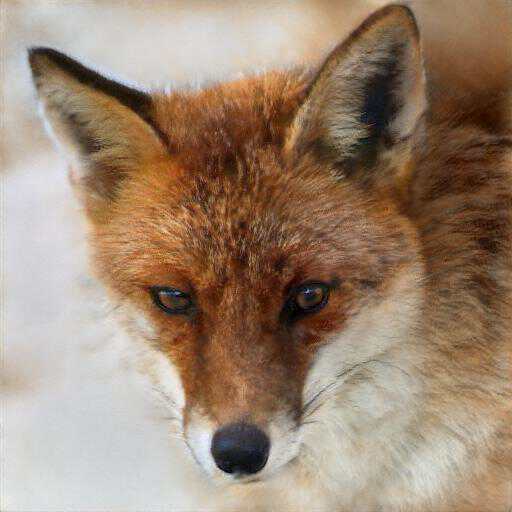}              &
  \includegraphics[width=0.145\linewidth]{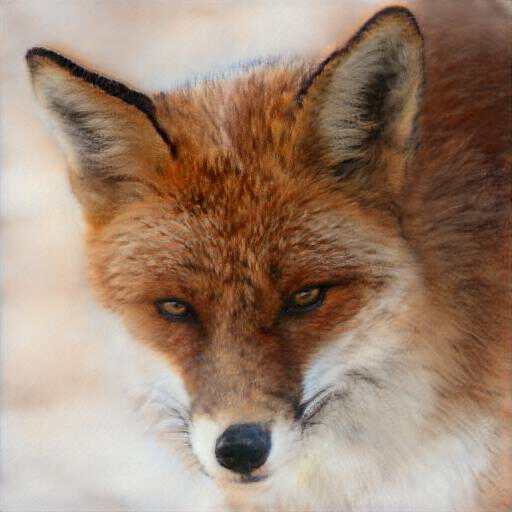}               &
  \includegraphics[width=0.145\linewidth]{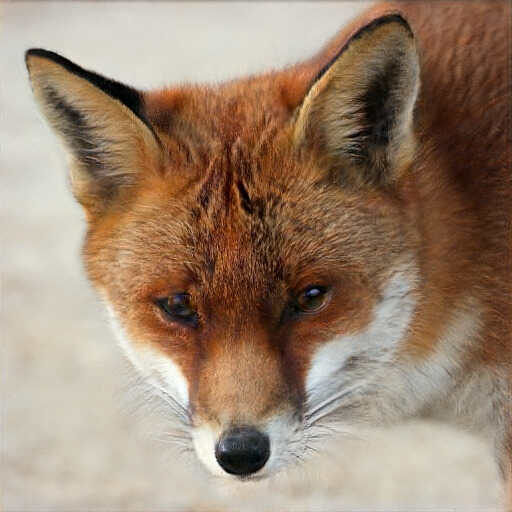}                      \\
  \includegraphics[width=0.145\linewidth]{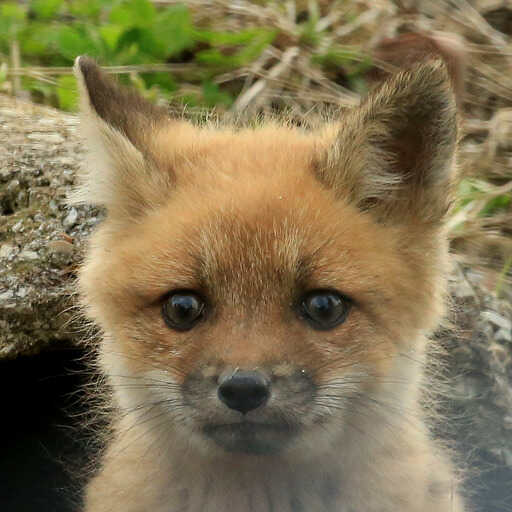} &
  \includegraphics[width=0.145\linewidth]{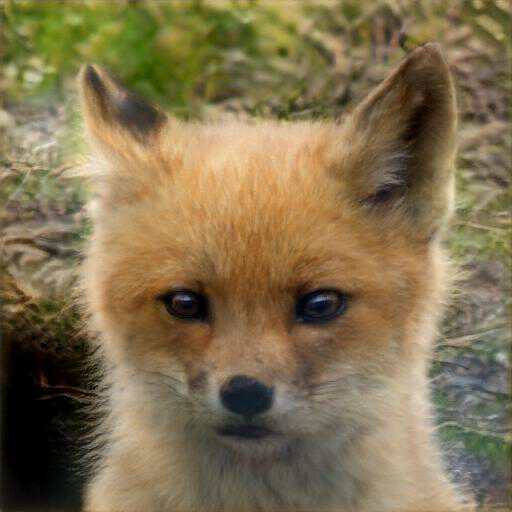}              &
  \includegraphics[width=0.145\linewidth]{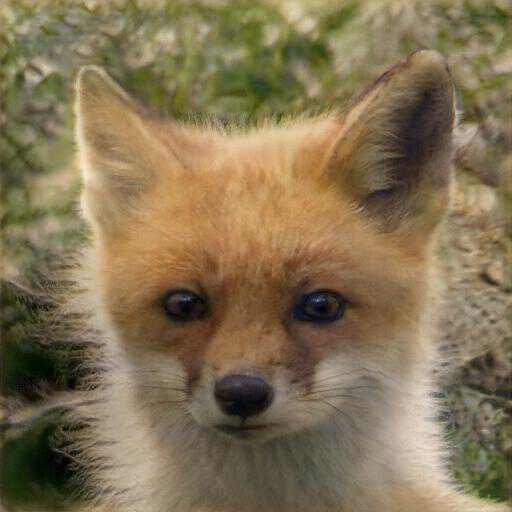}              &
  \includegraphics[width=0.145\linewidth]{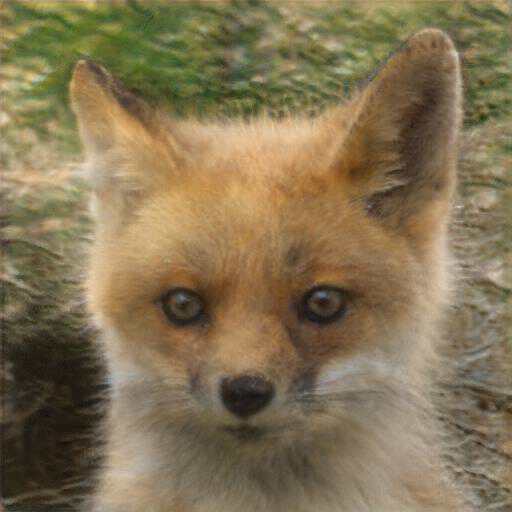}               &
  \includegraphics[width=0.145\linewidth]{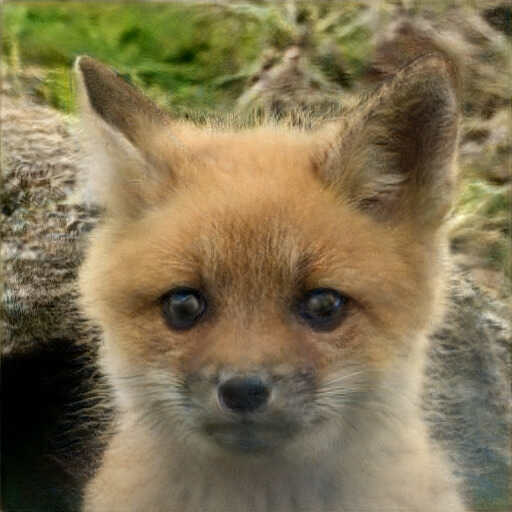}                      \\
  \includegraphics[width=0.145\linewidth]{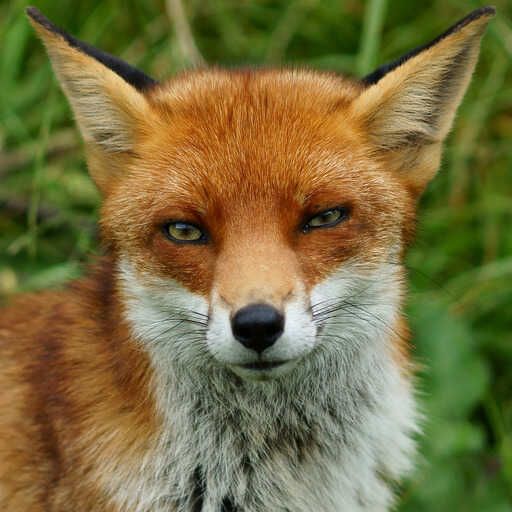} &
  \includegraphics[width=0.145\linewidth]{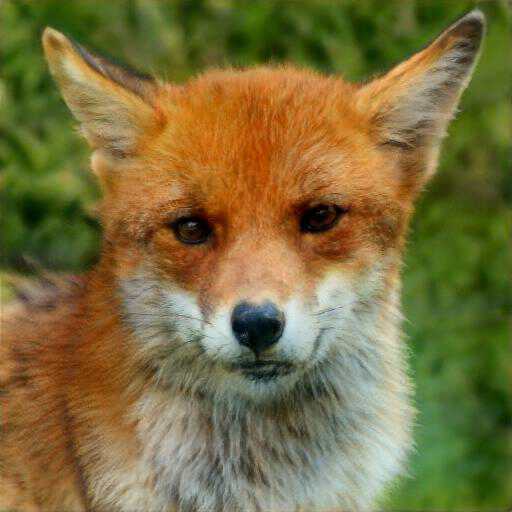}              &
  \includegraphics[width=0.145\linewidth]{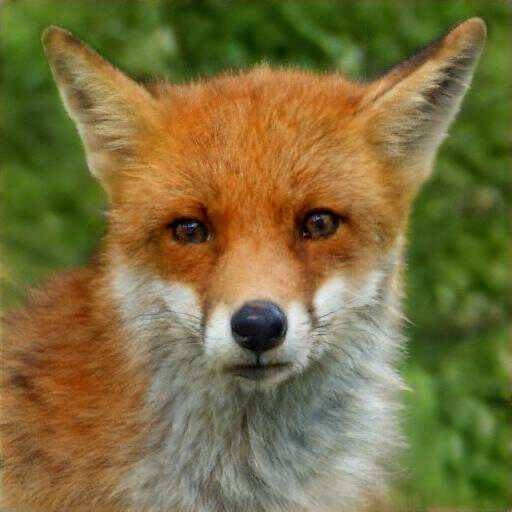}              &
  \includegraphics[width=0.145\linewidth]{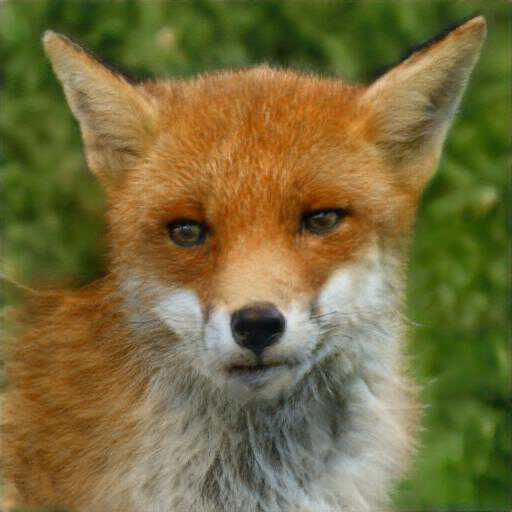}               &
  \includegraphics[width=0.145\linewidth]{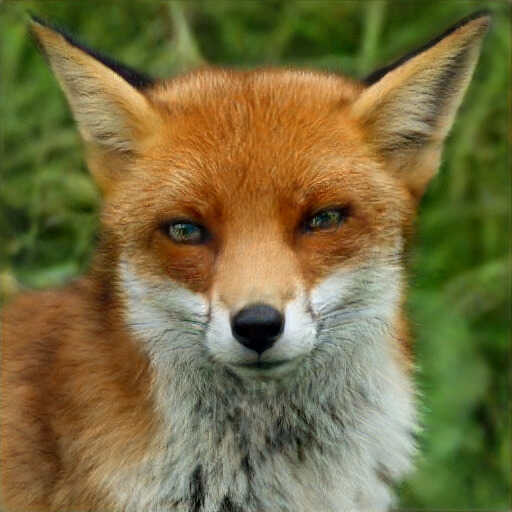}                      \\
  \includegraphics[width=0.145\linewidth]{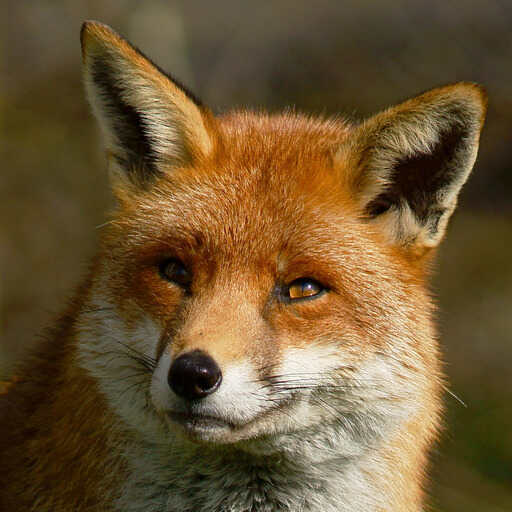} &
  \includegraphics[width=0.145\linewidth]{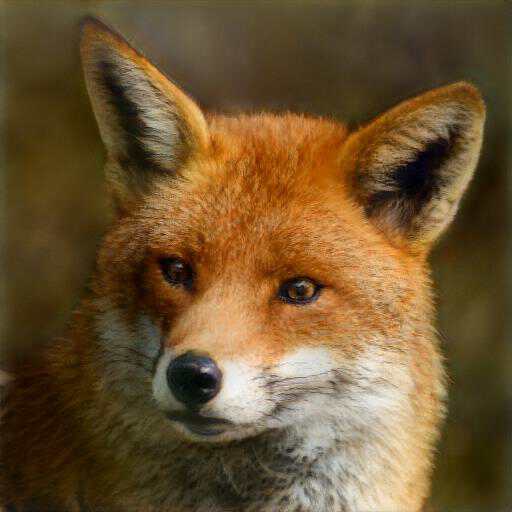}              &
  \includegraphics[width=0.145\linewidth]{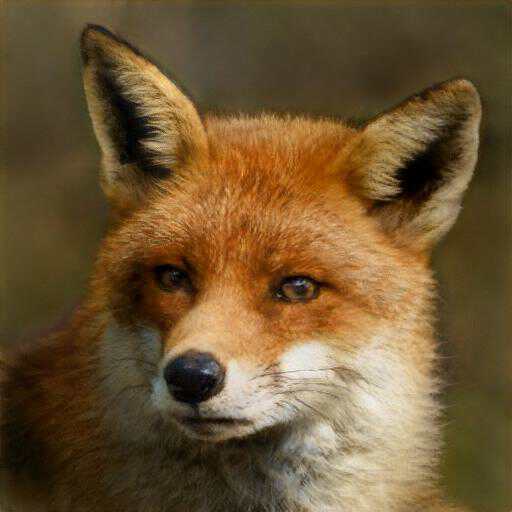}              &
  \includegraphics[width=0.145\linewidth]{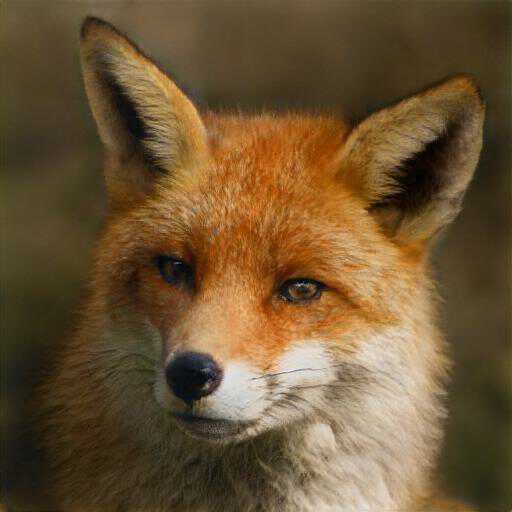}               &
  \includegraphics[width=0.145\linewidth]{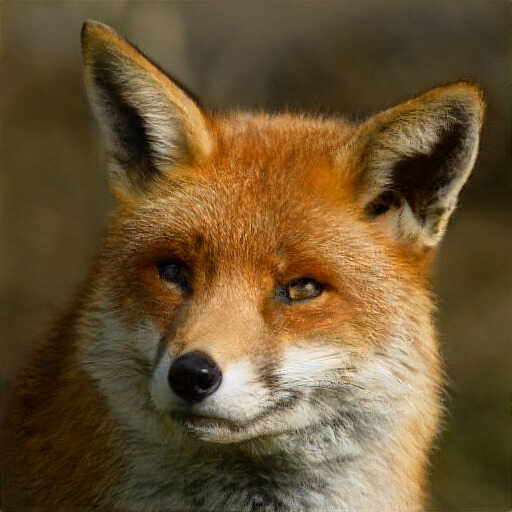}                      \\
  \includegraphics[width=0.145\linewidth]{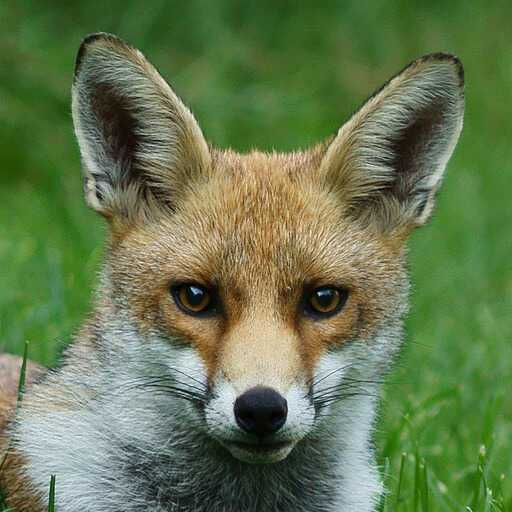} &
  \includegraphics[width=0.145\linewidth]{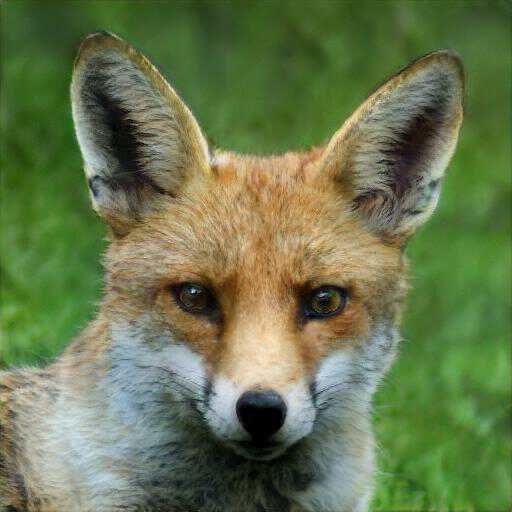}              &
  \includegraphics[width=0.145\linewidth]{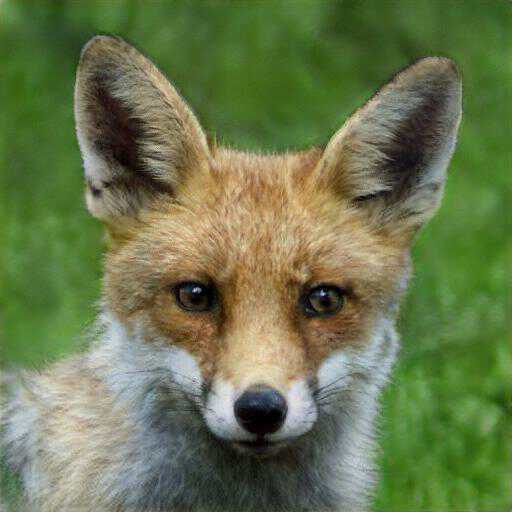}              &
  \includegraphics[width=0.145\linewidth]{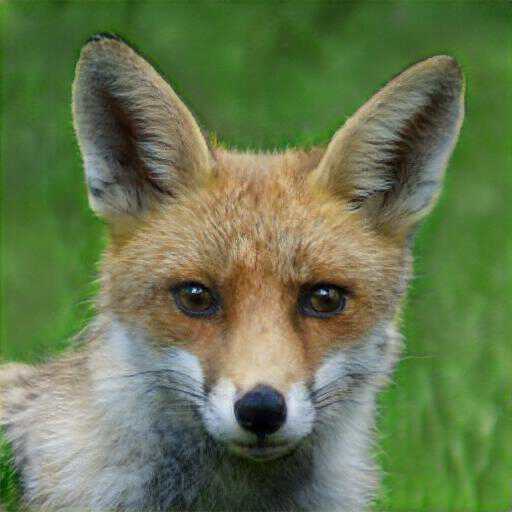}               &
  \includegraphics[width=0.145\linewidth]{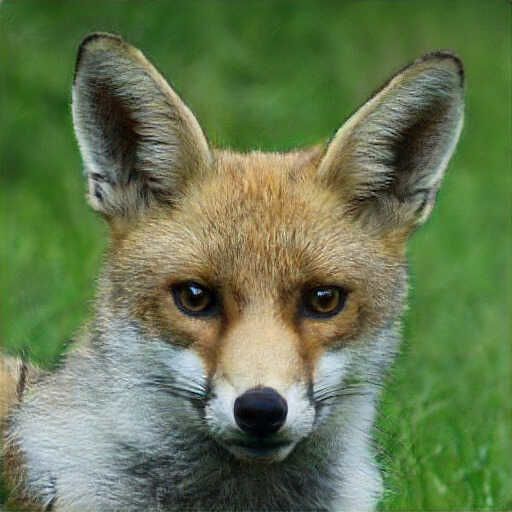}                      \\ [5pt]
  GT & $\text{ReStyle}_{\text{pSp}}$ & $\text{ReStyle}_{\text{e4e}}$ & HyperStyle & Ours \\
\end{tabular}
\caption{%
  Visual comparison of image reconstruction quality on the AFHQ-Wild dataset.}
\end{figure}

\begin{figure}[t]
\centering
\footnotesize
\renewcommand{\arraystretch}{0.0}
\begin{tabular}{@{}c@{\hspace{2pt}}c@{\hspace{2pt}}:c@{}c@{}c@{}c@{}}
  \rotatebox{90}{\hspace{12pt} +Age}                                                                                    &
  \includegraphics[width=0.15\linewidth]{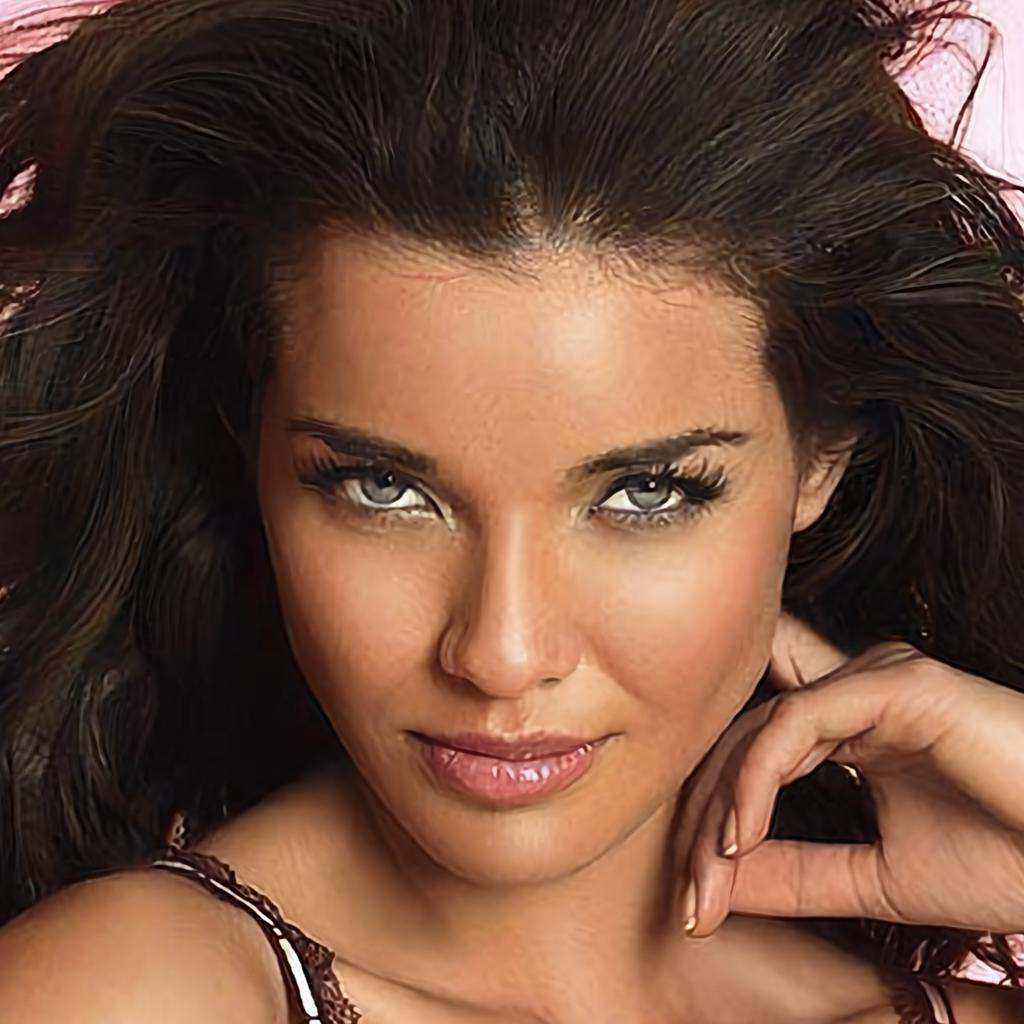}                                             &
  \includegraphics[width=0.15\linewidth]{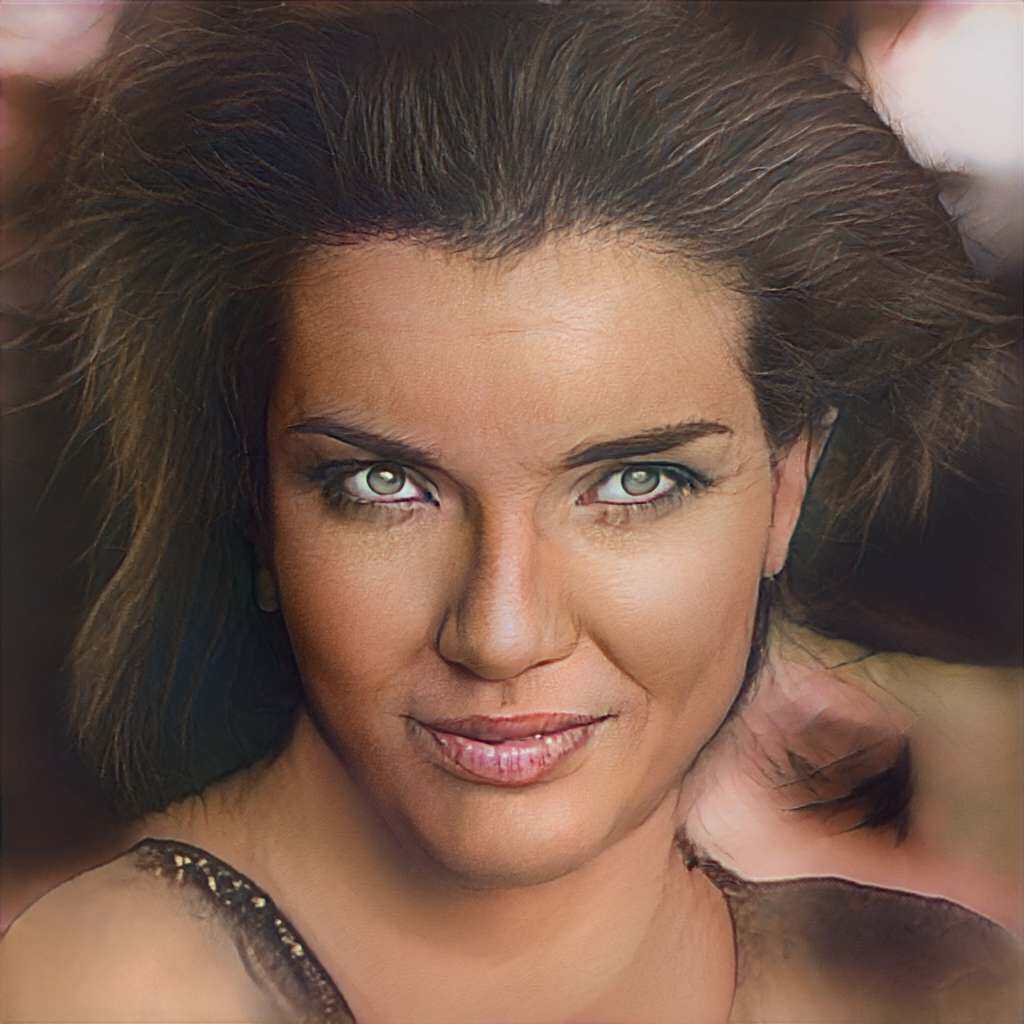}       &
  \includegraphics[width=0.15\linewidth]{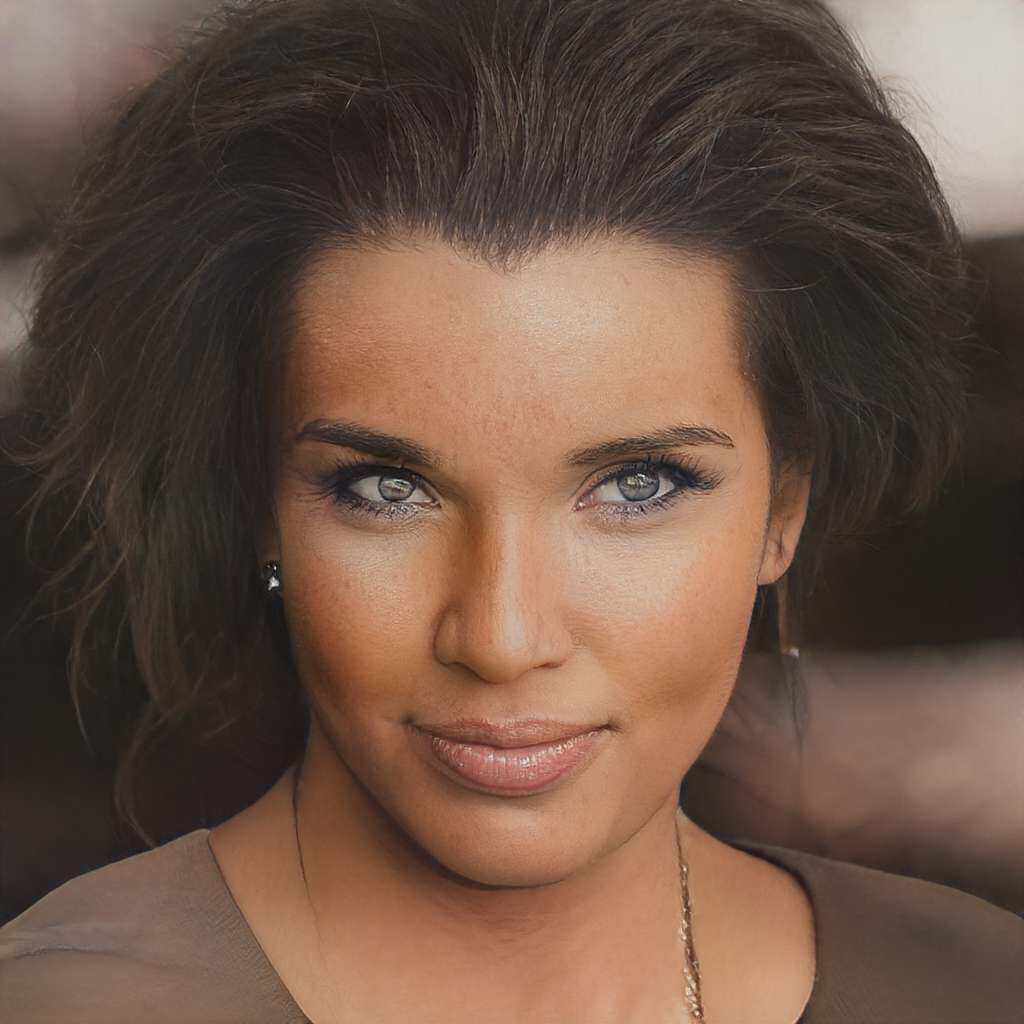}       &
  \includegraphics[width=0.15\linewidth]{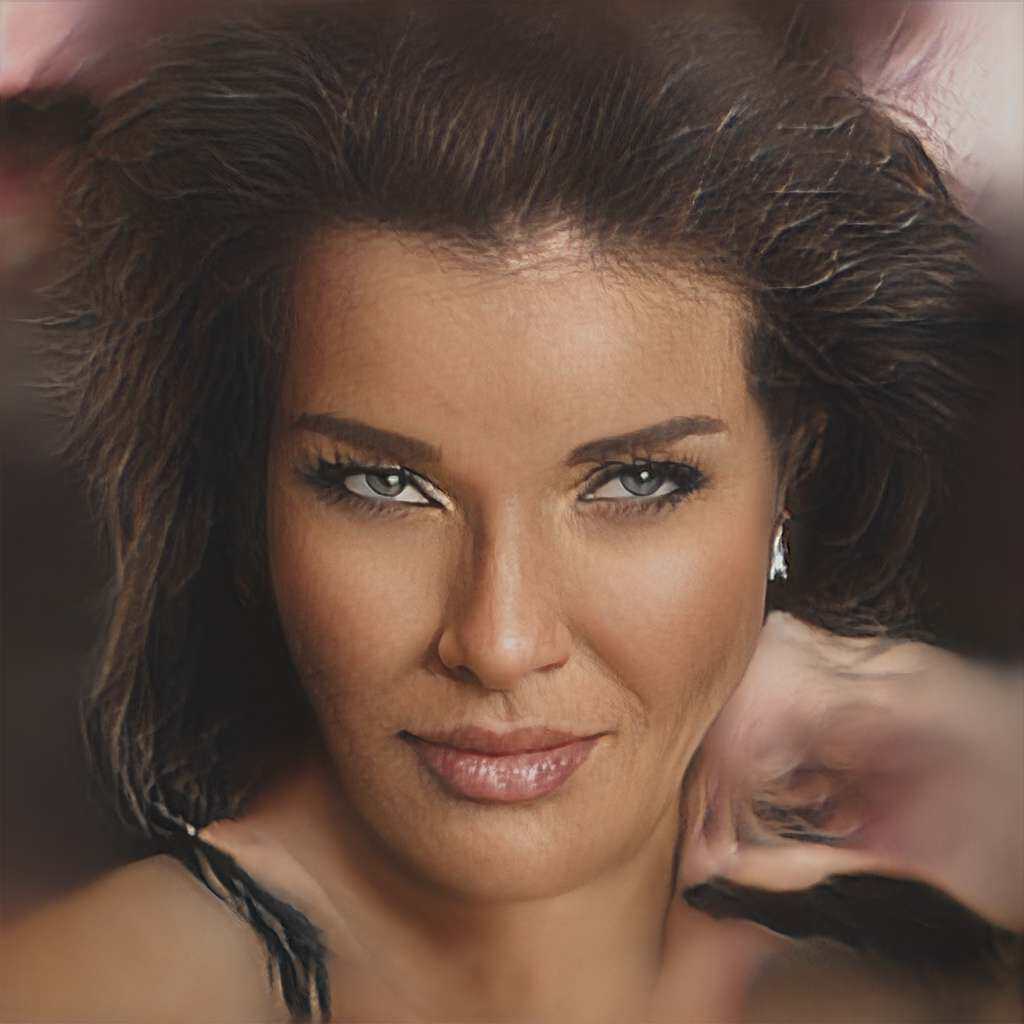}        &
  \includegraphics[width=0.15\linewidth]{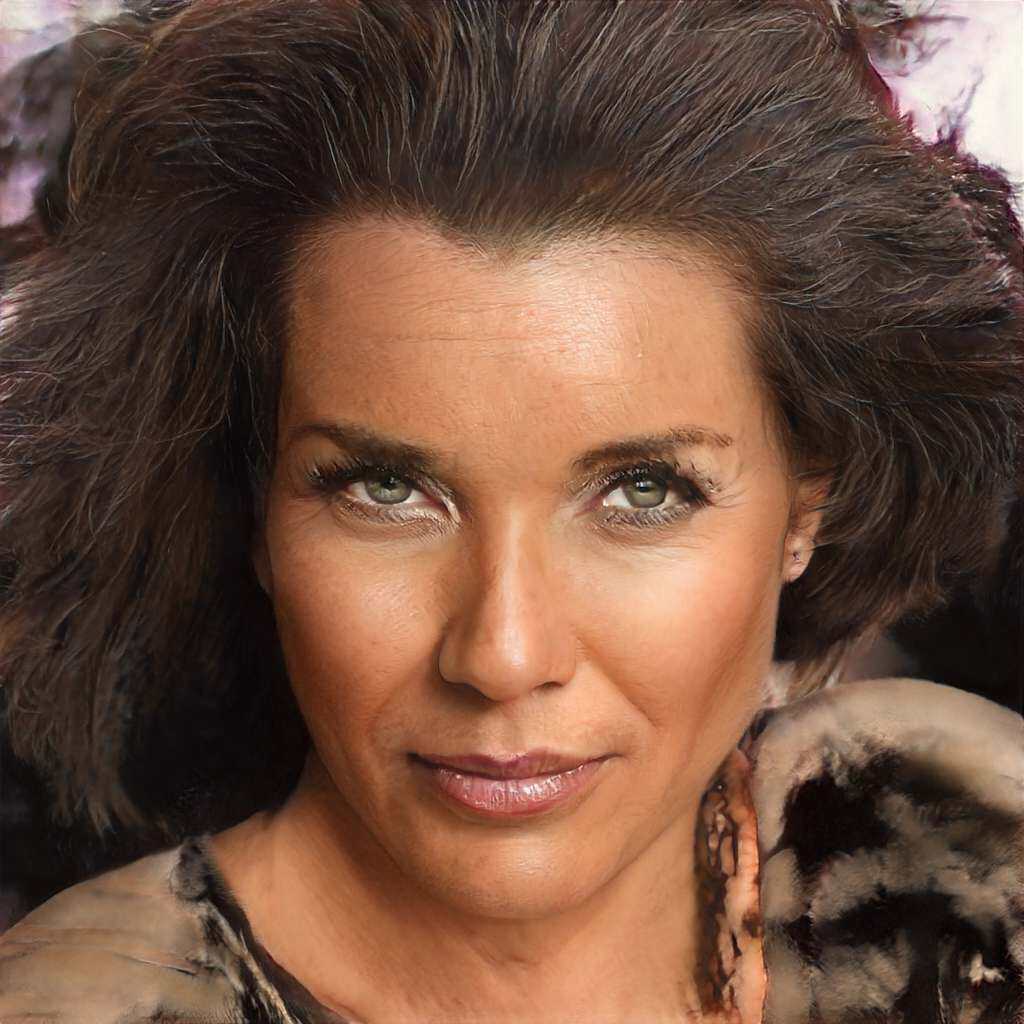}                                                                                                                                                                \\
  \rotatebox{90}{\hspace{12pt} -Age}                                                                                    &
  \includegraphics[width=0.15\linewidth]{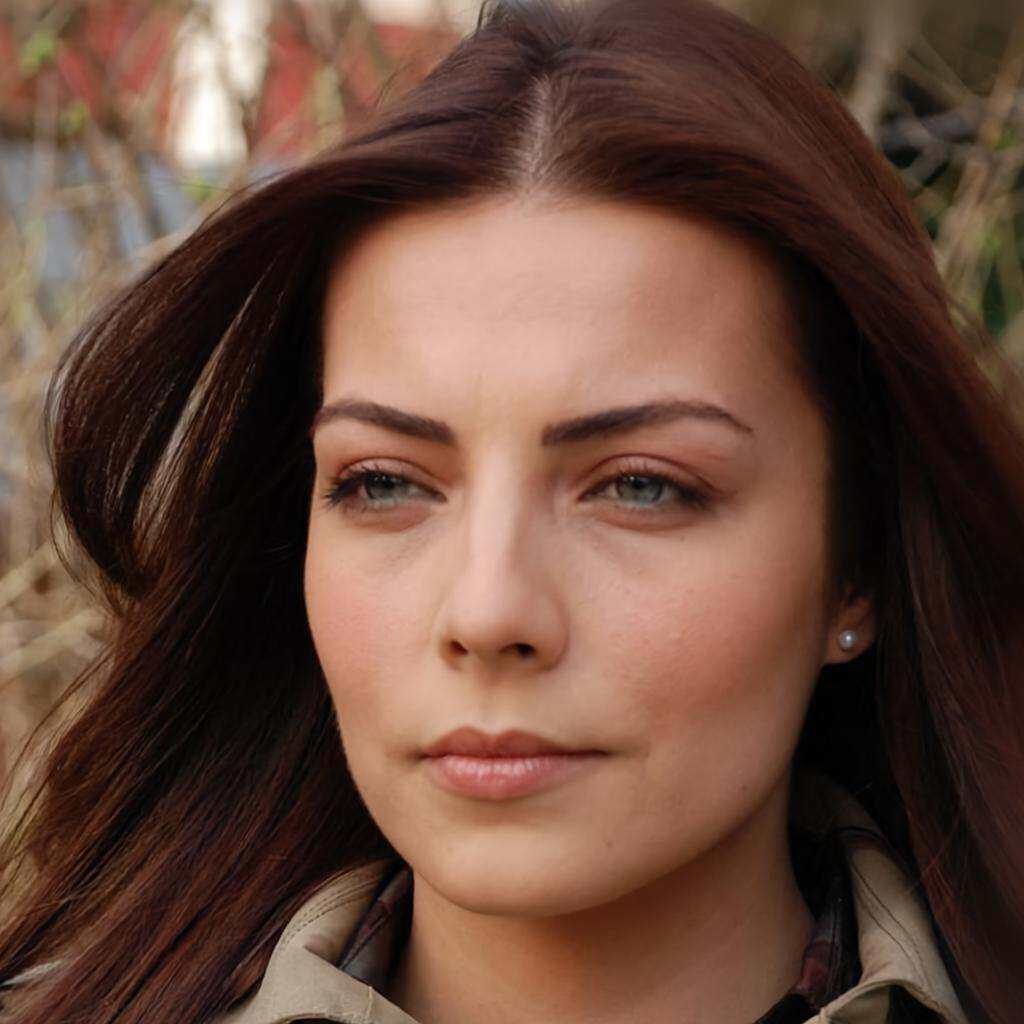}                                             &
  \includegraphics[width=0.15\linewidth]{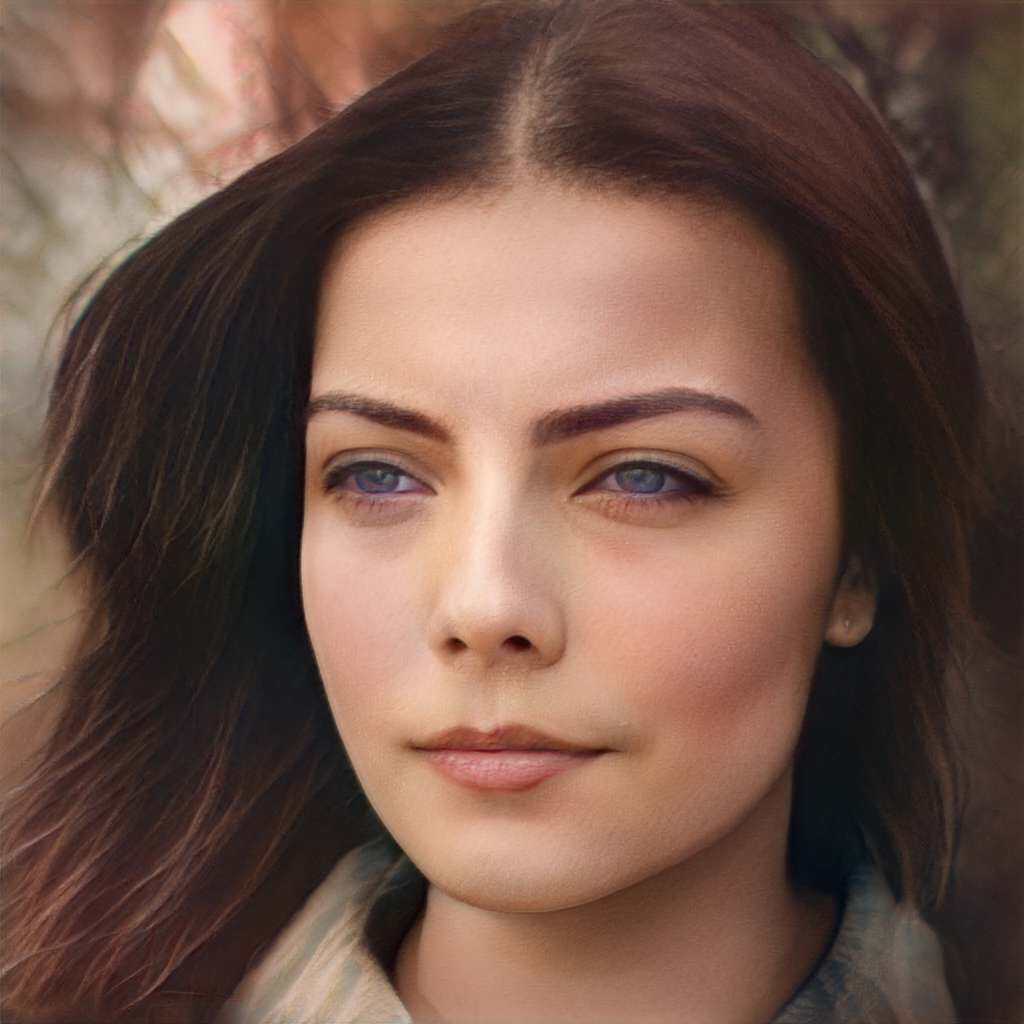}      &
  \includegraphics[width=0.15\linewidth]{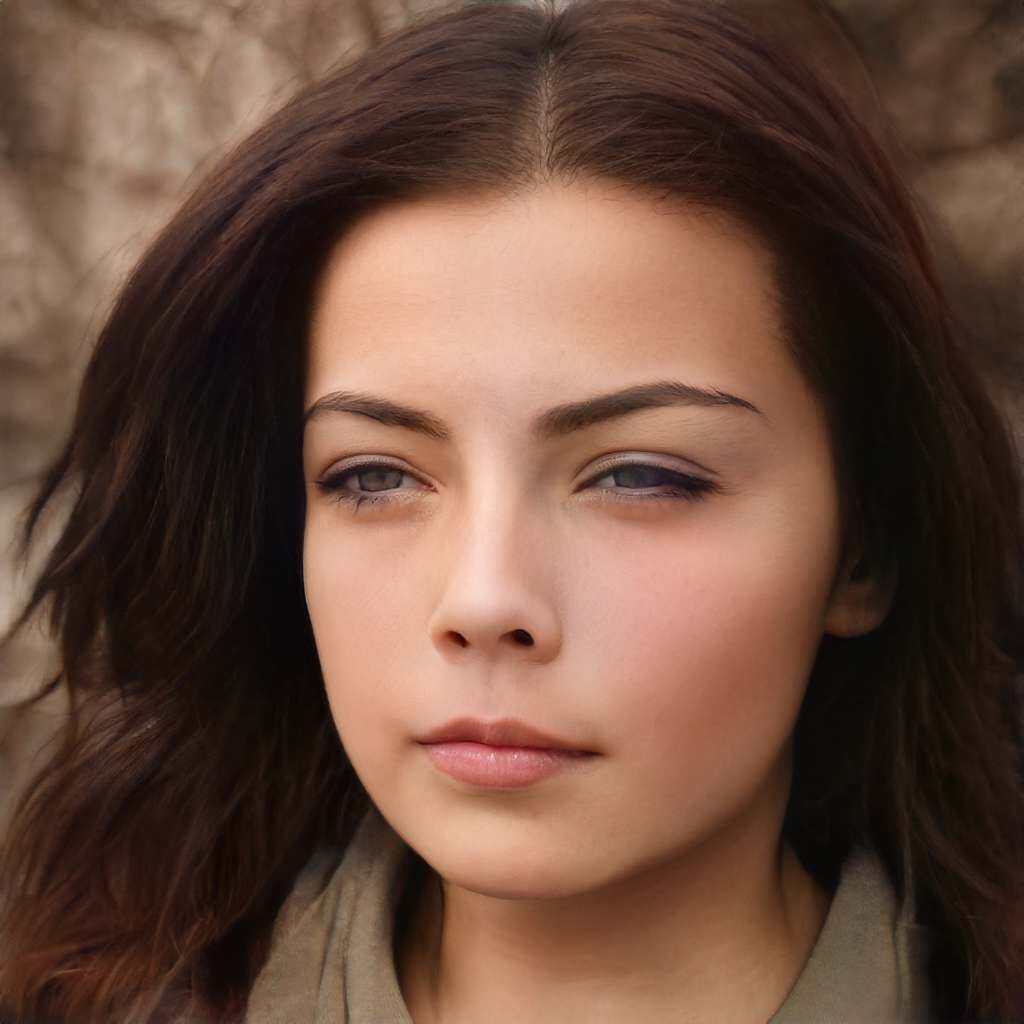}      &
  \includegraphics[width=0.15\linewidth]{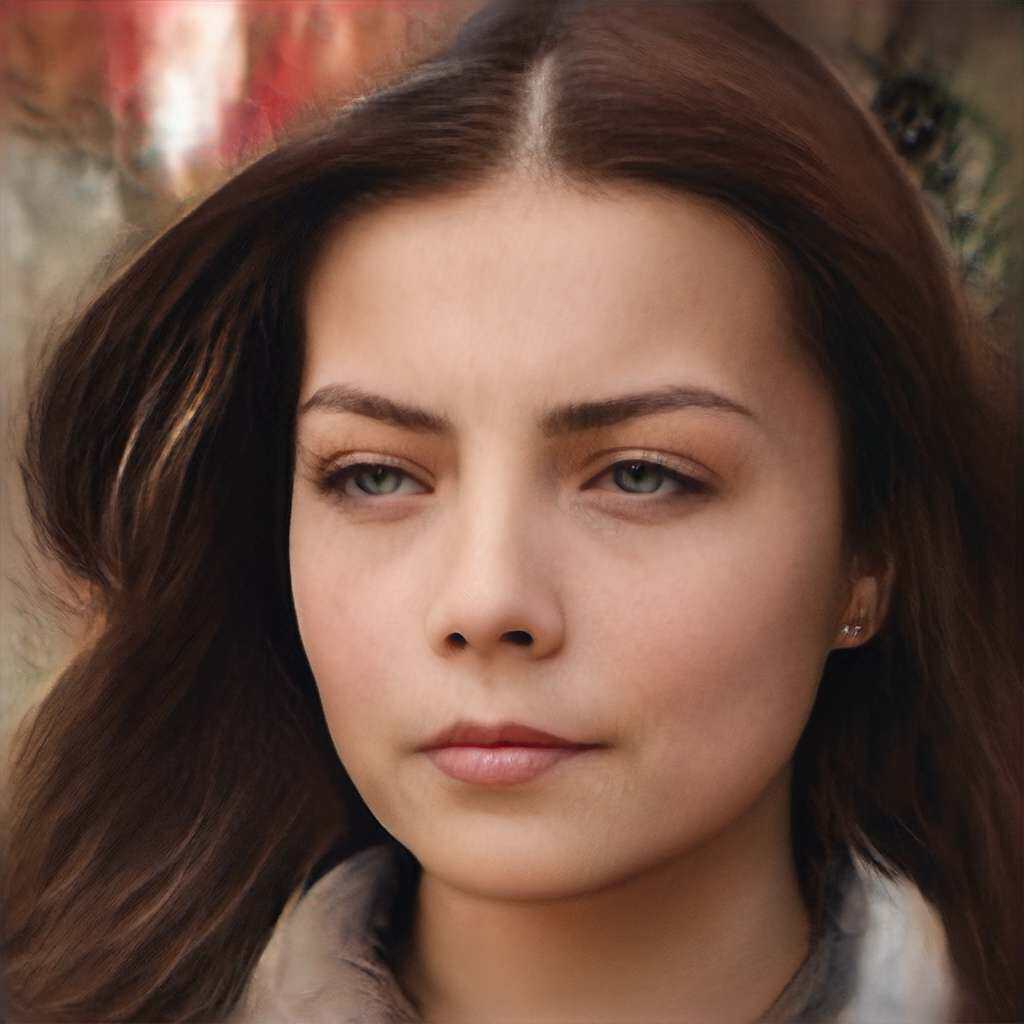}       &
  \includegraphics[width=0.15\linewidth]{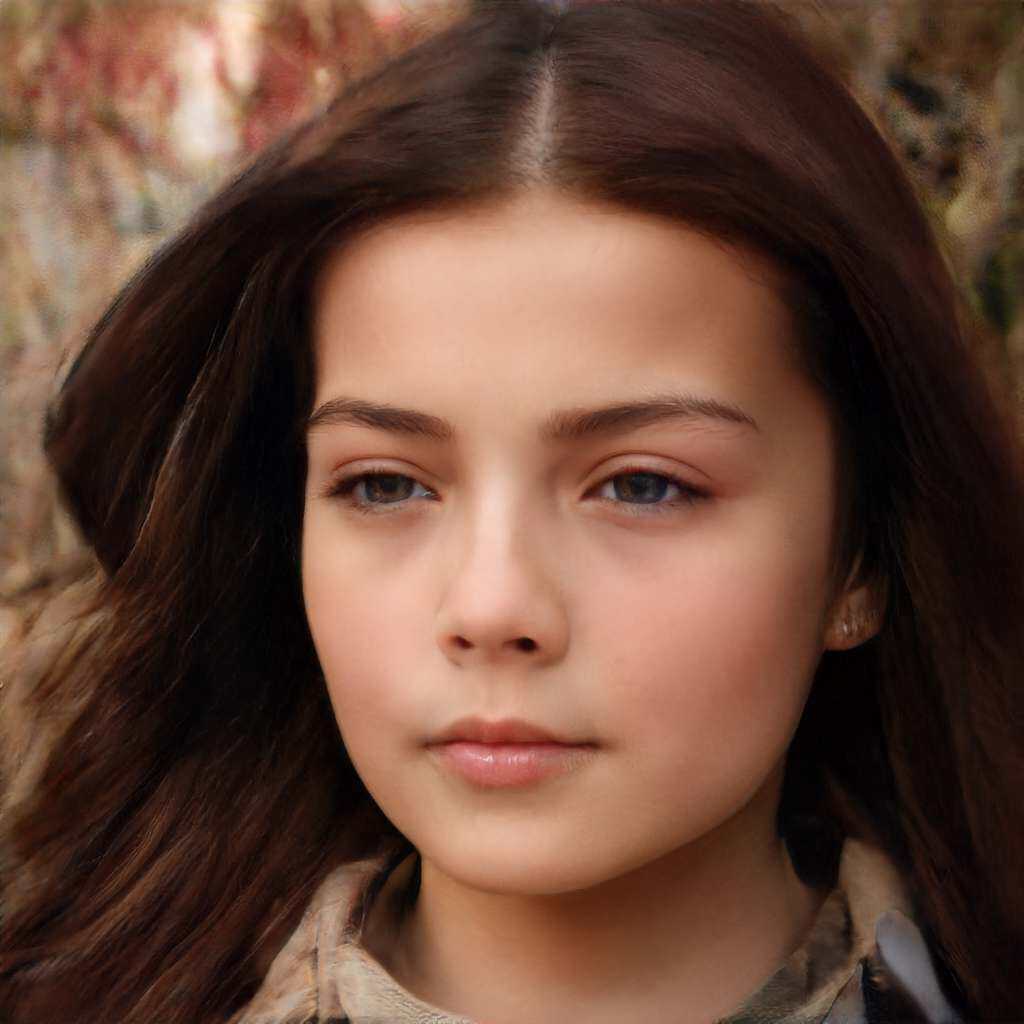}                                                                                                                                                               \\
  \rotatebox{90}{\hspace{12pt} +Pose}                                                                               &
  \includegraphics[width=0.15\linewidth]{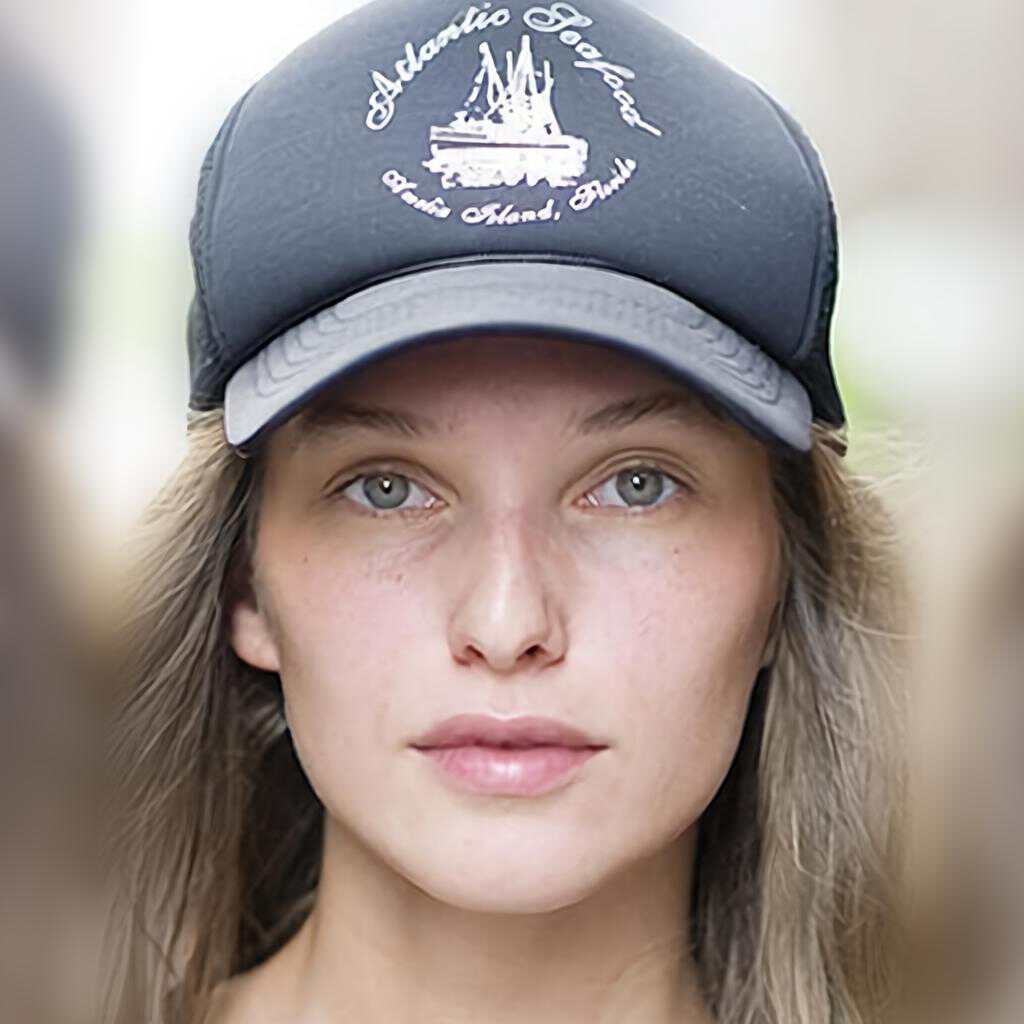}                                             &
  \includegraphics[width=0.15\linewidth]{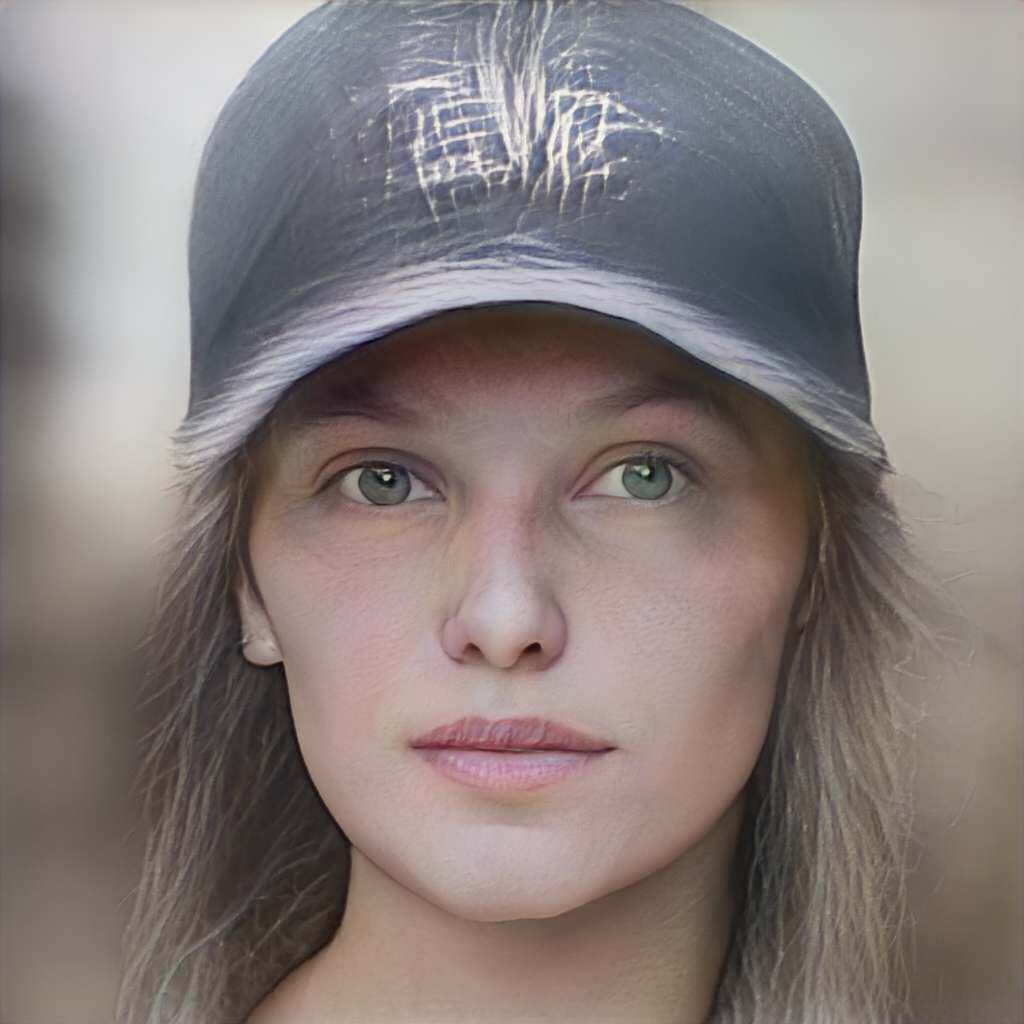}  &
  \includegraphics[width=0.15\linewidth]{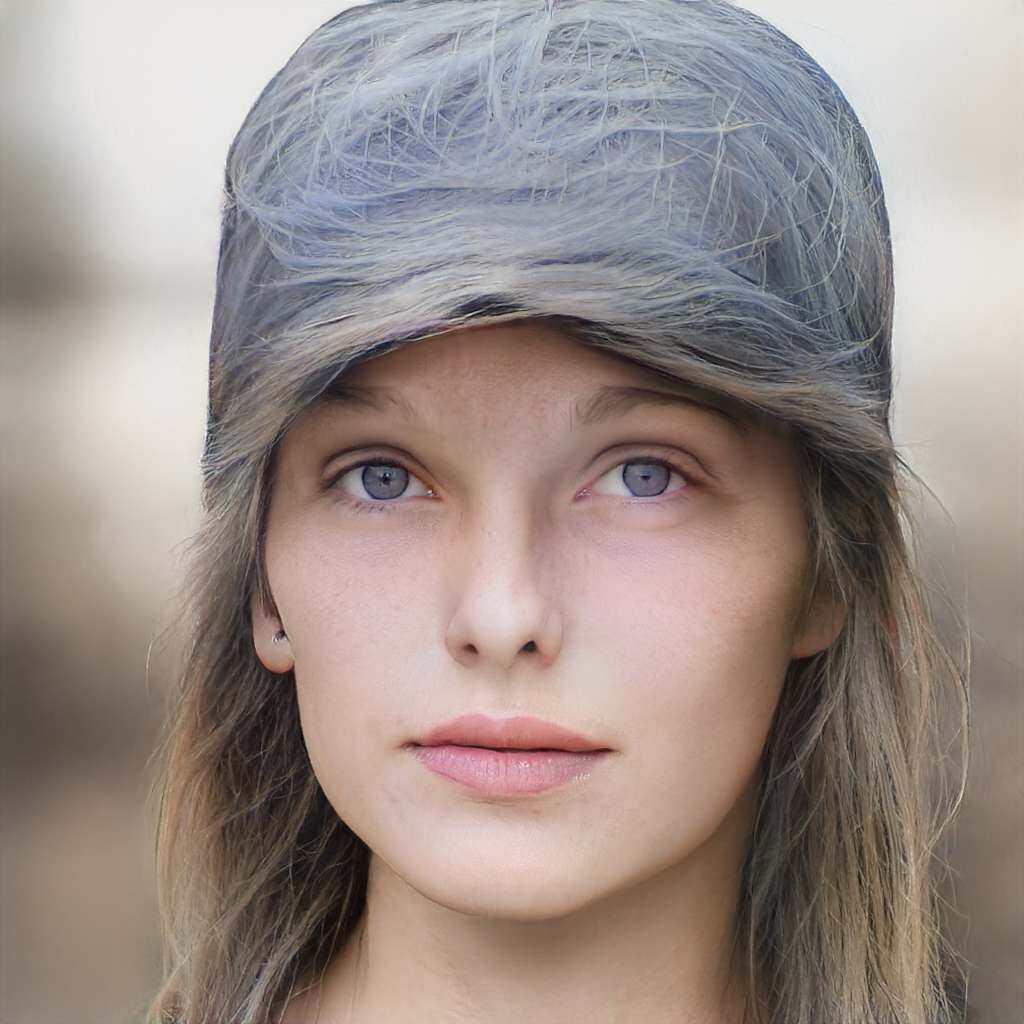}  &
  \includegraphics[width=0.15\linewidth]{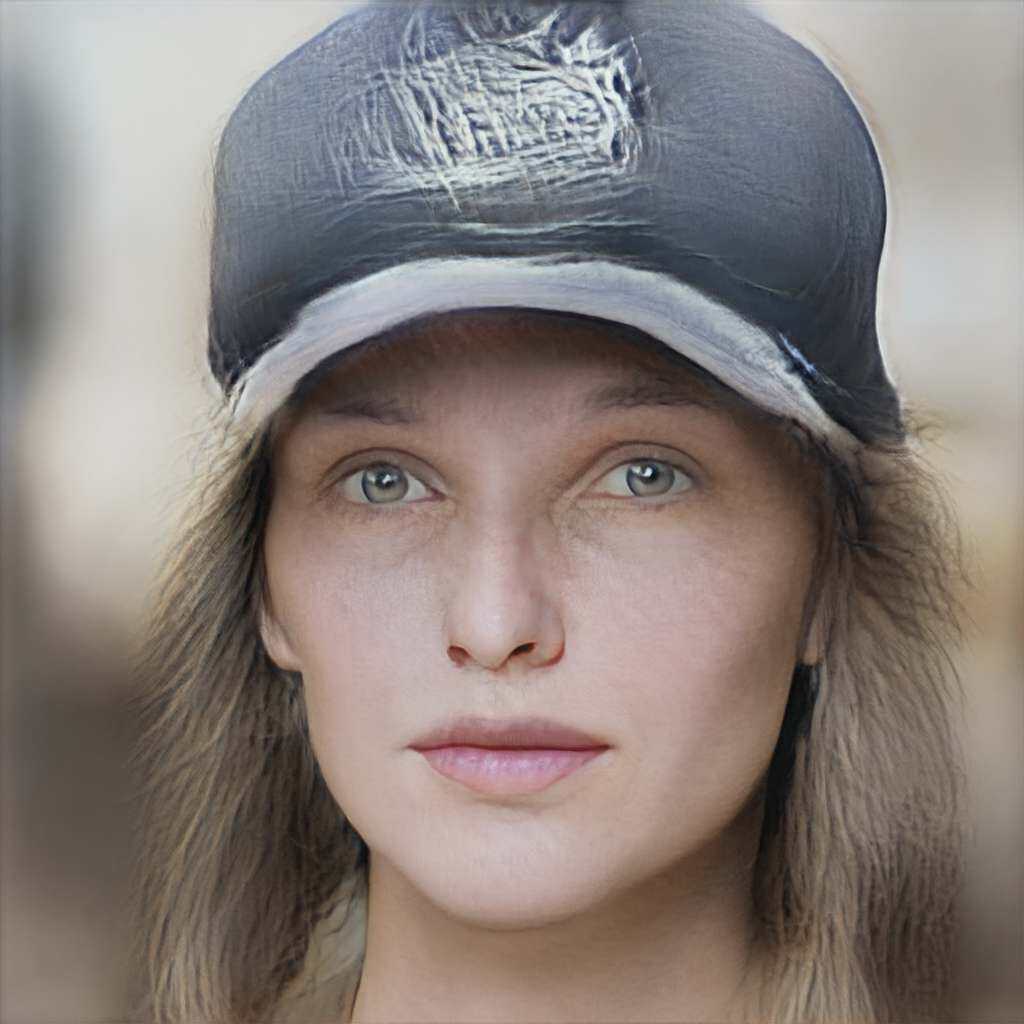}   &
  \includegraphics[width=0.15\linewidth]{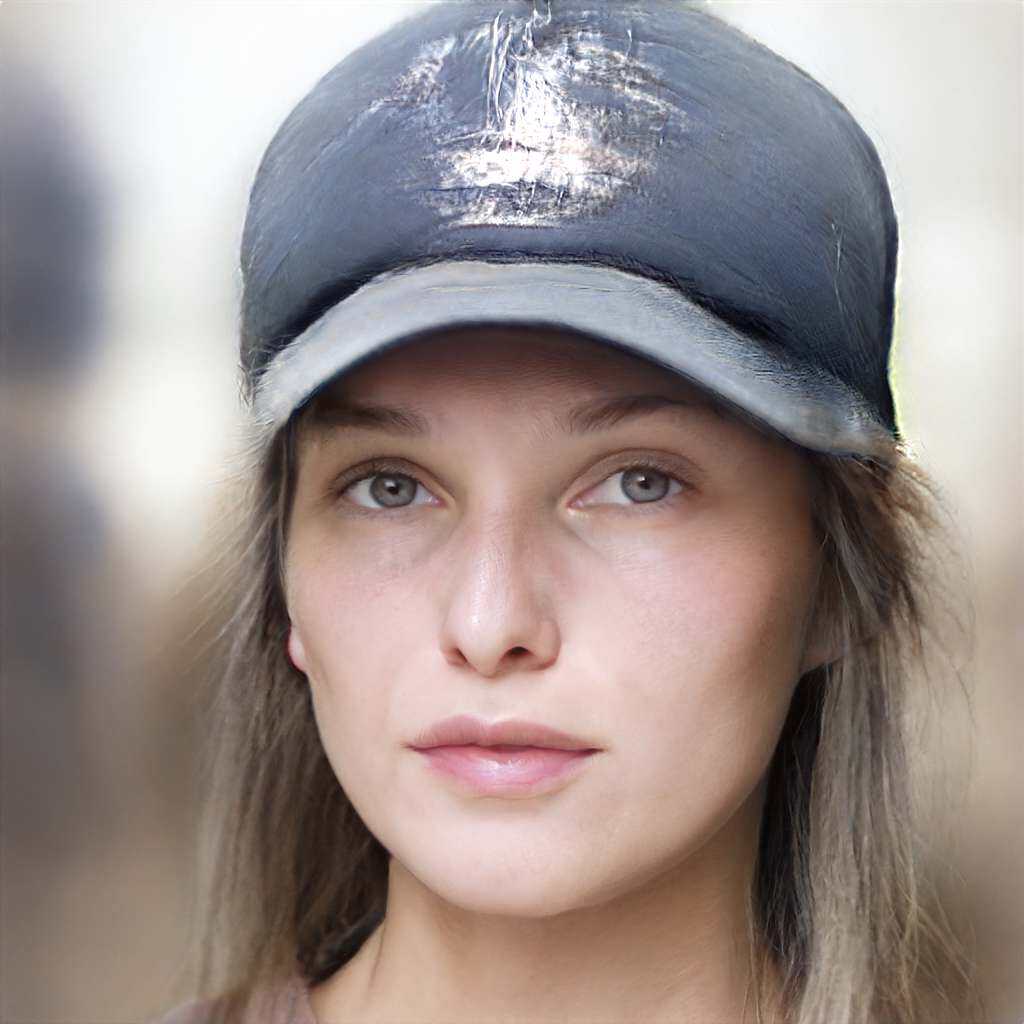}                                                                                                                                                           \\
  \rotatebox{90}{\hspace{12pt} -Pose}                                                                               &
  \includegraphics[width=0.15\linewidth]{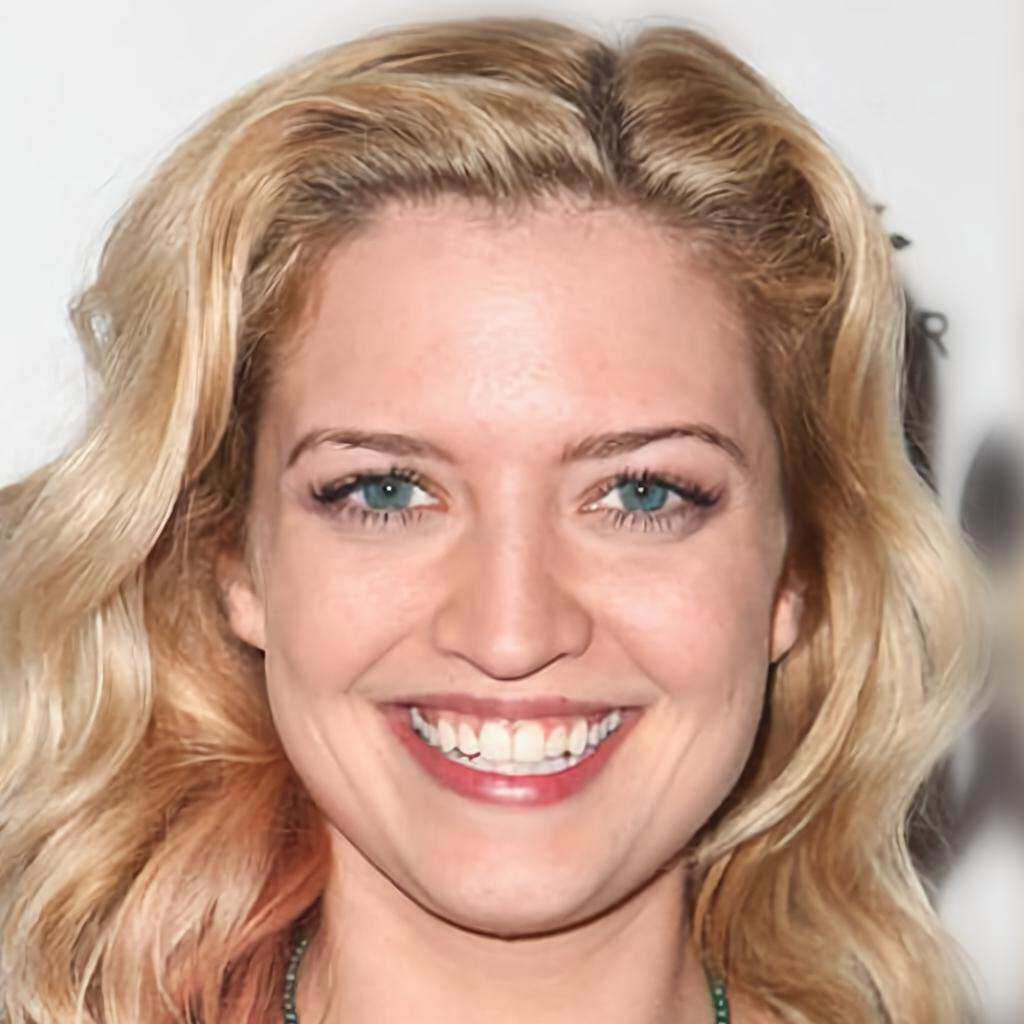}                                             &
  \includegraphics[width=0.15\linewidth]{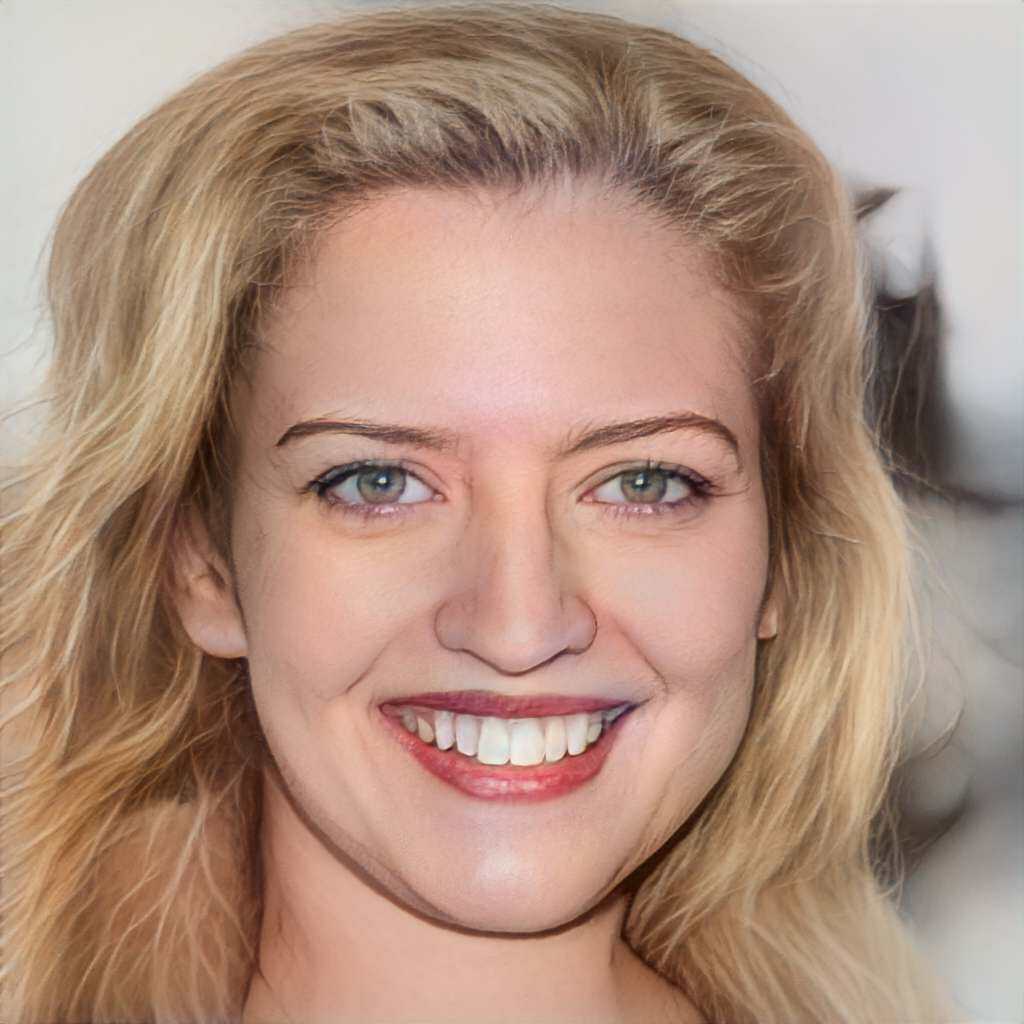} &
  \includegraphics[width=0.15\linewidth]{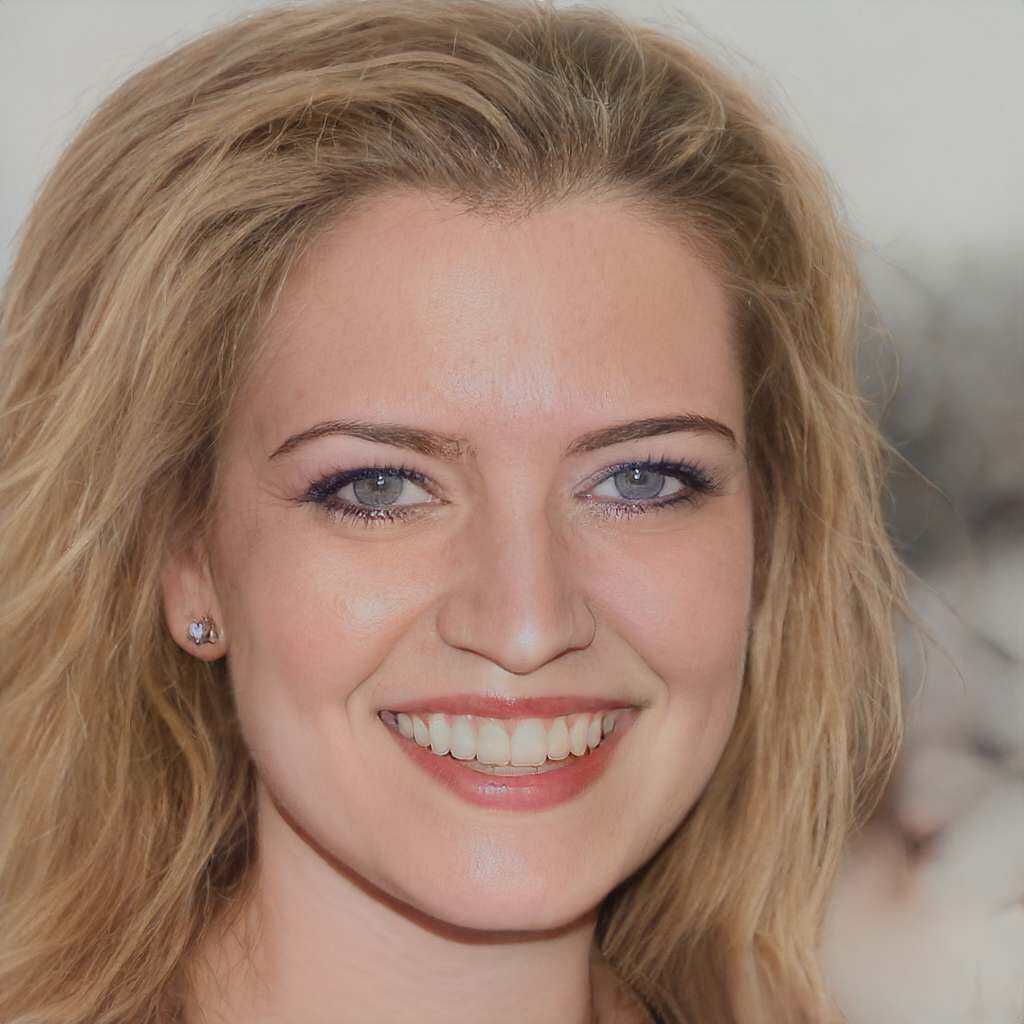} &
  \includegraphics[width=0.15\linewidth]{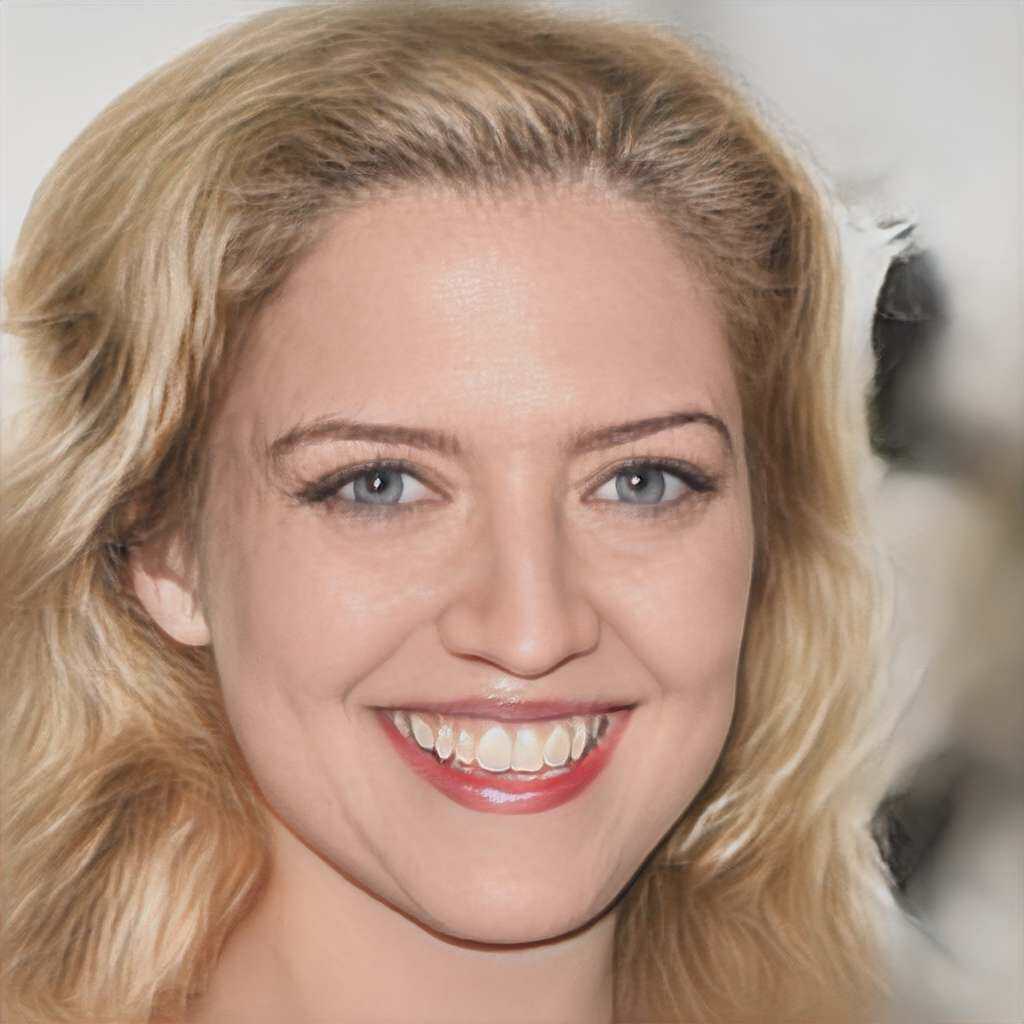}  &
  \includegraphics[width=0.15\linewidth]{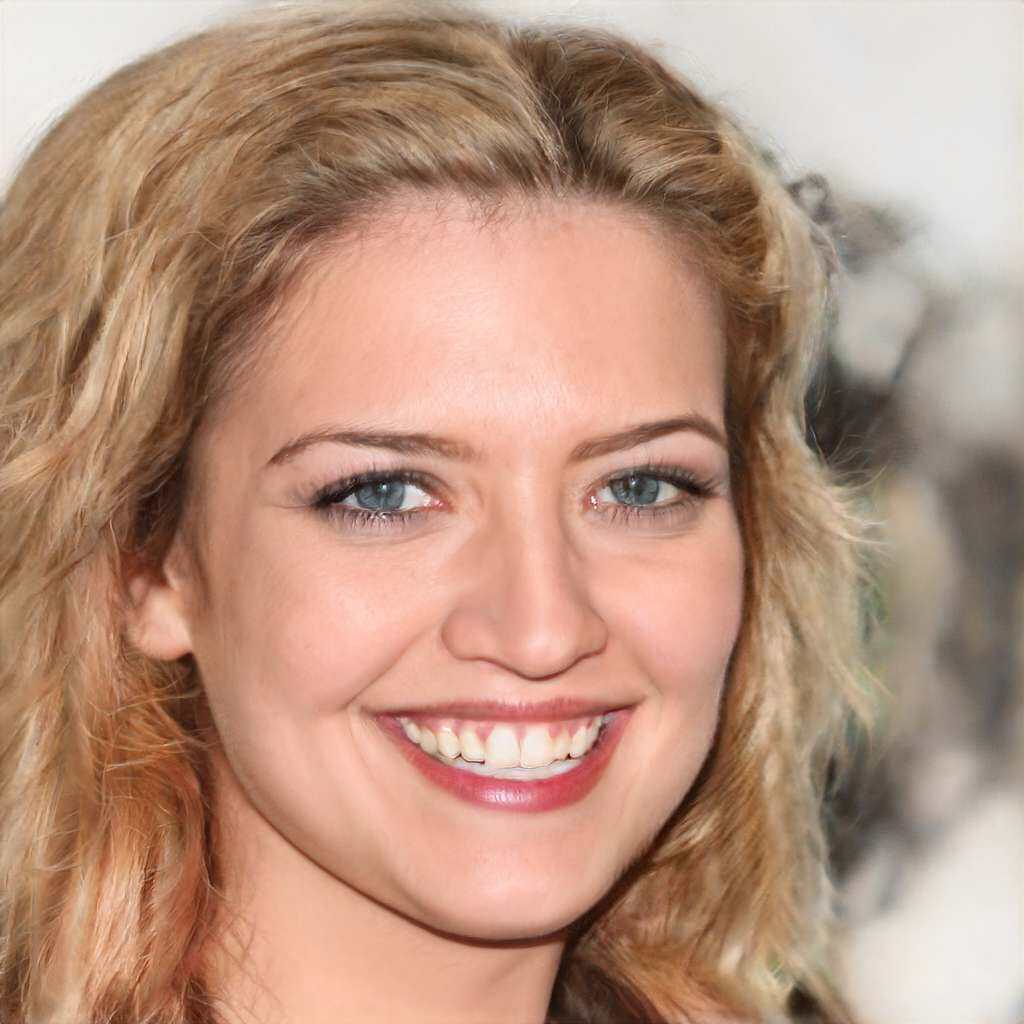}                                                                                                                                                          \\
  \rotatebox{90}{\hspace{10pt} +Smile}                                                                                  &
  \includegraphics[width=0.15\linewidth]{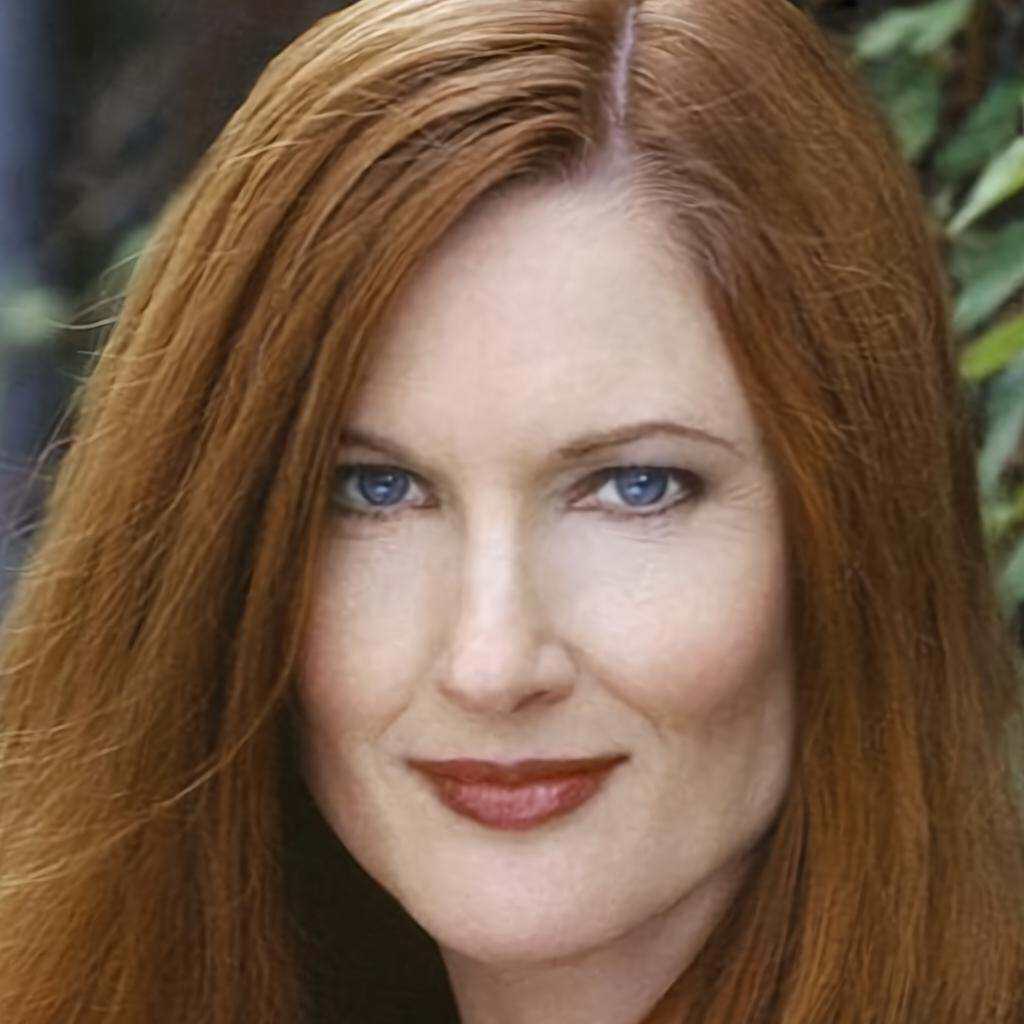}                                             &
  \includegraphics[width=0.15\linewidth]{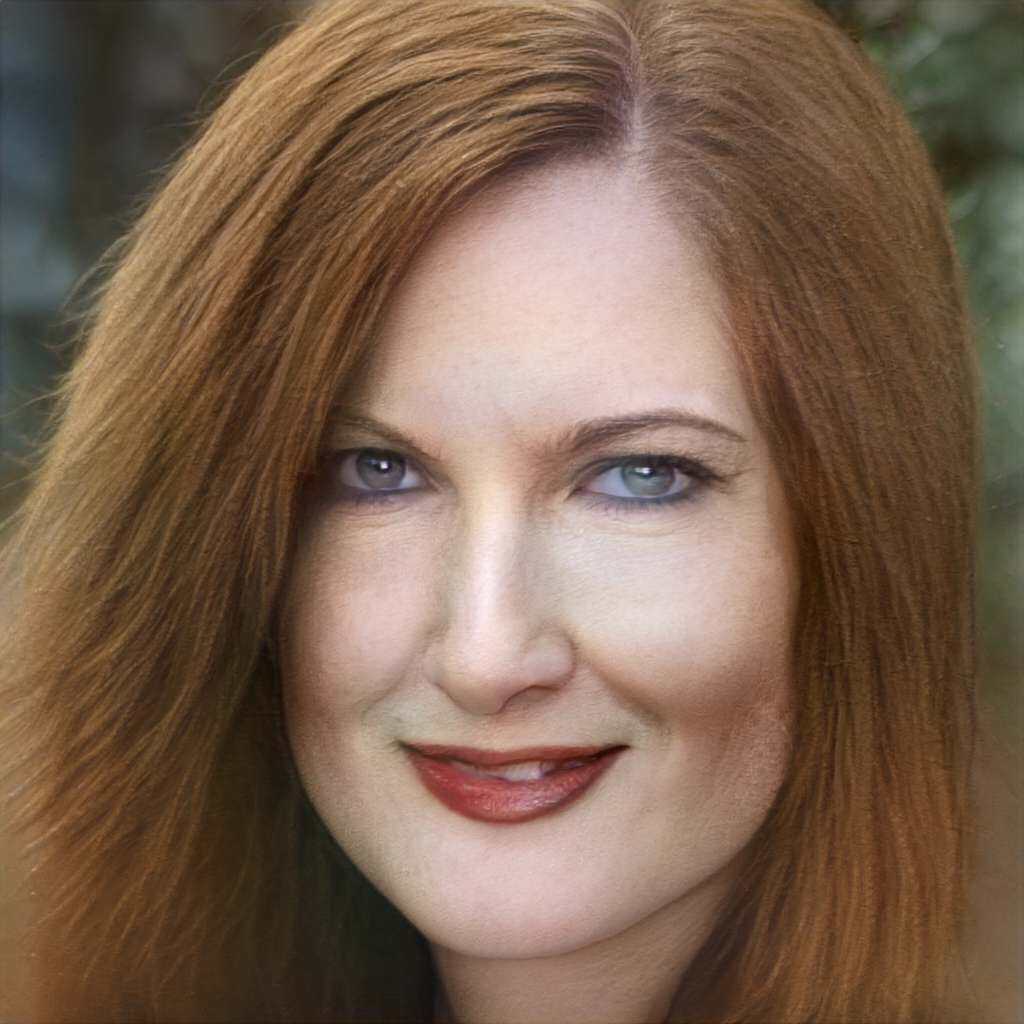}     &
  \includegraphics[width=0.15\linewidth]{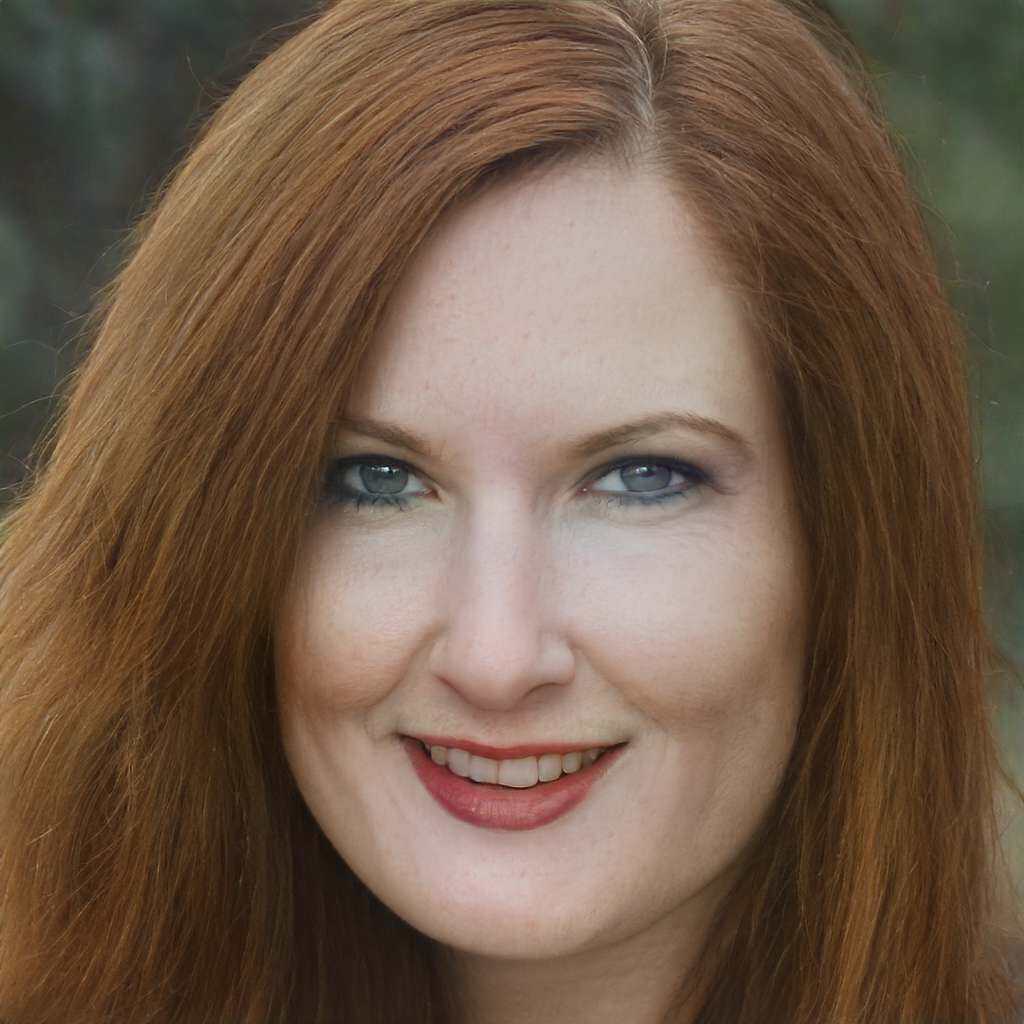}     &
  \includegraphics[width=0.15\linewidth]{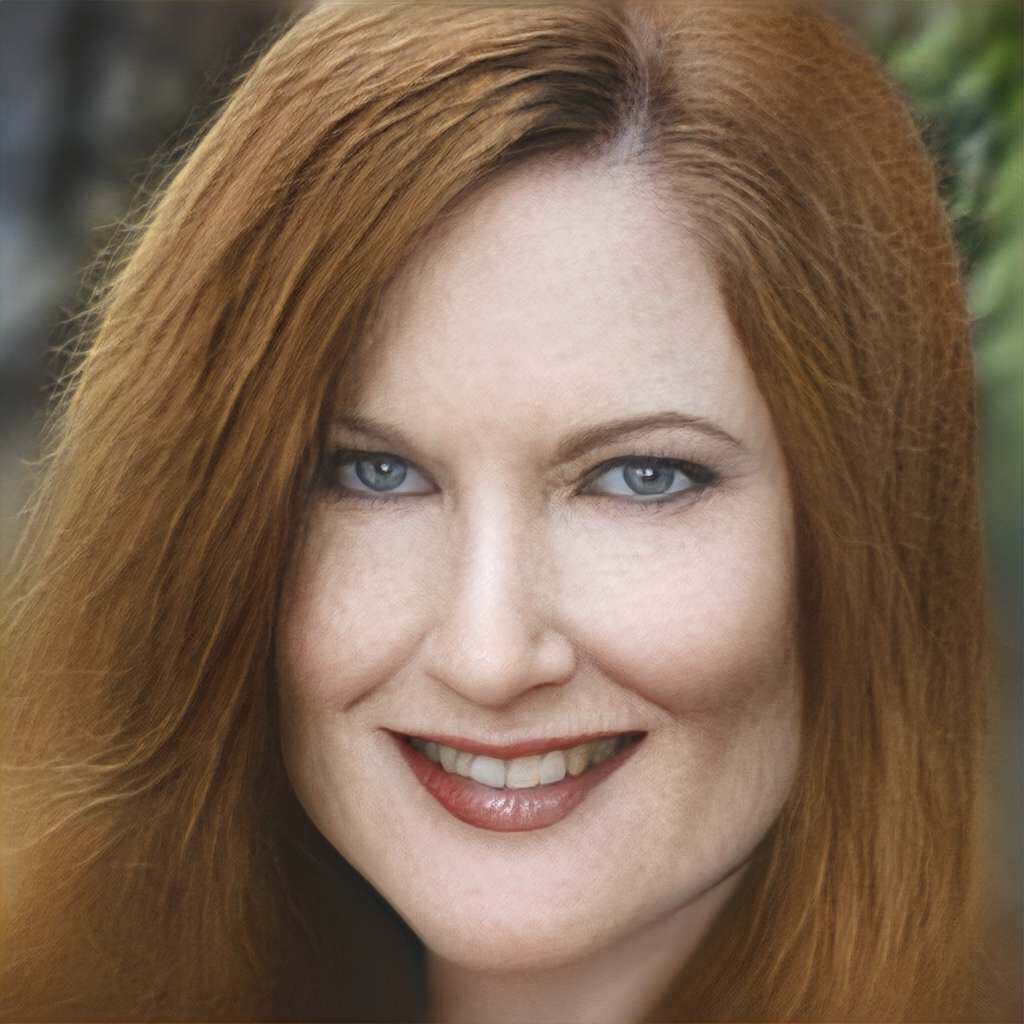}      &
  \includegraphics[width=0.15\linewidth]{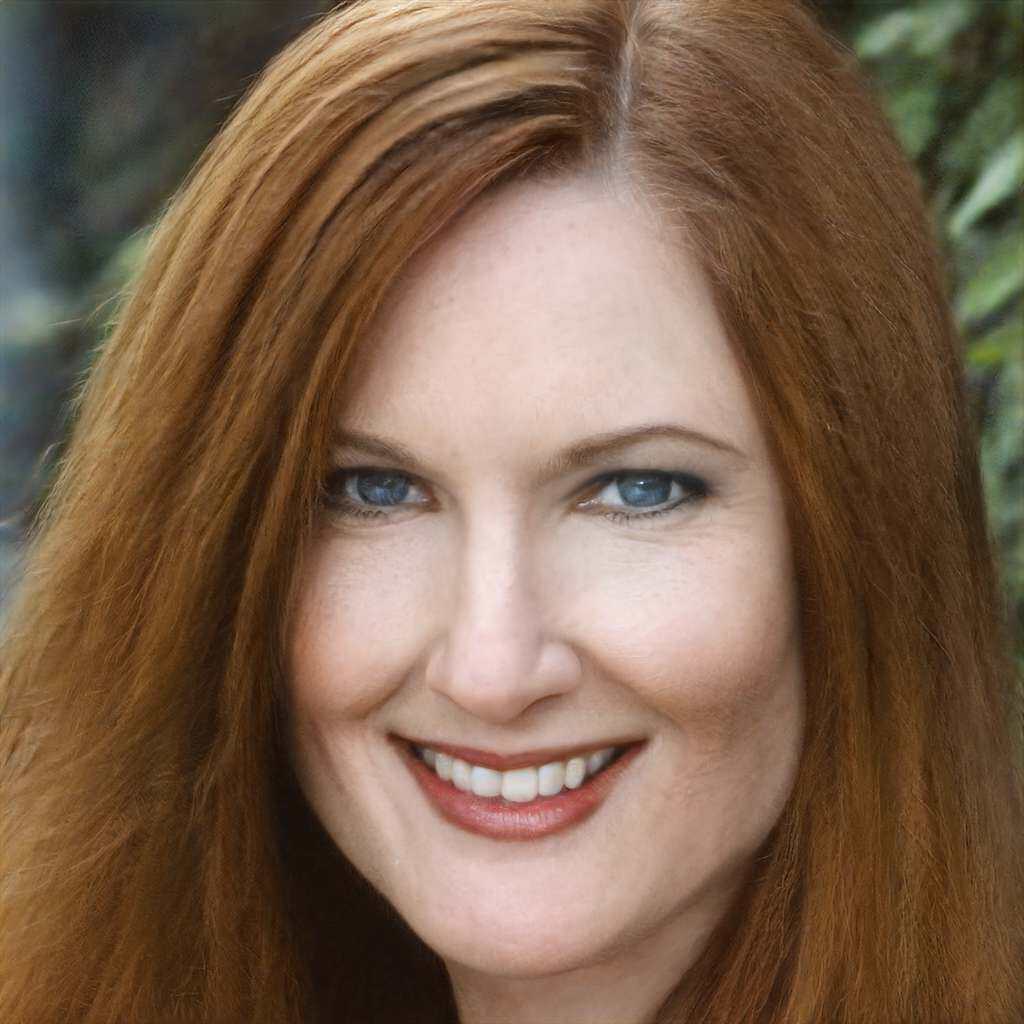}                                                                                                                                                              \\
  \rotatebox{90}{\hspace{10pt} -Smile}                                                                                  &
  \includegraphics[width=0.15\linewidth]{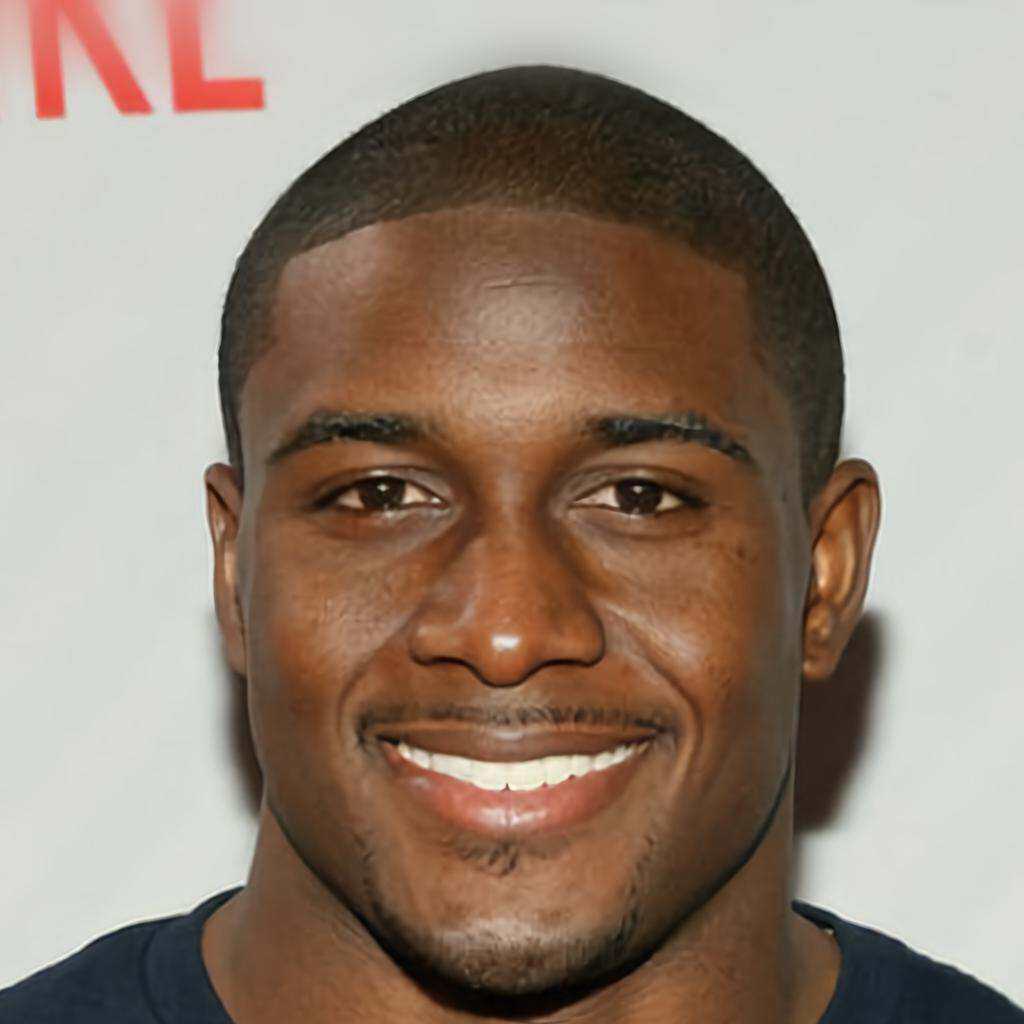}                                             &
  \includegraphics[width=0.15\linewidth]{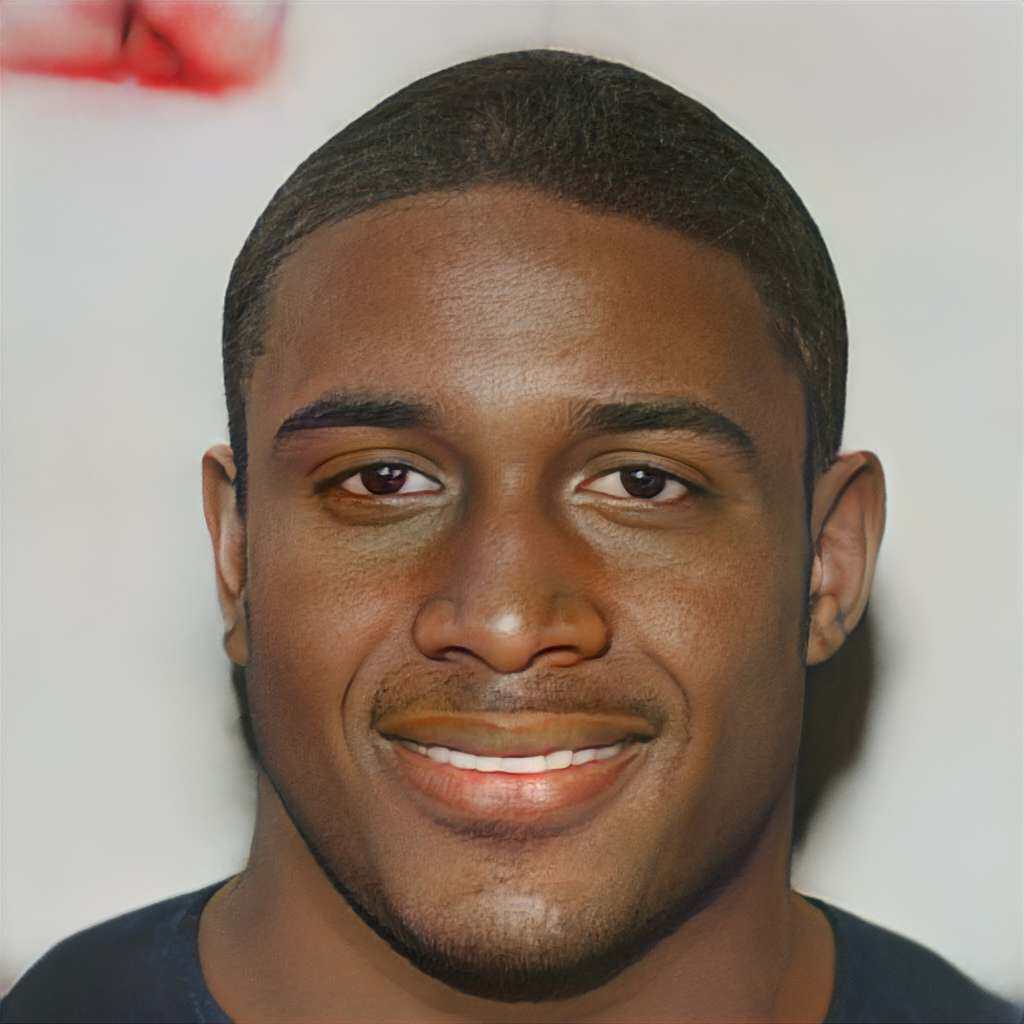}    &
  \includegraphics[width=0.15\linewidth]{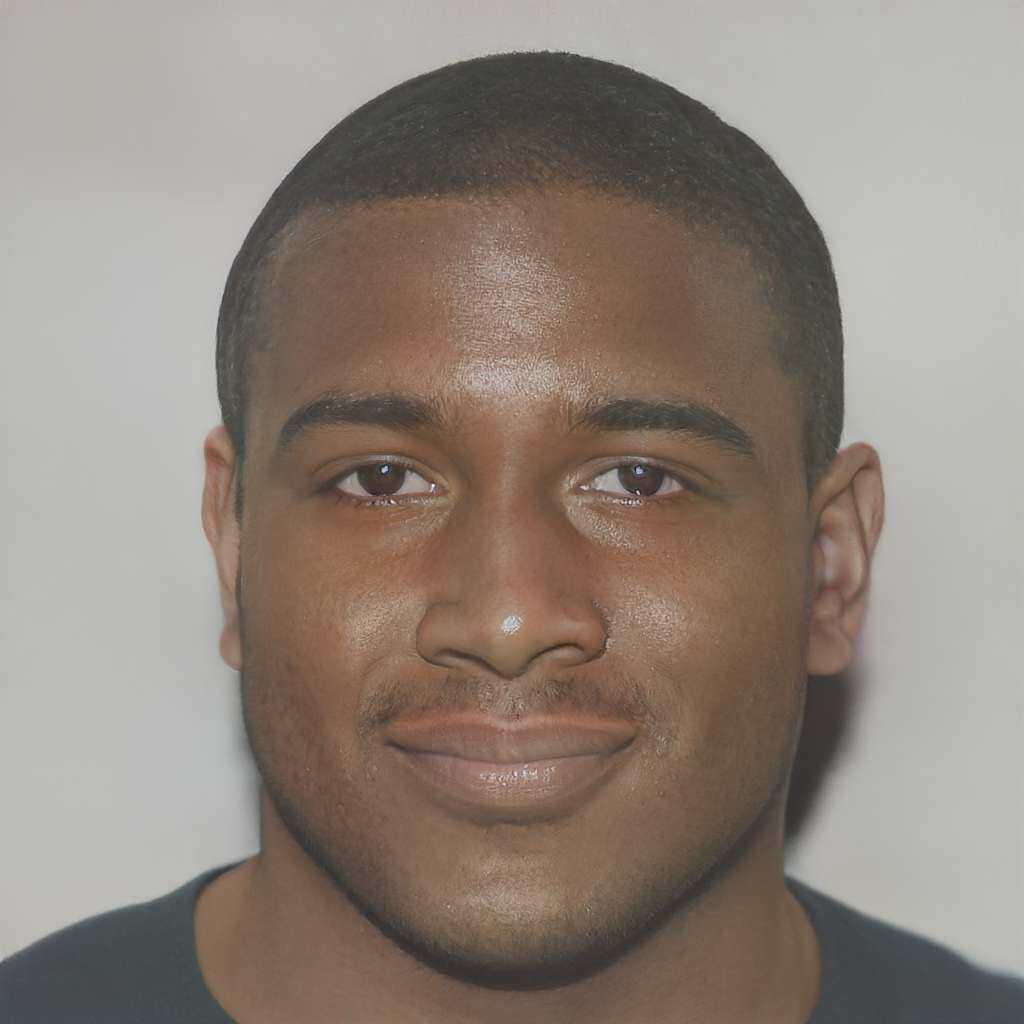}    &
  \includegraphics[width=0.15\linewidth]{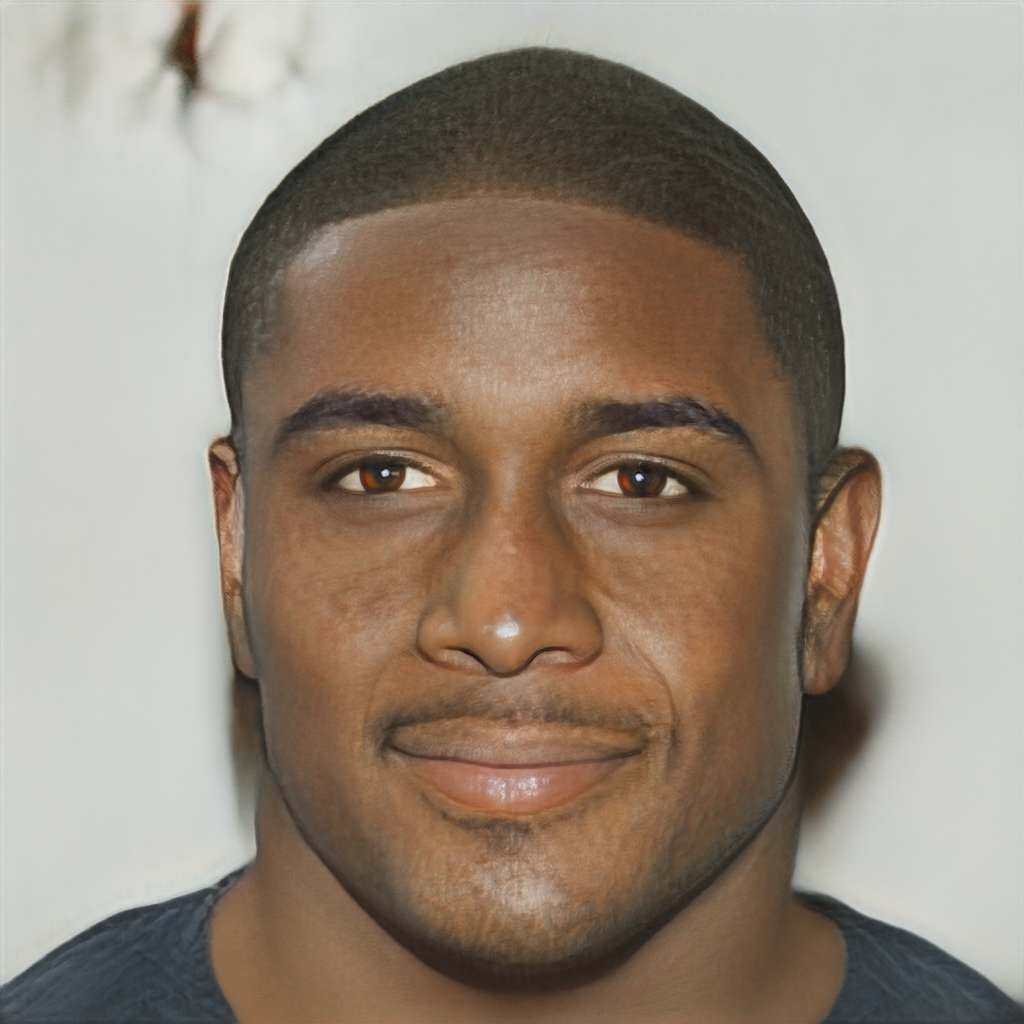}     &
  \includegraphics[width=0.15\linewidth]{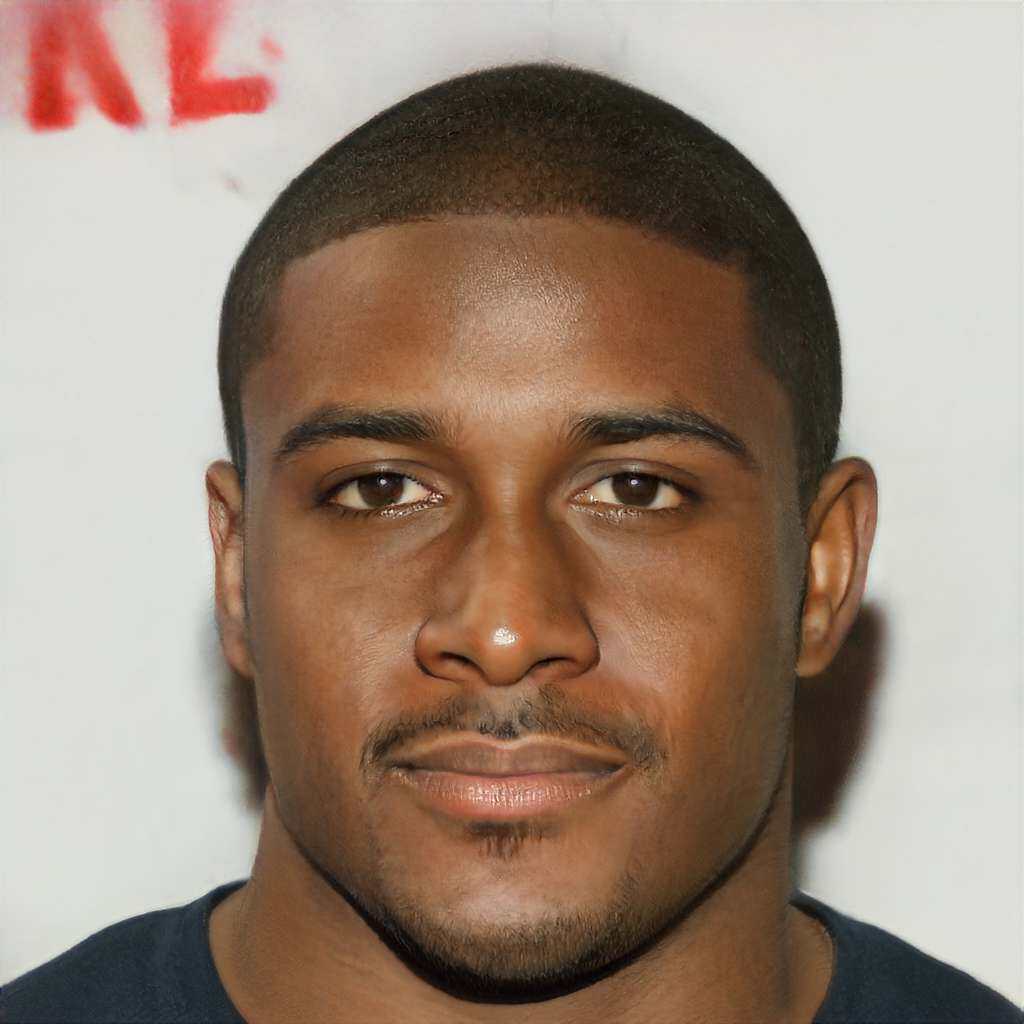}                                                                                                                                                             \\ [5pt]
  & GT & $\text{ReStyle}_{\text{pSp}}$ & $\text{ReStyle}_{\text{e4e}}$ & HyperStyle & \textbf{Ours} \\
\end{tabular}
\caption{%
  Additional Pose, smile and age editing using InterFaceGAN~\cite{shen2020interpreting} of images from the CelebA-HQ dataset.}
\label{fig:additional_edit_celeba_hq_pose_smile_edit}
\end{figure}

\begin{figure}[t]
\centering
\footnotesize
\renewcommand{\arraystretch}{0.0}
\begin{tabular}{@{}c@{\hspace{2pt}}c@{\hspace{2pt}}:c@{}c@{}c@{}c@{}}
  \rotatebox{90}{\hspace{4pt} +Grass}                                                                                     &
  \includegraphics[width=0.16\linewidth]{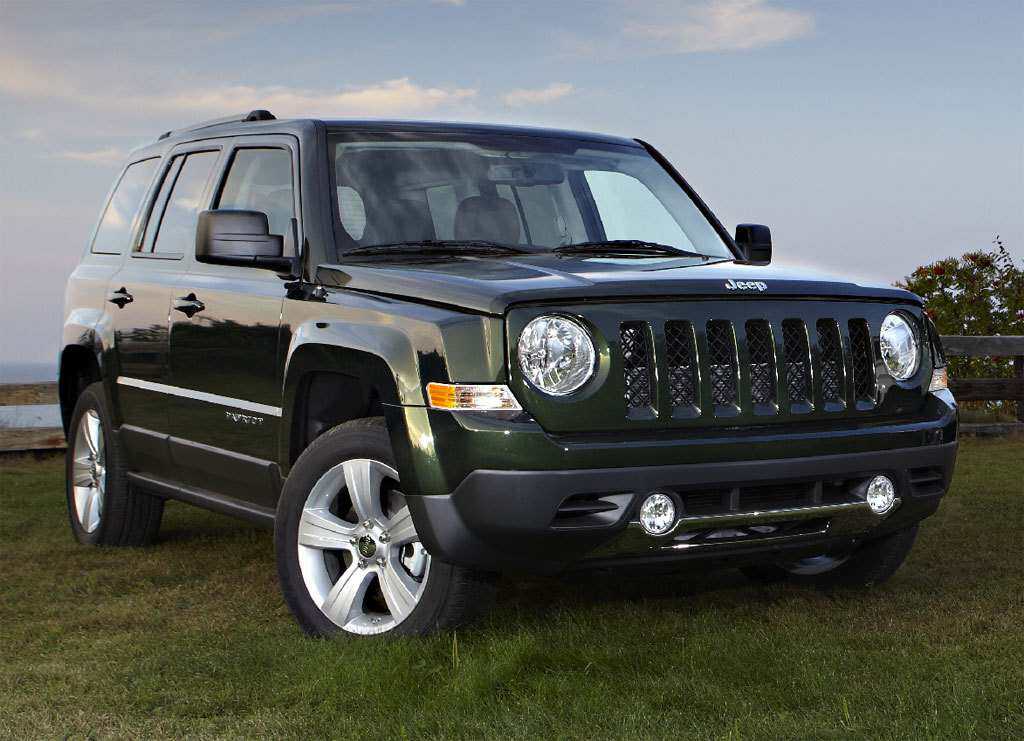}                                            &
  \includegraphics[width=0.16\linewidth]{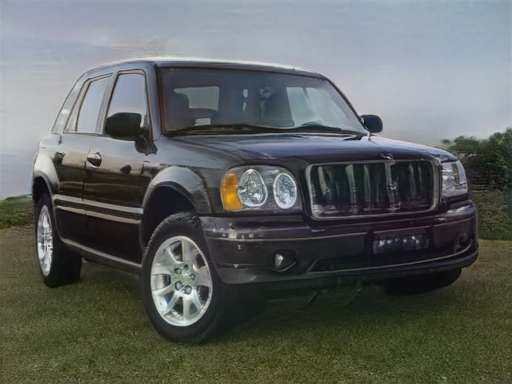}      &
  \includegraphics[width=0.16\linewidth]{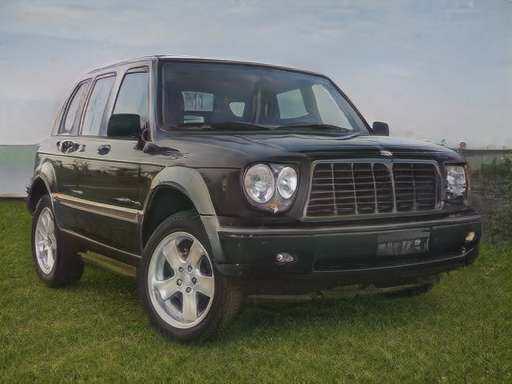}      &
  \includegraphics[width=0.16\linewidth]{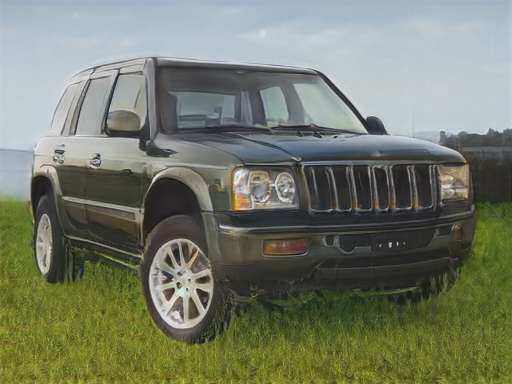}       &
  \includegraphics[width=0.16\linewidth]{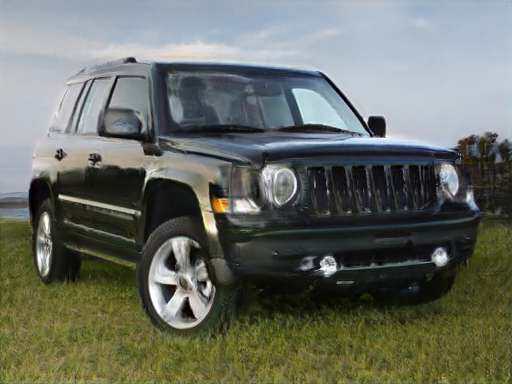}                                                                                                                                                     \\
  \rotatebox{90}{\hspace{2pt} +Reflect}                                                                                &
  \includegraphics[width=0.16\linewidth]{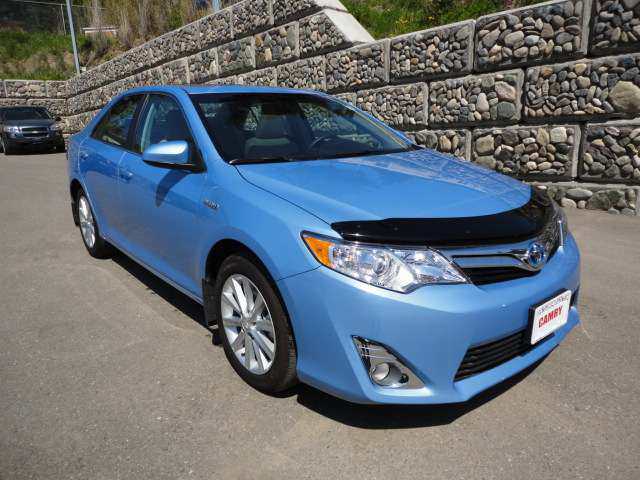}                                            &
  \includegraphics[width=0.16\linewidth]{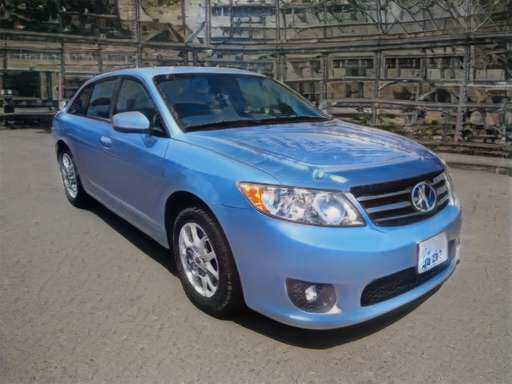} &
  \includegraphics[width=0.16\linewidth]{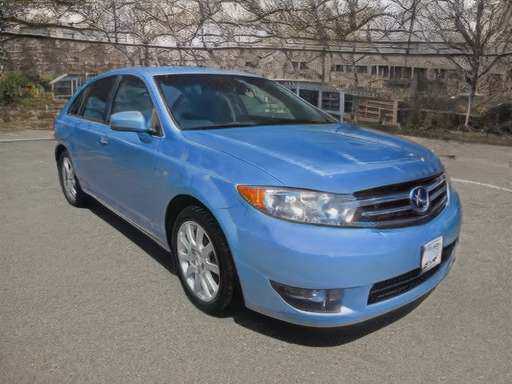} &
  \includegraphics[width=0.16\linewidth]{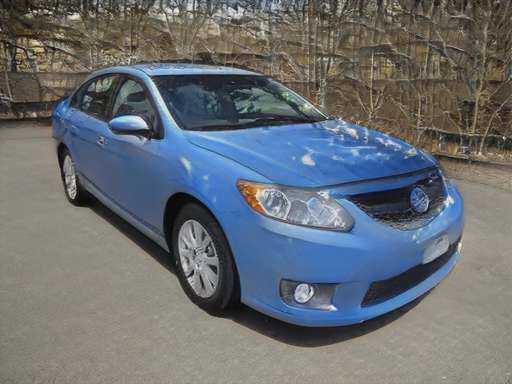}  &
  \includegraphics[width=0.16\linewidth]{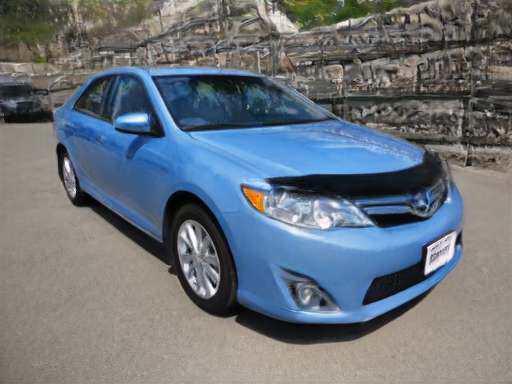}                                                                                                                                                \\
  \rotatebox{90}{\hspace{2pt} +Color}                                                                                     &
  \includegraphics[width=0.16\linewidth]{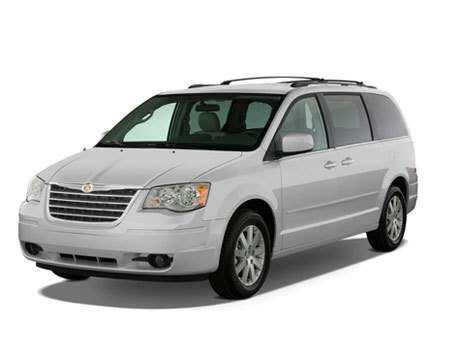}                                            &
  \includegraphics[width=0.16\linewidth]{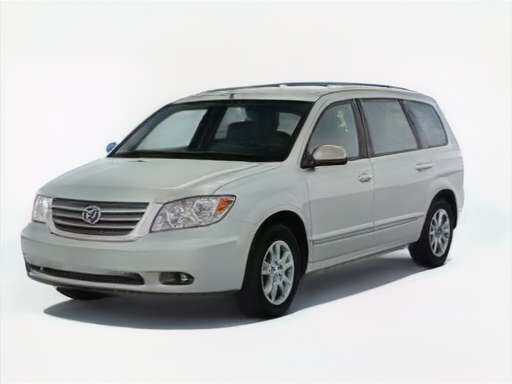}      &
  \includegraphics[width=0.16\linewidth]{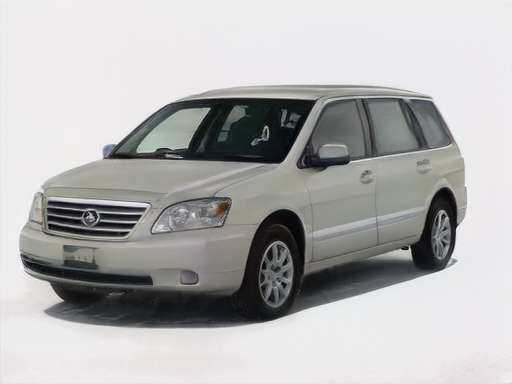}      &
  \includegraphics[width=0.16\linewidth]{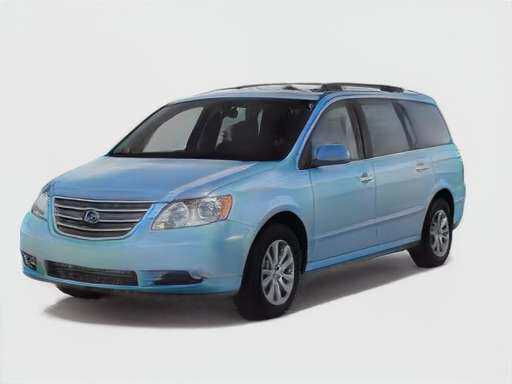}       &
  \includegraphics[width=0.16\linewidth]{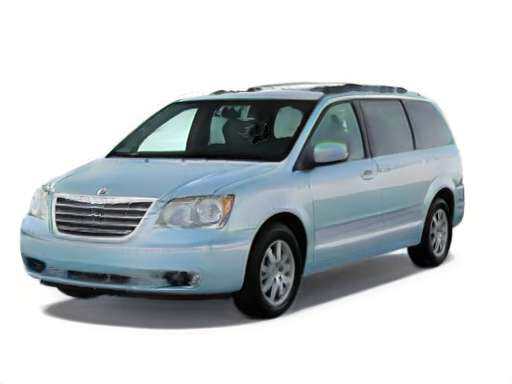}                                                                                                                                                     \\
  \rotatebox{90}{\hspace{2pt} +Model}                                                                                     &
  \includegraphics[width=0.16\linewidth]{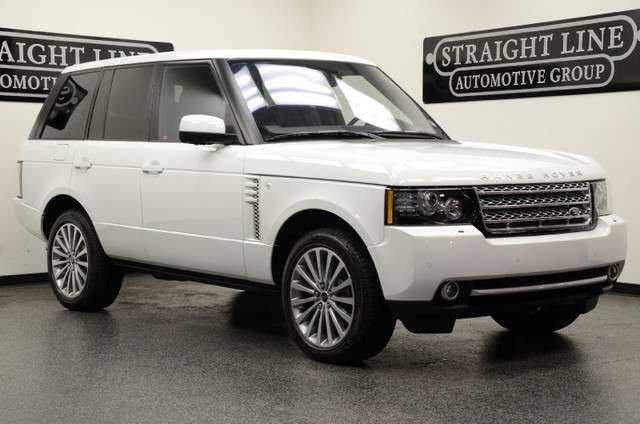}                                            &
  \includegraphics[width=0.16\linewidth]{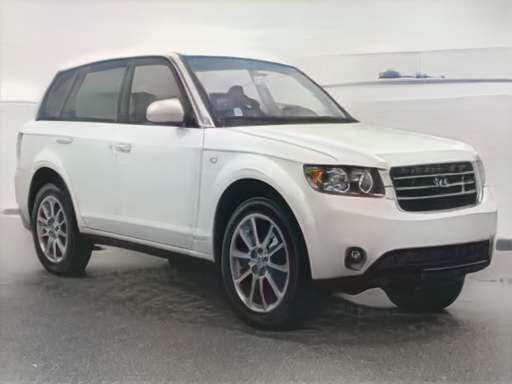}  &
  \includegraphics[width=0.16\linewidth]{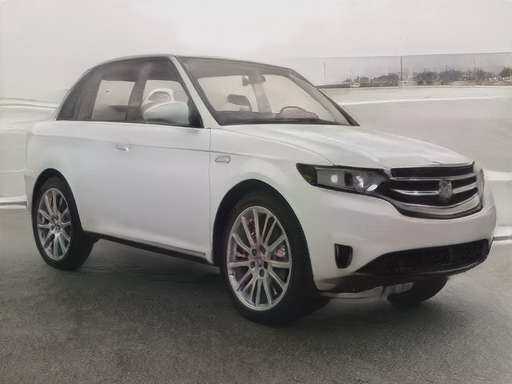}  &
  \includegraphics[width=0.16\linewidth]{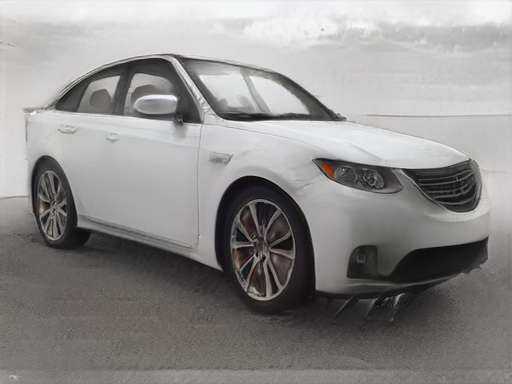}   &
  \includegraphics[width=0.16\linewidth]{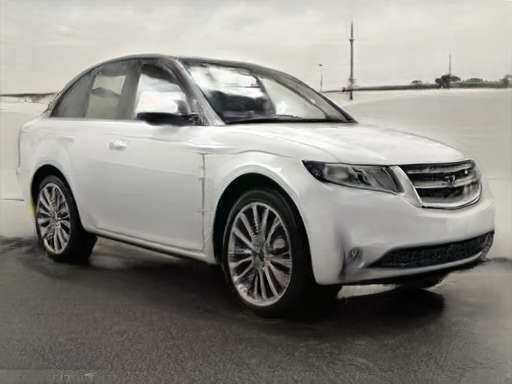}                                                                                                                                                 \\ [5pt]
                                                                                                                           & GT &  $\text{ReStyle}_{\text{pSp}}$ & $\text{ReStyle}_{\text{e4e}}$ & HyperStyle & \textbf{Ours} \\
\end{tabular}
\caption{%
  Additional visual comparison of editing quality (Stanford Cars). Edits were obtained using GANSpace~\cite{harkonen2020ganspace}}
\end{figure}

\begin{figure}[t]
\centering
\footnotesize
\renewcommand{\arraystretch}{0.0}
\begin{tabular}{@{}c@{\hspace{2pt}}c@{\hspace{2pt}}:c@{}c@{}c@{}}
  \rotatebox{90}{\hspace{26pt} Rider}                                                                                  &
  \includegraphics[width=0.2\linewidth]{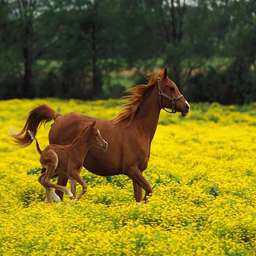}                                          &
  \includegraphics[width=0.2\linewidth]{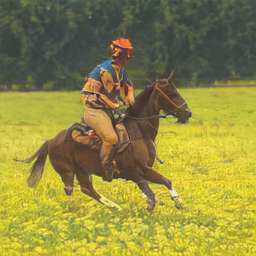}         &
  \includegraphics[width=0.2\linewidth]{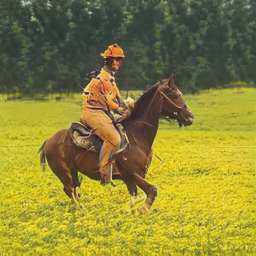} &
  \includegraphics[width=0.2\linewidth]{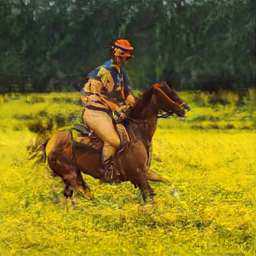}                                                                                           \\
  \rotatebox{90}{\hspace{26pt} Color}                                                                                  &
  \includegraphics[width=0.2\linewidth]{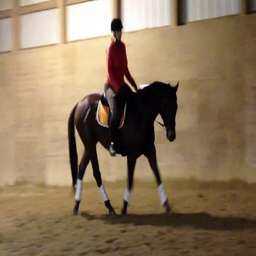}                                          &
  \includegraphics[width=0.2\linewidth]{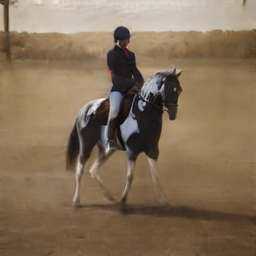}          &
  \includegraphics[width=0.2\linewidth]{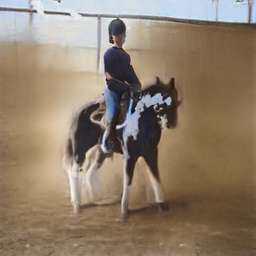}  &
  \includegraphics[width=0.2\linewidth]{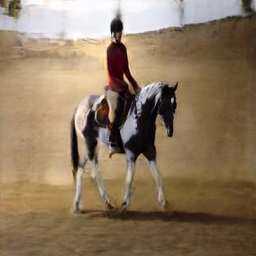}                                                                                            \\ [5pt]
                                                                                                                       & GT & e4e & $\text{ReStyle}_{\text{e4e}}$ & \textbf{Ours} \\
\end{tabular}
\caption{%
  Visual comparison of editing quality (LSUN Horses). Edits were obtained using GANSpace~\cite{harkonen2020ganspace}}
\label{fig:lsun_horse_editing_baseline}
\end{figure}

\end{document}